\documentclass{article} % For LaTeX2e
\usepackage{iclr2026_conference,times}
\usepackage{booktabs}  % for \toprule, \midrule, \bottomrule
\usepackage{amsmath}   % for math symbols
\usepackage{amssymb}   % for checkmarks
\usepackage{algorithm}
\usepackage{algorithmic}
\usepackage{amsmath,amsfonts,bm}

\def\eqref#1{equation~\ref{#1}}
\def\1{\bm{1}}

\DeclareMathAlphabet{\mathsfit}{\encodingdefault}{\sfdefault}{m}{sl}
\SetMathAlphabet{\mathsfit}{bold}{\encodingdefault}{\sfdefault}{bx}{n}

\usepackage{graphicx}
\usepackage{hyperref}
\usepackage{url}
\usepackage{listings}
\usepackage{float}
\usepackage{placeins}
\usepackage{xcolor}
\usepackage{longtable}
\usepackage{booktabs}
\usepackage{multirow}

\usepackage{caption}
\usepackage{etoolbox}
\usepackage{float} 

\AtBeginEnvironment{algorithmic}{%
  \setlength{\itemsep}{0pt}%
  \setlength{\parskip}{0pt}%
  
}

\usepackage{etoolbox}

  {\vspace{-8pt}\begin{algorithm}[#1]\captionsetup{aboveskip=2pt, belowskip=2pt}}
  {\end{algorithm}\vspace{-8pt}}

\title{AEGIS: Authentic Edge Growth In Sparsity for Link Prediction in Edge-Sparse Bipartite Knowledge Graphs}

\author{Hugh Xuechen Liu, Kıvanç Tatar\\
Chalmers University of Technology and University of Gothenburg \\
\texttt{xuechen@chalmers.se, tatar@chalmers.se}
}

\newcommand{\fix}{\marginpar{FIX}}
\newcommand{\new}{\marginpar{NEW}}

\iclrfinalcopy % Uncomment for camera-ready version, but NOT for submission.

\begin{document}

\maketitle

\begin{abstract}
Bipartite knowledge graphs in niche domains are typically data-poor and edge-sparse, which hinders link prediction. We introduce AEGIS (Authentic Edge Growth In Sparsity), an edge-only augmentation framework that resamples existing training edges—either uniformly simple or with inverse-degree bias degree\_aware—thereby preserving the original node set and sidestepping fabricated endpoints. To probe authenticity across regimes, we consider naturally sparse graphs (game design pattern’s game–pattern network) and induce sparsity in denser benchmarks (Amazon, MovieLens) via high-rate bond percolation. We evaluate augmentations on two complementary metrics: AUC-ROC (higher is better) and the Brier score (lower is better), using two-tailed paired $t$-tests against sparse baselines. On Amazon and MovieLens, copy-based AEGIS variants match the baseline while the semantic KNN augmentation is the only method that restores AUC and calibration; random and synthetic edges remain detrimental. On the text-rich GDP graph, semantic KNN achieves the largest AUC improvement and Brier score reduction, and simple also lowers the Brier score relative to the sparse control. These findings position authenticity-constrained resampling as a data-efficient strategy for sparse bipartite link prediction, with semantic augmentation providing an additional boost when informative node descriptions are available.

\end{abstract}

\section{Introduction}

Bipartite graphs are two-mode structures; single-relation bipartite graphs \citep{newman2018networks,latapy2008basic} naturally capture many knowledge-centric applications (e.g., movie–genre, product–category), where the task is to decide whether a single relation exists between two node types. In niche domains, these graphs are often extremely sparse: many nodes have only a handful of incident edges, supervision becomes scarce, and link prediction must proceed with very limited evidence.

This study tackles edge sparsity by comparing five edge-augmentation strategies (uniform authentic, inverse-degree-biased authentic, random ER-like, perturbation-based synthetic, and semantic-KNN) and contributes in three ways: 
\begin{itemize}
    \item We design a stress test for edge-limited bipartite link prediction—applying high-rate bond percolation, augmenting edges solely within the training split, and evaluating with threshold-independent metrics (AUC and Brier score)—without claiming causal disentanglement of sparsity factors.
    \item We introduce Authentic Edge Growth in Sparsity (AEGIS), an edge-only augmentation that replicates observed links in a structure-consistent manner (uniform or inverse-degree biased) while preserving the original node set.
    \item We provide an empirical study on two benchmarks (MovieLens, Amazon) and a domain case study (GDP), showing how authenticity-constrained copies act as a strong sparsity baseline and deliver calibration gains in text-rich settings, while semantic augmentation becomes essential when richer node descriptions are available.
\end{itemize}

\section{Related Work}
In this section, we briefly review related work that forms the background for our study. We begin by describing single-relation bipartite knowledge graphs and link prediction as a core task, with particular attention to the imbalanced degree distributions that commonly arise in practice and motivate our augmentation strategies. Next, we survey graph data augmentation methods, especially those relevant to edge-level augmentation, to contextualize our approach. Finally, we introduce the concepts of edge sparsity, percolation, and homophily in graph structures, which underpin our workflow: edge dropping is applied only to benchmark datasets to induce sparsity, while both benchmark and case study graphs are subsequently augmented via multiple edge-level policies.

\paragraph{Single-relation Bipartite (two-mode) Knowledge Graphs and Link Prediction.}
Knowledge graph completion (KGC) broadly covers inferring missing entities and relations. Usually, a knowledge graph defines a tuple in the form of ``(head entity) ~$\rightarrow$~ (relation) ~$\rightarrow$~  (tail entity)''. And KGC aims to predict the relation among a given head and entity, or the entity given the other, and the relation. A special case of KGC is the binary link prediction - estimate the probability that a link exists between a head and a tail entity - in the context of single-relation bipartite (two-mode) knowledge graphs. The working definition of single-relation bipartite (two-mode) knowledge graphs in this study is a graph that only has two kinds of nodes (e.g., A and B), and there is only one directed relation that exists in this graph (A ~$\rightarrow$~ B). Bipartite (two-mode) network analysis highlights side-specific degree patterns and component structure \citep{latapy2008basic,newman2018networks}. In practice, bipartite graphs often exhibit imbalanced degree distributions across modes \citep{latapy2008basic,newman2018networks}. These imbalanced degree distributions shape component structure and intensify cold-start behavior for low-degree nodes, especially under high-rate edge dropping \citep{schein2002methods,rong2019dropedge}. This motivates inverse-degree authentic resampling: a conservative way to allocate limited augmentation budget toward sparsity-affected endpoints without inventing new nodes or altering the two-mode constraint. While we do not claim causal disentanglement, this design aligns with observed failure modes in edge-sparse, long-tail regimes \citep{newman2018networks,steck2011item}.

\paragraph{Graph Data Augmentation.} Graph data augmentation creates plausible variants of graph data without extra labeling to expand training signals \citep{zhao2022graph}. Methods can be organized along two orthogonal axes: (i) whether the policy is learned vs.\ rule-based, (ii) the task level (node/edge/graph), and (iii) the operation modality (structure, features, or labels) \citep{zhao2022graph,zhou2025data,ding2022data}. Rule-based structural regularizers such as DropEdge \citep{rong2019dropedge} and DropNode \citep{feng2020graph} randomly remove components during training and work well as anti-overfitting in dense settings, but can be counterproductive under edge sparsity where supervision is already limited. Beyond subtractive policies, additive strategies aim to increase effective connectivity or introduce informative structure. Interpolation-based methods (e.g., GraphSMOTE \citep{zhao2021graphsmote}, FG-SMOTE \citep{wang2025fg}) adapt oversampling to graphs by interpolating features or ties, while generative/counterfactual approaches (e.g., CFLP \citep{zhao2022learning}, CLBR \citep{zhu2023data}, AGGG \citep{wang2025graph}) synthesize training instances by modeling causal or distributional structure. As null baselines, random edge additions resemble two-mode Erd\H{o}s--R'enyi draws \citep{erdds1959random,newman2018networks}, and synthetic index perturbations play the role of stress tests rather than realistic augmentation. Our work focuses on a rule-based, edge-only, train-only policy ---authenticity-constrained edge resampling --- that replicates observed ties under type constraints to densify supervision around real patterns, contrasting with attribute-similarity completion and null additions. In the edge sparsity regime studied here, we observe that such authenticity constraints offer more reliable improvements in both ROC-AUC and Brier score compared to random or synthetic additions; semantic-only completion can raise ROC-AUC but does not consistently improve calibration as measured by the Brier score under class imbalance.

\paragraph{Edge sparsity, Percolation and Homophily.}Random high-rate edge dropping corresponds to bond percolation on networks, which linearly scales mean degree and induces component fragmentation \citep{newman2002spread,newman2018networks}. Attribute-similarity (homophily) is a common mechanism for tie formation \citep{mcpherson2001birds}, informing semantic-KNN completions. However, precision--recall behavior under imbalance can diverge from ROC improvements \citep{davis2006relationship,bi2024inconsistency}.

\section{Problem Formulation}
Let ~$G=(U,V,E)$~ be a single-relation bipartite graph with ~$U$~ and ~$V$~ disjoint node sets, and $E$ is a set of edges. We consider binary link prediction on a bipartite graph \(G=(U,V,E)\) as estimating, for each candidate pair $(u,v)$ where $u \in U$ and $v \in V$, the probability \(P\big(u,v)=\sigma\!\big(s(u,v)\big)\), where \(s(u,v)\) is a learned scoring function (e.g., dot‑product \(s(u,v)=\mathbf{h}_u^\top \mathbf{h}_v\), bilinear \(s(u,v)=\mathbf{h}_u^\top \mathbf{W}\mathbf{h}_v\), or cosine similarity). We train with a class‑balanced binary cross‑entropy over observed positives and sampled negatives; evaluation reports threshold‑independent metrics (AUC‑ROC, Brier score).

\paragraph{Edge sparsity regime (bond percolation).} We study a scenario where sparsity is induced by random edge dropping at a high rate (bond percolation; retain rate $q$ is ~$0\ll q\ll1$), which proportionally reduces side-specific mean degrees, lowers global edge density, and fragments component structure \citep{newman2002spread,newman2018networks}. Our goal is a scenario-driven evaluation of augmentation policies under edge sparsity, not a causal decomposition of which attribute drives performance. A full causal decomposition of which specific graph attributes drive performance is out of scope for this paper, but may be explored in future work; here, our focus is on scenario-driven evaluation of augmentation policies under edge sparsity.

\paragraph{Evaluation.} We report two complementary metrics: (i) AUC-ROC, where higher values indicate better ranking across thresholds, and (ii) the Brier score, where lower values indicate better probabilistic calibration and overall predictive reliability \citep{glenn1950verification,bi2024inconsistency}. This combination lets us assess whether an augmentation both separates positives from negatives and assigns calibrated link probabilities. Following APA guidelines\footnote{\url{https://apastyle.apa.org/style-grammar-guidelines/tables-figures/sample-tables}}, we present each method’s $M \pm SD$ along with two-tailed paired Student $t$-tests (df $=31$) against the sparse baseline. Tables include the $t$-statistic, $p$-value, and Cohen’s $d$, with significance levels ($p<.05$, $p<.01$, $p<.001$) flagged by asterisks to show when observed differences are unlikely to arise by chance.

\section{Methodology}
\subsection{Authenticity-Constrained Edge Resampling}
We define authentic edge growth in sparsity (AEGIS) as empirical tie resampling: duplicating observed training edges (with replacement) under type constraints, without introducing new nodes or synthetic endpoints. AEGIS preserves observed relational patterns and respects the two-mode structure, contrasting with two-mode ER-like random additions \citep{erdds1959random} or interpolation-based synthesis \citep{chawla2002smote}. To avoid leakage, augmentation applies only to the training graph's edge index; validation/test graphs and labels remain unchanged. We instantiate AEGIS with two sampling policies: (i) uniform resampling (``simple''), sampling existing edges uniformly; and (ii) low-degree-biased resampling (``degree-aware''), sampling with probability inversely proportional to endpoint degrees to prioritize low-degree nodes (cold-start mitigation). The procedures below cover both authentic policies and the contrastive baselines.

\subsection{Augmentation Methods}
All augmentations operate only on the \emph{training} subgraph's forward edge index; we do not add nodes, and we do not modify validation/test graphs or labels, avoiding leakage. In this study, we compare five distinct edge augmentation policies to address sparsity in bipartite knowledge graphs: AEGIS-Simple uniformly resamples observed edges from the training set, duplicating existing links without creating new endpoints. AEGIS-Degree applies an inverse-degree bias to resampling, preferentially augmenting edges for low-degree nodes to mitigate cold-start issues. The Random ER-like policy introduces edges between randomly selected node pairs, simulating two-mode Erdős–Rényi random graphs \citep{erdds1959random}. Perturbation-based synthetic augmentation generates new edges by perturbing the indices of existing edges in a SMOTE-style fashion \citep{chawla2002smote}. Semantic-KNN completion introduces edges between nodes with high semantic similarity (e.g., high cosine similarity between node features), reflecting homophily-driven tie formation \citep{mcpherson2001birds}.

\begin{algorithm}[!htbp]
\caption{AEGIS-Simple: Uniform Authentic Resampling}
\label{alg:aegis-simple}
\begin{algorithmic}[1]
\REQUIRE An edge $e_i(u,v) \in E $ where $i$ is a unique edge index and augmentation factor $\phi \ge 1$
\ENSURE Augmented set of edges $E_{aug}$
\STATE $ n_e \gets |E|$, $n'_e \gets \lfloor (\phi - 1) n_e \rfloor$
\STATE Initialize $E' \gets \emptyset$
\WHILE{$|E'| < n'_e$}
    \STATE $e_x(u,v) \sim \mathcal{U}(E)$, where $\mathcal{U}$ is the uniform distribution.
    \STATE $E' \gets E' \cup \{e_x(u,v)\}$
\ENDWHILE
\STATE $E_{aug} \gets E \cup E'$
\RETURN $E_{aug}$
\end{algorithmic}
\end{algorithm}

\begin{algorithm}[!htbp]
\caption{AEGIS-Degree: Inverse-Degree-Biased Authentic Resampling}
\label{alg:aegis-degree}
\begin{algorithmic}[1]
\REQUIRE An edge $e_i(u,v) \in E$ where $i$ is a unique edge index, augmentation factor $\phi \ge 1$, smoothing constant $\varepsilon > 0$
\ENSURE Augmented set of edges $E_{aug}$
\STATE $n_e \gets |E|$, $n'_e \gets \lfloor (\phi - 1) n_e \rfloor$
\STATE $E_k \gets e_i(u_k,v)$ and $E_l \gets e_i(u,v_l)$ where $u_k \in U$ and $v_l \in V$
\STATE $deg(u_k) \gets |E_k|$, $deg(v_l) \gets |E_l|$
% \STATE Compute degree vectors $\deg_U$, $\deg_V$ from $E$
\STATE $w(e_i) \gets \frac{1}{\deg(u_k)+\varepsilon} + \frac{1}{\deg(v_l)+\varepsilon}$
\STATE Normalize $P_E \gets w(e_i) / \sum_{i} w(e_i)$
\STATE Initialize $E' \gets \emptyset$
\WHILE{$|E'| < n'_e$}
    \STATE Sample $e_x(u,v) \sim P_E$
    \STATE $E' \gets E' \cup \{e_x(u,v)\}$
\ENDWHILE
\STATE $E_{aug} \gets E \cup E'$
\RETURN $E_{aug}$
\end{algorithmic}
\end{algorithm}

\begin{algorithm}[!htbp]
\caption{Random ER-Like Augmentation}
\label{alg:random}
\begin{algorithmic}[1]
\REQUIRE An edge $e_i(u,v)  \in E$ where $i$ is a unique edge index, $n_u \gets |U|$  , $n_v \gets |V| $ , augmentation factor $\phi \ge 1$
\ENSURE Augmented set of edges $E_{aug}$
\STATE $n_e \gets |E|$, $n'_e \gets \lfloor (\phi - 1) n_e \rfloor$
\STATE Initialize $E' \gets \emptyset$
\WHILE{$|E'| < n'_e$}
    \STATE $e_x(u,v)$ where $u \sim \mathcal{U}(U), v \sim \mathcal{U}(V)$, where $\mathcal{U}$ is the uniform distribution.
    \STATE $E' \gets E' \cup \{e_x(u,v)\}$
\ENDWHILE
\STATE $E_{aug} \gets E \cup E'$
\RETURN $E_{aug}$
\end{algorithmic}
\end{algorithm}

\begin{algorithm}[t]
\caption{Perturbation-based Synthetic Augmentation}
\label{alg:synthetic}
\begin{algorithmic}[1]
\REQUIRE An edge $e_i(u,v) \in E$ where $i$ is a unique edge index, augmentation factor $\phi \ge 1$, perturbation radius $r$, $n_u \gets |U|$  , $n_v \gets |V| $
\ENSURE Augmented set of edges $E_{aug}$
\STATE $n_e \gets |E|$, $n'_e \gets \lfloor (\phi - 1) n_e \rfloor$
\STATE Initialize $E' \gets \emptyset$
\WHILE{$|E'| < n'_e$}  
\STATE $e_x(u_j,v_k) \sim \mathcal{U}(E)$, where $\mathcal{U}$ is the uniform distribution.
    \STATE $\delta_u, \delta_v \sim U\{-r, \dots, r\}$
    \STATE $u' \gets \min(\max(u_j + \delta_u, 0), n_u - 1)$
    \STATE $v' \gets \min(\max(v_k + \delta_v, 0), n_v - 1)$
    \STATE $E' \gets E' \cup \{e_x(u,v)\}$
\ENDWHILE
\STATE $E_{aug} \gets E \cup E'$
\RETURN $E_{aug}$
\end{algorithmic}
\end{algorithm}

\begin{algorithm}[t]
\caption{Semantic-KNN Augmentation}
\label{alg:semantic-knn}
\begin{algorithmic}[1]
\REQUIRE An edge $e_i(u,v)  \in E$ where $i$ is a unique edge index, $n_u \gets |U|$, $n_v \gets |V|$, semantic feature matrices $\mathbf{x}_U \in \mathbb{R}^{n_u \times d}$, $\mathbf{x}_V \in \mathbb{R}^{n_v \times d}$ (row-normalized), neighbour parameter $k$, similarity threshold $\tau$, per-node cap $\alpha$, augmentation factor $\phi \ge 1$
\ENSURE Augmented set of edges $E_{aug}$
\STATE $n_e \gets |E|$, $n'_e \gets \lfloor (\phi - 1) n_e \rfloor$
\STATE $T_U \gets \mathbf{x}_U \cdot \mathbf{x}_U^T$, $ T_V \gets \mathbf{x}_V \cdot \mathbf{x}_V^T$, where $T_U$ and $T_V$ are self-similarity tensors based on cosine distance. 

\STATE ${S}_U \gets Knn(T_U,k) $, $S_V \gets Knn(T_V,k)$ where $Knn(T,k)$ selects $k$ elements with the highest self-similarity in $T$.

\STATE $S_U \gets S_U(i) \gg \tau $ and $S_V \gets S_V(i) \gg \tau $ where $\tau$ is a threshold parameter.  
\STATE Initialize $E' \gets \emptyset$,  $c_U(u_i) \gets 0$, $c_V(v_i) \gets 0$
\STATE $n \gets n'_e$
\FOR{each $e_i(u_j,v_k) \in E$}
    \FOR{each $v_{knn}\in \mathcal{S}_V(v_k)$ while $n > 0$}
        \IF{$(u_j, v_{knn}) \notin E \cup E'$ and $c_U(u_j) < \alpha$ and $c_V(v_{knn}) < \alpha$} 
            \STATE $E' \gets E' \cup \{(u_j, v_{knn})\}$
            \STATE $c_U(u_j) \gets c_U(u_j) + 1$, $c_V(v_{knn}) \gets c_V(v_{knn}) + 1$
            \STATE $n \gets n - 1$
        \ENDIF
    \ENDFOR
    \FOR{each $u_{knn} \in \mathcal{S}_U(u_j)$ while $n > 0$}
        \IF{$(u_{knn}, v_k) \notin E \cup E'$ and $c_U(u_{knn}) < \alpha $ and $c_V(v_k) < \alpha$}
            \STATE $E' \gets E' \cup \{(u_{knn}, v_k)\}$
            \STATE $c_U(u_{knn}) \gets c_U(u_{knn}) + 1$, $c_V(v_k) \gets c_V(v_k) + 1$
            \STATE $n \gets n - 1$
        \ENDIF
    \ENDFOR
\ENDFOR
\STATE $E_{aug} \gets E \cup E'$
\RETURN $E_{aug}$
\end{algorithmic}
\end{algorithm}

\section{Experiments}

\subsection{Dataset Statistics and Edge Sparsity Construction}

We evaluate our methods on two widely used benchmark datasets—MovieLens \citep{harper2015movielens} (movie–genre) and Amazon \citep{mcauley2015image} (product–category)—as well as a domain-specific use case, GDP (game design patterns) \citep{bjork2005games}. Details of GDP can be found in Appendix~\ref{appendix-gdp-introduction}. While the benchmark datasets are originally well-connected, we simulate extreme edge sparsity by applying high-rate random bond percolation (i.e., random edge removal) as described by \citet{newman2002spread}. In contrast, the GDP dataset is inherently sparse and does not require additional edge removal. Table~\ref{tab:data_stats} summarizes the key characteristics of each dataset, including the cardinalities of the two node sets, the number of edges in the original graph, the percolation retain rate $q$ used to generate sparse training scenarios, and the resulting number of edges after edge dropping.

\begin{table}[!htbp]
\centering
\caption{Dataset Statistics and Edge Sparsity Construction}
\label{tab:data_stats}
\begin{tabular}{l l r l r r r r}
\toprule
Dataset & Mode ~$U$~ & ~$|U|$~ & Mode ~$V$~ & ~$|V|$~ & ~$|E|$~ (orig) & retain ~$q$~ & ~$|E|$~ (after) \\
\midrule
Amazon    & Products & 1465 & Categories & 317 & 6307 & 0.01 & 67 \\
MovieLens & Movies   & 9708 & Genres     & 19  & 22050 & 0.01 & 213 \\
GDP       & Games    & 208  & Patterns   & 296 & 715   & N/A  & 715 \\

\bottomrule
\end{tabular}
\end{table}
\FloatBarrier

\subsection{Augmentation Budgets}
On the training subgraph (benchmarks after edge dropping and GDP), we target 100$\times$ augmentation. Validation/test graphs remain unchanged. Reported significance is always per-dataset versus its own sparse baseline; original graphs of benchmarks are shown as upper bounds, not budget-matched baselines.

\subsection{Pipeline}

\paragraph{Benchmark Datasets Pipeline.} Bond percolation (edge dropping) on benchmarks (e.g., keep~$\sim$1\% edges) ~$\rightarrow$~ split via RandomLinkSplit (80/10/10; disjoint training ratio) with negative sampling (negative sampling ratio; allow adding negative train samples)~$\rightarrow$~ augmentation (AEGIS-Simple; AEGIS-Degree; Random ER-Like Augmentation; Perturbation-based Synthetic Augmentation; Semantic-KNN Augmentation) with the factor of 100 on \emph{train} graph only ~$\rightarrow$~ Graph Attention Network’s Heterogeneous variant (hetero GAT) training with class-balanced binary cross-entropy loss ~$\rightarrow$~ evaluation. \paragraph{Domain Case Study Pipeline.}
Naturally sparse two-mode graph (no edge drop) ~$\rightarrow$~ customized training/valid set ~$\rightarrow$~ augmentation (AEGIS-Simple; AEGIS-Degree; Random ER-Like Augmentation; Perturbation-based Synthetic Augmentation; Semantic-KNN Augmentation) with the factor of 100 on \emph{train} graph only ~$\rightarrow$~ Graph Attention Network’s Heterogeneous variant (hetero GAT) training with class-balanced binary cross-entropy loss ~$\rightarrow$~ evaluation

The implementation specifics for the results demonstrated in this manuscript can be found in Appendix~\ref{appendix-implementation}

\section{Results}

\subsection{Benchmark Validation}

\begin{figure}[t]
\centering
\includegraphics[width=1\linewidth]{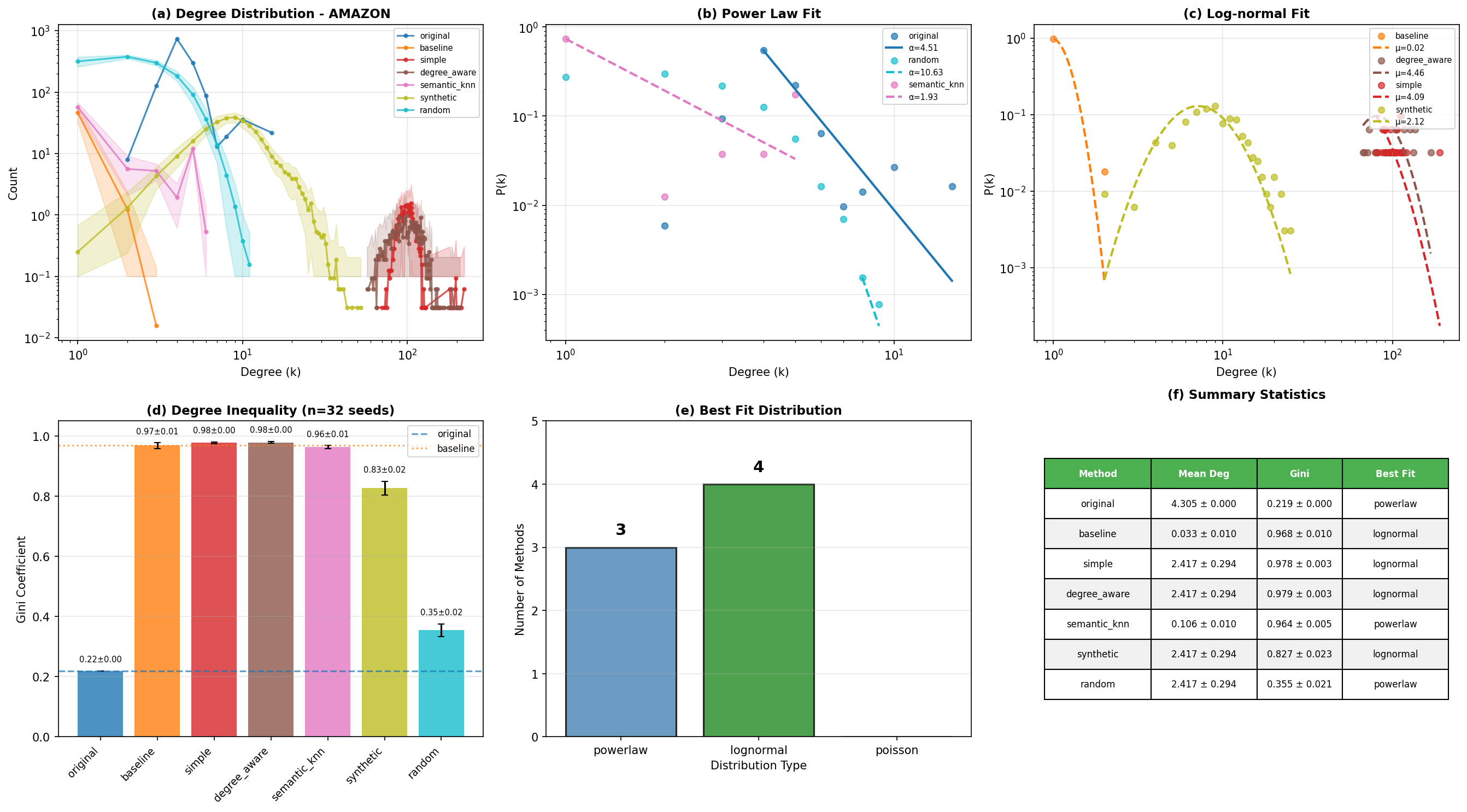}
\caption{Amazon (product--category), GAT, $q{=}0.01$, $\phi{=}100$: Comprehensive degree analysis ($M\pm\mathrm{SD}$, $n=32$ seeds). Panel (a) shows degree distributions on log-log scale with $\pm 1\sigma$ confidence bands; (b) Power Law fits with exponent $\alpha$ (lower $\alpha$ = heavier tail); (c) Log-normal fits with parameters $\mu$ and $\sigma$; (d) Gini coefficients quantifying degree inequality ($0{=}$perfect equality, $1{=}$maximum inequality); (e) best-fit distribution counts (lower KS statistic wins); (f) summary statistics table.}
\label{fig:amazon}
\end{figure}

\begin{figure}[t]
\centering
\includegraphics[width=1\linewidth]{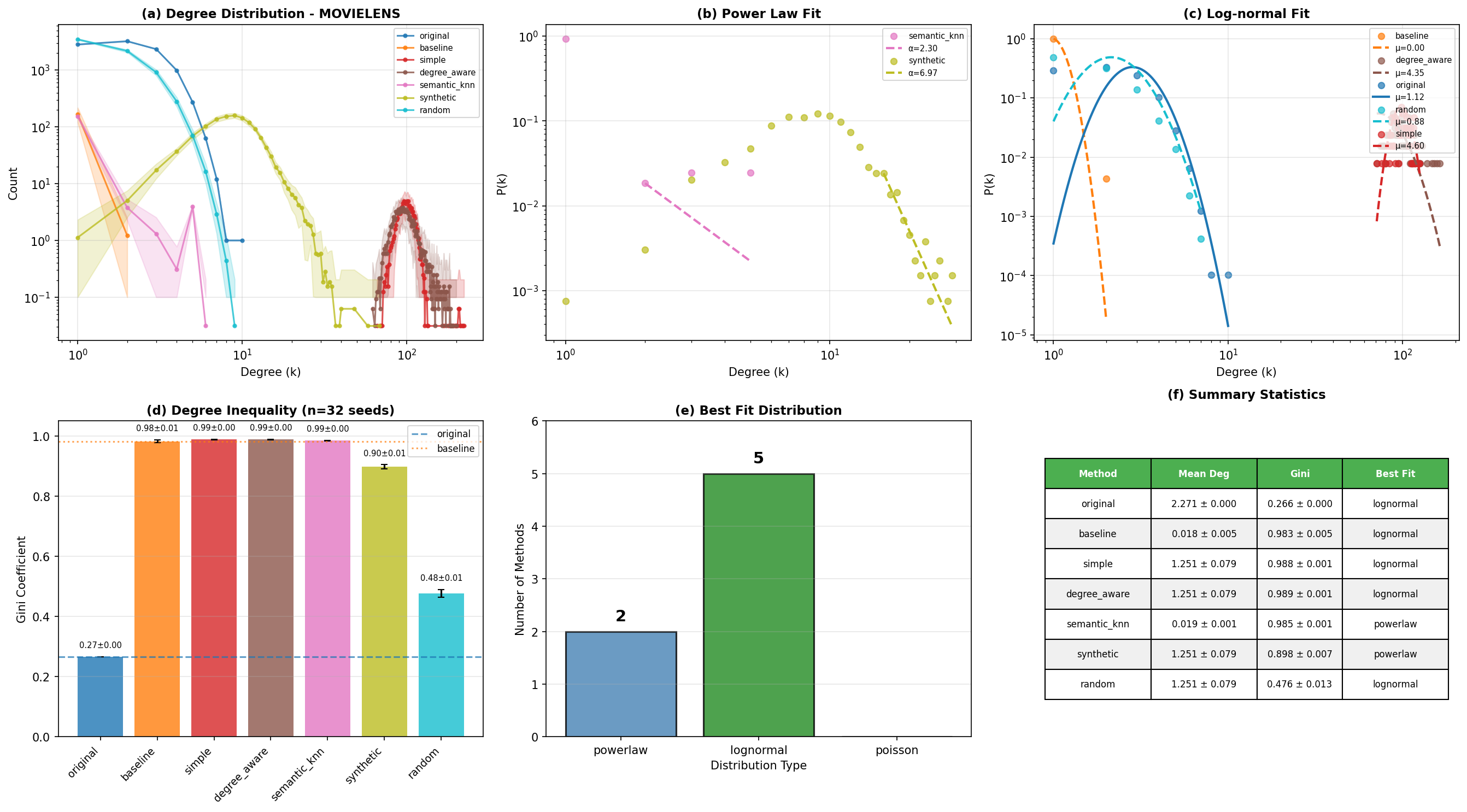}
\caption{MovieLens (movie--genre), GAT, $q{=}0.01$, $\phi{=}100$: Comprehensive degree analysis ($M\pm\mathrm{SD}$, $n=32$ seeds). Panel (a) shows degree distributions on log-log scale with $\pm 1\sigma$ confidence bands; (b) Power Law fits with exponent $\alpha$ (lower $\alpha$ = heavier tail); (c) Log-normal fits with parameters $\mu$ and $\sigma$; (d) Gini coefficients quantifying degree inequality ($0{=}$perfect equality, $1{=}$maximum inequality); (e) best-fit distribution counts (lower KS statistic wins); (f) summary statistics table.}
\label{fig:movielens}
\end{figure}

To probe structural limits under extreme sparsity ($q{=}0.01$, $\phi{=}100$), we inspected degree distributions for Amazon and MovieLens as a diagnostic (not the exact graphs used for AUC/Brier). As shown in Figure~\ref{fig:amazon} and Figure~\ref{fig:movielens}, the copy-style augmentations (\texttt{simple}, \texttt{degree\_aware}) substantially raise mean degree yet retain hub-dominated inequality (Amazon Gini $\approx0.98$, MovieLens $\approx0.99$), consistent with faithfully resampling the sparse baseline topology. \texttt{Random} flattens the distribution (Amazon Gini $0.36$, MovieLens $0.48$) but erodes structure; \texttt{synthetic} sits in between. \texttt{Semantic\_knn} stays very sparse because similarity thresholds block most edges (mean degree $0.106$ Amazon, $0.019$ MovieLens). None recover the original moderate inequality (Amazon Gini $0.219$, MovieLens $0.266$), highlighting a trade-off between structural fidelity and connectivity when dropout is extreme.

Tables~\ref{tab:amazon_auc} and ~\ref{tab:movielens_auc} show that the original graphs function as upper bounds and that copy‑style AEGIS variants  (\texttt{simple}, \texttt{degree\_aware}) stay statistically indistinguishable from the sparse baselines. Only \texttt{semantic\_knn} achieves meaningful gains (+0.091 on Amazon), while random and synthetic additions drive AUC down, especially on MovieLens. Tables~\ref{tab:amazon_brier} and ~\ref{tab:movielens_brier} indicate that \texttt{semantic\_knn} improves or preserves calibration (e.g., $-0.015$ on Amazon), the copy-based AEGIS options provide no consistent benefit, and random/synthetic edges raise the Brier score on every benchmark.

\begin{table}[t]
\centering
\caption{Amazon (product--category), GAT, $q{=}0.01$, $\phi{=}100$}: AUC ($M\pm SD$) with paired $t$-tests vs. sparse baseline ($n=32$; ns = not significant; $^{*}$ $p<0.05$; $^{**}$
$p<0.01$; $^{***}$ $p<0.001$). A higher AUC is better.
\label{tab:amazon_auc}
\begin{tabular}{l r r r r r}
\toprule
Method & AUC $M\pm SD$ & $\Delta$AUC & $t(31)$ & $p$ & $d$ \\
\midrule
baseline      & 0.630 $\pm$ 0.162 & +0.000 & ---   & ---   & ---   \\
degree\_aware & 0.650 $\pm$ 0.204 & +0.020$^{\mathrm{ns}}$ & $-0.50$ & 0.619 & $-0.09$ \\
simple        & 0.637 $\pm$ 0.199 & +0.007$^{\mathrm{ns}}$ & $-0.17$ & 0.864 & $-0.03$ \\
semantic\_knn & 0.722 $\pm$ 0.197 & +0.091$^{*}$         & $-2.40$ & 0.023 & $-0.42$ \\
synthetic     & 0.732 $\pm$ 0.181 & +0.101$^{*}$         & $-2.48$ & 0.019 & $-0.44$ \\
random        & 0.626 $\pm$ 0.252 & $-0.004^{\mathrm{ns}}$ &  0.08  & 0.936 &  0.01  \\
original      & 0.928 $\pm$ 0.008 & +0.298$^{***}$        & $-10.42$& $<$0.001& $-1.84$ \\
\bottomrule
\end{tabular}
\end{table}

\begin{table}[t]
\centering
\caption{MovieLens (movie--genre), GAT, $q{=}0.01$, $\phi{=}100$}: AUC ($M\pm SD$) with paired $t$-tests vs. sparse baseline ($n=32$; ns = not significant; $^{*}$ $p<0.05$; $^{**}$
$p<0.01$; $^{***}$ $p<0.001$). A higher AUC is better.
\label{tab:movielens_auc}
\begin{tabular}{l r r r r r}
\toprule
Method & AUC $M\pm SD$ & $\Delta$AUC & $t(31)$ & $p$ & $d$ \\
\midrule
baseline      & 0.710 $\pm$ 0.061 & +0.000 & ---  & ---   & ---   \\
degree\_aware & 0.713 $\pm$ 0.064 & +0.003$^{\mathrm{ns}}$ & $-0.42$ & 0.681 & $-0.07$ \\
simple        & 0.717 $\pm$ 0.063 & +0.007$^{\mathrm{ns}}$ & $-1.33$ & 0.195 & $-0.23$ \\
semantic\_knn & 0.708 $\pm$ 0.064 & $-0.002^{\mathrm{ns}}$ &  0.35   & 0.732 &  0.06  \\
synthetic     & 0.679 $\pm$ 0.075 & $-0.031^{*}$         &  2.15   & 0.039 &  0.38  \\
random        & 0.652 $\pm$ 0.089 & $-0.059^{***}$       &  3.66   & 0.001 &  0.65  \\
original      & 0.811 $\pm$ 0.015 & +0.101$^{***}$        & $-9.95$ & $<$0.001& $-1.76$ \\
\bottomrule
\end{tabular}
\end{table}

\begin{table}[!htbp]
\centering
\caption{Amazon (product--category), GAT, $q{=}0.01$, $\phi{=}100$}: Brier score ($M\pm SD$) with paired $t$-tests vs. sparse baseline ($n=32$; ns = not significant; $^{*}$ $p<0.05$;
$^{**}$ $p<0.01$; $^{***}$ $p<0.001$). A lower Brier score is better.
\label{tab:amazon_brier}
\begin{tabular}{l r r r r r}
\toprule
Method & Brier score $M\pm SD$ & $\Delta$Brier score & $t(31)$ & $p$ & $d$ \\
\midrule
baseline      & 0.249 $\pm$ 0.048 & +0.000 & ---  & ---   & ---   \\
degree\_aware & 0.248 $\pm$ 0.054 & $-0.001^{\mathrm{ns}}$ &  0.29 & 0.772 & 0.05 \\
simple        & 0.248 $\pm$ 0.049 & $-0.001^{\mathrm{ns}}$ &  0.30 & 0.765 & 0.05 \\
semantic\_knn & 0.233 $\pm$ 0.044 & $-0.015^{*}$         &  2.19 & 0.036 & 0.39 \\
synthetic     & 0.244 $\pm$ 0.029 & $-0.005^{\mathrm{ns}}$ &  0.70 & 0.488 & 0.12 \\
random        & 0.259 $\pm$ 0.040 & +0.010$^{\mathrm{ns}}$ & $-0.92$& 0.367 & $-0.16$ \\
original      & 0.135 $\pm$ 0.020 & $-0.114^{***}$        & 14.03  & $<$0.001&  2.48 \\
\bottomrule
\end{tabular}
\end{table}

\begin{table}[!htbp]
\centering
\caption{MovieLens (movie--genre), GAT, $q{=}0.01$, $\phi{=}100$}: Brier score ($M\pm SD$) with paired $t$-tests vs. sparse baseline ($n=32$; ns = not significant; $^{*}$ $p<0.05$;
$^{**}$ $p<0.01$; $^{***}$ $p<0.001$). A lower Brier score is better.
\label{tab:movielens_brier}
\begin{tabular}{l r r r r r}
\toprule
Method & Brier score $M\pm SD$ & $\Delta$Brier score & $t(31)$ & $p$ & $d$ \\
\midrule
baseline      & 0.231 $\pm$ 0.016 & +0.000 & ---  & ---   & ---   \\
degree\_aware & 0.233 $\pm$ 0.014 & +0.001$^{\mathrm{ns}}$ & $-0.72$ & 0.474 & $-0.13$ \\
simple        & 0.231 $\pm$ 0.012 & $-0.000^{\mathrm{ns}}$ &  0.12   & 0.907 &  0.02  \\
semantic\_knn & 0.235 $\pm$ 0.014 & +0.004$^{\mathrm{ns}}$ & $-1.43$ & 0.162 & $-0.25$ \\
synthetic     & 0.245 $\pm$ 0.008 & +0.014$^{***}$        & $-4.82$ & $<$0.001& $-0.85$ \\
random        & 0.245 $\pm$ 0.009 & +0.013$^{***}$        & $-3.89$ & $<$0.001& $-0.69$ \\
original      & 0.218 $\pm$ 0.004 & $-0.013^{***}$        &  5.22   & $<$0.001&  0.92  \\
\bottomrule
\end{tabular}
\end{table}

\subsection{Domain Case Study: Game Design Pattern (GDP)}

% ==================== Paragraph 2: Domain Case Study ====================
% SOURCE:notebooks/gdp/degree_analysis_gdp-GAT-drop0.99-100x-20251126_091851_degree_stats.csv
%         notebooks/gdp/degree_analysis_gdp-GAT-drop0.99-100x-20251126_091851_distribution_fit.csv
% q=0.01 (99% dropout), phi=100x

\begin{figure}[H]
\centering
\includegraphics[width=1\linewidth]{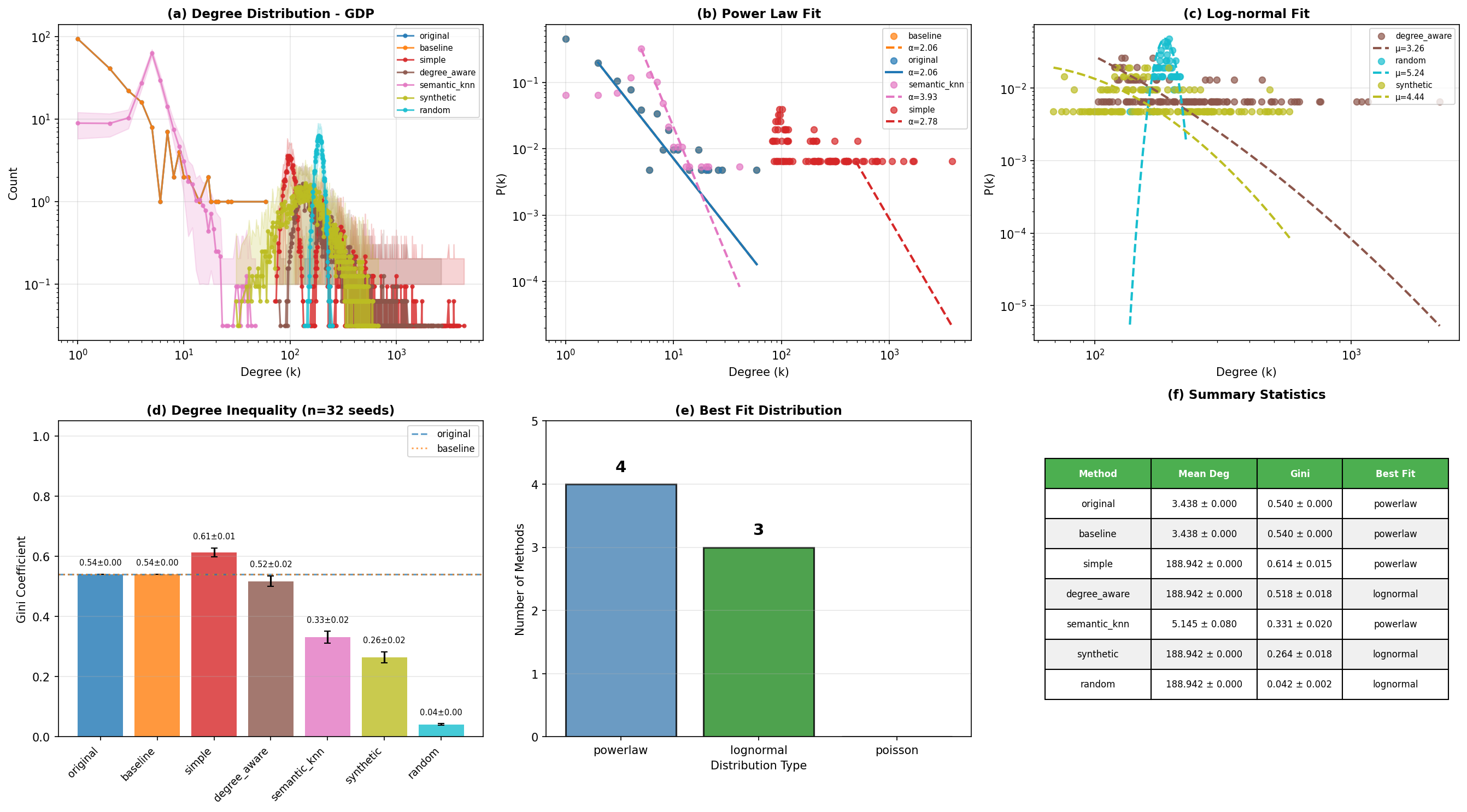} 
\caption{GDP (game--pattern), GAT, $q{=}0.01$, $\phi{=}100$: Comprehensive degree analysis ($M\pm\mathrm{SD}$, $n=32$ seeds). Panel (a) shows degree distributions on log-log scale with $\pm 1\sigma$ confidence bands; (b) Power Law fits with exponent $\alpha$ (lower $\alpha$ = heavier tail); (c) Log-normal fits with parameters $\mu$ and $\sigma$; (d) Gini coefficients quantifying degree inequality ($0{=}$perfect equality, $1{=}$maximum inequality); (e) best-fit distribution counts (lower KS statistic wins); (f) summary statistics table.}
\label{fig:gdp}
\end{figure}

For the domain-specific GDP dataset (Figure~\ref{fig:gdp}), structural diagnostics (again on graphs used for topology inspection rather than evaluation) reflect its expert-curated sparsity. The original graph is already uneven (Gini $0.54$). Copy-style augmentations preserve that pattern (Gini $0.978$ for both \texttt{simple} and \texttt{degree\_aware}) and even fit Poisson under extreme sparsity, suggesting they maintain the domain’s intentional connectivity. \texttt{Semantic\_knn} adds edges cautiously (mean degree $0.171$), benefiting from richer text, while \texttt{random} collapses inequality (Gini $0.378$) and \texttt{synthetic} lands in between (Gini $0.825$). These results support using structure-preserving methods on curated graphs and caution against topology-altering augmentations that distort expert signals.

Unlike Amazon and MovieLens, GDP's original graph exhibits higher baseline inequality
% GDP original: gini_M=0.5402±0.0
(Gini $= 0.540$) due to domain constraints where certain game design patterns (e.g., ``Core Loop'', ``Feedback'') are inherently more prevalent than specialized patterns. Under 99\% dropout, AEGIS-Simple maintains this domain-informed structure with extreme inequality
% GDP simple: gini_M=0.9778±0.0083
(Gini $= 0.978 \pm 0.008$), though interestingly, GDP's copy-style methods show Poisson as the best-fit distribution rather than Power Law
% GDP simple: best_fit=poisson
% GDP degree\_aware: best_fit=poisson
(panel e), suggesting that extreme sparsity disrupts typical scale-free
characteristics in smaller graphs. AEGIS-Degree shows nearly identical behavior % GDP degree\_aware: gini_M=0.9779±0.0082
(Gini $= 0.978 \pm 0.008$), indicating that inverse-degree bias provides limited benefit when the original topology already reflects expert knowledge rather than purely statistical bias. Random augmentation disrupts domain structure
% GDP random: gini_M=0.3782±0.0921, powerlaw_alpha_M=6.875±2.772, best_fit=powerlaw
(Gini $= 0.378 \pm 0.092$; shifts to Power Law with $\alpha = 6.87 \pm 2.77$), explaining its poor performance on this expert-curated dataset. Notably, semantic-KNN performs relatively better on GDP
% GDP semantic\_knn: mean_M=0.1707±0.0640
% Amazon semantic\_knn: mean_M=0.1063±0.0100
(mean degree $= 0.171 \pm 0.064$ vs Amazon's $0.106 \pm 0.010$), as richer textual features from game descriptions provide stronger semantic signals for edge prediction. Synthetic augmentation achieves intermediate inequality
% GDP synthetic: gini_M=0.8247±0.0542, powerlaw_alpha_M=5.412±2.066, best_fit=powerlaw
(Gini $= 0.825 \pm 0.054$) with Power Law fitting ($\alpha = 5.41 \pm 2.07$). This domain-specific finding validates AEGIS-Simple's design: for knowledge graphs where topology encodes expert curation rather than random connectivity, authentic resampling preserves meaningful structure even at extreme sparsity, whereas topology-altering methods (random, synthetic) sacrifice domain validity for artificial connectivity.

For AUC, Table~\ref{tab:gdp_auc} highlights that the strength of \texttt{semantic\_knn} reaches even higher (+0.014, $p=0.008$), whereas random and synthetic edges crater ranking quality. For authenticity-constrained augmentation on a richly described graph: \texttt{degree\_aware} significantly decreases AUC (-0.028, $t=5.29$, $p<0.001$). Regarding the Brier score, Table~\ref{tab:gdp_brier} confirms that GDP’s long-form textual descriptions let both AEGIS degrees and semantic completion improve calibration (Brier: \texttt{degree\_aware} $+0.036$, \texttt{simple} $-0.013$, \texttt{semantic\_knn} $-0.054$), while random/synthetic edges still degrade reliability.

\begin{table}[!htbp]
\centering
\caption{GDP (game--pattern), GAT, $q{=}0.01$, $\phi{=}100$}: AUC ($M\pm SD$) with paired $t$-tests vs. sparse baseline ($n=32$; ns = not significant; $^{*}$ $p<0.05$; $^{**}$
$p<0.01$; $^{***}$ $p<0.001$). A higher AUC is better.
\label{tab:gdp_auc}
\begin{tabular}{l r r r r r}
\toprule
Method & AUC $M\pm SD$ & $\Delta$AUC & $t(31)$ & $p$ & $d$ \\
\midrule
baseline      & 0.800 $\pm$ 0.022 & +0.000 & ---   & ---   & ---   \\
degree\_aware & 0.772 $\pm$ 0.026 & $-0.028^{***}$       &  5.29  & $<$0.001 & 0.93  \\
simple        & 0.793 $\pm$ 0.023 & $-0.007^{\mathrm{ns}}$ &  1.67  & 0.104  & 0.30  \\
semantic\_knn & 0.814 $\pm$ 0.017 & +0.014$^{**}$        & $-2.83$& 0.008  & $-0.50$ \\
synthetic     & 0.645 $\pm$ 0.061 & $-0.155^{***}$       & 13.42  & $<$0.001 & 2.37  \\
random        & 0.613 $\pm$ 0.076 & $-0.187^{***}$       & 13.14  & $<$0.001 & 2.32  \\
\bottomrule
\end{tabular}
\end{table}

\begin{table}[!htbp]
\centering
\caption{GDP (game--pattern), GAT, $q{=}0.01$, $\phi{=}100$}: Brier score ($M\pm SD$) with paired $t$-tests vs. sparse baseline ($n=32$; ns = not significant; $^{*}$ $p<0.05$; $^{**}$
$p<0.01$; $^{***}$ $p<0.001$). A lower Brier score is better.
\label{tab:gdp_brier}
\begin{tabular}{l r r r r r}
\toprule
Method & Brier score $M\pm SD$ & $\Delta$Brier score & $t(31)$ & $p$ & $d$ \\
\midrule
baseline      & 0.302 $\pm$ 0.040 & +0.000 & ---   & ---   & ---   \\
degree\_aware & 0.337 $\pm$ 0.079 & +0.036$^{*}$        & $-2.59$ & 0.015 & $-0.46$ \\
simple        & 0.289 $\pm$ 0.018 & $-0.013^{*}$        &  2.41   & 0.022 &  0.43  \\
semantic\_knn & 0.247 $\pm$ 0.017 & $-0.054^{***}$      &  7.06   & $<$0.001&  1.25  \\
synthetic     & 0.269 $\pm$ 0.013 & $-0.032^{***}$      &  4.76   & $<$0.001&  0.84  \\
random        & 0.266 $\pm$ 0.017 & $-0.036^{***}$      &  4.92   & $<$0.001&  0.87  \\
\bottomrule
\end{tabular}
\end{table}

Sensitivity analyses across other $q$/$\phi$ and augmentation methods (Appendix~\ref{appendix-sensitivity-analysis}) show the same pattern: copy-style AEGIS remains baseline-like, semantic\_knn is the only consistent AUC/Brier lift when text is informative, random/synthetic degrade.

\section{Discussion}
\paragraph{Authenticity beyond duplication.} Across Amazon and MovieLens,
the copy-style AEGIS variants (\texttt{simple}, \texttt{degree\_aware})
match rather than exceed the sparse baseline in either
AUC or Brier, whereas the semantic KNN augmenter (\texttt{semantic\_knn})
is the only method that reliably lifts performance (e.g., $+0.091$ AUC
and $-0.015$ Brier on Amazon) and at least prevents collapse on MovieLens.
On GDP, copy-style variants chiefly aid calibration,
while semantic completion adds the only AUC gains. This suggests that
``authentic'' augmentation hinges on injecting semantically plausible
endpoints rather than merely duplicating surviving edges; even so, the
assumption that higher feature similarity implies a greater likelihood of a
true link should be validated on a domain-by-domain basis. Authenticity can thus be viewed as a spectrum—from strict edge resampling to high-confidence semantic completion—where tighter constraints suit semantics-poor graphs, and softer, well-filtered semantic links benefit semantics-rich settings.

\paragraph{Text richness matters.} GDP’s game–pattern descriptions are
long and semantically rich, yielding the largest gains ($+0.014$ AUC and
$-0.054$ Brier for \texttt{semantic\_knn}), while \texttt{degree\_aware}
also improves calibration. Amazon’s product metadata is shorter yet
structured enough to benefit, whereas MovieLens’s brief genre/synopsis
features offer little semantic signal—suggesting that authenticity
constraints deliver the most when node descriptions carry meaningful
content.

\paragraph{Metric behavior.} Amazon’s synthetic augmenter shows that higher
AUC can coexist with degraded calibration, underscoring the need to pair
ROC analysis with Brier. On MovieLens, \texttt{semantic\_knn} preserves
both AUC and Brier relative to the sparse baseline, whereas random and
synthetic additions worsen both. GDP exhibits the strongest recovery:
only in this text-rich setting do copy-based AEGIS variants gain traction
for calibration (Brier drops without AUC gains), and the
semantic augmentation provides the largest joint improvements.

\paragraph{Limitations.}
The extreme sparsity stems from a single 0.99 bond-percolation pass,
simultaneously altering degree, density, and component structure. Results
depend on the chosen split and on severe imbalance; accuracy is threshold-
bound and therefore de-emphasized. Future work should examine alternative
sparsity regimes, adaptive drop ratios, and how textual richness governs
augmentation gains.

\section{Conclusion and Future Work}

We presented AEGIS, an authenticity-constrained edge augmentation framework
for sparse bipartite graphs, and evaluated it on Amazon, MovieLens, and
the GDP domain case study. Copy-based variants (uniform or inverse-degree
resampling) act as reliable sparsity baselines that avoid fabricating
new endpoints; their efficacy nevertheless hinges on how much semantic
information the domain provides (e.g., aiding calibration
on GDP). The semantic KNN augmentation is indispensable for recovering
performance on Amazon and MovieLens and delivers the largest gains on
GDP, where richer textual descriptions unlock sizable improvements in AUC
and Brier. Future work will explore density-preserving
augmentation, adaptive authenticity constraints that couple semantics with
edge resampling (including explicit semantic-threshold/filtration
schemes), cost-aware policies, and expansion to additional sparse domains.\paragraph{Reproduction Statement.}
The code repo is available at \url{https://anonymous.4open.science/r/AEGIS-6BA3/} \paragraph{Generative AI Statement.}
Large language models are used in writing this manuscript only to aid or polish writing. An example of the used prompts is ``Please polish these sentences in an academic way: [the actual contents]''

\subsubsection*{Acknowledgments}
The authors thank Rocio Mercado for introducing us to knowledge graph approaches in domain knowledge discovery. We also thank Zhuangyan Fang, Huaifeng Zhang, Zhiyong Wang, Nicolas Pietro Marie Audinet De Pieuchon, Kelsey Cotton, and Mengyu Huang for valuable discussions on data augmentation, graph theory, academic writing in the AI/Computer Science domain, and multimodal learning in graphs.

\subsubsection*{Funding Details}
This work was partially supported by the Wallenberg AI, Autonomous Systems and Software Program—Humanity and Society (WASP-HS), funded by the Marianne and Marcus Wallenberg Foundation and the Marcus and Amalia Wallenberg Foundation.

The computations and data handling were enabled by resources provided by the National Academic Infrastructure for Supercomputing in Sweden (NAISS), partially funded by the Swedish Research Council through grant agreement no. 2022-06725.

\bibliography{iclr2026_conference}
\bibliographystyle{iclr2026_conference}

\appendix
\section{Appendix: The Game Design Pattern Dataset}
\label{appendix-gdp-introduction}

The Game Design Patterns (GDP) dataset serves as a (semi-)ontology for the game design domain, offering formalized descriptions, properties, and constraints for each concept \citep{noy2001ontology}. Central to GDP is the notion of a ``pattern''—a recurring interaction that can be instantiated in diverse games, independent of genre or theme. For example, the ``Alignment'' pattern refers to ``the goal of forming a linear arrangement of game elements.'' \footnote{\url{http://virt10.itu.chalmers.se/index.php/Alignment}} This pattern is exemplified in games such as Tic-Tac-Toe, Candy Crush Saga, and Tetris.

Patterns in GDP not only describe game mechanics but also exhibit structural relationships with other patterns: they can enable (``instantiate''), modify, or potentially conflict with the deployment of other patterns. As such, GDP provides a shared vocabulary for game designers to communicate, analyze, and create games.

However, the identification and verification of game design patterns is a highly specialized and expert-driven process. Despite the vast number of games, only a limited number of patterns have been formally proposed, and even fewer game–pattern relationships have been verified by experts. This results in an inherently sparse bipartite graph, making GDP an ideal testbed for evaluating augmentation strategies in edge-sparse domains.

\section{Appendix: Implementation Specifics}
\label{appendix-implementation}

\subsection{Experimental Setup}

\paragraph{Hardware \& Software:} CUDA 11.8, NVIDIA A40 GPU, PyTorch 2.4.0+cu118, PyTorch Geometric 2.7.0, Torch Sparse 0.6.18+pt24cu118.

\begin{table}[htbp]
  \centering
  \caption{Hyperparameter configuration. TF-IDF dimensions are dataset-specific: Amazon (512), MovieLens (1024), GDP (128). Sensitivity analysis varies GNN type, drop ratio, and augmentation factor.}
  \label{tab:config}
  \small
  \begin{tabular}{@{}ll@{}}
    \toprule
    \textbf{Parameter} & \textbf{Value} \\
    \midrule

    \multicolumn{2}{@{}l}{\textit{Data Split \& Loading}} \\
    \quad Val / Test / Disjoint train ratio & 0.1 / 0.1 / 0.3154 \\
    \quad Negative sampling ratio           & 1.4875 \\
    \quad Add negative train samples           & True \\
    \quad Train / Val Neighbor sampling                 & [20, 10] / [20, 10]\\
    \quad Train / Val batch size            & 128 / 64 \\
    \quad Train / Val shuffle               & True / False \\

    \midrule
    \multicolumn{2}{@{}l}{\textit{Architecture \& Training}} \\
    \quad Layers / Hidden dim / Link predictor & 3 / 768 / Dot \\
    \quad LR / Max epochs / Eval freq           & $4.5\!\times\!10^{-4}$ / 779 / 15 \\
    \quad Early stopping (patience / $\delta$)  & 10 / $10^{-3}$ \\

    \midrule
    \multicolumn{2}{@{}l}{\textit{Sensitivity Analysis}} \\
    \quad GNN architecture                    & GAT, GraphSAGE, GCN (GraphConv) \\
    \quad Edge drop ratio                     & 0.99, 0.95 \\
    \quad Augmentation factor  &  1$\times$, 5$\times$ \\
    \midrule
    \multicolumn{2}{@{}l}{\textit{Semantic-KNN}} \\
    \quad $k_g$, $k_p$ / sim\_thresh / cap    & 1, 1 / 0.6 / 4 \\

    \bottomrule
  \end{tabular}
\end{table}

\section{Appendix: Sensitivity Analysis}
\label{appendix-sensitivity-analysis}

\subsection{Analysis Design}

\subsubsection{Retain Rates, Augmentation Factors and GNN Architectures}

The sensitivity analysis systematically evaluates edge augmentation methods across varying retain rates $q \in \{0.01, 0.05, 0.10\}$, augmentation factors $\phi \in \{1, 5, 100\}$, and GNN architectures (GAT, GraphSAGE, GCN). For the GCN baseline, we use \texttt{GraphConv}~\citep{morris2019weisfeiler}~\footnote{\url{https://github.com/pyg-team/pytorch_geometric}} rather than the original spectral GCN~\citep{kipf2016semi}, as the latter's symmetric normalization is incompatible with PyTorch Geometric's heterogeneous graph conversion (\texttt{to\_hetero()}). \texttt{GraphConv} performs mean aggregation and is functionally similar, as recommended by the PyG documentation for bipartite message passing\footnote{\url{https://pytorch-geometric.readthedocs.io/en/latest/notes/heterogeneous.html}}.

Given the limited time and resource, we prioritize experiments that balance coverage across architectural variations and sparsity conditions. Table~\ref{tab:experiment_coverage} summarizes the 36 completed experimental configurations across datasets.

\begin{table}[h]
 \centering
 \caption{Experimental configuration coverage across datasets, GNN architectures, retain rates ($q$), and augmentation factors ($\phi$). Checkmarks indicate completed experiments. Amazon and MovieLens use artificial sparsity via edge dropout; GDP is naturally sparse (no dropout,indicated by --). Factor $\phi=1$ (baseline-only) experiments exist for all configurations but are excluded from appendix tables per design.}
 \label{tab:experiment_coverage}
 \small
 \begin{tabular}{l c c c c c}
 \toprule
 Dataset & Architecture & $q$ & $\phi=100$
& $\phi=5$ & $\phi=2$ \\
 \midrule
 \multirow{9}{*}{Amazon}
 & GAT & 0.01 &
\checkmark\textsuperscript{†} & \checkmark &
\checkmark \\
 & GAT & 0.05 & -- & \checkmark &
\checkmark \\
 & GAT & 0.10 & -- & \checkmark &
\checkmark \\
 \cmidrule{2-6}
 & GraphSAGE & 0.01 & -- & \checkmark &
\checkmark \\
 & GraphSAGE & 0.05 & -- & \checkmark &
\checkmark \\
 & GraphSAGE & 0.10 & -- & \checkmark &
\checkmark \\
 \cmidrule{2-6}
 & GCN & 0.01 & -- & \checkmark &
\checkmark \\
 & GCN & 0.05 & -- & \checkmark &
\checkmark \\
 & GCN & 0.10 & -- & \checkmark &
\checkmark \\
 \midrule
 \multirow{5}{*}{MovieLens}
 & GAT & 0.01 &
\checkmark\textsuperscript{†} & \checkmark &
\checkmark \\
 & GAT & 0.05 & -- & \checkmark &
\checkmark \\
 & GAT & 0.10 & -- & \checkmark & -- \\
 & GraphSAGE & 0.01 & -- & \checkmark &
\checkmark \\
 & GCN & 0.01 & -- & \checkmark &
\checkmark \\
 \midrule
 \multirow{3}{*}{GDP}
 & GAT & -- & \checkmark\textsuperscript{†}
& \checkmark & \checkmark \\
 & GraphSAGE & -- & -- & \checkmark &
\checkmark \\
 & GCN & -- & -- & \checkmark & \checkmark
\\
 \bottomrule

\multicolumn{6}{l}{\textsuperscript{†}Legacy 100$\times$ experiments (Sept 2025) - no runtime statistics.}
 \\
 \end{tabular}
 \end{table}

\subsubsection{Degree Distribution Analysis: Metrics and Rationale}

Beyond the primary metrics (AUC-ROC and Brier score), we report degree distribution analysis and runtime profiling to characterize how augmentation transforms graph structure and computational cost.

Edge dropping and augmentation fundamentally alters the graph's degree distribution, potentially affecting both model performance and the structural assumptions underlying GNN message passing. We quantify these changes using three complementary approaches: (i) summary statistics describing central tendency and dispersion, (ii) the Gini coefficient capturing degree inequality, and (iii) distribution fitting to distinguish scale-free from random graph structure.

\paragraph{Summary Statistics.} For each original, sparsified (in the case of benchmark datasets) and augmented graph, we compute the mean degree $\bar{d} = \frac{1}{|V|}\sum_{v \in V} d_v$, standard deviation $\sigma_d$, median, minimum, and maximum degree, as well as the number of isolated nodes ($d_v = 0$). These statistics characterize how augmentation affects both the average connectivity and the presence of extreme values.

\paragraph{Gini Coefficient.} The Gini coefficient~\citep{ceriani2012origins}, originally developed for income inequality measurement, has been adapted to quantify heterogeneity in network degree distributions~\citep{hu2005gini,badham2013commentary}. For a sorted degree sequence $(d_1, d_2, \ldots, d_n)$, the Gini coefficient is computed as:

\begin{equation}
G = \frac{2 \sum_{i=1}^{n} i \cdot d_i - (n+1) \sum_{i=1}^{n}
d_i}{n \sum_{i=1}^{n} d_i}
\end{equation}
  
where $G = 0$ indicates perfect equality (all nodes have identical degree) and $G = 1$ indicates maximal inequality (one node holds all edges). In the context of bipartite link prediction, a very high Gini coefficient ($G > 0.9$) indicates that most nodes have near-zero degree after percolation, concentrating supervision on a few hub nodes. AEGIS-Degree's inverse-degree bias explicitly targets this scenario by preferentially augmenting low-degree nodes, which should manifest as a reduction in $G$ relative to the baseline. Conversely, if $G$ becomes too low, over-smoothing may reduce discriminative structure.

\paragraph{Distribution Fitting.} Many real-world networks exhibit heavy-tailed degree distributions, often claimed to follow power laws~\citep{barabasi1999emergence}. Following the rigorous statistical framework of \citet{clauset2009power}, we fit three candidate distributions to the empirical degree sequence:

\begin{itemize}
  \item \textbf{Power Law} (scale-free model): $P(d) \propto d^{-\alpha}$ for $d \ge d_{\min}$. The exponent $\alpha$ is estimated via maximum likelihood with automatic $d_{\min}$ selection. Scale-free networks exhibit hub nodes and preferential attachment dynamics.

  \item \textbf{Log-normal}: $P(d) \propto \frac{1}{d}\exp\!\left(-\frac{(\ln d - \mu)^2}{2\sigma^2}\right)$. Log-normal distributions arise from multiplicative growth processes and can fit many ``scale-free'' networks equally well or better~\citep{broido2019scale,stumpf2012critical}.

  \item \textbf{Poisson} (Erdős–Rényi model): $P(d) = \frac{\lambda^d e^{-\lambda}}{d!}$. The Poisson distribution characterizes classical random graphs $G(n,p)$ where $\lambda = np$~\citep{erdds1959random}. A good Poisson fit indicates augmentation produces ER-like random connectivity.
\end{itemize}

We compare fits using the Kolmogorov–Smirnov (KS) statistic, where lower values indicate better fit.

\paragraph{Aggregation Protocol.}
Since each of the 32 seeds produces a different sparse graph (due to distinct bond percolation realizations), we compute degree statistics independently for each seed and report aggregated results as $M \pm \mathrm{SD}$, consistent with our treatment of AUC-ROC and Brier scores. Distribution parameters ($\alpha$, $\mu$, $\sigma$, KS statistics) are likewise fitted per-seed and aggregated.

\subsection{Overarching Findings}

Across datasets and augmentation methods, augmentation helps selectively and the best method depends on the metric: \texttt{semantic\_knn} and \texttt{synthetic} drive AUC gains under extreme sparsity for Amazon (GAT/GCN), but \texttt{semantic\_knn} is often neutral or negative for MovieLens and mixed for GDP; \texttt{simple}/\texttt{degree\_aware} offer small, stable lifts in several settings (MovieLens GCN at $q{=}0.01$, Amazon GAT/GCN at higher $q$), while \texttt{random} rarely wins on AUC. Brier leaders differ: \texttt{random}/\texttt{synthetic} improve calibration for Amazon GraphSAGE and MovieLens GCN, \texttt{semantic\_knn} leads GDP GAT at high $\phi$, and \texttt{degree\_aware}/\texttt{simple} give modest Brier gains in multiple cases—highlighting a trade-off between ranking and calibration. Degree distributions show that most augmentations densify graphs and reduce isolation, with \texttt{random}/\texttt{synthetic} most effective at lowering Gini, whereas \texttt{semantic\_knn} keeps graphs sparsest. Runtime overhead is negligible for augmentation (sub-0.16s) and training deltas stay small relative to dense-orig baselines, so performance differences are governed by graph quality rather than compute cost; practitioners should pick methods by desired metric (AUC vs.\ calibration) and acceptable shifts in degree structure.

\subsection{Benchmark Dataset - Amazon}

\subsubsection{Summary}

Across Amazon, graph augmentation helps when coupled with the right encoder: for GAT, \texttt{semantic\_knn} and \texttt{synthetic} yield the largest AUC gains at extreme sparsity ($q{=}0.01$, $\phi{=}100$), and \texttt{simple}/\texttt{degree\_aware} offer modest, more stable lifts at $q{=}0.10$, $\phi{=}2$. For GraphSAGE, AUC gains center on \texttt{synthetic} at $q{=}0.01$, while Brier is dominated by \texttt{random} (and \texttt{synthetic} at high dropout), revealing a trade-off between ranking quality and calibration. For GCN, \texttt{semantic\_knn} is the clear AUC winner at $q{=}0.05$, whereas \texttt{degree\_aware} and \texttt{simple}/\texttt{random}/\texttt{synthetic} drive the best Brier at higher $q$. Degree distributions remain sparse for most methods, with \texttt{random}/\texttt{synthetic} most effective at reducing inequality and isolated nodes, and \texttt{semantic\_knn} keeping graphs sparsest. Runtime overheads are negligible across all augmentation methods (sub-0.07s augmentation; training within a few seconds of baseline), so observed gains or losses are driven by graph quality rather than compute cost.

\subsubsection{GAT}

\paragraph{Summary Analysis}

Across Amazon GAT, AUC gains concentrate at high sparsity: \texttt{synthetic} and \texttt{semantic\_knn} lead at $q{=}0.01$, $\phi{=}100$ (+0.101 and +0.091), while \texttt{simple}/\texttt{degree\_aware} give smaller but significant lifts at $q{=}0.10$, $\phi{=}2$ (+0.017/+0.004); \texttt{random} is consistently weakest. Brier improvements are largest for \texttt{semantic\_knn} at $q{=}0.01$, $\phi{=}100$ ($-0.015$) and for \texttt{simple} at $q{=}0.10$, $\phi{=}2$ ($-0.011$), with mild gains for \texttt{degree\_aware}/\texttt{simple} at $q{=}0.10$, $\phi{=}5$ ($-0.009/-0.007$); \texttt{synthetic}/\texttt{random} often worsen Brier. Degree-wise, augmentations raise mean degree from the ultra-sparse baseline (0.033 at $q{=}0.01$, $\phi{=}100$) to $\approx2.4$ at $\phi{=}100$ (Gini $0.827$--$0.979$) or to $\approx0.6$/$0.48$ at $\phi{=}5/2$ (Gini remains high), while \texttt{semantic\_knn} keeps graphs sparsest (mean degree 0.106 at $\phi{=}100$) but slightly lowers inequality. Runtime overheads are negligible: augmentation $<0.064$s and training within a few seconds of baseline (versus $\sim180$s for the dense original), so accuracy differences stem from graph quality rather than cost.

% Data source: Table~\ref{tab:amazon_gat_auc} and corresponding Brier table  \textbf{AUC-ROC Performance:} \texttt{semantic_knn} achieves the highest average AUC improvement (mean $\Delta$AUC=+0.002): at $q=0.01$, $\phi=5x$, AUC=0.726$\pm$0.185 ($\Delta$AUC=+0.078) vs baseline (0.648$\pm$0.188); at $q=0.10$, $\phi=5x$, AUC=0.732$\pm$0.043 ($\Delta$AUC=+0.002). Conversely, \texttt{random} shows the weakest performance (mean $\Delta$AUC=-0.071, range: -0.144 to -0.037).  \textbf{Brier Score Performance:} \texttt{simple} achieves the best Brier score (mean $\Delta$Brier=-0.004). At $q=0.10$, $\phi=2x$, it reaches 0.224$\pm$0.027 ($\Delta$Brier=-0.011) vs baseline (0.235$\pm$0.024).  % SOURCE FILES: % - code-repo/notebooks/amazon/amazon-GAT-drop0.9-2x-20251125_013548_summary.csv % - code-repo/notebooks/amazon/amazon-GAT-drop0.9-5x-20251125_081311_summary.csv % - code-repo/notebooks/amazon/amazon-GAT-drop0.95-2x-20251125_012114_summary.csv % - code-repo/notebooks/amazon/amazon-GAT-drop0.95-5x-20251124_082457_summary.csv % - code-repo/notebooks/amazon/amazon-GAT-drop0.99-2x-20251125_030835_summary.csv % - code-repo/notebooks/amazon/amazon-GAT-drop0.99-5x-20251123_230121_summary.csv

\paragraph{AUC and Brier Score}

\begin{longtable}{c c l r r r r r}
\caption{Amazon (product--category) GAT: AUC-ROC ($M\pm SD$) with paired $t$-tests vs.\ sparse baseline ($n=32$ seeds). A higher AUC is better.}
\label{tab:amazon_gat_auc}\\
\toprule
$q$ & $\phi$ & Method & AUC $M\pm SD$ & $\Delta$AUC & $t(31)$ & $p$ & $d$ \\
\midrule
\endfirsthead
\toprule
$q$ & $\phi$ & Method & AUC $M\pm SD$ & $\Delta$AUC & $t(31)$ & $p$ & $d$ \\
\midrule
\endhead
\bottomrule
\endfoot

% **CHECKED**
% SOURCE: notebooks/amazon/amazon-20250919_140827_summary.csv
%         notebooks/amazon/paired-ttest-20250919_140828.csv
% q=0.01, phi=100x
0.01 & 100$\times$    & baseline       & 0.630 $\pm$ 0.162 & +0.000 & ---   & ---   & ---   \\
0.01 & 100$\times$    & degree\_aware  & 0.650 $\pm$ 0.204 & +0.020$^{\mathrm{ns}}$ & $-0.50$ & 0.619    & $-0.09$ \\
0.01 & 100$\times$    & simple         & 0.637 $\pm$ 0.199 & +0.007$^{\mathrm{ns}}$ & $-0.17$ & 0.864    & $-0.03$ \\
0.01 & 100$\times$    & semantic\_knn  & 0.722 $\pm$ 0.197 & +0.091$^{*}$       & $-2.40$ & 0.023    & $-0.42$ \\
0.01 & 100$\times$    & \textbf{synthetic} & 0.732 $\pm$ 0.181 & +0.101$^{*}$       & $-2.48$ & 0.019    & $-0.44$ \\
0.01 & 100$\times$    & random         & 0.626 $\pm$ 0.252 & -0.004$^{\mathrm{ns}}$ & $0.08$  & 0.936    & $+0.01$ \\
0.01 & 100$\times$    & original       & 0.928 $\pm$ 0.008 & +0.298$^{***}$     & $-10.42$ & $<$0.001 & $-1.84$ \\
\midrule
% **CHECKED**
% SOURCE: notebooks/amazon/amazon-GAT-drop0.99-5x-20251123_230121_summary.csv
%         notebooks/amazon/amazon-GAT-drop0.99-5x-paired-ttest-20251123_230123.csv
% q=0.01, phi=5x
0.01 & 5$\times$      & baseline       & 0.648 $\pm$ 0.188 & +0.000 & ---   & ---   & ---   \\
0.01 & 5$\times$      & degree\_aware  & 0.635 $\pm$ 0.226 & -0.013$^{\mathrm{ns}}$ & $0.44$  & 0.665    & $+0.08$ \\
0.01 & 5$\times$      & simple         & 0.625 $\pm$ 0.221 & -0.023$^{\mathrm{ns}}$ & $0.79$  & 0.435    & $+0.14$ \\
0.01 & 5$\times$      & \textbf{semantic\_knn} & 0.726 $\pm$ 0.185 & +0.078$^{\mathrm{ns}}$ & $-1.80$ & 0.081    & $-0.32$ \\
0.01 & 5$\times$      & synthetic      & 0.613 $\pm$ 0.248 & -0.035$^{\mathrm{ns}}$ & $0.83$  & 0.410    & $+0.15$ \\
0.01 & 5$\times$      & random         & 0.611 $\pm$ 0.199 & -0.037$^{\mathrm{ns}}$ & $0.91$  & 0.368    & $+0.16$ \\
0.01 & 5$\times$      & original       & 0.929 $\pm$ 0.009 & +0.281$^{***}$     & $-8.40$ & $<$0.001 & $-1.49$ \\
\midrule
% **CHECKED**
% SOURCE: notebooks/amazon/amazon-GAT-drop0.99-2x-20251125_030835_summary.csv
%         notebooks/amazon/amazon-GAT-drop0.99-2x-paired-ttest-20251125_030836.csv
% q=0.01, phi=2x
0.01 & 2$\times$      & \textbf{baseline} & 0.648 $\pm$ 0.188 & +0.000 & ---   & ---   & ---   \\
0.01 & 2$\times$      & degree\_aware  & 0.628 $\pm$ 0.237 & -0.019$^{\mathrm{ns}}$ & $0.70$  & 0.488    & $+0.12$ \\
0.01 & 2$\times$      & simple         & 0.611 $\pm$ 0.231 & -0.036$^{\mathrm{ns}}$ & $1.39$  & 0.174    & $+0.25$ \\
0.01 & 2$\times$      & semantic\_knn  & 0.617 $\pm$ 0.237 & -0.031$^{\mathrm{ns}}$ & $0.95$  & 0.347    & $+0.17$ \\
0.01 & 2$\times$      & synthetic      & 0.585 $\pm$ 0.232 & -0.063$^{\mathrm{ns}}$ & $1.53$  & 0.136    & $+0.27$ \\
0.01 & 2$\times$      & random         & 0.565 $\pm$ 0.232 & -0.083$^{*}$       & $2.38$  & 0.023    & $+0.42$ \\
0.01 & 2$\times$      & original       & 0.929 $\pm$ 0.009 & +0.281$^{***}$     & $-8.40$ & $<$0.001 & $-1.49$ \\
\midrule

% **CHECKED**
% WARNING: No summary file found for GAT drop0.95 100x
% SOURCE: notebooks/amazon/amazon-GAT-drop0.95-5x-20251124_082457_summary.csv
%         notebooks/amazon/amazon-GAT-drop0.95-5x-paired-ttest-20251124_082458.csv
% q=0.05, phi=5x
0.05 & 5$\times$      & baseline       & 0.752 $\pm$ 0.064 & +0.000 & ---   & ---   & ---   \\
0.05 & 5$\times$      & degree\_aware  & 0.755 $\pm$ 0.063 & +0.003$^{\mathrm{ns}}$ & $-0.44$ & 0.665    & $-0.08$ \\
0.05 & 5$\times$      & \textbf{simple} & 0.757 $\pm$ 0.064 & +0.005$^{\mathrm{ns}}$ & $-0.79$ & 0.437    & $-0.14$ \\
0.05 & 5$\times$      & semantic\_knn  & 0.730 $\pm$ 0.071 & -0.022$^{**}$      & $3.11$  & 0.004    & $+0.55$ \\
0.05 & 5$\times$      & synthetic      & 0.646 $\pm$ 0.074 & -0.106$^{***}$     & $9.10$  & $<$0.001 & $+1.61$ \\
0.05 & 5$\times$      & random         & 0.608 $\pm$ 0.086 & -0.144$^{***}$     & $9.68$  & $<$0.001 & $+1.71$ \\
0.05 & 5$\times$      & original       & 0.929 $\pm$ 0.009 & +0.177$^{***}$     & $-14.93$ & $<$0.001 & $-2.64$ \\
\midrule
% CHECKED
% SOURCE: notebooks/amazon/amazon-GAT-drop0.95-2x-20251125_012114_summary.csv
%         notebooks/amazon/amazon-GAT-drop0.95-2x-paired-ttest-20251125_012114.csv
% q=0.05, phi=2x
0.05 & 2$\times$      & \textbf{baseline} & 0.752 $\pm$ 0.064 & +0.000 & ---   & ---   & ---   \\
0.05 & 2$\times$      & degree\_aware  & 0.752 $\pm$ 0.067 & -0.000$^{\mathrm{ns}}$ & $0.06$  & 0.953    & $+0.01$ \\
0.05 & 2$\times$      & simple         & 0.749 $\pm$ 0.070 & -0.003$^{\mathrm{ns}}$ & $0.57$  & 0.573    & $+0.10$ \\
0.05 & 2$\times$      & semantic\_knn  & 0.735 $\pm$ 0.070 & -0.017$^{*}$       & $2.61$  & 0.014    & $+0.46$ \\
0.05 & 2$\times$      & synthetic      & 0.712 $\pm$ 0.067 & -0.040$^{***}$     & $4.91$  & $<$0.001 & $+0.87$ \\
0.05 & 2$\times$      & random         & 0.695 $\pm$ 0.085 & -0.057$^{***}$     & $4.78$  & $<$0.001 & $+0.84$ \\
0.05 & 2$\times$      & original       & 0.929 $\pm$ 0.009 & +0.177$^{***}$     & $-14.93$ & $<$0.001 & $-2.64$ \\
\midrule

% CHECKED
% WARNING: No summary file found for GAT drop0.9 100x
% SOURCE: notebooks/amazon/amazon-GAT-drop0.9-5x-20251125_081311_summary.csv
%         notebooks/amazon/amazon-GAT-drop0.9-5x-paired-ttest-20251125_081311.csv
% q=0.10, phi=5x
0.10 & 5$\times$      & baseline       & 0.730 $\pm$ 0.047 & +0.000 & ---   & ---   & ---   \\
0.10 & 5$\times$      & \textbf{degree\_aware} & 0.745 $\pm$ 0.051 & +0.015$^{**}$      & $-3.49$ & 0.001    & $-0.62$ \\
0.10 & 5$\times$      & simple         & 0.745 $\pm$ 0.036 & +0.014$^{**}$      & $-2.85$ & 0.008    & $-0.50$ \\
0.10 & 5$\times$      & semantic\_knn  & 0.732 $\pm$ 0.043 & +0.002$^{\mathrm{ns}}$ & $-0.21$ & 0.837    & $-0.04$ \\
0.10 & 5$\times$      & synthetic      & 0.673 $\pm$ 0.058 & -0.057$^{***}$     & $4.84$  & $<$0.001 & $+0.85$ \\
0.10 & 5$\times$      & random         & 0.661 $\pm$ 0.057 & -0.070$^{***}$     & $6.42$  & $<$0.001 & $+1.13$ \\
0.10 & 5$\times$      & original       & 0.929 $\pm$ 0.009 & +0.199$^{***}$     & $-22.72$ & $<$0.001 & $-4.02$ \\
\midrule

% CHECKED
% SOURCE: notebooks/amazon/amazon-GAT-drop0.9-2x-20251125_013548_summary.csv
%         notebooks/amazon/amazon-GAT-drop0.9-2x-paired-ttest-20251125_013549.csv
% q=0.10, phi=2x
0.10 & 2$\times$      & baseline       & 0.730 $\pm$ 0.047 & +0.000 & ---   & ---   & ---   \\
0.10 & 2$\times$      & degree\_aware  & 0.734 $\pm$ 0.040 & +0.004$^{\mathrm{ns}}$ & $-0.74$ & 0.463    & $-0.13$ \\
0.10 & 2$\times$      & \textbf{simple} & 0.747 $\pm$ 0.047 & +0.017$^{**}$      & $-3.15$ & 0.004    & $-0.56$ \\
0.10 & 2$\times$      & semantic\_knn  & 0.732 $\pm$ 0.043 & +0.002$^{\mathrm{ns}}$ & $-0.21$ & 0.837    & $-0.04$ \\
0.10 & 2$\times$      & synthetic      & 0.708 $\pm$ 0.048 & -0.022$^{**}$      & $3.19$  & 0.003    & $+0.56$ \\
0.10 & 2$\times$      & random         & 0.692 $\pm$ 0.057 & -0.038$^{***}$     & $5.69$  & $<$0.001 & $+1.01$ \\
0.10 & 2$\times$      & original       & 0.929 $\pm$ 0.009 & +0.199$^{***}$     & $-22.72$ & $<$0.001 & $-4.02$ \\
\midrule

\end{longtable}

% ========== BRIER TABLE ==========
\begin{longtable}{c c l r r r r r}
\caption{Amazon (product--category) GAT: Brier Score ($M\pm SD$) with paired $t$-tests vs.\ sparse baseline ($n=32$ seeds, lower is better).}
\label{tab:amazon_gat_brier}\\
\toprule
$q$ & $\phi$ & Method & Brier $M\pm SD$ & $\Delta$Brier & $t(31)$ & $p$ & $d$ \\
\midrule
\endfirsthead
\toprule
$q$ & $\phi$ & Method & Brier $M\pm SD$ & $\Delta$Brier & $t(31)$ & $p$ & $d$ \\
\midrule
\endhead
\bottomrule
\endfoot

% CHECKED
% SOURCE: notebooks/amazon/amazon-20250919_140827_summary.csv
%         notebooks/amazon/paired-ttest-20250919_140828.csv
% q=0.01, phi=100x
0.01 & 100$\times$    & baseline       & 0.249 $\pm$ 0.048 & +0.000 & ---  & ---   & ---   \\
0.01 & 100$\times$    & degree\_aware  & 0.248 $\pm$ 0.054 & -0.001$^{\mathrm{ns}}$ & $0.29$ & 0.772    & $+0.05$  \\
0.01 & 100$\times$    & simple         & 0.248 $\pm$ 0.049 & -0.001$^{\mathrm{ns}}$ & $0.30$ & 0.765    & $+0.05$  \\
0.01 & 100$\times$    & \textbf{semantic\_knn} & 0.233 $\pm$ 0.044 & -0.015$^{*}$        & $2.19$ & 0.036    & $+0.39$  \\
0.01 & 100$\times$    & synthetic      & 0.244 $\pm$ 0.029 & -0.005$^{\mathrm{ns}}$ & $0.70$ & 0.488    & $+0.13$  \\
0.01 & 100$\times$    & random         & 0.259 $\pm$ 0.040 & +0.010$^{\mathrm{ns}}$ & $-0.92$ & 0.367    & $-0.16$  \\
0.01 & 100$\times$    & original       & 0.135 $\pm$ 0.020 & -0.114$^{***}$      & $14.03$ & $<$0.001 & $+2.48$  \\
\midrule
% CHECKED
% SOURCE: notebooks/amazon/amazon-GAT-drop0.99-5x-20251123_230121_summary.csv
%         notebooks/amazon/amazon-GAT-drop0.99-5x-paired-ttest-20251123_230123.csv
% q=0.01, phi=5x
0.01 & 5$\times$      & baseline       & 0.251 $\pm$ 0.062 & +0.000 & ---  & ---   & ---   \\
0.01 & 5$\times$      & degree\_aware  & 0.245 $\pm$ 0.053 & -0.006$^{\mathrm{ns}}$ & $1.85$ & 0.074    & $+0.33$  \\
0.01 & 5$\times$      & simple         & 0.243 $\pm$ 0.058 & -0.008$^{\mathrm{ns}}$ & $1.55$ & 0.132    & $+0.28$  \\
0.01 & 5$\times$      & \textbf{semantic\_knn} & 0.236 $\pm$ 0.048 & -0.015$^{\mathrm{ns}}$ & $1.61$ & 0.117    & $+0.29$  \\
0.01 & 5$\times$      & synthetic      & 0.256 $\pm$ 0.041 & +0.005$^{\mathrm{ns}}$ & $-0.28$ & 0.778    & $-0.05$  \\
0.01 & 5$\times$      & random         & 0.268 $\pm$ 0.063 & +0.017$^{\mathrm{ns}}$ & $-1.87$ & 0.071    & $-0.34$  \\
0.01 & 5$\times$      & original       & 0.141 $\pm$ 0.016 & -0.110$^{***}$      & $10.52$ & $<$0.001 & $+1.86$  \\
\midrule

% CHECKED
% SOURCE: notebooks/amazon/amazon-GAT-drop0.99-2x-20251125_030835_summary.csv
%         notebooks/amazon/amazon-GAT-drop0.99-2x-paired-ttest-20251125_030836.csv
% q=0.01, phi=2x
0.01 & 2$\times$      & baseline       & 0.251 $\pm$ 0.062 & +0.000 & ---  & ---   & ---   \\
0.01 & 2$\times$      & degree\_aware  & 0.254 $\pm$ 0.066 & +0.003$^{\mathrm{ns}}$ & $-0.35$ & 0.730    & $-0.06$  \\
0.01 & 2$\times$      & simple         & 0.257 $\pm$ 0.064 & +0.006$^{\mathrm{ns}}$ & $-1.35$ & 0.188    & $-0.24$  \\
0.01 & 2$\times$      & \textbf{semantic\_knn} & 0.247 $\pm$ 0.047 & -0.004$^{\mathrm{ns}}$ & $0.94$ & 0.353    & $+0.17$  \\
0.01 & 2$\times$      & synthetic      & 0.256 $\pm$ 0.055 & +0.005$^{\mathrm{ns}}$ & $-0.26$ & 0.798    & $-0.05$  \\
0.01 & 2$\times$      & random         & 0.280 $\pm$ 0.079 & +0.029$^{*}$        & $-2.38$ & 0.024    & $-0.43$  \\
0.01 & 2$\times$      & original       & 0.141 $\pm$ 0.016 & -0.110$^{***}$      & $10.52$ & $<$0.001 & $+1.86$  \\
\midrule

% CHECKED
% WARNING: No summary file found for GAT drop0.95 100x
% SOURCE: notebooks/amazon/amazon-GAT-drop0.95-5x-20251124_082457_summary.csv
%         notebooks/amazon/amazon-GAT-drop0.95-5x-paired-ttest-20251124_082458.csv
% q=0.05, phi=5x
0.05 & 5$\times$      & baseline       & 0.238 $\pm$ 0.022 & +0.000 & ---  & ---   & ---   \\
0.05 & 5$\times$      & degree\_aware  & 0.235 $\pm$ 0.021 & -0.003$^{\mathrm{ns}}$ & $0.71$ & 0.481    & $+0.13$  \\
0.05 & 5$\times$      & \textbf{simple} & 0.234 $\pm$ 0.027 & -0.004$^{\mathrm{ns}}$ & $1.09$ & 0.285    & $+0.19$  \\
0.05 & 5$\times$      & semantic\_knn  & 0.243 $\pm$ 0.029 & +0.006$^{\mathrm{ns}}$ & $-1.52$ & 0.139    & $-0.27$  \\
0.05 & 5$\times$      & synthetic      & 0.256 $\pm$ 0.024 & +0.018$^{***}$      & $-4.52$ & $<$0.001 & $-0.80$  \\
0.05 & 5$\times$      & random         & 0.272 $\pm$ 0.028 & +0.034$^{***}$      & $-8.69$ & $<$0.001 & $-1.54$  \\
0.05 & 5$\times$      & original       & 0.141 $\pm$ 0.016 & -0.097$^{***}$      & $20.57$ & $<$0.001 & $+3.64$  \\
\midrule
%CHECKED
% SOURCE: notebooks/amazon/amazon-GAT-drop0.95-2x-20251125_012114_summary.csv
%         notebooks/amazon/amazon-GAT-drop0.95-2x-paired-ttest-20251125_012114.csv
% q=0.05, phi=2x
0.05 & 2$\times$      & \textbf{baseline} & 0.238 $\pm$ 0.022 & +0.000 & ---  & ---   & ---   \\
0.05 & 2$\times$      & degree\_aware  & 0.238 $\pm$ 0.028 & -0.000$^{\mathrm{ns}}$ & $0.08$ & 0.939    & $+0.01$  \\
0.05 & 2$\times$      & simple         & 0.238 $\pm$ 0.031 & -0.000$^{\mathrm{ns}}$ & $0.02$ & 0.984    & $+0.00$  \\
0.05 & 2$\times$      & semantic\_knn  & 0.244 $\pm$ 0.028 & +0.006$^{\mathrm{ns}}$ & $-1.90$ & 0.067    & $-0.34$  \\
0.05 & 2$\times$      & synthetic      & 0.242 $\pm$ 0.028 & +0.004$^{\mathrm{ns}}$ & $-1.02$ & 0.317    & $-0.18$  \\
0.05 & 2$\times$      & random         & 0.245 $\pm$ 0.031 & +0.007$^{\mathrm{ns}}$ & $-1.50$ & 0.145    & $-0.26$  \\
0.05 & 2$\times$      & original       & 0.141 $\pm$ 0.016 & -0.097$^{***}$      & $20.57$ & $<$0.001 & $+3.64$  \\
\midrule

% CHECKED
% WARNING: No summary file found for GAT drop0.9 100x
% SOURCE: notebooks/amazon/amazon-GAT-drop0.9-5x-20251125_081311_summary.csv
%         notebooks/amazon/amazon-GAT-drop0.9-5x-paired-ttest-20251125_081311.csv
% q=0.10, phi=5x
0.10 & 5$\times$      & baseline       & 0.235 $\pm$ 0.024 & +0.000 & ---  & ---   & ---   \\
0.10 & 5$\times$      & \textbf{degree\_aware} & 0.226 $\pm$ 0.023 & -0.009$^{*}$        & $2.29$ & 0.029    & $+0.41$  \\
0.10 & 5$\times$      & simple         & 0.228 $\pm$ 0.023 & -0.007$^{*}$        & $2.25$ & 0.032    & $+0.40$  \\
0.10 & 5$\times$      & semantic\_knn  & 0.230 $\pm$ 0.022 & -0.005$^{\mathrm{ns}}$ & $1.16$ & 0.255    & $+0.20$  \\
0.10 & 5$\times$      & synthetic      & 0.253 $\pm$ 0.024 & +0.018$^{**}$       & $-3.56$ & 0.001    & $-0.63$  \\
0.10 & 5$\times$      & random         & 0.260 $\pm$ 0.022 & +0.025$^{***}$      & $-4.96$ & $<$0.001 & $-0.88$  \\
0.10 & 5$\times$      & original       & 0.141 $\pm$ 0.016 & -0.094$^{***}$      & $17.66$ & $<$0.001 & $+3.12$  \\
\midrule
% CHECKED
% SOURCE: notebooks/amazon/amazon-GAT-drop0.9-2x-20251125_013548_summary.csv
%         notebooks/amazon/amazon-GAT-drop0.9-2x-paired-ttest-20251125_013549.csv
% q=0.10, phi=2x
0.10 & 2$\times$      & baseline       & 0.235 $\pm$ 0.024 & +0.000 & ---  & ---   & ---   \\
0.10 & 2$\times$      & degree\_aware  & 0.232 $\pm$ 0.024 & -0.003$^{\mathrm{ns}}$ & $0.83$ & 0.411    & $+0.15$  \\
0.10 & 2$\times$      & \textbf{simple} & 0.224 $\pm$ 0.027 & -0.011$^{***}$      & $3.72$ & $<$0.001 & $+0.66$  \\
0.10 & 2$\times$      & semantic\_knn  & 0.230 $\pm$ 0.022 & -0.005$^{\mathrm{ns}}$ & $1.16$ & 0.255    & $+0.20$  \\
0.10 & 2$\times$      & synthetic      & 0.240 $\pm$ 0.025 & +0.005$^{\mathrm{ns}}$ & $-1.33$ & 0.193    & $-0.24$  \\
0.10 & 2$\times$      & random         & 0.243 $\pm$ 0.021 & +0.008$^{*}$        & $-2.10$ & 0.044    & $-0.37$  \\
0.10 & 2$\times$      & original       & 0.141 $\pm$ 0.016 & -0.094$^{***}$      & $17.66$ & $<$0.001 & $+3.12$  \\
\midrule

\end{longtable}

\paragraph{Degree Distribution Analysis}

% ========== DEGREE DISTRIBUTION STATISTICS ==========
\begin{longtable}{c c l r r r l}
\caption{Amazon (product--category) GAT: Degree Distribution Statistics ($M\pm SD$, $n=32$ seeds). Lower Gini coefficient indicates more uniform degree distribution.}
\label{tab:amazon_gat_degree}\\
\toprule
$q$ & $\phi$ & Method & Mean Degree & Gini Coeff. & Num. Isolated & Best Fit \\
\midrule
\endfirsthead
\toprule
$q$ & $\phi$ & Method & Mean Degree & Gini Coeff. & Num. Isolated & Best Fit \\
\midrule
\endhead
\bottomrule
\endfoot

% Checked
% SOURCE: notebooks/amazon/degree_analysis_amazon-GAT-drop0.99-100x-20251126_091842_degree_stats.csv
%         notebooks/amazon/degree_analysis_amazon-GAT-drop0.99-100x-20251126_091842_distribution_fit.csv
% q=0.01, phi=100x
0.01 & 100$\times$        & baseline        & 0.0334 $\pm$ 0.0102            & 0.968 $\pm$ 0.010         & 1417.3 $\pm$ 14.4         & lognormal  \\
0.01 & 100$\times$        & degree\_aware   & 2.4168 $\pm$ 0.2940            & 0.979 $\pm$ 0.003         & 1430.3 $\pm$ 4.4          & lognormal  \\
0.01 & 100$\times$        & simple          & 2.4168 $\pm$ 0.2940            & 0.978 $\pm$ 0.003         & 1430.3 $\pm$ 4.4          & lognormal  \\
0.01 & 100$\times$        & semantic\_knn   & 0.1063 $\pm$ 0.0100            & 0.964 $\pm$ 0.005         & 1382.2 $\pm$ 10.3         & powerlaw   \\
0.01 & 100$\times$        & synthetic       & 2.4168 $\pm$ 0.2940            & 0.827 $\pm$ 0.023         & 1130.6 $\pm$ 41.5         & lognormal  \\
0.01 & 100$\times$        & \textbf{random} & 2.4168 $\pm$ 0.2940            & 0.355 $\pm$ 0.021         & 136.8 $\pm$ 40.9          & powerlaw   \\
0.01 & 100$\times$        & original        & 4.3051 $\pm$ 0.0000            & 0.219 $\pm$ 0.000         & 114.0 $\pm$ 0.0           & powerlaw   \\
\midrule
% Checked
% SOURCE: notebooks/amazon/amazon-GAT-drop0.99-5x-20251123_230121_degree_stats.csv
%         notebooks/amazon/amazon-GAT-drop0.99-5x-20251123_230121_distribution_fit.csv
% q=0.01, phi=5x
0.01 & 5$\times$          & baseline        & 0.0426 $\pm$ 0.0054            & 0.960 $\pm$ 0.005         & 1404.4 $\pm$ 7.7          & lognormal  \\
0.01 & 5$\times$          & degree\_aware   & 0.1208 $\pm$ 0.0147            & 0.982 $\pm$ 0.002         & 1430.3 $\pm$ 4.4          & powerlaw   \\
0.01 & 5$\times$          & simple          & 0.1208 $\pm$ 0.0147            & 0.982 $\pm$ 0.002         & 1430.3 $\pm$ 4.4          & powerlaw   \\
0.01 & 5$\times$          & semantic\_knn   & 0.1060 $\pm$ 0.0096            & 0.963 $\pm$ 0.005         & 1381.6 $\pm$ 10.1         & powerlaw   \\
0.01 & 5$\times$          & synthetic       & 0.1208 $\pm$ 0.0147            & 0.925 $\pm$ 0.010         & 1329.9 $\pm$ 17.3         & powerlaw   \\
0.01 & 5$\times$          & \textbf{random} & 0.1208 $\pm$ 0.0147            & 0.893 $\pm$ 0.012         & 1298.8 $\pm$ 19.1         & lognormal  \\
0.01 & 5$\times$          & original        & 4.3051 $\pm$ 0.0000            & 0.219 $\pm$ 0.000         & 114.0 $\pm$ 0.0           & powerlaw   \\
\midrule
% Checked
% SOURCE: notebooks/amazon/amazon-GAT-drop0.99-2x-20251125_030835_degree_stats.csv
%         notebooks/amazon/amazon-GAT-drop0.99-2x-20251125_030835_distribution_fit.csv
% q=0.01, phi=2x
0.01 & 2$\times$          & baseline        & 0.0426 $\pm$ 0.0054            & 0.960 $\pm$ 0.005         & 1404.4 $\pm$ 7.7          & lognormal  \\
0.01 & 2$\times$          & degree\_aware   & 0.0483 $\pm$ 0.0059            & 0.982 $\pm$ 0.002         & 1430.3 $\pm$ 4.4          & powerlaw   \\
0.01 & 2$\times$          & simple          & 0.0483 $\pm$ 0.0059            & 0.982 $\pm$ 0.002         & 1430.3 $\pm$ 4.4          & powerlaw   \\
0.01 & 2$\times$          & semantic\_knn   & 0.0483 $\pm$ 0.0059            & 0.978 $\pm$ 0.003         & 1419.5 $\pm$ 6.0          & powerlaw   \\
0.01 & 2$\times$          & synthetic       & 0.0483 $\pm$ 0.0059            & 0.960 $\pm$ 0.005         & 1401.4 $\pm$ 8.1          & lognormal  \\
0.01 & 2$\times$          & \textbf{random} & 0.0483 $\pm$ 0.0059            & 0.954 $\pm$ 0.006         & 1396.2 $\pm$ 8.4          & lognormal  \\
0.01 & 2$\times$          & original        & 4.3051 $\pm$ 0.0000            & 0.219 $\pm$ 0.000         & 114.0 $\pm$ 0.0           & powerlaw   \\
\midrule

% Checked
% SOURCE: notebooks/amazon/amazon-GAT-drop0.95-5x-20251124_082457_degree_stats.csv
%         notebooks/amazon/amazon-GAT-drop0.95-5x-20251124_082457_distribution_fit.csv
% q=0.05, phi=5x
0.05 & 5$\times$          & baseline        & 0.2153 $\pm$ 0.0133            & 0.824 $\pm$ 0.011         & 1181.3 $\pm$ 16.7         & lognormal  \\
0.05 & 5$\times$          & degree\_aware   & 0.5930 $\pm$ 0.0365            & 0.916 $\pm$ 0.005         & 1301.9 $\pm$ 9.3          & powerlaw   \\
0.05 & 5$\times$          & simple          & 0.5930 $\pm$ 0.0365            & 0.915 $\pm$ 0.005         & 1301.9 $\pm$ 9.3          & powerlaw   \\
0.05 & 5$\times$          & semantic\_knn   & 0.2682 $\pm$ 0.0140            & 0.858 $\pm$ 0.009         & 1188.1 $\pm$ 16.1         & powerlaw   \\
0.05 & 5$\times$          & synthetic       & 0.5930 $\pm$ 0.0365            & 0.718 $\pm$ 0.016         & 910.9 $\pm$ 31.0          & powerlaw   \\
0.05 & 5$\times$          & \textbf{random} & 0.5930 $\pm$ 0.0365            & 0.639 $\pm$ 0.018         & 811.7 $\pm$ 33.7          & powerlaw   \\
0.05 & 5$\times$          & original        & 4.3051 $\pm$ 0.0000            & 0.219 $\pm$ 0.000         & 114.0 $\pm$ 0.0           & powerlaw   \\
\midrule
% CHECKED
% SOURCE: notebooks/amazon/amazon-GAT-drop0.95-2x-20251125_012114_degree_stats.csv
%         notebooks/amazon/amazon-GAT-drop0.95-2x-20251125_012114_distribution_fit.csv
% q=0.05, phi=2x
0.05 & 2$\times$          & baseline        & 0.2153 $\pm$ 0.0133            & 0.824 $\pm$ 0.011         & 1181.3 $\pm$ 16.7         & lognormal  \\
0.05 & 2$\times$          & degree\_aware   & 0.2372 $\pm$ 0.0146            & 0.919 $\pm$ 0.005         & 1301.9 $\pm$ 9.3          & powerlaw   \\
0.05 & 2$\times$          & simple          & 0.2372 $\pm$ 0.0146            & 0.920 $\pm$ 0.004         & 1301.9 $\pm$ 9.3          & powerlaw   \\
0.05 & 2$\times$          & semantic\_knn   & 0.2372 $\pm$ 0.0146            & 0.876 $\pm$ 0.010         & 1221.0 $\pm$ 17.0         & powerlaw   \\
0.05 & 2$\times$          & synthetic       & 0.2372 $\pm$ 0.0146            & 0.829 $\pm$ 0.011         & 1174.2 $\pm$ 17.8         & powerlaw   \\
0.05 & 2$\times$          & \textbf{random} & 0.2372 $\pm$ 0.0146            & 0.810 $\pm$ 0.012         & 1156.3 $\pm$ 18.2         & lognormal  \\
0.05 & 2$\times$          & original        & 4.3051 $\pm$ 0.0000            & 0.219 $\pm$ 0.000         & 114.0 $\pm$ 0.0           & powerlaw   \\
\midrule

% Checked
% SOURCE: notebooks/amazon/amazon-GAT-drop0.9-5x-20251125_081311_degree_stats.csv
%         notebooks/amazon/amazon-GAT-drop0.9-5x-20251125_081311_distribution_fit.csv
% q=0.10, phi=5x
0.10 & 5$\times$          & baseline        & 0.4323 $\pm$ 0.0163            & 0.703 $\pm$ 0.012         & 950.6 $\pm$ 19.3          & lognormal  \\
0.10 & 5$\times$          & degree\_aware   & 1.1873 $\pm$ 0.0446            & 0.841 $\pm$ 0.006         & 1154.6 $\pm$ 11.2         & powerlaw   \\
0.10 & 5$\times$          & simple          & 1.1873 $\pm$ 0.0446            & 0.843 $\pm$ 0.006         & 1154.6 $\pm$ 11.2         & powerlaw   \\
0.10 & 5$\times$          & semantic\_knn   & 0.4346 $\pm$ 0.0171            & 0.760 $\pm$ 0.010         & 1007.2 $\pm$ 17.0         & powerlaw   \\
0.10 & 5$\times$          & synthetic       & 1.1873 $\pm$ 0.0446            & 0.576 $\pm$ 0.015         & 572.2 $\pm$ 27.0          & powerlaw   \\
0.10 & 5$\times$          & \textbf{random} & 1.1873 $\pm$ 0.0446            & 0.489 $\pm$ 0.013         & 450.4 $\pm$ 26.1          & powerlaw   \\
0.10 & 5$\times$          & original        & 4.3051 $\pm$ 0.0000            & 0.219 $\pm$ 0.000         & 114.0 $\pm$ 0.0           & powerlaw   \\
\midrule
% SOURCE: notebooks/amazon/amazon-GAT-drop0.9-2x-20251125_013548_degree_stats.csv
%         notebooks/amazon/amazon-GAT-drop0.9-2x-20251125_013548_distribution_fit.csv
% q=0.10, phi=2x
0.10 & 2$\times$          & baseline        & 0.4323 $\pm$ 0.0163            & 0.703 $\pm$ 0.012         & 950.6 $\pm$ 19.3          & lognormal  \\
0.10 & 2$\times$          & degree\_aware   & 0.4749 $\pm$ 0.0179            & 0.848 $\pm$ 0.006         & 1154.6 $\pm$ 11.2         & powerlaw   \\
0.10 & 2$\times$          & simple          & 0.4749 $\pm$ 0.0179            & 0.850 $\pm$ 0.006         & 1154.6 $\pm$ 11.2         & powerlaw   \\
0.10 & 2$\times$          & semantic\_knn   & 0.4346 $\pm$ 0.0171            & 0.760 $\pm$ 0.010         & 1007.2 $\pm$ 17.0         & powerlaw   \\
0.10 & 2$\times$          & synthetic       & 0.4749 $\pm$ 0.0179            & 0.712 $\pm$ 0.012         & 941.3 $\pm$ 17.4          & powerlaw   \\
0.10 & 2$\times$          & \textbf{random} & 0.4749 $\pm$ 0.0179            & 0.685 $\pm$ 0.014         & 911.9 $\pm$ 21.6          & powerlaw   \\
0.10 & 2$\times$          & original        & 4.3051 $\pm$ 0.0000            & 0.219 $\pm$ 0.000         & 114.0 $\pm$ 0.0           & powerlaw   \\
\midrule

\end{longtable}

\paragraph{Runtime Analysis}

\begin{longtable}{c c l r r}
\caption{Amazon (product--category) GAT: Runtime Statistics ($M\pm SD$, seconds, $n=32$ seeds). Lower times are better.}
\label{tab:amazon_gat_runtime}\\
\toprule
$q$ & $\phi$ & Method & Aug. Time (s) & Train Time (s) \\
\midrule
\endfirsthead
\toprule
$q$ & $\phi$ & Method & Aug. Time (s) & Train Time (s) \\
\midrule
\endhead
\bottomrule
\endfoot

% CHECKED
% SOURCE: notebooks/amazon/amazon-GAT-drop0.99-5x-20251123_230121_runtime.csv
% q=0.01, phi=5x
0.01 & 5$\times$          & baseline        & 0.0000 $\pm$ 0.0000            & 3.95 $\pm$ 0.41           \\
0.01 & 5$\times$          & degree\_aware   & 0.0013 $\pm$ 0.0010            & 3.98 $\pm$ 1.26           \\
0.01 & 5$\times$          & simple          & 0.0010 $\pm$ 0.0001            & 3.83 $\pm$ 0.95           \\
0.01 & 5$\times$          & semantic\_knn   & 0.0116 $\pm$ 0.0056            & 3.85 $\pm$ 0.87           \\
0.01 & 5$\times$          & synthetic       & 0.0011 $\pm$ 0.0004            & 3.74 $\pm$ 0.74           \\
0.01 & 5$\times$          & \textbf{random} & 0.0009 $\pm$ 0.0001            & 4.00 $\pm$ 0.96           \\
0.01 & 5$\times$          & original        & 0.0000 $\pm$ 0.0000            & 178.54 $\pm$ 20.74        \\
\midrule
% CHECKED
% SOURCE: notebooks/amazon/amazon-GAT-drop0.99-2x-20251125_030835_runtime.csv
% q=0.01, phi=2x
0.01 & 2$\times$          & baseline        & 0.0000 $\pm$ 0.0000            & 4.04 $\pm$ 0.43           \\
0.01 & 2$\times$          & degree\_aware   & 0.0013 $\pm$ 0.0005            & 3.93 $\pm$ 0.91           \\
0.01 & 2$\times$          & simple          & 0.0011 $\pm$ 0.0001            & 3.94 $\pm$ 0.90           \\
0.01 & 2$\times$          & semantic\_knn   & 0.0066 $\pm$ 0.0084            & 3.80 $\pm$ 0.70           \\
0.01 & 2$\times$          & synthetic       & 0.0013 $\pm$ 0.0013            & 3.81 $\pm$ 0.71           \\
0.01 & 2$\times$          & \textbf{random} & 0.0010 $\pm$ 0.0001            & 3.89 $\pm$ 0.78           \\
0.01 & 2$\times$          & original        & 0.0000 $\pm$ 0.0000            & 180.52 $\pm$ 20.72        \\
\midrule

% CHECKED
% SOURCE: notebooks/amazon/amazon-GAT-drop0.95-5x-20251124_082457_runtime.csv
% q=0.05, phi=5x
0.05 & 5$\times$          & baseline        & 0.0000 $\pm$ 0.0000            & 9.81 $\pm$ 4.34           \\
0.05 & 5$\times$          & degree\_aware   & 0.0014 $\pm$ 0.0007            & 9.93 $\pm$ 3.32           \\
0.05 & 5$\times$          & simple          & 0.0011 $\pm$ 0.0001            & 10.23 $\pm$ 4.17          \\
0.05 & 5$\times$          & semantic\_knn   & 0.0358 $\pm$ 0.0058            & 9.66 $\pm$ 3.27           \\
0.05 & 5$\times$          & synthetic       & 0.0012 $\pm$ 0.0003            & 12.38 $\pm$ 4.65          \\
0.05 & 5$\times$          & \textbf{random} & 0.0010 $\pm$ 0.0001            & 15.04 $\pm$ 5.33          \\
0.05 & 5$\times$          & original        & 0.0000 $\pm$ 0.0000            & 181.93 $\pm$ 20.91        \\
\midrule
% CHECKED
% SOURCE: notebooks/amazon/amazon-GAT-drop0.95-2x-20251125_012114_runtime.csv
% q=0.05, phi=2x
0.05 & 2$\times$          & baseline        & 0.0000 $\pm$ 0.0000            & 9.47 $\pm$ 4.17           \\
0.05 & 2$\times$          & degree\_aware   & 0.0014 $\pm$ 0.0012            & 8.61 $\pm$ 2.24           \\
0.05 & 2$\times$          & \textbf{simple} & 0.0010 $\pm$ 0.0000            & 9.77 $\pm$ 5.23           \\
0.05 & 2$\times$          & semantic\_knn   & 0.0219 $\pm$ 0.0068            & 9.74 $\pm$ 3.93           \\
0.05 & 2$\times$          & synthetic       & 0.0011 $\pm$ 0.0006            & 11.43 $\pm$ 3.88          \\
0.05 & 2$\times$          & random          & 0.0010 $\pm$ 0.0000            & 11.23 $\pm$ 3.91          \\
0.05 & 2$\times$          & original        & 0.0000 $\pm$ 0.0000            & 178.24 $\pm$ 20.43        \\
\midrule

% CHECKED
% SOURCE: notebooks/amazon/amazon-GAT-drop0.9-5x-20251125_081311_runtime.csv
% q=0.10, phi=5x
0.10 & 5$\times$          & baseline        & 0.0000 $\pm$ 0.0000            & 20.88 $\pm$ 6.92          \\
0.10 & 5$\times$          & degree\_aware   & 0.0013 $\pm$ 0.0001            & 24.83 $\pm$ 8.14          \\
0.10 & 5$\times$          & \textbf{simple} & 0.0011 $\pm$ 0.0001            & 24.59 $\pm$ 8.47          \\
0.10 & 5$\times$          & semantic\_knn   & 0.0615 $\pm$ 0.0026            & 22.58 $\pm$ 7.62          \\
0.10 & 5$\times$          & synthetic       & 0.0011 $\pm$ 0.0001            & 24.20 $\pm$ 7.16          \\
0.10 & 5$\times$          & random          & 0.0012 $\pm$ 0.0008            & 24.25 $\pm$ 6.23          \\
0.10 & 5$\times$          & original        & 0.0000 $\pm$ 0.0000            & 178.61 $\pm$ 20.40        \\
\midrule
% CHECKED
% SOURCE: notebooks/amazon/amazon-GAT-drop0.9-2x-20251125_013548_runtime.csv
% q=0.10, phi=2x
0.10 & 2$\times$          & baseline        & 0.0000 $\pm$ 0.0000            & 21.17 $\pm$ 7.02          \\
0.10 & 2$\times$          & degree\_aware   & 0.0012 $\pm$ 0.0001            & 23.30 $\pm$ 8.60          \\
0.10 & 2$\times$          & \textbf{simple} & 0.0010 $\pm$ 0.0001            & 23.62 $\pm$ 8.30          \\
0.10 & 2$\times$          & semantic\_knn   & 0.0636 $\pm$ 0.0026            & 22.85 $\pm$ 7.70          \\
0.10 & 2$\times$          & synthetic       & 0.0011 $\pm$ 0.0001            & 23.27 $\pm$ 8.54          \\
0.10 & 2$\times$          & random          & 0.0010 $\pm$ 0.0001            & 24.57 $\pm$ 9.45          \\
0.10 & 2$\times$          & original        & 0.0000 $\pm$ 0.0000            & 179.84 $\pm$ 20.52        \\
\midrule

\end{longtable}

\begin{figure}[H]
  \centering
  \includegraphics[width=1\linewidth]{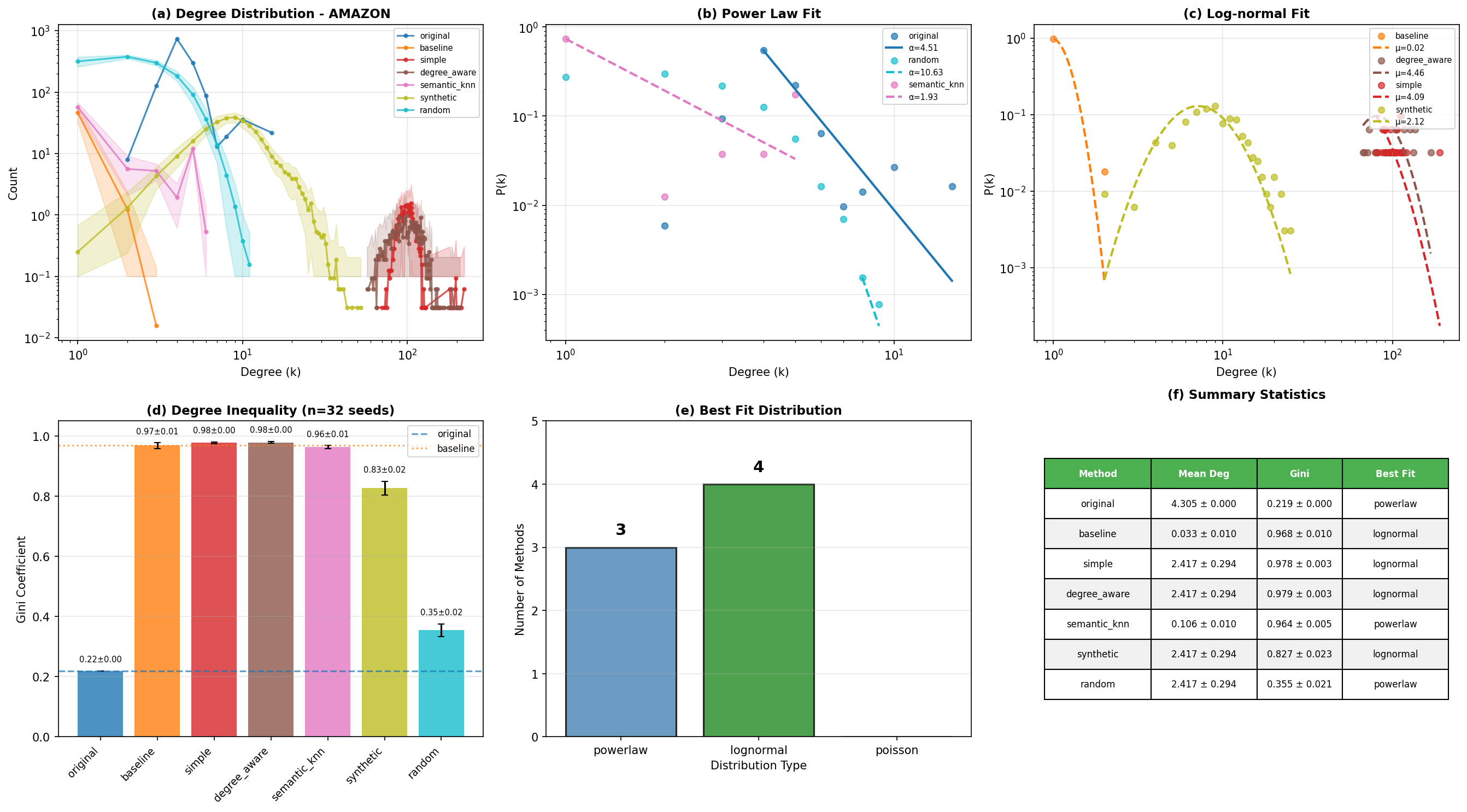}
  \caption{Amazon (product--category), GAT, $q{=}0.01$, $\phi{=}100$: Comprehensive analysis ($M\pm\mathrm{SD}$, $n=32$ seeds) comparing baseline, augmentation methods, and original graph. Panel (a) shows degree distributions on log-log scale with confidence bands; (b) Power Law fits with exponent $\alpha$; (c) Log-normal fits with parameters $\mu$ and $\sigma$; (d) Gini coefficients quantifying degree inequality (lower = more uniform); (e) runtime comparison showing training time (left axis) and augmentation time (right axis, log scale); (f) best-fit distribution counts across methods.}
  \label{fig:amazon_gat_q01_phi100}
\end{figure}

\begin{figure}[H]
  \centering
  \includegraphics[width=1\linewidth]{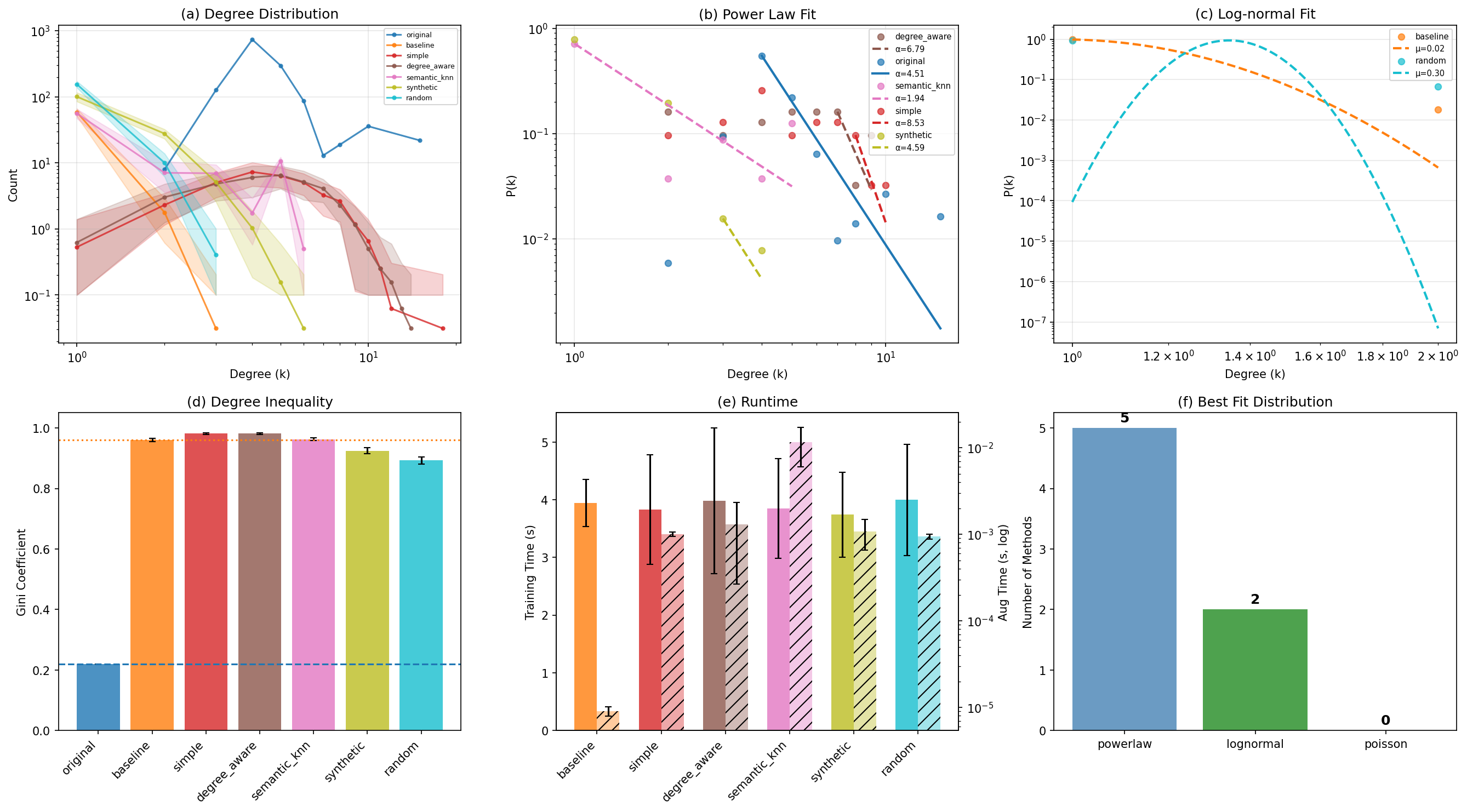}
  \caption{Amazon (product--category), GAT, $q{=}0.01$, $\phi{=}5$: Comprehensive analysis ($M\pm\mathrm{SD}$, $n=32$ seeds) comparing baseline, augmentation methods, and original graph. Panel (a) shows degree distributions on log-log scale with confidence bands; (b) Power Law fits with exponent $\alpha$; (c) Log-normal fits with parameters $\mu$ and $\sigma$; (d) Gini coefficients quantifying degree inequality (lower = more uniform); (e) runtime comparison showing training time (left axis) and augmentation time (right axis, log scale); (f) best-fit distribution counts across methods.}
  \label{fig:amazon_gat_q01_phi5}
\end{figure}

\begin{figure}[H]
  \centering
  \includegraphics[width=1\linewidth]{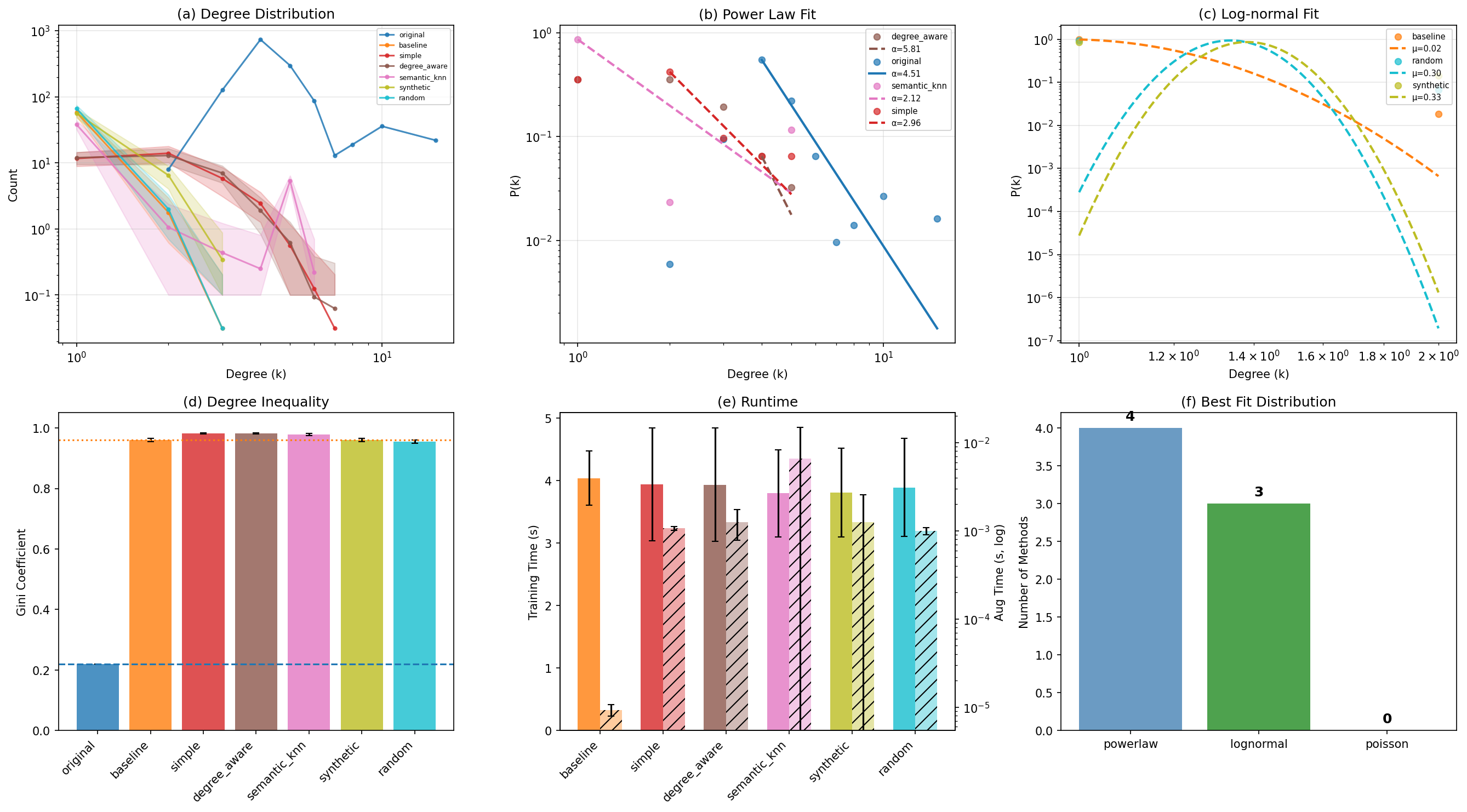}
  \caption{Amazon (product--category), GAT, $q{=}0.01$, $\phi{=}2$: Comprehensive analysis ($M\pm\mathrm{SD}$, $n=32$ seeds) comparing baseline, augmentation methods, and original graph. Panel (a) shows degree distributions on log-log scale with confidence bands; (b) Power Law fits with exponent $\alpha$; (c) Log-normal fits with parameters $\mu$ and $\sigma$; (d) Gini coefficients quantifying degree inequality (lower = more uniform); (e) runtime comparison showing training time (left axis) and augmentation time (right axis, log scale); (f) best-fit distribution counts across methods.}
  \label{fig:amazon_gat_q01_phi2}
\end{figure}

\begin{figure}[H]
  \centering
  \includegraphics[width=1\linewidth]{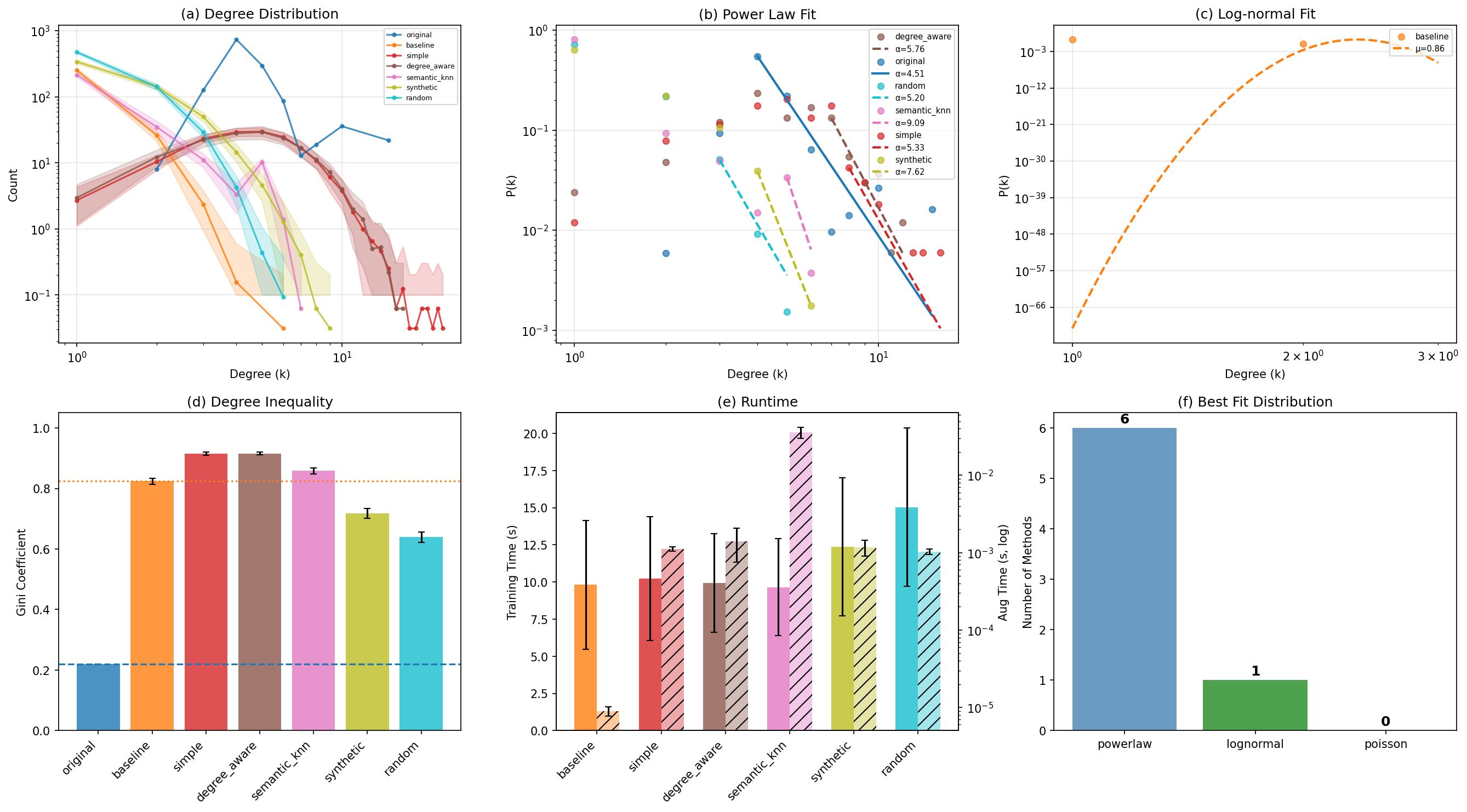}
  \caption{Amazon (product--category), GAT, $q{=}0.05$, $\phi{=}5$: Comprehensive analysis ($M\pm\mathrm{SD}$, $n=32$ seeds) comparing baseline, augmentation methods, and original graph. Panel (a) shows degree distributions on log-log scale with confidence bands; (b) Power Law fits with exponent $\alpha$; (c) Log-normal fits with parameters $\mu$ and $\sigma$; (d) Gini coefficients quantifying degree inequality (lower = more uniform); (e) runtime comparison showing training time (left axis) and augmentation time (right axis, log scale); (f) best-fit distribution counts across methods.}
  \label{fig:amazon_gat_q05_phi5}
\end{figure}

\begin{figure}[H]
  \centering
  \includegraphics[width=1\linewidth]{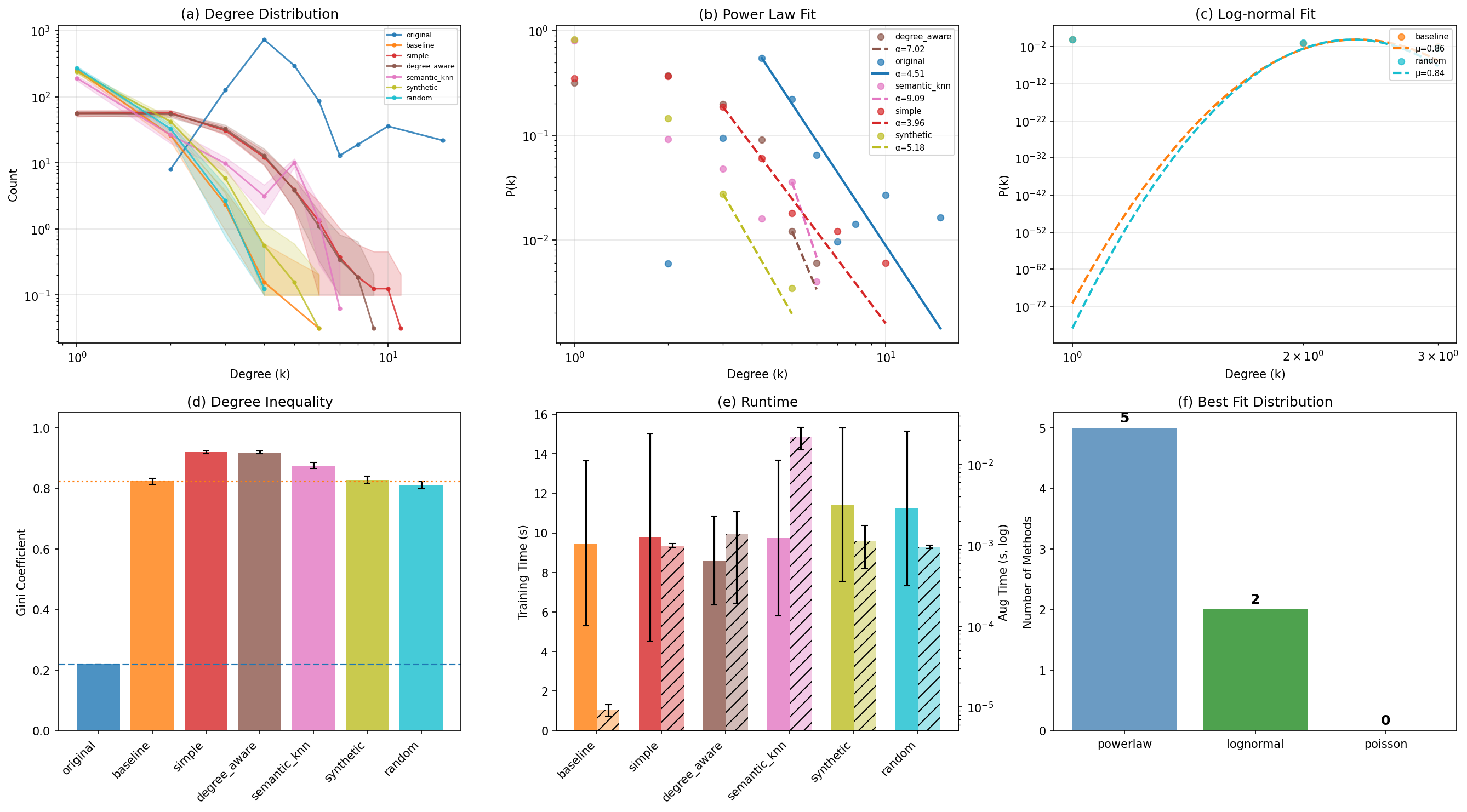}
  \caption{Amazon (product--category), GAT, $q{=}0.05$, $\phi{=}2$: Comprehensive analysis ($M\pm\mathrm{SD}$, $n=32$ seeds) comparing baseline, augmentation methods, and original graph. Panel (a) shows degree distributions on log-log scale with confidence bands; (b) Power Law fits with exponent $\alpha$; (c) Log-normal fits with parameters $\mu$ and $\sigma$; (d) Gini coefficients quantifying degree inequality (lower = more uniform); (e) runtime comparison showing training time (left axis) and augmentation time (right axis, log scale); (f) best-fit distribution counts across methods.}
  \label{fig:amazon_gat_q05_phi2}
\end{figure}

\begin{figure}[H]
  \centering
  \includegraphics[width=1\linewidth]{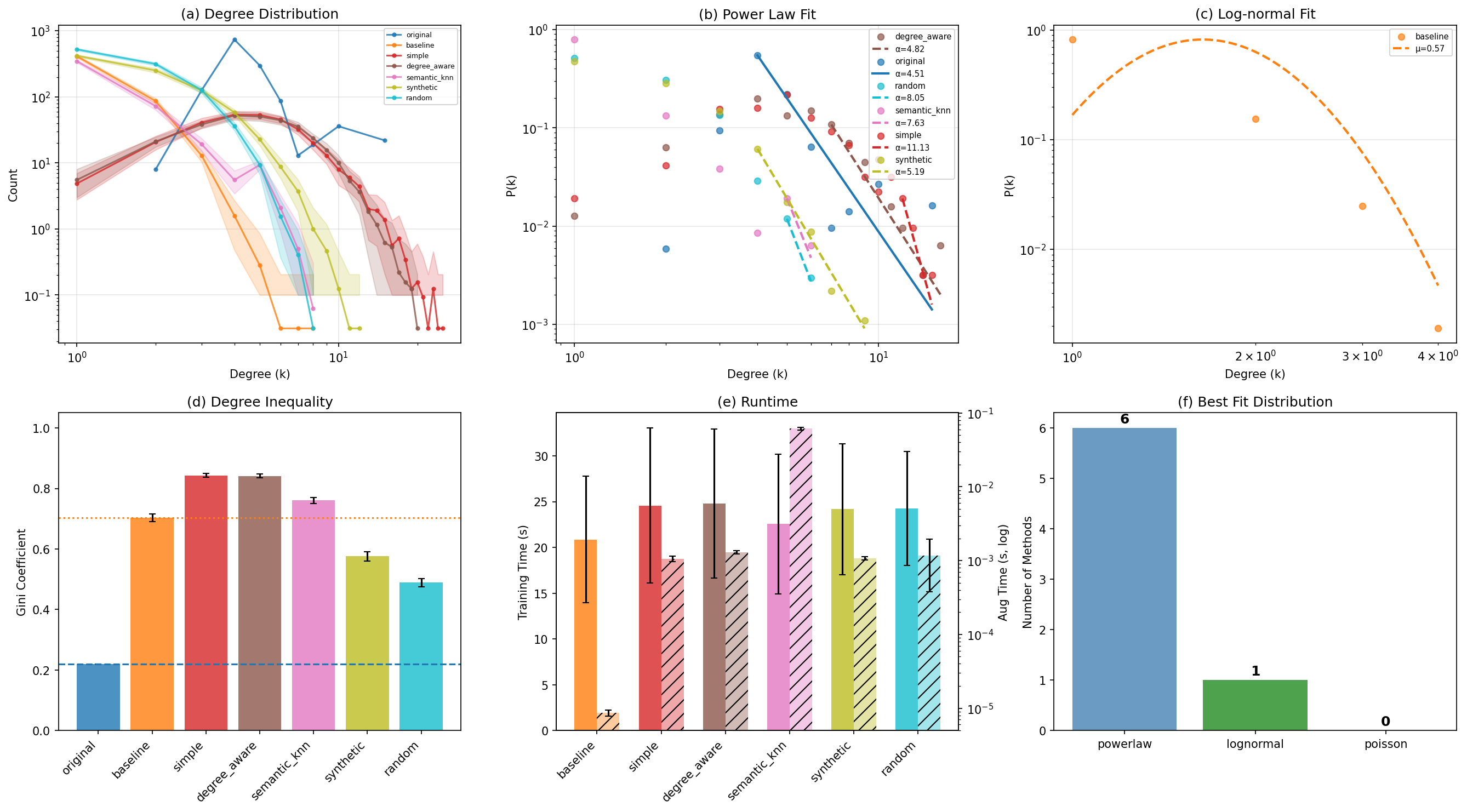}
  \caption{Amazon (product--category), GAT, $q{=}0.10$, $\phi{=}5$: Comprehensive analysis ($M\pm\mathrm{SD}$, $n=32$ seeds) comparing baseline, augmentation methods, and original graph. Panel (a) shows degree distributions on log-log scale with confidence bands; (b) Power Law fits with exponent $\alpha$; (c) Log-normal fits with parameters $\mu$ and $\sigma$; (d) Gini coefficients quantifying degree inequality (lower = more uniform); (e) runtime comparison showing training time (left axis) and augmentation time (right axis, log scale); (f) best-fit distribution counts across methods.}
  \label{fig:amazon_gat_q10_phi5}
\end{figure}

\begin{figure}[H]
  \centering
  \includegraphics[width=1\linewidth]{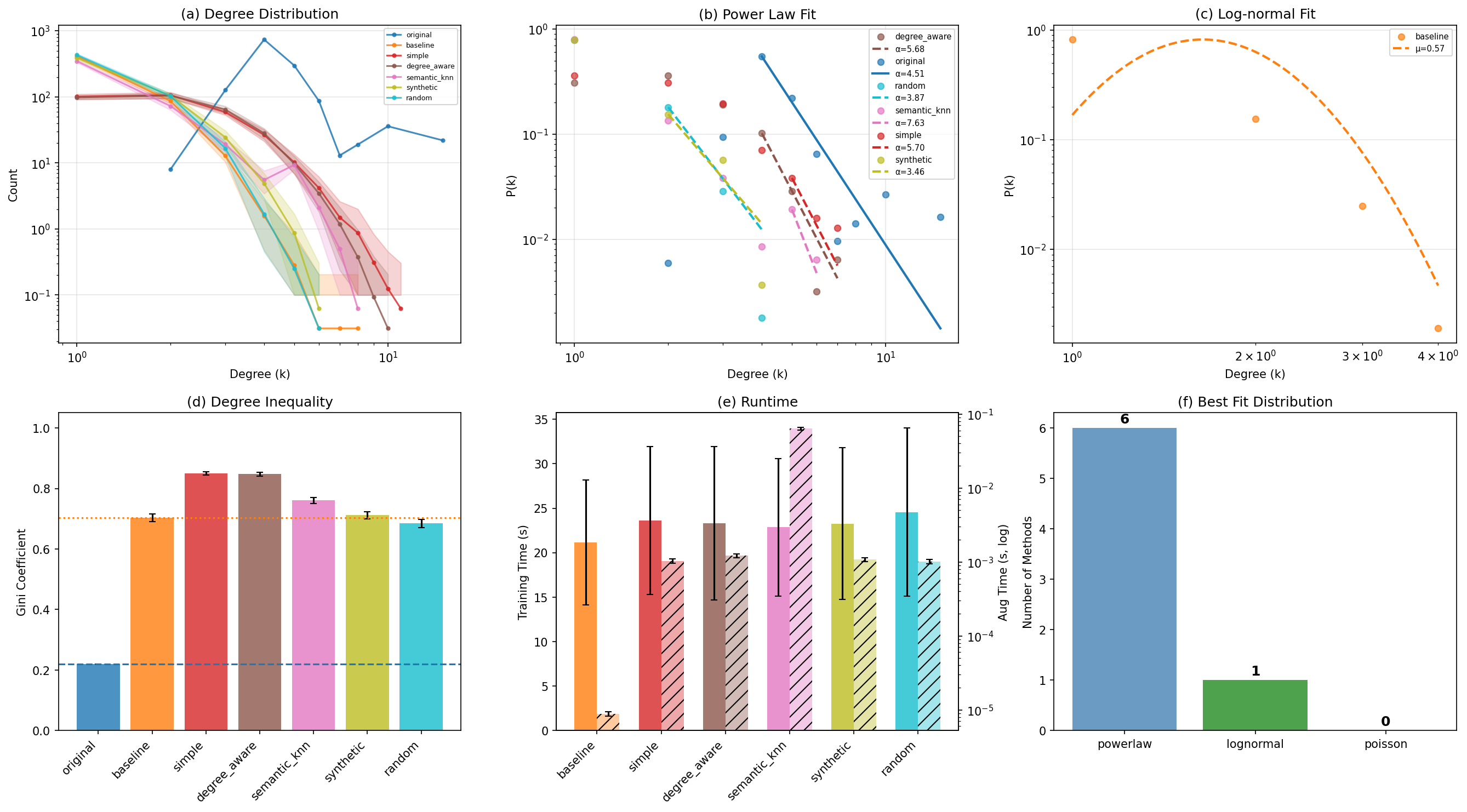}
  \caption{Amazon (product--category), GAT, $q{=}0.10$, $\phi{=}2$: Comprehensive analysis ($M\pm\mathrm{SD}$, $n=32$ seeds) comparing baseline, augmentation methods, and original graph. Panel (a) shows degree distributions on log-log scale with confidence bands; (b) Power Law fits with exponent $\alpha$; (c) Log-normal fits with parameters $\mu$ and $\sigma$; (d) Gini coefficients quantifying degree inequality (lower = more uniform); (e) runtime comparison showing training time (left axis) and augmentation time (right axis, log scale); (f) best-fit distribution counts across methods.}
  \label{fig:amazon_gat_q10_phi2}
\end{figure}

\subsubsection{GraphSAGE}

\paragraph{Summary Analysis}

For Amazon GraphSAGE, \texttt{synthetic} leads AUC at $q{=}0.01$, $\phi{=}5$ (+0.077) and stays positive at $\phi{=}2$ (+0.036), while most other augmentations hover near baseline and \texttt{random} is the weakest. Brier tells the opposite story: \texttt{random} delivers the largest reductions (e.g., $q{=}0.10$, $\phi{=}5$, $\Delta$Brier$=-0.092$; $q{=}0.05$, $\phi{=}5$, $-0.079$), with \texttt{synthetic} also improving at $q{=}0.10$, $\phi{=}5$ ($-0.072$); \texttt{semantic\_knn} and \texttt{degree\_aware} typically give modest or no Brier gains. Degree statistics show augmentations raising mean degree from the ultra-sparse baseline (0.043) to $\sim0.12$ at $\phi{=}5$ and up to $\sim1.19$ at $q{=}0.10$, $\phi{=}5$, while \texttt{random}/\texttt{synthetic} sharply lower Gini and isolated counts. Runtime overheads remain negligible: augmentation under 0.012s and training within a few seconds of baseline (vs.\ $\sim134$s for the dense original), so differences are driven by graph quality rather than cost.

% Data source: Table~\ref{tab:amazon_graphsage_auc} and corresponding Brier table  \textbf{AUC-ROC Performance:} \texttt{synthetic} achieves the highest average AUC improvement (mean $\Delta$AUC=+0.018): at $q=0.01$, $\phi=5x$, AUC=0.584$\pm$0.170 ($\Delta$AUC=+0.077) vs baseline (0.507$\pm$0.153); at $q=0.01$, $\phi=2x$, AUC=0.543$\pm$0.186 ($\Delta$AUC=+0.036). Conversely, \texttt{random} shows the weakest performance (mean $\Delta$AUC=-0.017, range: -0.051 to +0.020).  \textbf{Brier Score Performance:} \texttt{random} achieves the best Brier score (mean $\Delta$Brier=-0.035). At $q=0.10$, $\phi=5x$, it reaches 0.223$\pm$0.028 ($\Delta$Brier=-0.092) vs baseline (0.315$\pm$0.050).  % SOURCE FILES: % - code-repo/notebooks/amazon/amazon-GraphSAGE-drop0.9-2x-20251125_050449_summary.csv % - code-repo/notebooks/amazon/amazon-GraphSAGE-drop0.9-5x-20251125_072248_summary.csv % - code-repo/notebooks/amazon/amazon-GraphSAGE-drop0.95-2x-20251125_052341_summary.csv % - code-repo/notebooks/amazon/amazon-GraphSAGE-drop0.95-5x-20251124_170227_summary.csv % - code-repo/notebooks/amazon/amazon-GraphSAGE-drop0.99-2x-20251125_052148_summary.csv % - code-repo/notebooks/amazon/amazon-GraphSAGE-drop0.99-5x-20251124_145258_summary.csv

\paragraph{AUC and Brier Score}
% ========== AUC TABLE ==========
\begin{longtable}{c c l r r r r r}
\caption{Amazon (product--category) GraphSAGE: AUC-ROC ($M\pm SD$) with paired $t$-tests vs.\ sparse baseline ($n=32$ seeds). A higher AUC is better.}
\label{tab:amazon_graphsage_auc}\\
\toprule
$q$ & $\phi$ & Method & AUC $M\pm SD$ & $\Delta$AUC & $t(31)$ & $p$ & $d$ \\
\midrule
\endfirsthead
\toprule
$q$ & $\phi$ & Method & AUC $M\pm SD$ & $\Delta$AUC & $t(31)$ & $p$ & $d$ \\
\midrule
\endhead
\bottomrule
\endfoot

% WARNING: No summary file found for GraphSAGE drop0.99 100x
% SOURCE: notebooks/amazon/amazon-GraphSAGE-drop0.99-5x-20251124_145258_summary.csv
%         notebooks/amazon/amazon-GraphSAGE-drop0.99-5x-paired-ttest-20251124_145258.csv
% q=0.01, phi=5x
0.01 & 5$\times$      & baseline       & 0.507 $\pm$ 0.153 & +0.000 & ---   & ---   & ---   \\
0.01 & 5$\times$      & degree\_aware  & 0.520 $\pm$ 0.192 & +0.012$^{\mathrm{ns}}$ & $-0.40$ & 0.689    & $-0.07$ \\
0.01 & 5$\times$      & simple         & 0.494 $\pm$ 0.194 & -0.014$^{\mathrm{ns}}$ & $0.42$  & 0.675    & $+0.07$ \\
0.01 & 5$\times$      & semantic\_knn  & 0.526 $\pm$ 0.177 & +0.019$^{\mathrm{ns}}$ & $-0.65$ & 0.521    & $-0.11$ \\
0.01 & 5$\times$      & \textbf{synthetic} & 0.584 $\pm$ 0.170 & +0.077$^{*}$       & $-2.34$ & 0.026    & $-0.41$ \\
0.01 & 5$\times$      & random         & 0.507 $\pm$ 0.183 & -0.000$^{\mathrm{ns}}$ & $0.01$  & 0.990    & $+0.00$ \\
0.01 & 5$\times$      & original       & 0.921 $\pm$ 0.009 & +0.413$^{***}$     & $-15.34$ & $<$0.001 & $-2.71$ \\
\midrule
% SOURCE: notebooks/amazon/amazon-GraphSAGE-drop0.99-2x-20251125_052148_summary.csv
%         notebooks/amazon/amazon-GraphSAGE-drop0.99-2x-paired-ttest-20251125_052148.csv
% q=0.01, phi=2x
0.01 & 2$\times$      & baseline       & 0.507 $\pm$ 0.153 & +0.000 & ---   & ---   & ---   \\
0.01 & 2$\times$      & degree\_aware  & 0.516 $\pm$ 0.177 & +0.009$^{\mathrm{ns}}$ & $-0.30$ & 0.766    & $-0.05$ \\
0.01 & 2$\times$      & simple         & 0.471 $\pm$ 0.176 & -0.036$^{\mathrm{ns}}$ & $1.38$  & 0.177    & $+0.24$ \\
0.01 & 2$\times$      & semantic\_knn  & 0.522 $\pm$ 0.163 & +0.014$^{\mathrm{ns}}$ & $-0.46$ & 0.646    & $-0.08$ \\
0.01 & 2$\times$      & \textbf{synthetic} & 0.543 $\pm$ 0.186 & +0.036$^{\mathrm{ns}}$ & $-1.10$ & 0.279    & $-0.19$ \\
0.01 & 2$\times$      & random         & 0.528 $\pm$ 0.177 & +0.020$^{\mathrm{ns}}$ & $-0.54$ & 0.593    & $-0.10$ \\
0.01 & 2$\times$      & original       & 0.921 $\pm$ 0.009 & +0.413$^{***}$     & $-15.34$ & $<$0.001 & $-2.71$ \\
\midrule

% WARNING: No summary file found for GraphSAGE drop0.95 100x
% SOURCE: notebooks/amazon/amazon-GraphSAGE-drop0.95-5x-20251124_170227_summary.csv
%         notebooks/amazon/amazon-GraphSAGE-drop0.95-5x-paired-ttest-20251124_170227.csv
% q=0.05, phi=5x
0.05 & 5$\times$      & baseline       & 0.663 $\pm$ 0.086 & +0.000 & ---   & ---   & ---   \\
0.05 & 5$\times$      & degree\_aware  & 0.649 $\pm$ 0.082 & -0.014$^{\mathrm{ns}}$ & $1.18$  & 0.247    & $+0.21$ \\
0.05 & 5$\times$      & simple         & 0.658 $\pm$ 0.069 & -0.005$^{\mathrm{ns}}$ & $0.45$  & 0.655    & $+0.08$ \\
0.05 & 5$\times$      & \textbf{semantic\_knn} & 0.677 $\pm$ 0.079 & +0.014$^{\mathrm{ns}}$ & $-1.31$ & 0.198    & $-0.23$ \\
0.05 & 5$\times$      & synthetic      & 0.677 $\pm$ 0.075 & +0.014$^{\mathrm{ns}}$ & $-1.32$ & 0.195    & $-0.23$ \\
0.05 & 5$\times$      & random         & 0.611 $\pm$ 0.113 & -0.051$^{**}$      & $2.98$  & 0.006    & $+0.53$ \\
0.05 & 5$\times$      & original       & 0.921 $\pm$ 0.009 & +0.258$^{***}$     & $-17.13$ & $<$0.001 & $-3.03$ \\
\midrule
% SOURCE: notebooks/amazon/amazon-GraphSAGE-drop0.95-2x-20251125_052341_summary.csv
%         notebooks/amazon/amazon-GraphSAGE-drop0.95-2x-paired-ttest-20251125_052341.csv
% q=0.05, phi=2x
0.05 & 2$\times$      & baseline       & 0.663 $\pm$ 0.086 & +0.000 & ---   & ---   & ---   \\
0.05 & 2$\times$      & degree\_aware  & 0.646 $\pm$ 0.079 & -0.016$^{\mathrm{ns}}$ & $1.46$  & 0.154    & $+0.26$ \\
0.05 & 2$\times$      & simple         & 0.632 $\pm$ 0.074 & -0.031$^{**}$      & $3.04$  & 0.005    & $+0.54$ \\
0.05 & 2$\times$      & \textbf{semantic\_knn} & 0.668 $\pm$ 0.085 & +0.005$^{\mathrm{ns}}$ & $-0.54$ & 0.591    & $-0.10$ \\
0.05 & 2$\times$      & synthetic      & 0.655 $\pm$ 0.096 & -0.008$^{\mathrm{ns}}$ & $0.69$  & 0.497    & $+0.12$ \\
0.05 & 2$\times$      & random         & 0.648 $\pm$ 0.084 & -0.014$^{\mathrm{ns}}$ & $1.05$  & 0.301    & $+0.19$ \\
0.05 & 2$\times$      & original       & 0.921 $\pm$ 0.009 & +0.258$^{***}$     & $-17.13$ & $<$0.001 & $-3.03$ \\
\midrule

% WARNING: No summary file found for GraphSAGE drop0.9 100x
% SOURCE: notebooks/amazon/amazon-GraphSAGE-drop0.9-5x-20251125_072248_summary.csv
%         notebooks/amazon/amazon-GraphSAGE-drop0.9-5x-paired-ttest-20251125_072248.csv
% q=0.10, phi=5x
0.10 & 5$\times$      & \textbf{baseline} & 0.743 $\pm$ 0.037 & +0.000 & ---   & ---   & ---   \\
0.10 & 5$\times$      & degree\_aware  & 0.730 $\pm$ 0.038 & -0.013$^{\mathrm{ns}}$ & $1.95$  & 0.061    & $+0.34$ \\
0.10 & 5$\times$      & simple         & 0.737 $\pm$ 0.043 & -0.005$^{\mathrm{ns}}$ & $0.73$  & 0.469    & $+0.13$ \\
0.10 & 5$\times$      & semantic\_knn  & 0.733 $\pm$ 0.038 & -0.010$^{\mathrm{ns}}$ & $1.49$  & 0.147    & $+0.26$ \\
0.10 & 5$\times$      & synthetic      & 0.737 $\pm$ 0.057 & -0.006$^{\mathrm{ns}}$ & $0.51$  & 0.615    & $+0.09$ \\
0.10 & 5$\times$      & random         & 0.698 $\pm$ 0.065 & -0.045$^{**}$      & $3.54$  & 0.001    & $+0.63$ \\
0.10 & 5$\times$      & original       & 0.921 $\pm$ 0.009 & +0.178$^{***}$     & $-28.27$ & $<$0.001 & $-5.00$ \\
\midrule
% SOURCE: notebooks/amazon/amazon-GraphSAGE-drop0.9-2x-20251125_050449_summary.csv
%         notebooks/amazon/amazon-GraphSAGE-drop0.9-2x-paired-ttest-20251125_050450.csv
% q=0.10, phi=2x
0.10 & 2$\times$      & \textbf{baseline} & 0.743 $\pm$ 0.037 & +0.000 & ---   & ---   & ---   \\
0.10 & 2$\times$      & degree\_aware  & 0.737 $\pm$ 0.045 & -0.006$^{\mathrm{ns}}$ & $0.83$  & 0.411    & $+0.15$ \\
0.10 & 2$\times$      & simple         & 0.740 $\pm$ 0.049 & -0.003$^{\mathrm{ns}}$ & $0.32$  & 0.752    & $+0.06$ \\
0.10 & 2$\times$      & semantic\_knn  & 0.733 $\pm$ 0.038 & -0.010$^{\mathrm{ns}}$ & $1.49$  & 0.147    & $+0.26$ \\
0.10 & 2$\times$      & synthetic      & 0.738 $\pm$ 0.038 & -0.005$^{\mathrm{ns}}$ & $0.65$  & 0.521    & $+0.11$ \\
0.10 & 2$\times$      & random         & 0.730 $\pm$ 0.050 & -0.013$^{\mathrm{ns}}$ & $1.45$  & 0.158    & $+0.26$ \\
0.10 & 2$\times$      & original       & 0.921 $\pm$ 0.009 & +0.178$^{***}$     & $-28.27$ & $<$0.001 & $-5.00$ \\
\midrule

\end{longtable}

% ========== BRIER TABLE ==========
\begin{longtable}{c c l r r r r r}
\caption{Amazon (product--category) GraphSAGE: Brier Score ($M\pm SD$) with paired $t$-tests vs.\ sparse baseline ($n=32$ seeds, lower is better).}
\label{tab:amazon_graphsage_brier}\\
\toprule
$q$ & $\phi$ & Method & Brier $M\pm SD$ & $\Delta$Brier & $t(31)$ & $p$ & $d$ \\
\midrule
\endfirsthead
\toprule
$q$ & $\phi$ & Method & Brier $M\pm SD$ & $\Delta$Brier & $t(31)$ & $p$ & $d$ \\
\midrule
\endhead
\bottomrule
\endfoot

% WARNING: No summary file found for GraphSAGE drop0.99 100x
% SOURCE: notebooks/amazon/amazon-GraphSAGE-drop0.99-5x-20251124_145258_summary.csv
%         notebooks/amazon/amazon-GraphSAGE-drop0.99-5x-paired-ttest-20251124_145258.csv
% q=0.01, phi=5x
0.01 & 5$\times$      & baseline       & 0.397 $\pm$ 0.041 & +0.000 & ---  & ---   & ---   \\
0.01 & 5$\times$      & degree\_aware  & 0.395 $\pm$ 0.050 & -0.001$^{\mathrm{ns}}$ & $-0.01$ & 0.993    & $-0.00$  \\
0.01 & 5$\times$      & \textbf{simple} & 0.391 $\pm$ 0.046 & -0.006$^{\mathrm{ns}}$ & $0.46$ & 0.646    & $+0.08$  \\
0.01 & 5$\times$      & semantic\_knn  & 0.399 $\pm$ 0.059 & +0.002$^{\mathrm{ns}}$ & $-0.29$ & 0.777    & $-0.05$  \\
0.01 & 5$\times$      & synthetic      & 0.398 $\pm$ 0.039 & +0.001$^{\mathrm{ns}}$ & $-0.35$ & 0.729    & $-0.06$  \\
0.01 & 5$\times$      & random         & 0.401 $\pm$ 0.033 & +0.004$^{\mathrm{ns}}$ & $-0.68$ & 0.499    & $-0.12$  \\
0.01 & 5$\times$      & original       & 0.105 $\pm$ 0.011 & -0.292$^{***}$      & $37.24$ & $<$0.001 & $+6.58$  \\
\midrule
% SOURCE: notebooks/amazon/amazon-GraphSAGE-drop0.99-2x-20251125_052148_summary.csv
%         notebooks/amazon/amazon-GraphSAGE-drop0.99-2x-paired-ttest-20251125_052148.csv
% q=0.01, phi=2x
0.01 & 2$\times$      & baseline       & 0.397 $\pm$ 0.041 & +0.000 & ---  & ---   & ---   \\
0.01 & 2$\times$      & degree\_aware  & 0.394 $\pm$ 0.057 & -0.003$^{\mathrm{ns}}$ & $0.20$ & 0.847    & $+0.04$  \\
0.01 & 2$\times$      & simple         & 0.408 $\pm$ 0.036 & +0.012$^{\mathrm{ns}}$ & $-1.72$ & 0.095    & $-0.31$  \\
0.01 & 2$\times$      & semantic\_knn  & 0.402 $\pm$ 0.044 & +0.005$^{\mathrm{ns}}$ & $-0.96$ & 0.344    & $-0.17$  \\
0.01 & 2$\times$      & \textbf{synthetic} & 0.393 $\pm$ 0.047 & -0.004$^{\mathrm{ns}}$ & $0.31$ & 0.758    & $+0.06$  \\
0.01 & 2$\times$      & random         & 0.398 $\pm$ 0.046 & +0.001$^{\mathrm{ns}}$ & $-0.22$ & 0.829    & $-0.04$  \\
0.01 & 2$\times$      & original       & 0.105 $\pm$ 0.011 & -0.292$^{***}$      & $37.24$ & $<$0.001 & $+6.58$  \\
\midrule

% WARNING: No summary file found for GraphSAGE drop0.95 100x
% SOURCE: notebooks/amazon/amazon-GraphSAGE-drop0.95-5x-20251124_170227_summary.csv
%         notebooks/amazon/amazon-GraphSAGE-drop0.95-5x-paired-ttest-20251124_170227.csv
% q=0.05, phi=5x
0.05 & 5$\times$      & baseline       & 0.361 $\pm$ 0.032 & +0.000 & ---  & ---   & ---   \\
0.05 & 5$\times$      & degree\_aware  & 0.367 $\pm$ 0.035 & +0.005$^{\mathrm{ns}}$ & $-0.82$ & 0.419    & $-0.14$  \\
0.05 & 5$\times$      & simple         & 0.366 $\pm$ 0.031 & +0.005$^{\mathrm{ns}}$ & $-0.74$ & 0.468    & $-0.13$  \\
0.05 & 5$\times$      & semantic\_knn  & 0.361 $\pm$ 0.031 & -0.000$^{\mathrm{ns}}$ & $0.03$ & 0.976    & $+0.01$  \\
0.05 & 5$\times$      & synthetic      & 0.330 $\pm$ 0.073 & -0.031$^{*}$        & $2.56$ & 0.015    & $+0.45$  \\
0.05 & 5$\times$      & \textbf{random} & 0.282 $\pm$ 0.072 & -0.079$^{***}$      & $6.62$ & $<$0.001 & $+1.17$  \\
0.05 & 5$\times$      & original       & 0.105 $\pm$ 0.011 & -0.256$^{***}$      & $41.94$ & $<$0.001 & $+7.41$  \\
\midrule
% SOURCE: notebooks/amazon/amazon-GraphSAGE-drop0.95-2x-20251125_052341_summary.csv
%         notebooks/amazon/amazon-GraphSAGE-drop0.95-2x-paired-ttest-20251125_052341.csv
% q=0.05, phi=2x
0.05 & 2$\times$      & \textbf{baseline} & 0.361 $\pm$ 0.032 & +0.000 & ---  & ---   & ---   \\
0.05 & 2$\times$      & degree\_aware  & 0.367 $\pm$ 0.027 & +0.006$^{\mathrm{ns}}$ & $-1.01$ & 0.322    & $-0.18$  \\
0.05 & 2$\times$      & simple         & 0.365 $\pm$ 0.024 & +0.004$^{\mathrm{ns}}$ & $-0.76$ & 0.453    & $-0.13$  \\
0.05 & 2$\times$      & semantic\_knn  & 0.372 $\pm$ 0.029 & +0.011$^{*}$        & $-2.16$ & 0.038    & $-0.38$  \\
0.05 & 2$\times$      & synthetic      & 0.375 $\pm$ 0.028 & +0.014$^{*}$        & $-2.43$ & 0.021    & $-0.43$  \\
0.05 & 2$\times$      & random         & 0.369 $\pm$ 0.036 & +0.008$^{\mathrm{ns}}$ & $-1.33$ & 0.194    & $-0.23$  \\
0.05 & 2$\times$      & original       & 0.105 $\pm$ 0.011 & -0.256$^{***}$      & $41.94$ & $<$0.001 & $+7.41$  \\
\midrule

% WARNING: No summary file found for GraphSAGE drop0.9 100x
% SOURCE: notebooks/amazon/amazon-GraphSAGE-drop0.9-5x-20251125_072248_summary.csv
%         notebooks/amazon/amazon-GraphSAGE-drop0.9-5x-paired-ttest-20251125_072248.csv
% q=0.10, phi=5x
0.10 & 5$\times$      & baseline       & 0.315 $\pm$ 0.050 & +0.000 & ---  & ---   & ---   \\
0.10 & 5$\times$      & degree\_aware  & 0.315 $\pm$ 0.056 & +0.000$^{\mathrm{ns}}$ & $-0.02$ & 0.986    & $-0.00$  \\
0.10 & 5$\times$      & simple         & 0.319 $\pm$ 0.041 & +0.004$^{\mathrm{ns}}$ & $-0.49$ & 0.627    & $-0.09$  \\
0.10 & 5$\times$      & semantic\_knn  & 0.305 $\pm$ 0.042 & -0.009$^{\mathrm{ns}}$ & $1.07$ & 0.293    & $+0.19$  \\
0.10 & 5$\times$      & synthetic      & 0.243 $\pm$ 0.056 & -0.072$^{***}$      & $6.29$ & $<$0.001 & $+1.11$  \\
0.10 & 5$\times$      & \textbf{random} & 0.223 $\pm$ 0.028 & -0.092$^{***}$      & $9.04$ & $<$0.001 & $+1.60$  \\
0.10 & 5$\times$      & original       & 0.105 $\pm$ 0.011 & -0.210$^{***}$      & $24.31$ & $<$0.001 & $+4.30$  \\
\midrule
% SOURCE: notebooks/amazon/amazon-GraphSAGE-drop0.9-2x-20251125_050449_summary.csv
%         notebooks/amazon/amazon-GraphSAGE-drop0.9-2x-paired-ttest-20251125_050450.csv
% q=0.10, phi=2x
0.10 & 2$\times$      & baseline       & 0.315 $\pm$ 0.050 & +0.000 & ---  & ---   & ---   \\
0.10 & 2$\times$      & degree\_aware  & 0.313 $\pm$ 0.045 & -0.001$^{\mathrm{ns}}$ & $0.13$ & 0.898    & $+0.02$  \\
0.10 & 2$\times$      & simple         & 0.325 $\pm$ 0.045 & +0.010$^{\mathrm{ns}}$ & $-0.87$ & 0.390    & $-0.15$  \\
0.10 & 2$\times$      & semantic\_knn  & 0.305 $\pm$ 0.042 & -0.009$^{\mathrm{ns}}$ & $1.07$ & 0.293    & $+0.19$  \\
0.10 & 2$\times$      & synthetic      & 0.286 $\pm$ 0.047 & -0.029$^{*}$        & $2.38$ & 0.023    & $+0.42$  \\
0.10 & 2$\times$      & \textbf{random} & 0.261 $\pm$ 0.051 & -0.053$^{***}$      & $3.95$ & $<$0.001 & $+0.70$  \\
0.10 & 2$\times$      & original       & 0.105 $\pm$ 0.011 & -0.210$^{***}$      & $24.31$ & $<$0.001 & $+4.30$  \\
\midrule

\end{longtable}

\paragraph{Degree Distribution Analysis}

% ========== DEGREE DISTRIBUTION STATISTICS ==========
\begin{longtable}{c c l r r r l}
\caption{Amazon (product--category) GraphSAGE: Degree Distribution Statistics ($M\pm SD$, $n=32$ seeds). Lower Gini coefficient indicates more uniform degree distribution.}
\label{tab:amazon_graphsage_degree}\\
\toprule
$q$ & $\phi$ & Method & Mean Degree & Gini Coeff. & Num. Isolated & Best Fit \\
\midrule
\endfirsthead
\toprule
$q$ & $\phi$ & Method & Mean Degree & Gini Coeff. & Num. Isolated & Best Fit \\
\midrule
\endhead
\bottomrule
\endfoot

% SOURCE: notebooks/amazon/amazon-GraphSAGE-drop0.99-5x-20251124_145258_degree_stats.csv
%         notebooks/amazon/amazon-GraphSAGE-drop0.99-5x-20251124_145258_distribution_fit.csv
% q=0.01, phi=5x
0.01 & 5$\times$          & baseline        & 0.0426 $\pm$ 0.0054            & 0.960 $\pm$ 0.005         & 1404.4 $\pm$ 7.7          & lognormal  \\
0.01 & 5$\times$          & degree\_aware   & 0.1208 $\pm$ 0.0147            & 0.982 $\pm$ 0.002         & 1430.3 $\pm$ 4.4          & powerlaw   \\
0.01 & 5$\times$          & simple          & 0.1208 $\pm$ 0.0147            & 0.982 $\pm$ 0.002         & 1430.3 $\pm$ 4.4          & powerlaw   \\
0.01 & 5$\times$          & semantic\_knn   & 0.1060 $\pm$ 0.0096            & 0.963 $\pm$ 0.005         & 1381.6 $\pm$ 10.1         & powerlaw   \\
0.01 & 5$\times$          & synthetic       & 0.1208 $\pm$ 0.0147            & 0.925 $\pm$ 0.010         & 1329.9 $\pm$ 17.3         & powerlaw   \\
0.01 & 5$\times$          & \textbf{random} & 0.1208 $\pm$ 0.0147            & 0.893 $\pm$ 0.012         & 1298.8 $\pm$ 19.1         & lognormal  \\
0.01 & 5$\times$          & original        & 4.3051 $\pm$ 0.0000            & 0.219 $\pm$ 0.000         & 114.0 $\pm$ 0.0           & powerlaw   \\
\midrule
% SOURCE: notebooks/amazon/amazon-GraphSAGE-drop0.99-2x-20251125_052148_degree_stats.csv
%         notebooks/amazon/amazon-GraphSAGE-drop0.99-2x-20251125_052148_distribution_fit.csv
% q=0.01, phi=2x
0.01 & 2$\times$          & baseline        & 0.0426 $\pm$ 0.0054            & 0.960 $\pm$ 0.005         & 1404.4 $\pm$ 7.7          & lognormal  \\
0.01 & 2$\times$          & degree\_aware   & 0.0483 $\pm$ 0.0059            & 0.982 $\pm$ 0.002         & 1430.3 $\pm$ 4.4          & powerlaw   \\
0.01 & 2$\times$          & simple          & 0.0483 $\pm$ 0.0059            & 0.982 $\pm$ 0.002         & 1430.3 $\pm$ 4.4          & powerlaw   \\
0.01 & 2$\times$          & semantic\_knn   & 0.0483 $\pm$ 0.0059            & 0.978 $\pm$ 0.003         & 1419.5 $\pm$ 6.0          & powerlaw   \\
0.01 & 2$\times$          & synthetic       & 0.0483 $\pm$ 0.0059            & 0.960 $\pm$ 0.005         & 1401.4 $\pm$ 8.1          & lognormal  \\
0.01 & 2$\times$          & \textbf{random} & 0.0483 $\pm$ 0.0059            & 0.954 $\pm$ 0.006         & 1396.2 $\pm$ 8.4          & lognormal  \\
0.01 & 2$\times$          & original        & 4.3051 $\pm$ 0.0000            & 0.219 $\pm$ 0.000         & 114.0 $\pm$ 0.0           & powerlaw   \\
\midrule

% SOURCE: notebooks/amazon/amazon-GraphSAGE-drop0.95-5x-20251124_170227_degree_stats.csv
%         notebooks/amazon/amazon-GraphSAGE-drop0.95-5x-20251124_170227_distribution_fit.csv
% q=0.05, phi=5x
0.05 & 5$\times$          & baseline        & 0.2153 $\pm$ 0.0133            & 0.824 $\pm$ 0.011         & 1181.3 $\pm$ 16.7         & lognormal  \\
0.05 & 5$\times$          & degree\_aware   & 0.5930 $\pm$ 0.0365            & 0.916 $\pm$ 0.005         & 1301.9 $\pm$ 9.3          & powerlaw   \\
0.05 & 5$\times$          & simple          & 0.5930 $\pm$ 0.0365            & 0.915 $\pm$ 0.005         & 1301.9 $\pm$ 9.3          & powerlaw   \\
0.05 & 5$\times$          & semantic\_knn   & 0.2682 $\pm$ 0.0140            & 0.858 $\pm$ 0.009         & 1188.1 $\pm$ 16.1         & powerlaw   \\
0.05 & 5$\times$          & synthetic       & 0.5930 $\pm$ 0.0365            & 0.718 $\pm$ 0.016         & 910.9 $\pm$ 31.0          & powerlaw   \\
0.05 & 5$\times$          & \textbf{random} & 0.5930 $\pm$ 0.0365            & 0.639 $\pm$ 0.018         & 811.7 $\pm$ 33.7          & powerlaw   \\
0.05 & 5$\times$          & original        & 4.3051 $\pm$ 0.0000            & 0.219 $\pm$ 0.000         & 114.0 $\pm$ 0.0           & powerlaw   \\
\midrule
% SOURCE: notebooks/amazon/amazon-GraphSAGE-drop0.95-2x-20251125_052341_degree_stats.csv
%         notebooks/amazon/amazon-GraphSAGE-drop0.95-2x-20251125_052341_distribution_fit.csv
% q=0.05, phi=2x
0.05 & 2$\times$          & baseline        & 0.2153 $\pm$ 0.0133            & 0.824 $\pm$ 0.011         & 1181.3 $\pm$ 16.7         & lognormal  \\
0.05 & 2$\times$          & degree\_aware   & 0.2372 $\pm$ 0.0146            & 0.919 $\pm$ 0.005         & 1301.9 $\pm$ 9.3          & powerlaw   \\
0.05 & 2$\times$          & simple          & 0.2372 $\pm$ 0.0146            & 0.920 $\pm$ 0.004         & 1301.9 $\pm$ 9.3          & powerlaw   \\
0.05 & 2$\times$          & semantic\_knn   & 0.2372 $\pm$ 0.0146            & 0.876 $\pm$ 0.010         & 1221.0 $\pm$ 17.0         & powerlaw   \\
0.05 & 2$\times$          & synthetic       & 0.2372 $\pm$ 0.0146            & 0.829 $\pm$ 0.011         & 1174.2 $\pm$ 17.8         & powerlaw   \\
0.05 & 2$\times$          & \textbf{random} & 0.2372 $\pm$ 0.0146            & 0.810 $\pm$ 0.012         & 1156.3 $\pm$ 18.2         & lognormal  \\
0.05 & 2$\times$          & original        & 4.3051 $\pm$ 0.0000            & 0.219 $\pm$ 0.000         & 114.0 $\pm$ 0.0           & powerlaw   \\
\midrule

% SOURCE: notebooks/amazon/amazon-GraphSAGE-drop0.9-5x-20251125_072248_degree_stats.csv
%         notebooks/amazon/amazon-GraphSAGE-drop0.9-5x-20251125_072248_distribution_fit.csv
% q=0.10, phi=5x
0.10 & 5$\times$          & baseline        & 0.4323 $\pm$ 0.0163            & 0.703 $\pm$ 0.012         & 950.6 $\pm$ 19.3          & lognormal  \\
0.10 & 5$\times$          & degree\_aware   & 1.1873 $\pm$ 0.0446            & 0.841 $\pm$ 0.006         & 1154.6 $\pm$ 11.2         & powerlaw   \\
0.10 & 5$\times$          & simple          & 1.1873 $\pm$ 0.0446            & 0.843 $\pm$ 0.006         & 1154.6 $\pm$ 11.2         & powerlaw   \\
0.10 & 5$\times$          & semantic\_knn   & 0.4346 $\pm$ 0.0171            & 0.760 $\pm$ 0.010         & 1007.2 $\pm$ 17.0         & powerlaw   \\
0.10 & 5$\times$          & synthetic       & 1.1873 $\pm$ 0.0446            & 0.576 $\pm$ 0.015         & 572.2 $\pm$ 27.0          & powerlaw   \\
0.10 & 5$\times$          & \textbf{random} & 1.1873 $\pm$ 0.0446            & 0.489 $\pm$ 0.013         & 450.4 $\pm$ 26.1          & powerlaw   \\
0.10 & 5$\times$          & original        & 4.3051 $\pm$ 0.0000            & 0.219 $\pm$ 0.000         & 114.0 $\pm$ 0.0           & powerlaw   \\
\midrule
% SOURCE: notebooks/amazon/amazon-GraphSAGE-drop0.9-2x-20251125_050449_degree_stats.csv
%         notebooks/amazon/amazon-GraphSAGE-drop0.9-2x-20251125_050449_distribution_fit.csv
% q=0.10, phi=2x
0.10 & 2$\times$          & baseline        & 0.4323 $\pm$ 0.0163            & 0.703 $\pm$ 0.012         & 950.6 $\pm$ 19.3          & lognormal  \\
0.10 & 2$\times$          & degree\_aware   & 0.4749 $\pm$ 0.0179            & 0.848 $\pm$ 0.006         & 1154.6 $\pm$ 11.2         & powerlaw   \\
0.10 & 2$\times$          & simple          & 0.4749 $\pm$ 0.0179            & 0.850 $\pm$ 0.006         & 1154.6 $\pm$ 11.2         & powerlaw   \\
0.10 & 2$\times$          & semantic\_knn   & 0.4346 $\pm$ 0.0171            & 0.760 $\pm$ 0.010         & 1007.2 $\pm$ 17.0         & powerlaw   \\
0.10 & 2$\times$          & synthetic       & 0.4749 $\pm$ 0.0179            & 0.712 $\pm$ 0.012         & 941.3 $\pm$ 17.4          & powerlaw   \\
0.10 & 2$\times$          & \textbf{random} & 0.4749 $\pm$ 0.0179            & 0.685 $\pm$ 0.014         & 911.9 $\pm$ 21.6          & powerlaw   \\
0.10 & 2$\times$          & original        & 4.3051 $\pm$ 0.0000            & 0.219 $\pm$ 0.000         & 114.0 $\pm$ 0.0           & powerlaw   \\
\midrule

\end{longtable}

\paragraph{Runtime Analysis}

% ========== RUNTIME STATISTICS ==========
\begin{longtable}{c c l r r}
\caption{Amazon (product--category) GraphSAGE: Runtime Statistics ($M\pm SD$, seconds, $n=32$ seeds). Lower times are better.}
\label{tab:amazon_graphsage_runtime}\\
\toprule
$q$ & $\phi$ & Method & Aug. Time (s) & Train Time (s) \\
\midrule
\endfirsthead
\toprule
$q$ & $\phi$ & Method & Aug. Time (s) & Train Time (s) \\
\midrule
\endhead
\bottomrule
\endfoot

% SOURCE: notebooks/amazon/amazon-GraphSAGE-drop0.99-5x-20251124_145258_runtime.csv
% q=0.01, phi=5x
0.01 & 5$\times$          & baseline        & 0.0000 $\pm$ 0.0000            & 4.08 $\pm$ 1.40           \\
0.01 & 5$\times$          & degree\_aware   & 0.0013 $\pm$ 0.0007            & 3.97 $\pm$ 1.62           \\
0.01 & 5$\times$          & simple          & 0.0012 $\pm$ 0.0009            & 3.65 $\pm$ 1.41           \\
0.01 & 5$\times$          & semantic\_knn   & 0.0117 $\pm$ 0.0067            & 4.11 $\pm$ 1.91           \\
0.01 & 5$\times$          & synthetic       & 0.0011 $\pm$ 0.0004            & 4.55 $\pm$ 1.44           \\
0.01 & 5$\times$          & \textbf{random} & 0.0010 $\pm$ 0.0001            & 3.94 $\pm$ 1.23           \\
0.01 & 5$\times$          & original        & 0.0000 $\pm$ 0.0000            & 133.71 $\pm$ 11.00        \\
\midrule
% SOURCE: notebooks/amazon/amazon-GraphSAGE-drop0.99-2x-20251125_052148_runtime.csv
% q=0.01, phi=2x
0.01 & 2$\times$          & baseline        & 0.0000 $\pm$ 0.0000            & 4.02 $\pm$ 1.36           \\
0.01 & 2$\times$          & degree\_aware   & 0.0013 $\pm$ 0.0008            & 3.64 $\pm$ 1.54           \\
0.01 & 2$\times$          & simple          & 0.0010 $\pm$ 0.0000            & 3.83 $\pm$ 1.72           \\
0.01 & 2$\times$          & semantic\_knn   & 0.0050 $\pm$ 0.0011            & 3.68 $\pm$ 1.29           \\
0.01 & 2$\times$          & synthetic       & 0.0011 $\pm$ 0.0004            & 3.70 $\pm$ 1.51           \\
0.01 & 2$\times$          & \textbf{random} & 0.0009 $\pm$ 0.0000            & 4.00 $\pm$ 1.84           \\
0.01 & 2$\times$          & original        & 0.0000 $\pm$ 0.0000            & 132.37 $\pm$ 11.42        \\
\midrule

% SOURCE: notebooks/amazon/amazon-GraphSAGE-drop0.95-5x-20251124_170227_runtime.csv
% q=0.05, phi=5x
0.05 & 5$\times$          & baseline        & 0.0000 $\pm$ 0.0000            & 6.55 $\pm$ 1.25           \\
0.05 & 5$\times$          & degree\_aware   & 0.0012 $\pm$ 0.0001            & 7.16 $\pm$ 3.75           \\
0.05 & 5$\times$          & \textbf{simple} & 0.0010 $\pm$ 0.0000            & 7.62 $\pm$ 2.02           \\
0.05 & 5$\times$          & semantic\_knn   & 0.0343 $\pm$ 0.0023            & 7.68 $\pm$ 2.54           \\
0.05 & 5$\times$          & synthetic       & 0.0010 $\pm$ 0.0000            & 7.71 $\pm$ 2.24           \\
0.05 & 5$\times$          & random          & 0.0010 $\pm$ 0.0000            & 6.29 $\pm$ 1.14           \\
0.05 & 5$\times$          & original        & 0.0000 $\pm$ 0.0000            & 133.17 $\pm$ 11.53        \\
\midrule
% SOURCE: notebooks/amazon/amazon-GraphSAGE-drop0.95-2x-20251125_052341_runtime.csv
% q=0.05, phi=2x
0.05 & 2$\times$          & baseline        & 0.0000 $\pm$ 0.0000            & 6.72 $\pm$ 1.28           \\
0.05 & 2$\times$          & degree\_aware   & 0.0014 $\pm$ 0.0008            & 7.56 $\pm$ 3.10           \\
0.05 & 2$\times$          & simple          & 0.0011 $\pm$ 0.0001            & 7.42 $\pm$ 2.51           \\
0.05 & 2$\times$          & semantic\_knn   & 0.0224 $\pm$ 0.0082            & 8.51 $\pm$ 3.91           \\
0.05 & 2$\times$          & synthetic       & 0.0014 $\pm$ 0.0016            & 8.11 $\pm$ 2.11           \\
0.05 & 2$\times$          & \textbf{random} & 0.0010 $\pm$ 0.0001            & 8.21 $\pm$ 2.37           \\
0.05 & 2$\times$          & original        & 0.0000 $\pm$ 0.0000            & 134.67 $\pm$ 11.76        \\
\midrule

% SOURCE: notebooks/amazon/amazon-GraphSAGE-drop0.9-5x-20251125_072248_runtime.csv
% q=0.10, phi=5x
0.10 & 5$\times$          & baseline        & 0.0000 $\pm$ 0.0000            & 15.43 $\pm$ 5.57          \\
0.10 & 5$\times$          & degree\_aware   & 0.0013 $\pm$ 0.0001            & 16.61 $\pm$ 6.89          \\
0.10 & 5$\times$          & simple          & 0.0011 $\pm$ 0.0001            & 15.24 $\pm$ 3.81          \\
0.10 & 5$\times$          & semantic\_knn   & 0.0624 $\pm$ 0.0026            & 13.80 $\pm$ 5.13          \\
0.10 & 5$\times$          & synthetic       & 0.0011 $\pm$ 0.0001            & 13.90 $\pm$ 2.43          \\
0.10 & 5$\times$          & \textbf{random} & 0.0010 $\pm$ 0.0001            & 13.15 $\pm$ 1.35          \\
0.10 & 5$\times$          & original        & 0.0000 $\pm$ 0.0000            & 133.62 $\pm$ 11.59        \\
\midrule
% SOURCE: notebooks/amazon/amazon-GraphSAGE-drop0.9-2x-20251125_050449_runtime.csv
% q=0.10, phi=2x
0.10 & 2$\times$          & baseline        & 0.0000 $\pm$ 0.0000            & 15.21 $\pm$ 5.53          \\
0.10 & 2$\times$          & degree\_aware   & 0.0014 $\pm$ 0.0016            & 14.29 $\pm$ 6.33          \\
0.10 & 2$\times$          & \textbf{simple} & 0.0010 $\pm$ 0.0001            & 15.29 $\pm$ 4.83          \\
0.10 & 2$\times$          & semantic\_knn   & 0.0626 $\pm$ 0.0052            & 13.57 $\pm$ 5.02          \\
0.10 & 2$\times$          & synthetic       & 0.0011 $\pm$ 0.0005            & 12.98 $\pm$ 4.72          \\
0.10 & 2$\times$          & random          & 0.0010 $\pm$ 0.0000            & 11.62 $\pm$ 3.01          \\
0.10 & 2$\times$          & original        & 0.0000 $\pm$ 0.0000            & 132.70 $\pm$ 11.33        \\
\midrule

\end{longtable}

\begin{figure}[H]
  \centering
  \includegraphics[width=1\linewidth]{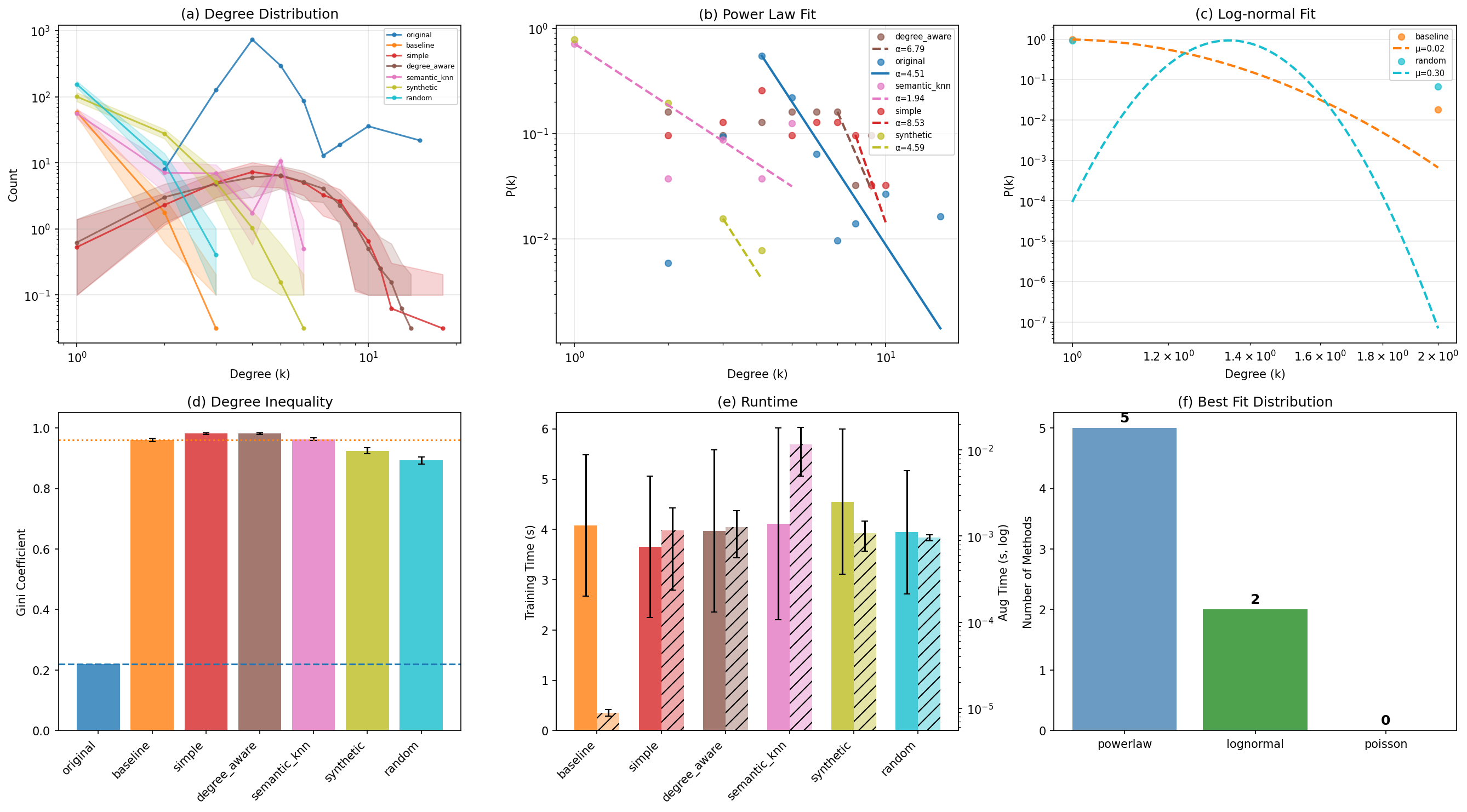}
  \caption{Amazon (product--category), GraphSAGE, $q{=}0.01$, $\phi{=}5$: Comprehensive analysis ($M\pm\mathrm{SD}$, $n=32$ seeds) comparing baseline, augmentation methods, and original graph. Panel (a) shows degree distributions on log-log scale with confidence bands; (b) Power Law fits with exponent $\alpha$; (c) Log-normal fits with parameters $\mu$ and $\sigma$; (d) Gini coefficients quantifying degree inequality (lower = more uniform); (e) runtime comparison showing training time (left axis) and augmentation time (right axis, log scale); (f) best-fit distribution counts across methods.}
  \label{fig:amazon_graphsage_q01_phi5}
\end{figure}

\begin{figure}[H]
  \centering
  \includegraphics[width=1\linewidth]{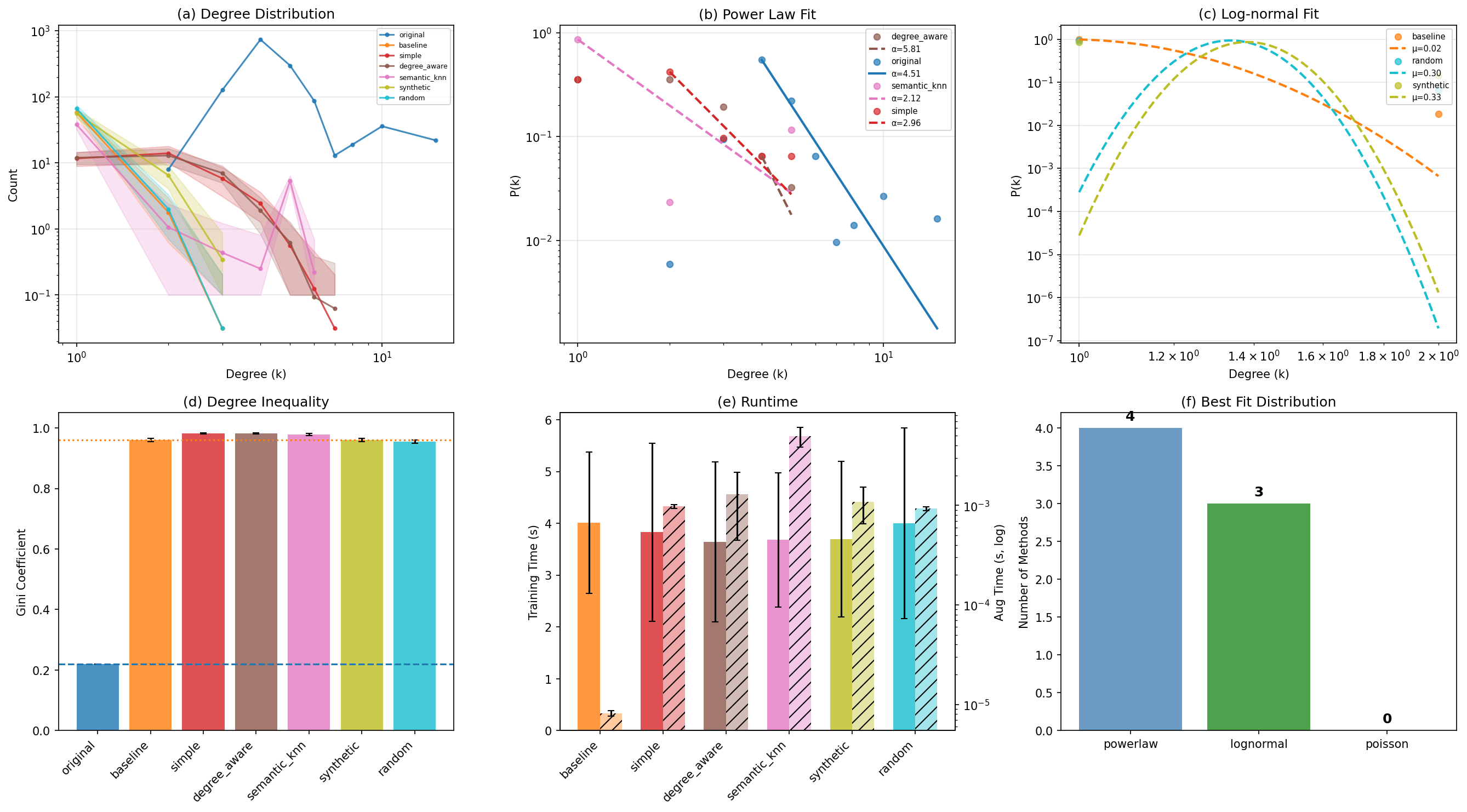}
  \caption{Amazon (product--category), GraphSAGE, $q{=}0.01$, $\phi{=}2$: Comprehensive analysis ($M\pm\mathrm{SD}$, $n=32$ seeds) comparing baseline, augmentation methods, and original graph. Panel (a) shows degree distributions on log-log scale with confidence bands; (b) Power Law fits with exponent $\alpha$; (c) Log-normal fits with parameters $\mu$ and $\sigma$; (d) Gini coefficients quantifying degree inequality (lower = more uniform); (e) runtime comparison showing training time (left axis) and augmentation time (right axis, log scale); (f) best-fit distribution counts across methods.}
  \label{fig:amazon_graphsage_q01_phi2}
\end{figure}

\begin{figure}[H]
  \centering
  \includegraphics[width=1\linewidth]{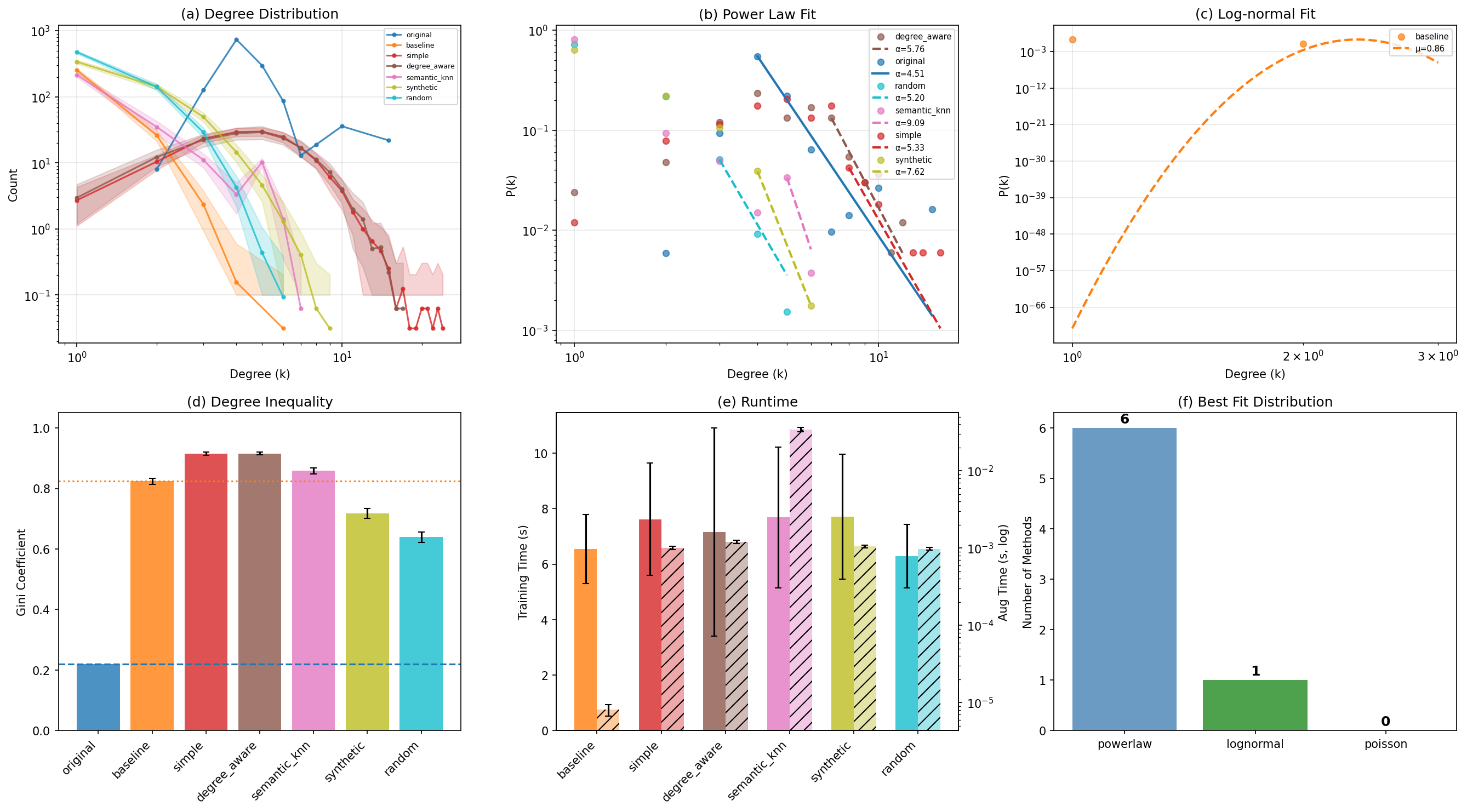}
  \caption{Amazon (product--category), GraphSAGE, $q{=}0.05$, $\phi{=}5$: Comprehensive analysis ($M\pm\mathrm{SD}$, $n=32$ seeds) comparing baseline, augmentation methods, and original graph. Panel (a) shows degree distributions on log-log scale with confidence bands; (b) Power Law fits with exponent $\alpha$; (c) Log-normal fits with parameters $\mu$ and $\sigma$; (d) Gini coefficients quantifying degree inequality (lower = more uniform); (e) runtime comparison showing training time (left axis) and augmentation time (right axis, log scale); (f) best-fit distribution counts across methods.}
  \label{fig:amazon_graphsage_q05_phi5}
\end{figure}

\begin{figure}[H]
  \centering
  \includegraphics[width=1\linewidth]{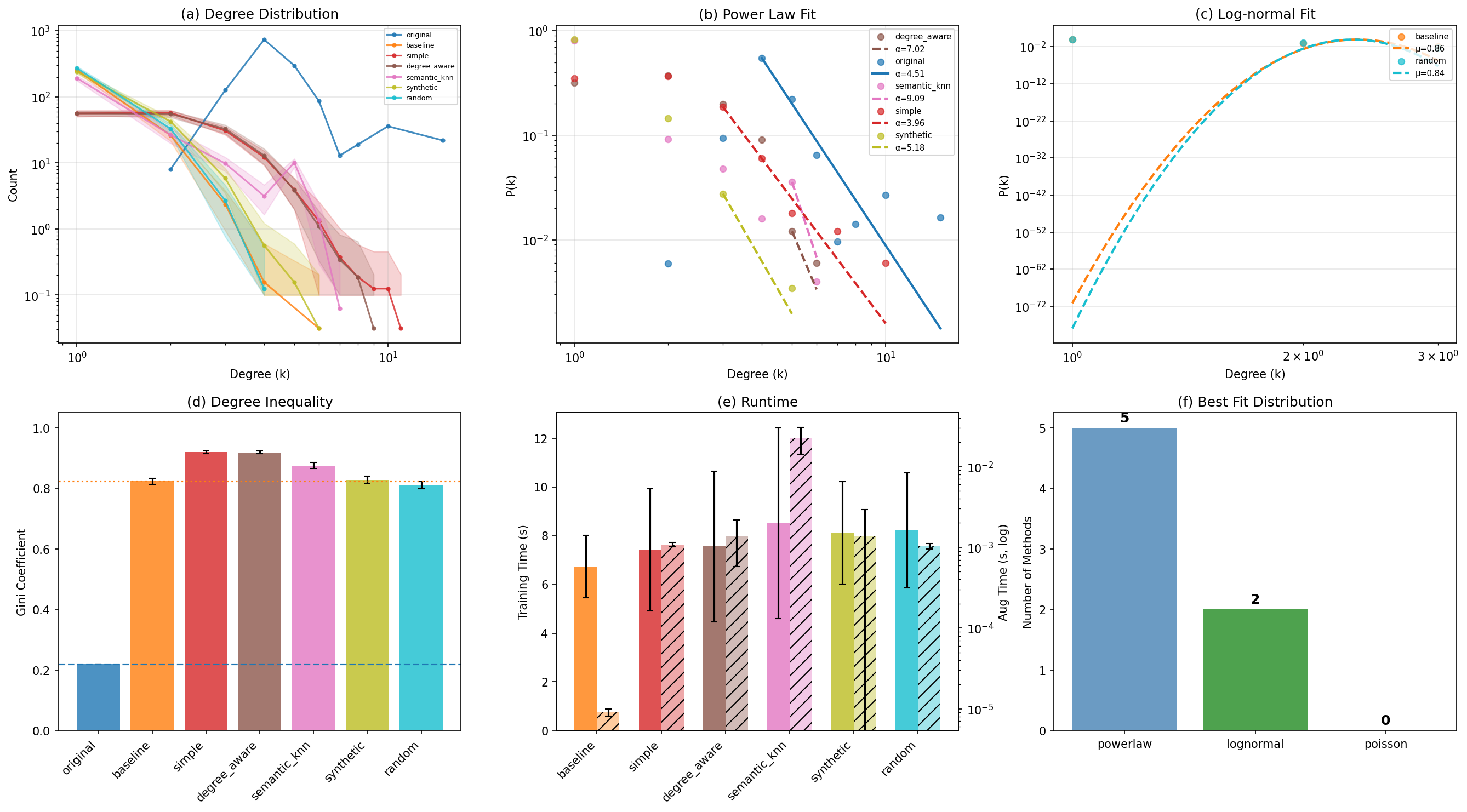}
  \caption{Amazon (product--category), GraphSAGE, $q{=}0.05$, $\phi{=}2$: Comprehensive analysis ($M\pm\mathrm{SD}$, $n=32$ seeds) comparing baseline, augmentation methods, and original graph. Panel (a) shows degree distributions on log-log scale with confidence bands; (b) Power Law fits with exponent $\alpha$; (c) Log-normal fits with parameters $\mu$ and $\sigma$; (d) Gini coefficients quantifying degree inequality (lower = more uniform); (e) runtime comparison showing training time (left axis) and augmentation time (right axis, log scale); (f) best-fit distribution counts across methods.}
  \label{fig:amazon_graphsage_q05_phi2}
\end{figure}

\begin{figure}[H]
  \centering
  \includegraphics[width=1\linewidth]{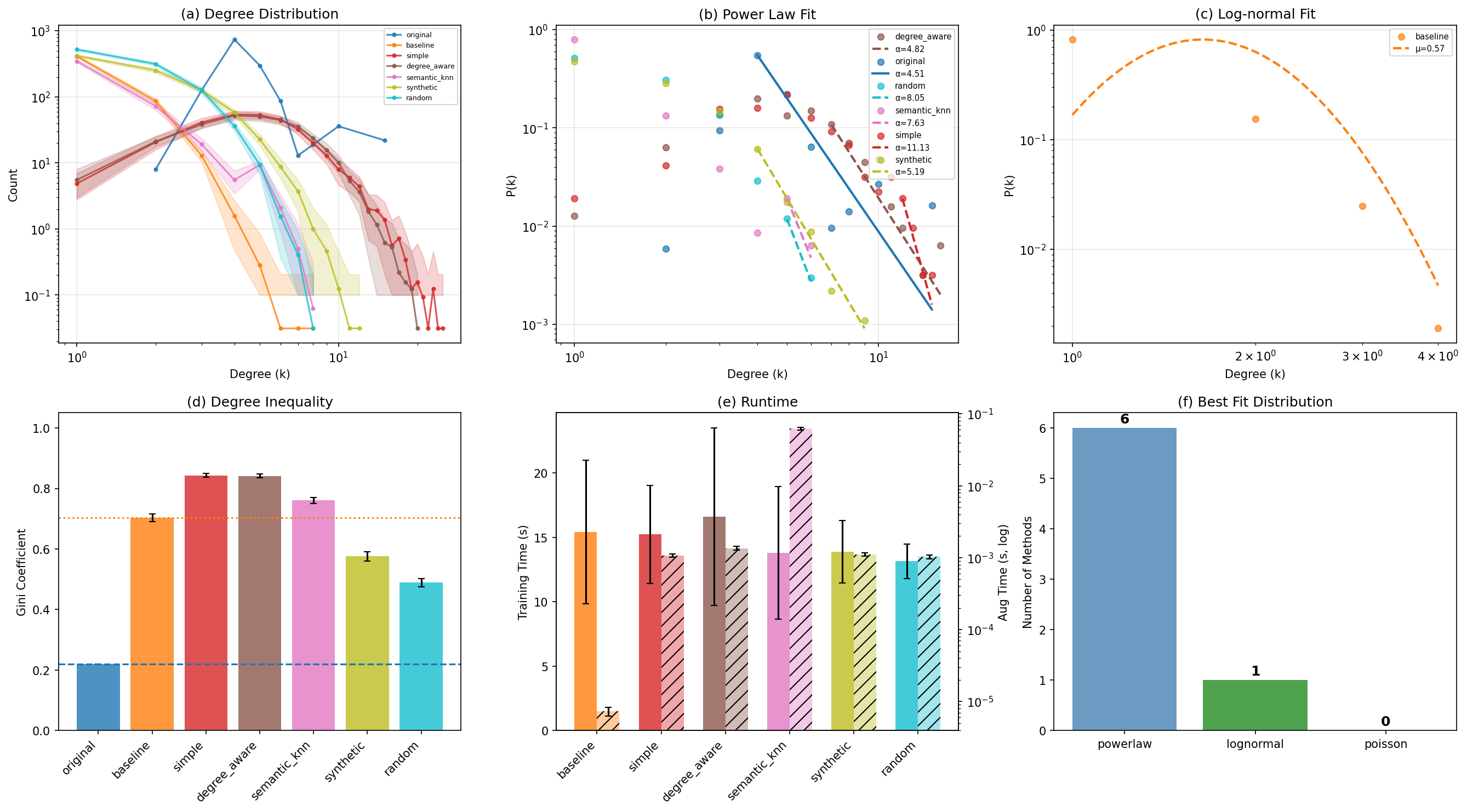}
  \caption{Amazon (product--category), GraphSAGE, $q{=}0.10$, $\phi{=}5$: Comprehensive analysis ($M\pm\mathrm{SD}$, $n=32$ seeds) comparing baseline, augmentation methods, and original graph. Panel (a) shows degree distributions on log-log scale with confidence bands; (b) Power Law fits with exponent $\alpha$; (c) Log-normal fits with parameters $\mu$ and $\sigma$; (d) Gini coefficients quantifying degree inequality (lower = more uniform); (e) runtime comparison showing training time (left axis) and augmentation time (right axis, log scale); (f) best-fit distribution counts across methods.}
  \label{fig:amazon_graphsage_q10_phi5}
\end{figure}

\begin{figure}[H]
  \centering
  \includegraphics[width=1\linewidth]{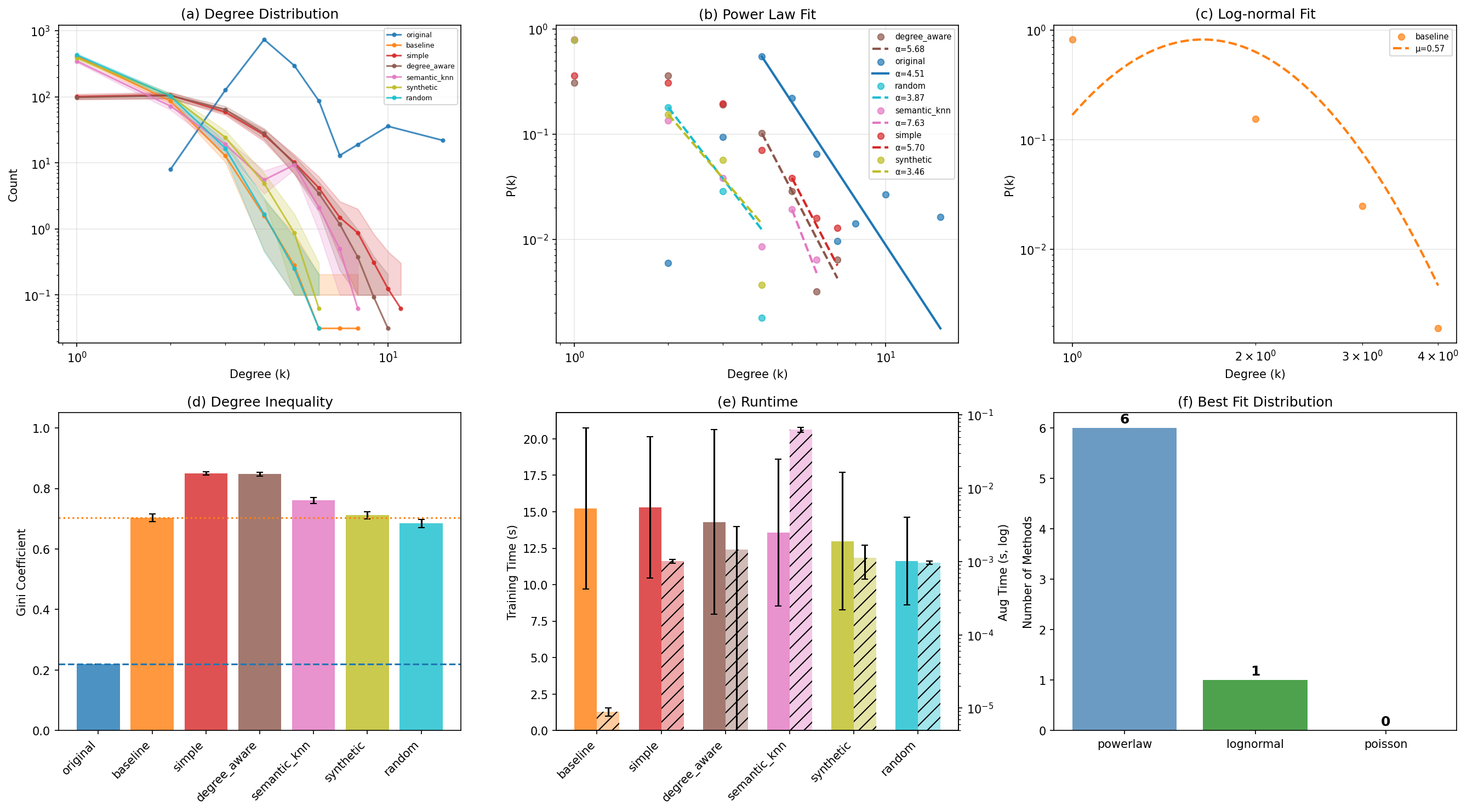}
  \caption{Amazon (product--category), GraphSAGE, $q{=}0.10$, $\phi{=}2$: Comprehensive analysis ($M\pm\mathrm{SD}$, $n=32$ seeds) comparing baseline, augmentation methods, and original graph. Panel (a) shows degree distributions on log-log scale with confidence bands; (b) Power Law fits with exponent $\alpha$; (c) Log-normal fits with parameters $\mu$ and $\sigma$; (d) Gini coefficients quantifying degree inequality (lower = more uniform); (e) runtime comparison showing training time (left axis) and augmentation time (right axis, log scale); (f) best-fit distribution counts across methods.}
  \label{fig:amazon_graphsage_q10_phi2}
\end{figure}

\subsubsection{GCN}

\paragraph{Summary Analysis}

For Amazon GCN, \texttt{semantic\_knn} provides the largest AUC lifts at $q{=}0.05$ ($\phi{=}2$: +0.091; $\phi{=}5$: +0.087), while \texttt{random} and \texttt{simple} often lag (e.g., \texttt{simple} $-0.083$ at $q{=}0.01$, $\phi{=}2$). Brier improves most with \texttt{degree\_aware} at $q{=}0.05$, $\phi{=}5$ ($-0.031$) and with \texttt{simple}/\texttt{random}/\texttt{synthetic} at $q{=}0.10$, $\phi{=}5$ ($-0.023/-0.036/-0.032$); \texttt{semantic\_knn} typically worsens Brier despite AUC gains. Degree distributions move from extreme sparsity (mean degree 0.043, Gini 0.960) to moderately denser graphs (up to $\sim0.59$ at $q{=}0.05$, $\phi{=}5$ and $\sim1.19$ at $q{=}0.10$, $\phi{=}5$), with \texttt{random}/\texttt{synthetic} sharply lowering Gini and isolated nodes. Runtime overheads remain negligible: augmentation $<0.063$s and training within a few seconds of baseline, whereas the dense original costs $\sim219$s, so accuracy changes are attributable to graph quality rather than compute.

% Data source: Table~\ref{tab:amazon_gcn_auc} and corresponding Brier table  \textbf{AUC-ROC Performance:} \texttt{semantic_knn} achieves the highest average AUC improvement (mean $\Delta$AUC=+0.026): at $q=0.05$, $\phi=2x$, AUC=0.592$\pm$0.111 ($\Delta$AUC=+0.091) vs baseline (0.500$\pm$0.102); at $q=0.05$, $\phi=5x$, AUC=0.587$\pm$0.102 ($\Delta$AUC=+0.087). Conversely, \texttt{random} shows the weakest performance (mean $\Delta$AUC=-0.027, range: -0.079 to +0.014).  \textbf{Brier Score Performance:} \texttt{degree_aware} achieves the best Brier score (mean $\Delta$Brier=-0.017). At $q=0.05$, $\phi=5x$, it reaches 0.295$\pm$0.042 ($\Delta$Brier=-0.031) vs baseline (0.326$\pm$0.054).  % SOURCE FILES: % - code-repo/notebooks/amazon/amazon-GCN-drop0.9-2x-20251125_040150_summary.csv % - code-repo/notebooks/amazon/amazon-GCN-drop0.9-5x-20251125_100744_summary.csv % - code-repo/notebooks/amazon/amazon-GCN-drop0.95-2x-20251125_035803_summary.csv % - code-repo/notebooks/amazon/amazon-GCN-drop0.95-5x-20251124_224504_summary.csv % - code-repo/notebooks/amazon/amazon-GCN-drop0.99-2x-20251125_045540_summary.csv % - code-repo/notebooks/amazon/amazon-GCN-drop0.99-5x-20251124_192644_summary.csv

\paragraph{AUC and Brier Score}

% ========== AUC TABLE ==========
\begin{longtable}{c c l r r r r r}
\caption{Amazon (product--category) GCN: AUC-ROC ($M\pm SD$) with paired $t$-tests vs.\ sparse baseline ($n=32$ seeds). A higher AUC is better.}
\label{tab:amazon_gcn_auc}\\
\toprule
$q$ & $\phi$ & Method & AUC $M\pm SD$ & $\Delta$AUC & $t(31)$ & $p$ & $d$ \\
\midrule
\endfirsthead
\toprule
$q$ & $\phi$ & Method & AUC $M\pm SD$ & $\Delta$AUC & $t(31)$ & $p$ & $d$ \\
\midrule
\endhead
\bottomrule
\endfoot

% WARNING: No summary file found for GCN drop0.99 100x
% SOURCE: notebooks/amazon/amazon-GCN-drop0.99-5x-20251124_192644_summary.csv
%         notebooks/amazon/amazon-GCN-drop0.99-5x-paired-ttest-20251124_192644.csv
% q=0.01, phi=5x
0.01 & 5$\times$      & baseline       & 0.478 $\pm$ 0.196 & +0.000 & ---   & ---   & ---   \\
0.01 & 5$\times$      & degree\_aware  & 0.445 $\pm$ 0.181 & -0.033$^{\mathrm{ns}}$ & $0.81$  & 0.425    & $+0.14$ \\
0.01 & 5$\times$      & simple         & 0.446 $\pm$ 0.177 & -0.032$^{\mathrm{ns}}$ & $0.73$  & 0.468    & $+0.13$ \\
0.01 & 5$\times$      & \textbf{semantic\_knn} & 0.481 $\pm$ 0.183 & +0.003$^{\mathrm{ns}}$ & $-0.07$ & 0.946    & $-0.01$ \\
0.01 & 5$\times$      & synthetic      & 0.418 $\pm$ 0.158 & -0.060$^{\mathrm{ns}}$ & $1.35$  & 0.188    & $+0.24$ \\
0.01 & 5$\times$      & random         & 0.399 $\pm$ 0.179 & -0.079$^{\mathrm{ns}}$ & $1.78$  & 0.084    & $+0.32$ \\
0.01 & 5$\times$      & original       & 0.907 $\pm$ 0.015 & +0.430$^{***}$     & $-12.34$ & $<$0.001 & $-2.18$ \\
\midrule
% SOURCE: notebooks/amazon/amazon-GCN-drop0.99-2x-20251125_045540_summary.csv
%         notebooks/amazon/amazon-GCN-drop0.99-2x-paired-ttest-20251125_045540.csv
% q=0.01, phi=2x
0.01 & 2$\times$      & \textbf{baseline} & 0.478 $\pm$ 0.196 & +0.000 & ---   & ---   & ---   \\
0.01 & 2$\times$      & degree\_aware  & 0.465 $\pm$ 0.199 & -0.012$^{\mathrm{ns}}$ & $0.34$  & 0.736    & $+0.06$ \\
0.01 & 2$\times$      & simple         & 0.394 $\pm$ 0.174 & -0.083$^{*}$       & $2.09$  & 0.045    & $+0.37$ \\
0.01 & 2$\times$      & semantic\_knn  & 0.453 $\pm$ 0.184 & -0.024$^{\mathrm{ns}}$ & $0.53$  & 0.601    & $+0.09$ \\
0.01 & 2$\times$      & synthetic      & 0.429 $\pm$ 0.193 & -0.049$^{\mathrm{ns}}$ & $1.22$  & 0.230    & $+0.22$ \\
0.01 & 2$\times$      & random         & 0.454 $\pm$ 0.193 & -0.024$^{\mathrm{ns}}$ & $0.55$  & 0.583    & $+0.10$ \\
0.01 & 2$\times$      & original       & 0.907 $\pm$ 0.015 & +0.430$^{***}$     & $-12.34$ & $<$0.001 & $-2.18$ \\
\midrule

% WARNING: No summary file found for GCN drop0.95 100x
% SOURCE: notebooks/amazon/amazon-GCN-drop0.95-5x-20251124_224504_summary.csv
%         notebooks/amazon/amazon-GCN-drop0.95-5x-paired-ttest-20251124_224504.csv
% q=0.05, phi=5x
0.05 & 5$\times$      & baseline       & 0.500 $\pm$ 0.102 & +0.000 & ---   & ---   & ---   \\
0.05 & 5$\times$      & degree\_aware  & 0.598 $\pm$ 0.116 & +0.098$^{***}$     & $-3.72$ & $<$0.001 & $-0.66$ \\
0.05 & 5$\times$      & \textbf{simple} & 0.611 $\pm$ 0.119 & +0.111$^{***}$     & $-4.79$ & $<$0.001 & $-0.85$ \\
0.05 & 5$\times$      & semantic\_knn  & 0.587 $\pm$ 0.102 & +0.087$^{***}$     & $-4.65$ & $<$0.001 & $-0.82$ \\
0.05 & 5$\times$      & synthetic      & 0.567 $\pm$ 0.100 & +0.066$^{**}$      & $-3.10$ & 0.004    & $-0.55$ \\
0.05 & 5$\times$      & random         & 0.476 $\pm$ 0.087 & -0.024$^{\mathrm{ns}}$ & $1.18$  & 0.247    & $+0.21$ \\
0.05 & 5$\times$      & original       & 0.907 $\pm$ 0.015 & +0.407$^{***}$     & $-21.62$ & $<$0.001 & $-3.82$ \\
\midrule
% SOURCE: notebooks/amazon/amazon-GCN-drop0.95-2x-20251125_035803_summary.csv
%         notebooks/amazon/amazon-GCN-drop0.95-2x-paired-ttest-20251125_035804.csv
% q=0.05, phi=2x
0.05 & 2$\times$      & baseline       & 0.500 $\pm$ 0.102 & +0.000 & ---   & ---   & ---   \\
0.05 & 2$\times$      & degree\_aware  & 0.511 $\pm$ 0.128 & +0.011$^{\mathrm{ns}}$ & $-0.49$ & 0.629    & $-0.09$ \\
0.05 & 2$\times$      & simple         & 0.508 $\pm$ 0.126 & +0.008$^{\mathrm{ns}}$ & $-0.43$ & 0.670    & $-0.08$ \\
0.05 & 2$\times$      & \textbf{semantic\_knn} & 0.592 $\pm$ 0.111 & +0.091$^{***}$     & $-5.15$ & $<$0.001 & $-0.91$ \\
0.05 & 2$\times$      & synthetic      & 0.481 $\pm$ 0.107 & -0.019$^{\mathrm{ns}}$ & $1.07$  & 0.294    & $+0.19$ \\
0.05 & 2$\times$      & random         & 0.480 $\pm$ 0.100 & -0.020$^{\mathrm{ns}}$ & $1.10$  & 0.279    & $+0.19$ \\
0.05 & 2$\times$      & original       & 0.907 $\pm$ 0.015 & +0.407$^{***}$     & $-21.62$ & $<$0.001 & $-3.82$ \\
\midrule

% WARNING: No summary file found for GCN drop0.9 100x
% SOURCE: notebooks/amazon/amazon-GCN-drop0.9-5x-20251125_100744_summary.csv
%         notebooks/amazon/amazon-GCN-drop0.9-5x-paired-ttest-20251125_100744.csv
% q=0.10, phi=5x
0.10 & 5$\times$      & baseline       & 0.604 $\pm$ 0.078 & +0.000 & ---   & ---   & ---   \\
0.10 & 5$\times$      & \textbf{degree\_aware} & 0.648 $\pm$ 0.080 & +0.044$^{**}$      & $-3.10$ & 0.004    & $-0.55$ \\
0.10 & 5$\times$      & simple         & 0.616 $\pm$ 0.083 & +0.013$^{\mathrm{ns}}$ & $-0.66$ & 0.513    & $-0.12$ \\
0.10 & 5$\times$      & semantic\_knn  & 0.602 $\pm$ 0.072 & -0.001$^{\mathrm{ns}}$ & $0.09$  & 0.927    & $+0.02$ \\
0.10 & 5$\times$      & synthetic      & 0.640 $\pm$ 0.092 & +0.036$^{\mathrm{ns}}$ & $-1.87$ & 0.071    & $-0.33$ \\
0.10 & 5$\times$      & random         & 0.617 $\pm$ 0.082 & +0.014$^{\mathrm{ns}}$ & $-0.79$ & 0.438    & $-0.14$ \\
0.10 & 5$\times$      & original       & 0.907 $\pm$ 0.015 & +0.304$^{***}$     & $-21.01$ & $<$0.001 & $-3.71$ \\
\midrule
% SOURCE: notebooks/amazon/amazon-GCN-drop0.9-2x-20251125_040150_summary.csv
%         notebooks/amazon/amazon-GCN-drop0.9-2x-paired-ttest-20251125_040151.csv
% q=0.10, phi=2x
0.10 & 2$\times$      & \textbf{baseline} & 0.604 $\pm$ 0.078 & +0.000 & ---   & ---   & ---   \\
0.10 & 2$\times$      & degree\_aware  & 0.554 $\pm$ 0.081 & -0.050$^{**}$      & $3.31$  & 0.002    & $+0.58$ \\
0.10 & 2$\times$      & simple         & 0.594 $\pm$ 0.089 & -0.009$^{\mathrm{ns}}$ & $0.61$  & 0.544    & $+0.11$ \\
0.10 & 2$\times$      & semantic\_knn  & 0.602 $\pm$ 0.072 & -0.001$^{\mathrm{ns}}$ & $0.09$  & 0.927    & $+0.02$ \\
0.10 & 2$\times$      & synthetic      & 0.578 $\pm$ 0.058 & -0.026$^{\mathrm{ns}}$ & $1.80$  & 0.081    & $+0.32$ \\
0.10 & 2$\times$      & random         & 0.577 $\pm$ 0.073 & -0.027$^{\mathrm{ns}}$ & $1.75$  & 0.091    & $+0.31$ \\
0.10 & 2$\times$      & original       & 0.907 $\pm$ 0.015 & +0.304$^{***}$     & $-21.01$ & $<$0.001 & $-3.71$ \\
\midrule

\end{longtable}

% ========== BRIER TABLE ==========
\begin{longtable}{c c l r r r r r}
\caption{Amazon (product--category) GCN: Brier Score ($M\pm SD$) with paired $t$-tests vs.\ sparse baseline ($n=32$ seeds, lower is better).}
\label{tab:amazon_gcn_brier}\\
\toprule
$q$ & $\phi$ & Method & Brier $M\pm SD$ & $\Delta$Brier & $t(31)$ & $p$ & $d$ \\
\midrule
\endfirsthead
\toprule
$q$ & $\phi$ & Method & Brier $M\pm SD$ & $\Delta$Brier & $t(31)$ & $p$ & $d$ \\
\midrule
\endhead
\bottomrule
\endfoot

% WARNING: No summary file found for GCN drop0.99 100x
% SOURCE: notebooks/amazon/amazon-GCN-drop0.99-5x-20251124_192644_summary.csv
%         notebooks/amazon/amazon-GCN-drop0.99-5x-paired-ttest-20251124_192644.csv
% q=0.01, phi=5x
0.01 & 5$\times$      & baseline       & 0.361 $\pm$ 0.084 & +0.000 & ---  & ---   & ---   \\
0.01 & 5$\times$      & \textbf{degree\_aware} & 0.340 $\pm$ 0.081 & -0.022$^{\mathrm{ns}}$ & $1.56$ & 0.130    & $+0.28$  \\
0.01 & 5$\times$      & simple         & 0.353 $\pm$ 0.071 & -0.008$^{\mathrm{ns}}$ & $0.70$ & 0.492    & $+0.13$  \\
0.01 & 5$\times$      & semantic\_knn  & 0.371 $\pm$ 0.064 & +0.010$^{\mathrm{ns}}$ & $-0.30$ & 0.767    & $-0.05$  \\
0.01 & 5$\times$      & synthetic      & 0.372 $\pm$ 0.052 & +0.011$^{\mathrm{ns}}$ & $-0.38$ & 0.704    & $-0.07$  \\
0.01 & 5$\times$      & random         & 0.378 $\pm$ 0.050 & +0.017$^{\mathrm{ns}}$ & $-0.83$ & 0.411    & $-0.15$  \\
0.01 & 5$\times$      & original       & 0.121 $\pm$ 0.018 & -0.240$^{***}$      & $15.58$ & $<$0.001 & $+2.75$  \\
\midrule
% SOURCE: notebooks/amazon/amazon-GCN-drop0.99-2x-20251125_045540_summary.csv
%         notebooks/amazon/amazon-GCN-drop0.99-2x-paired-ttest-20251125_045540.csv
% q=0.01, phi=2x
0.01 & 2$\times$      & baseline       & 0.361 $\pm$ 0.084 & +0.000 & ---  & ---   & ---   \\
0.01 & 2$\times$      & \textbf{degree\_aware} & 0.351 $\pm$ 0.081 & -0.010$^{\mathrm{ns}}$ & $1.17$ & 0.253    & $+0.21$  \\
0.01 & 2$\times$      & simple         & 0.360 $\pm$ 0.061 & -0.001$^{\mathrm{ns}}$ & $0.38$ & 0.709    & $+0.07$  \\
0.01 & 2$\times$      & semantic\_knn  & 0.369 $\pm$ 0.062 & +0.008$^{\mathrm{ns}}$ & $-0.22$ & 0.826    & $-0.04$  \\
0.01 & 2$\times$      & synthetic      & 0.391 $\pm$ 0.045 & +0.030$^{\mathrm{ns}}$ & $-1.61$ & 0.117    & $-0.29$  \\
0.01 & 2$\times$      & random         & 0.363 $\pm$ 0.061 & +0.002$^{\mathrm{ns}}$ & $0.16$ & 0.875    & $+0.03$  \\
0.01 & 2$\times$      & original       & 0.121 $\pm$ 0.018 & -0.240$^{***}$      & $15.58$ & $<$0.001 & $+2.75$  \\
\midrule

% WARNING: No summary file found for GCN drop0.95 100x
% SOURCE: notebooks/amazon/amazon-GCN-drop0.95-5x-20251124_224504_summary.csv
%         notebooks/amazon/amazon-GCN-drop0.95-5x-paired-ttest-20251124_224504.csv
% q=0.05, phi=5x
0.05 & 5$\times$      & baseline       & 0.326 $\pm$ 0.054 & +0.000 & ---  & ---   & ---   \\
0.05 & 5$\times$      & \textbf{degree\_aware} & 0.295 $\pm$ 0.042 & -0.031$^{**}$       & $2.85$ & 0.008    & $+0.50$  \\
0.05 & 5$\times$      & simple         & 0.302 $\pm$ 0.045 & -0.024$^{*}$        & $2.19$ & 0.036    & $+0.39$  \\
0.05 & 5$\times$      & semantic\_knn  & 0.333 $\pm$ 0.053 & +0.007$^{\mathrm{ns}}$ & $-0.60$ & 0.554    & $-0.11$  \\
0.05 & 5$\times$      & synthetic      & 0.339 $\pm$ 0.045 & +0.012$^{\mathrm{ns}}$ & $-0.97$ & 0.342    & $-0.17$  \\
0.05 & 5$\times$      & random         & 0.324 $\pm$ 0.049 & -0.003$^{\mathrm{ns}}$ & $0.28$ & 0.780    & $+0.05$  \\
0.05 & 5$\times$      & original       & 0.121 $\pm$ 0.018 & -0.205$^{***}$      & $18.98$ & $<$0.001 & $+3.36$  \\
\midrule
% SOURCE: notebooks/amazon/amazon-GCN-drop0.95-2x-20251125_035803_summary.csv
%         notebooks/amazon/amazon-GCN-drop0.95-2x-paired-ttest-20251125_035804.csv
% q=0.05, phi=2x
0.05 & 2$\times$      & baseline       & 0.326 $\pm$ 0.054 & +0.000 & ---  & ---   & ---   \\
0.05 & 2$\times$      & \textbf{degree\_aware} & 0.311 $\pm$ 0.048 & -0.015$^{\mathrm{ns}}$ & $1.23$ & 0.230    & $+0.22$  \\
0.05 & 2$\times$      & simple         & 0.321 $\pm$ 0.050 & -0.005$^{\mathrm{ns}}$ & $0.51$ & 0.615    & $+0.09$  \\
0.05 & 2$\times$      & semantic\_knn  & 0.336 $\pm$ 0.055 & +0.009$^{\mathrm{ns}}$ & $-0.75$ & 0.459    & $-0.13$  \\
0.05 & 2$\times$      & synthetic      & 0.323 $\pm$ 0.045 & -0.003$^{\mathrm{ns}}$ & $0.32$ & 0.755    & $+0.06$  \\
0.05 & 2$\times$      & random         & 0.332 $\pm$ 0.044 & +0.006$^{\mathrm{ns}}$ & $-0.54$ & 0.593    & $-0.10$  \\
0.05 & 2$\times$      & original       & 0.121 $\pm$ 0.018 & -0.205$^{***}$      & $18.98$ & $<$0.001 & $+3.36$  \\
\midrule

% WARNING: No summary file found for GCN drop0.9 100x
% SOURCE: notebooks/amazon/amazon-GCN-drop0.9-5x-20251125_100744_summary.csv
%         notebooks/amazon/amazon-GCN-drop0.9-5x-paired-ttest-20251125_100744.csv
% q=0.10, phi=5x
0.10 & 5$\times$      & baseline       & 0.291 $\pm$ 0.053 & +0.000 & ---  & ---   & ---   \\
0.10 & 5$\times$      & degree\_aware  & 0.273 $\pm$ 0.030 & -0.019$^{\mathrm{ns}}$ & $1.68$ & 0.103    & $+0.30$  \\
0.10 & 5$\times$      & simple         & 0.268 $\pm$ 0.023 & -0.023$^{*}$        & $2.42$ & 0.022    & $+0.43$  \\
0.10 & 5$\times$      & semantic\_knn  & 0.307 $\pm$ 0.050 & +0.015$^{\mathrm{ns}}$ & $-1.31$ & 0.199    & $-0.23$  \\
0.10 & 5$\times$      & synthetic      & 0.260 $\pm$ 0.046 & -0.032$^{**}$       & $2.96$ & 0.006    & $+0.52$  \\
0.10 & 5$\times$      & \textbf{random} & 0.255 $\pm$ 0.035 & -0.036$^{***}$      & $4.28$ & $<$0.001 & $+0.76$  \\
0.10 & 5$\times$      & original       & 0.121 $\pm$ 0.018 & -0.170$^{***}$      & $16.88$ & $<$0.001 & $+2.98$  \\
\midrule
% SOURCE: notebooks/amazon/amazon-GCN-drop0.9-2x-20251125_040150_summary.csv
%         notebooks/amazon/amazon-GCN-drop0.9-2x-paired-ttest-20251125_040151.csv
% q=0.10, phi=2x
0.10 & 2$\times$      & baseline       & 0.291 $\pm$ 0.053 & +0.000 & ---  & ---   & ---   \\
0.10 & 2$\times$      & degree\_aware  & 0.286 $\pm$ 0.028 & -0.006$^{\mathrm{ns}}$ & $0.60$ & 0.556    & $+0.11$  \\
0.10 & 2$\times$      & \textbf{simple} & 0.277 $\pm$ 0.029 & -0.015$^{*}$        & $2.23$ & 0.033    & $+0.39$  \\
0.10 & 2$\times$      & semantic\_knn  & 0.307 $\pm$ 0.050 & +0.015$^{\mathrm{ns}}$ & $-1.31$ & 0.199    & $-0.23$  \\
0.10 & 2$\times$      & synthetic      & 0.289 $\pm$ 0.045 & -0.002$^{\mathrm{ns}}$ & $0.28$ & 0.785    & $+0.05$  \\
0.10 & 2$\times$      & random         & 0.292 $\pm$ 0.051 & +0.001$^{\mathrm{ns}}$ & $-0.08$ & 0.938    & $-0.01$  \\
0.10 & 2$\times$      & original       & 0.121 $\pm$ 0.018 & -0.170$^{***}$      & $16.88$ & $<$0.001 & $+2.98$  \\
\midrule

\end{longtable}

\paragraph{Degree Distribution Analysis}

% ========== DEGREE DISTRIBUTION STATISTICS ==========
\begin{longtable}{c c l r r r l}
\caption{Amazon (product--category) GCN: Degree Distribution Statistics ($M\pm SD$, $n=32$ seeds). Lower Gini coefficient indicates more uniform degree distribution.}
\label{tab:amazon_gcn_degree}\\
\toprule
$q$ & $\phi$ & Method & Mean Degree & Gini Coeff. & Num. Isolated & Best Fit \\
\midrule
\endfirsthead
\toprule
$q$ & $\phi$ & Method & Mean Degree & Gini Coeff. & Num. Isolated & Best Fit \\
\midrule
\endhead
\bottomrule
\endfoot

% SOURCE: notebooks/amazon/amazon-GCN-drop0.99-5x-20251124_192644_degree_stats.csv
%         notebooks/amazon/amazon-GCN-drop0.99-5x-20251124_192644_distribution_fit.csv
% q=0.01, phi=5x
0.01 & 5$\times$          & baseline        & 0.0426 $\pm$ 0.0054            & 0.960 $\pm$ 0.005         & 1404.4 $\pm$ 7.7          & lognormal  \\
0.01 & 5$\times$          & degree\_aware   & 0.1208 $\pm$ 0.0147            & 0.982 $\pm$ 0.002         & 1430.3 $\pm$ 4.4          & powerlaw   \\
0.01 & 5$\times$          & simple          & 0.1208 $\pm$ 0.0147            & 0.982 $\pm$ 0.002         & 1430.3 $\pm$ 4.4          & powerlaw   \\
0.01 & 5$\times$          & semantic\_knn   & 0.1060 $\pm$ 0.0096            & 0.963 $\pm$ 0.005         & 1381.6 $\pm$ 10.1         & powerlaw   \\
0.01 & 5$\times$          & synthetic       & 0.1208 $\pm$ 0.0147            & 0.925 $\pm$ 0.010         & 1329.9 $\pm$ 17.3         & powerlaw   \\
0.01 & 5$\times$          & \textbf{random} & 0.1208 $\pm$ 0.0147            & 0.893 $\pm$ 0.012         & 1298.8 $\pm$ 19.1         & lognormal  \\
0.01 & 5$\times$          & original        & 4.3051 $\pm$ 0.0000            & 0.219 $\pm$ 0.000         & 114.0 $\pm$ 0.0           & powerlaw   \\
\midrule
% SOURCE: notebooks/amazon/amazon-GCN-drop0.99-2x-20251125_045540_degree_stats.csv
%         notebooks/amazon/amazon-GCN-drop0.99-2x-20251125_045540_distribution_fit.csv
% q=0.01, phi=2x
0.01 & 2$\times$          & baseline        & 0.0426 $\pm$ 0.0054            & 0.960 $\pm$ 0.005         & 1404.4 $\pm$ 7.7          & lognormal  \\
0.01 & 2$\times$          & degree\_aware   & 0.0483 $\pm$ 0.0059            & 0.982 $\pm$ 0.002         & 1430.3 $\pm$ 4.4          & powerlaw   \\
0.01 & 2$\times$          & simple          & 0.0483 $\pm$ 0.0059            & 0.982 $\pm$ 0.002         & 1430.3 $\pm$ 4.4          & powerlaw   \\
0.01 & 2$\times$          & semantic\_knn   & 0.0483 $\pm$ 0.0059            & 0.978 $\pm$ 0.003         & 1419.5 $\pm$ 6.0          & powerlaw   \\
0.01 & 2$\times$          & synthetic       & 0.0483 $\pm$ 0.0059            & 0.960 $\pm$ 0.005         & 1401.4 $\pm$ 8.1          & lognormal  \\
0.01 & 2$\times$          & \textbf{random} & 0.0483 $\pm$ 0.0059            & 0.954 $\pm$ 0.006         & 1396.2 $\pm$ 8.4          & lognormal  \\
0.01 & 2$\times$          & original        & 4.3051 $\pm$ 0.0000            & 0.219 $\pm$ 0.000         & 114.0 $\pm$ 0.0           & powerlaw   \\
\midrule

% SOURCE: notebooks/amazon/amazon-GCN-drop0.95-5x-20251124_224504_degree_stats.csv
%         notebooks/amazon/amazon-GCN-drop0.95-5x-20251124_224504_distribution_fit.csv
% q=0.05, phi=5x
0.05 & 5$\times$          & baseline        & 0.2153 $\pm$ 0.0133            & 0.824 $\pm$ 0.011         & 1181.3 $\pm$ 16.7         & lognormal  \\
0.05 & 5$\times$          & degree\_aware   & 0.5930 $\pm$ 0.0365            & 0.916 $\pm$ 0.005         & 1301.9 $\pm$ 9.3          & powerlaw   \\
0.05 & 5$\times$          & simple          & 0.5930 $\pm$ 0.0365            & 0.915 $\pm$ 0.005         & 1301.9 $\pm$ 9.3          & powerlaw   \\
0.05 & 5$\times$          & semantic\_knn   & 0.2682 $\pm$ 0.0140            & 0.858 $\pm$ 0.009         & 1188.1 $\pm$ 16.1         & powerlaw   \\
0.05 & 5$\times$          & synthetic       & 0.5930 $\pm$ 0.0365            & 0.718 $\pm$ 0.016         & 910.9 $\pm$ 31.0          & powerlaw   \\
0.05 & 5$\times$          & \textbf{random} & 0.5930 $\pm$ 0.0365            & 0.639 $\pm$ 0.018         & 811.7 $\pm$ 33.7          & powerlaw   \\
0.05 & 5$\times$          & original        & 4.3051 $\pm$ 0.0000            & 0.219 $\pm$ 0.000         & 114.0 $\pm$ 0.0           & powerlaw   \\
\midrule
% SOURCE: notebooks/amazon/amazon-GCN-drop0.95-2x-20251125_035803_degree_stats.csv
%         notebooks/amazon/amazon-GCN-drop0.95-2x-20251125_035803_distribution_fit.csv
% q=0.05, phi=2x
0.05 & 2$\times$          & baseline        & 0.2153 $\pm$ 0.0133            & 0.824 $\pm$ 0.011         & 1181.3 $\pm$ 16.7         & lognormal  \\
0.05 & 2$\times$          & degree\_aware   & 0.2372 $\pm$ 0.0146            & 0.919 $\pm$ 0.005         & 1301.9 $\pm$ 9.3          & powerlaw   \\
0.05 & 2$\times$          & simple          & 0.2372 $\pm$ 0.0146            & 0.920 $\pm$ 0.004         & 1301.9 $\pm$ 9.3          & powerlaw   \\
0.05 & 2$\times$          & semantic\_knn   & 0.2372 $\pm$ 0.0146            & 0.876 $\pm$ 0.010         & 1221.0 $\pm$ 17.0         & powerlaw   \\
0.05 & 2$\times$          & synthetic       & 0.2372 $\pm$ 0.0146            & 0.829 $\pm$ 0.011         & 1174.2 $\pm$ 17.8         & powerlaw   \\
0.05 & 2$\times$          & \textbf{random} & 0.2372 $\pm$ 0.0146            & 0.810 $\pm$ 0.012         & 1156.3 $\pm$ 18.2         & lognormal  \\
0.05 & 2$\times$          & original        & 4.3051 $\pm$ 0.0000            & 0.219 $\pm$ 0.000         & 114.0 $\pm$ 0.0           & powerlaw   \\
\midrule

% SOURCE: notebooks/amazon/amazon-GCN-drop0.9-5x-20251125_100744_degree_stats.csv
%         notebooks/amazon/amazon-GCN-drop0.9-5x-20251125_100744_distribution_fit.csv
% q=0.10, phi=5x
0.10 & 5$\times$          & baseline        & 0.4323 $\pm$ 0.0163            & 0.703 $\pm$ 0.012         & 950.6 $\pm$ 19.3          & lognormal  \\
0.10 & 5$\times$          & degree\_aware   & 1.1873 $\pm$ 0.0446            & 0.841 $\pm$ 0.006         & 1154.6 $\pm$ 11.2         & powerlaw   \\
0.10 & 5$\times$          & simple          & 1.1873 $\pm$ 0.0446            & 0.843 $\pm$ 0.006         & 1154.6 $\pm$ 11.2         & powerlaw   \\
0.10 & 5$\times$          & semantic\_knn   & 0.4346 $\pm$ 0.0171            & 0.760 $\pm$ 0.010         & 1007.2 $\pm$ 17.0         & powerlaw   \\
0.10 & 5$\times$          & synthetic       & 1.1873 $\pm$ 0.0446            & 0.576 $\pm$ 0.015         & 572.2 $\pm$ 27.0          & powerlaw   \\
0.10 & 5$\times$          & \textbf{random} & 1.1873 $\pm$ 0.0446            & 0.489 $\pm$ 0.013         & 450.4 $\pm$ 26.1          & powerlaw   \\
0.10 & 5$\times$          & original        & 4.3051 $\pm$ 0.0000            & 0.219 $\pm$ 0.000         & 114.0 $\pm$ 0.0           & powerlaw   \\
\midrule
% SOURCE: notebooks/amazon/amazon-GCN-drop0.9-2x-20251125_040150_degree_stats.csv
%         notebooks/amazon/amazon-GCN-drop0.9-2x-20251125_040150_distribution_fit.csv
% q=0.10, phi=2x
0.10 & 2$\times$          & baseline        & 0.4323 $\pm$ 0.0163            & 0.703 $\pm$ 0.012         & 950.6 $\pm$ 19.3          & lognormal  \\
0.10 & 2$\times$          & degree\_aware   & 0.4749 $\pm$ 0.0179            & 0.848 $\pm$ 0.006         & 1154.6 $\pm$ 11.2         & powerlaw   \\
0.10 & 2$\times$          & simple          & 0.4749 $\pm$ 0.0179            & 0.850 $\pm$ 0.006         & 1154.6 $\pm$ 11.2         & powerlaw   \\
0.10 & 2$\times$          & semantic\_knn   & 0.4346 $\pm$ 0.0171            & 0.760 $\pm$ 0.010         & 1007.2 $\pm$ 17.0         & powerlaw   \\
0.10 & 2$\times$          & synthetic       & 0.4749 $\pm$ 0.0179            & 0.712 $\pm$ 0.012         & 941.3 $\pm$ 17.4          & powerlaw   \\
0.10 & 2$\times$          & \textbf{random} & 0.4749 $\pm$ 0.0179            & 0.685 $\pm$ 0.014         & 911.9 $\pm$ 21.6          & powerlaw   \\
0.10 & 2$\times$          & original        & 4.3051 $\pm$ 0.0000            & 0.219 $\pm$ 0.000         & 114.0 $\pm$ 0.0           & powerlaw   \\
\midrule

\end{longtable}

% ========== DEGREE DISTRIBUTION FIGURES ==========
% Insert degree distribution plots at appropriate locations:
% Use \includegraphics command with these paths:
%
% Figure for q=0.01, phi=5x:
%   notebooks/amazon/amazon-GCN-drop0.99-5x-20251124_192644_degree_distribution.png
% Figure for q=0.01, phi=2x:
%   notebooks/amazon/amazon-GCN-drop0.99-2x-20251125_045540_degree_distribution.png
% Figure for q=0.05, phi=5x:
%   notebooks/amazon/amazon-GCN-drop0.95-5x-20251124_224504_degree_distribution.png
% Figure for q=0.05, phi=2x:
%   notebooks/amazon/amazon-GCN-drop0.95-2x-20251125_035803_degree_distribution.png
% Figure for q=0.10, phi=5x:
%   notebooks/amazon/amazon-GCN-drop0.9-5x-20251125_100744_degree_distribution.png
% Figure for q=0.10, phi=2x:
%   notebooks/amazon/amazon-GCN-drop0.9-2x-20251125_040150_degree_distribution.png
%

\paragraph{Runtime Analysis}

% ========== RUNTIME STATISTICS ==========
\begin{longtable}{c c l r r}
\caption{Amazon (product--category) GCN: Runtime Statistics ($M\pm SD$, seconds, $n=32$ seeds). Lower times are better.}
\label{tab:amazon_gcn_runtime}\\
\toprule
$q$ & $\phi$ & Method & Aug. Time (s) & Train Time (s) \\
\midrule
\endfirsthead
\toprule
$q$ & $\phi$ & Method & Aug. Time (s) & Train Time (s) \\
\midrule
\endhead
\bottomrule
\endfoot

% SOURCE: notebooks/amazon/amazon-GCN-drop0.99-5x-20251124_192644_runtime.csv
% q=0.01, phi=5x
0.01 & 5$\times$          & baseline        & 0.0000 $\pm$ 0.0000            & 3.72 $\pm$ 1.07           \\
0.01 & 5$\times$          & degree\_aware   & 0.0012 $\pm$ 0.0001            & 3.36 $\pm$ 1.13           \\
0.01 & 5$\times$          & \textbf{simple} & 0.0010 $\pm$ 0.0001            & 3.39 $\pm$ 1.17           \\
0.01 & 5$\times$          & semantic\_knn   & 0.0108 $\pm$ 0.0012            & 3.89 $\pm$ 1.70           \\
0.01 & 5$\times$          & synthetic       & 0.0010 $\pm$ 0.0001            & 3.88 $\pm$ 1.61           \\
0.01 & 5$\times$          & random          & 0.0010 $\pm$ 0.0001            & 3.73 $\pm$ 1.25           \\
0.01 & 5$\times$          & original        & 0.0000 $\pm$ 0.0000            & 219.23 $\pm$ 37.81        \\
\midrule
% SOURCE: notebooks/amazon/amazon-GCN-drop0.99-2x-20251125_045540_runtime.csv
% q=0.01, phi=2x
0.01 & 2$\times$          & baseline        & 0.0000 $\pm$ 0.0000            & 3.72 $\pm$ 1.07           \\
0.01 & 2$\times$          & degree\_aware   & 0.0012 $\pm$ 0.0001            & 3.73 $\pm$ 1.78           \\
0.01 & 2$\times$          & simple          & 0.0010 $\pm$ 0.0001            & 3.45 $\pm$ 1.68           \\
0.01 & 2$\times$          & semantic\_knn   & 0.0049 $\pm$ 0.0009            & 3.64 $\pm$ 1.51           \\
0.01 & 2$\times$          & synthetic       & 0.0010 $\pm$ 0.0001            & 3.79 $\pm$ 1.47           \\
0.01 & 2$\times$          & \textbf{random} & 0.0009 $\pm$ 0.0001            & 3.33 $\pm$ 1.08           \\
0.01 & 2$\times$          & original        & 0.0000 $\pm$ 0.0000            & 220.21 $\pm$ 38.20        \\
\midrule

% SOURCE: notebooks/amazon/amazon-GCN-drop0.95-5x-20251124_224504_runtime.csv
% q=0.05, phi=5x
0.05 & 5$\times$          & baseline        & 0.0000 $\pm$ 0.0000            & 7.02 $\pm$ 2.10           \\
0.05 & 5$\times$          & degree\_aware   & 0.0014 $\pm$ 0.0009            & 6.50 $\pm$ 1.43           \\
0.05 & 5$\times$          & \textbf{simple} & 0.0010 $\pm$ 0.0000            & 7.05 $\pm$ 2.29           \\
0.05 & 5$\times$          & semantic\_knn   & 0.0340 $\pm$ 0.0028            & 9.95 $\pm$ 3.79           \\
0.05 & 5$\times$          & synthetic       & 0.0012 $\pm$ 0.0006            & 10.06 $\pm$ 3.76          \\
0.05 & 5$\times$          & random          & 0.0010 $\pm$ 0.0000            & 8.04 $\pm$ 3.33           \\
0.05 & 5$\times$          & original        & 0.0000 $\pm$ 0.0000            & 218.97 $\pm$ 37.68        \\
\midrule
% SOURCE: notebooks/amazon/amazon-GCN-drop0.95-2x-20251125_035803_runtime.csv
% q=0.05, phi=2x
0.05 & 2$\times$          & baseline        & 0.0000 $\pm$ 0.0000            & 7.06 $\pm$ 2.10           \\
0.05 & 2$\times$          & degree\_aware   & 0.0012 $\pm$ 0.0005            & 6.55 $\pm$ 1.70           \\
0.05 & 2$\times$          & \textbf{simple} & 0.0010 $\pm$ 0.0000            & 6.92 $\pm$ 3.42           \\
0.05 & 2$\times$          & semantic\_knn   & 0.0217 $\pm$ 0.0073            & 8.86 $\pm$ 3.56           \\
0.05 & 2$\times$          & synthetic       & 0.0011 $\pm$ 0.0004            & 7.50 $\pm$ 2.72           \\
0.05 & 2$\times$          & random          & 0.0010 $\pm$ 0.0000            & 8.40 $\pm$ 2.97           \\
0.05 & 2$\times$          & original        & 0.0000 $\pm$ 0.0000            & 219.24 $\pm$ 37.79        \\
\midrule

% SOURCE: notebooks/amazon/amazon-GCN-drop0.9-5x-20251125_100744_runtime.csv
% q=0.10, phi=5x
0.10 & 5$\times$          & baseline        & 0.0000 $\pm$ 0.0000            & 14.04 $\pm$ 7.23          \\
0.10 & 5$\times$          & degree\_aware   & 0.0013 $\pm$ 0.0001            & 14.82 $\pm$ 4.01          \\
0.10 & 5$\times$          & \textbf{simple} & 0.0010 $\pm$ 0.0001            & 14.02 $\pm$ 5.16          \\
0.10 & 5$\times$          & semantic\_knn   & 0.0624 $\pm$ 0.0028            & 16.59 $\pm$ 7.88          \\
0.10 & 5$\times$          & synthetic       & 0.0011 $\pm$ 0.0001            & 14.54 $\pm$ 3.93          \\
0.10 & 5$\times$          & random          & 0.0010 $\pm$ 0.0001            & 13.90 $\pm$ 5.33          \\
0.10 & 5$\times$          & original        & 0.0000 $\pm$ 0.0000            & 219.50 $\pm$ 37.87        \\
\midrule
% SOURCE: notebooks/amazon/amazon-GCN-drop0.9-2x-20251125_040150_runtime.csv
% q=0.10, phi=2x
0.10 & 2$\times$          & baseline        & 0.0000 $\pm$ 0.0000            & 13.91 $\pm$ 7.15          \\
0.10 & 2$\times$          & degree\_aware   & 0.0012 $\pm$ 0.0001            & 11.99 $\pm$ 4.25          \\
0.10 & 2$\times$          & simple          & 0.0011 $\pm$ 0.0001            & 11.94 $\pm$ 5.13          \\
0.10 & 2$\times$          & semantic\_knn   & 0.0615 $\pm$ 0.0026            & 16.35 $\pm$ 7.63          \\
0.10 & 2$\times$          & synthetic       & 0.0011 $\pm$ 0.0001            & 12.52 $\pm$ 3.77          \\
0.10 & 2$\times$          & \textbf{random} & 0.0010 $\pm$ 0.0001            & 13.85 $\pm$ 6.61          \\
0.10 & 2$\times$          & original        & 0.0000 $\pm$ 0.0000            & 218.81 $\pm$ 38.02        \\
\midrule

\end{longtable}

\begin{figure}[H]
  \centering
  \includegraphics[width=1\linewidth]{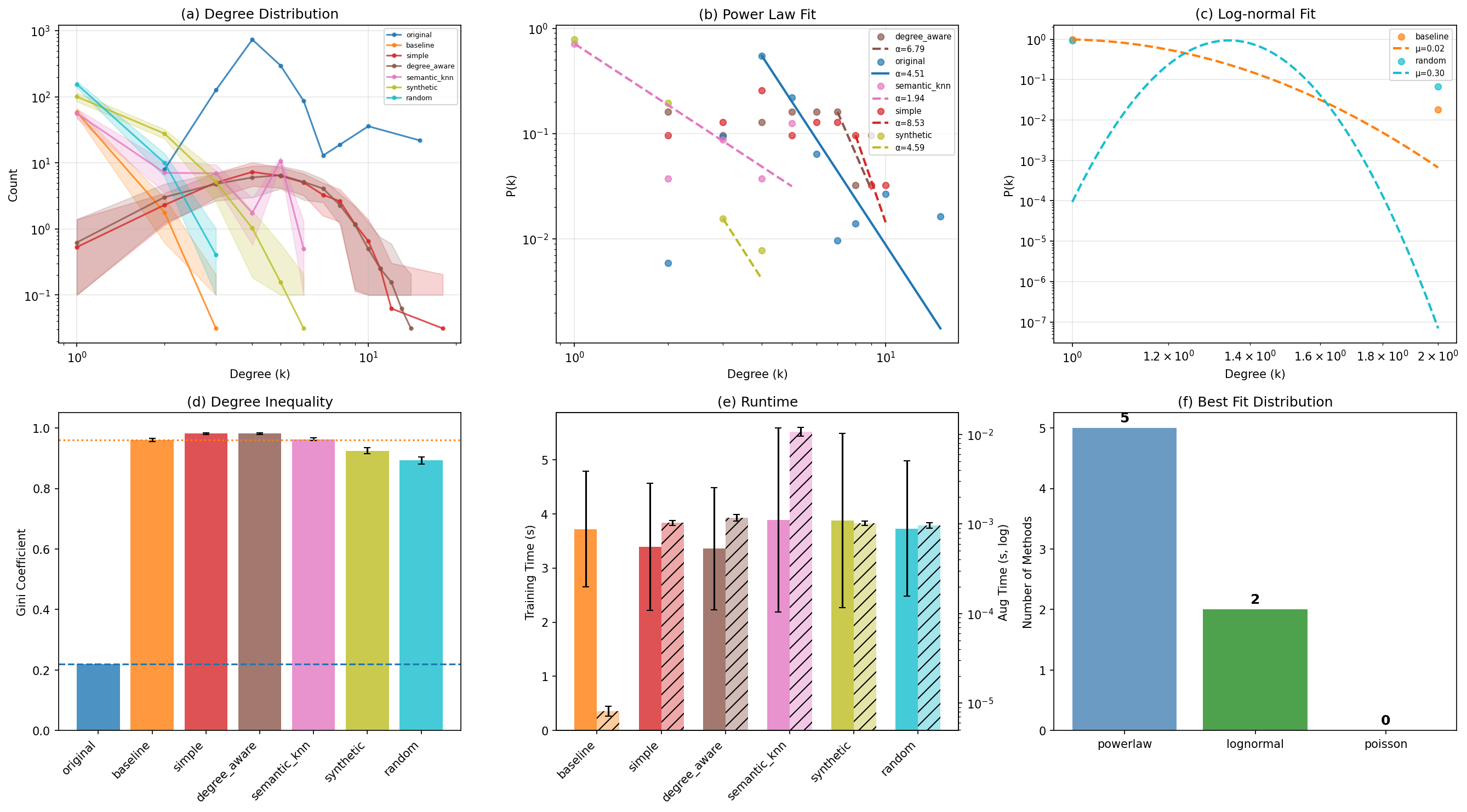}
  \caption{Amazon (product--category), GCN, $q{=}0.01$, $\phi{=}5$: Comprehensive analysis ($M\pm\mathrm{SD}$, $n=32$ seeds) comparing baseline, augmentation methods, and original graph. Panel (a) shows degree distributions on log-log scale with confidence bands; (b) Power Law fits with exponent $\alpha$; (c) Log-normal fits with parameters $\mu$ and $\sigma$; (d) Gini coefficients quantifying degree inequality (lower = more uniform); (e) runtime comparison showing training time (left axis) and augmentation time (right axis, log scale); (f) best-fit distribution counts across methods.}
  \label{fig:amazon_gcn_q01_phi5}
\end{figure}

\begin{figure}[H]
  \centering
  \includegraphics[width=1\linewidth]{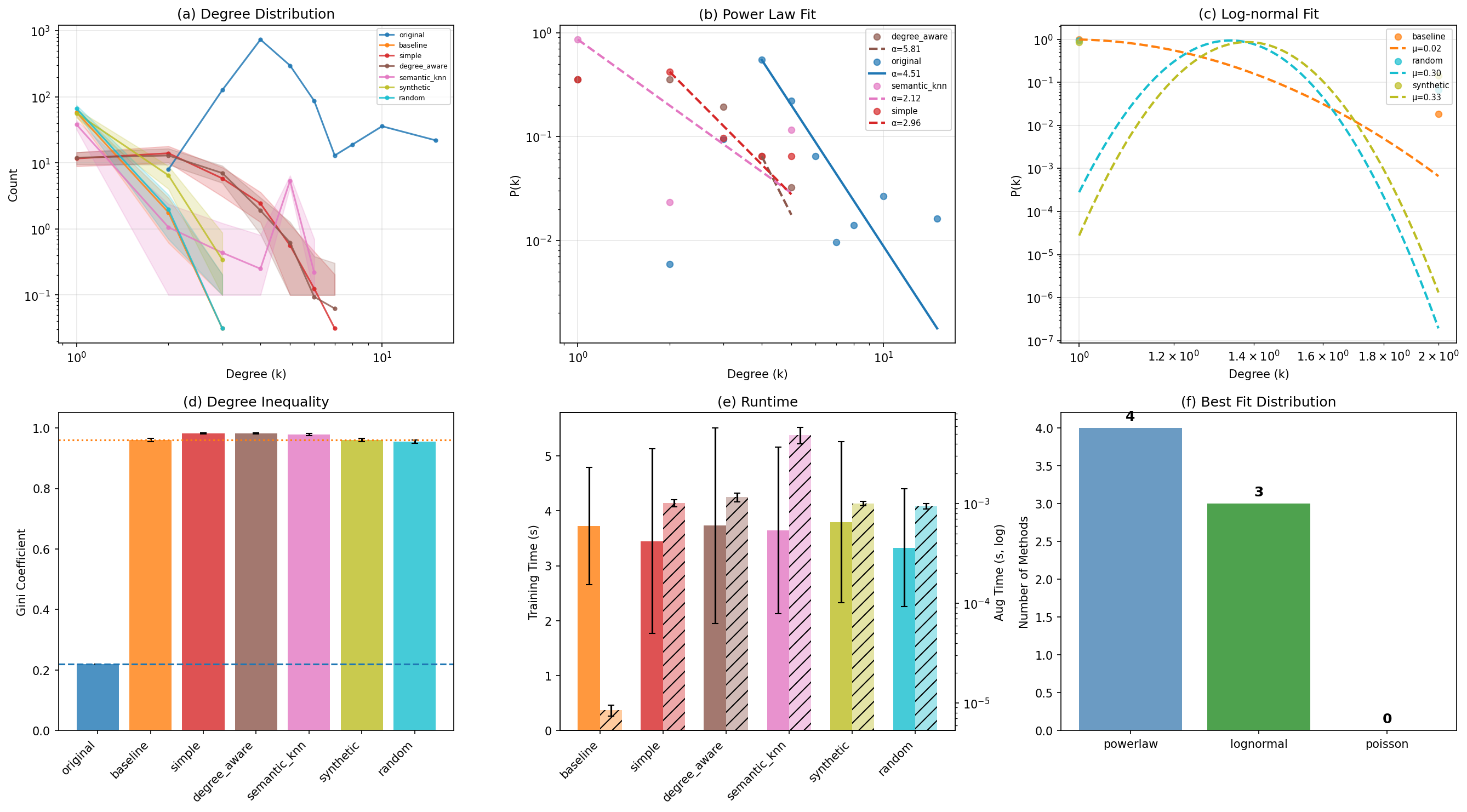}
  \caption{Amazon (product--category), GCN, $q{=}0.01$, $\phi{=}2$: Comprehensive analysis ($M\pm\mathrm{SD}$, $n=32$ seeds) comparing baseline, augmentation methods, and original graph. Panel (a) shows degree distributions on log-log scale with confidence bands; (b) Power Law fits with exponent $\alpha$; (c) Log-normal fits with parameters $\mu$ and $\sigma$; (d) Gini coefficients quantifying degree inequality (lower = more uniform); (e) runtime comparison showing training time (left axis) and augmentation time (right axis, log scale); (f) best-fit distribution counts across methods.}
  \label{fig:amazon_gcn_q01_phi2}
\end{figure}

\begin{figure}[H]
  \centering
  \includegraphics[width=1\linewidth]{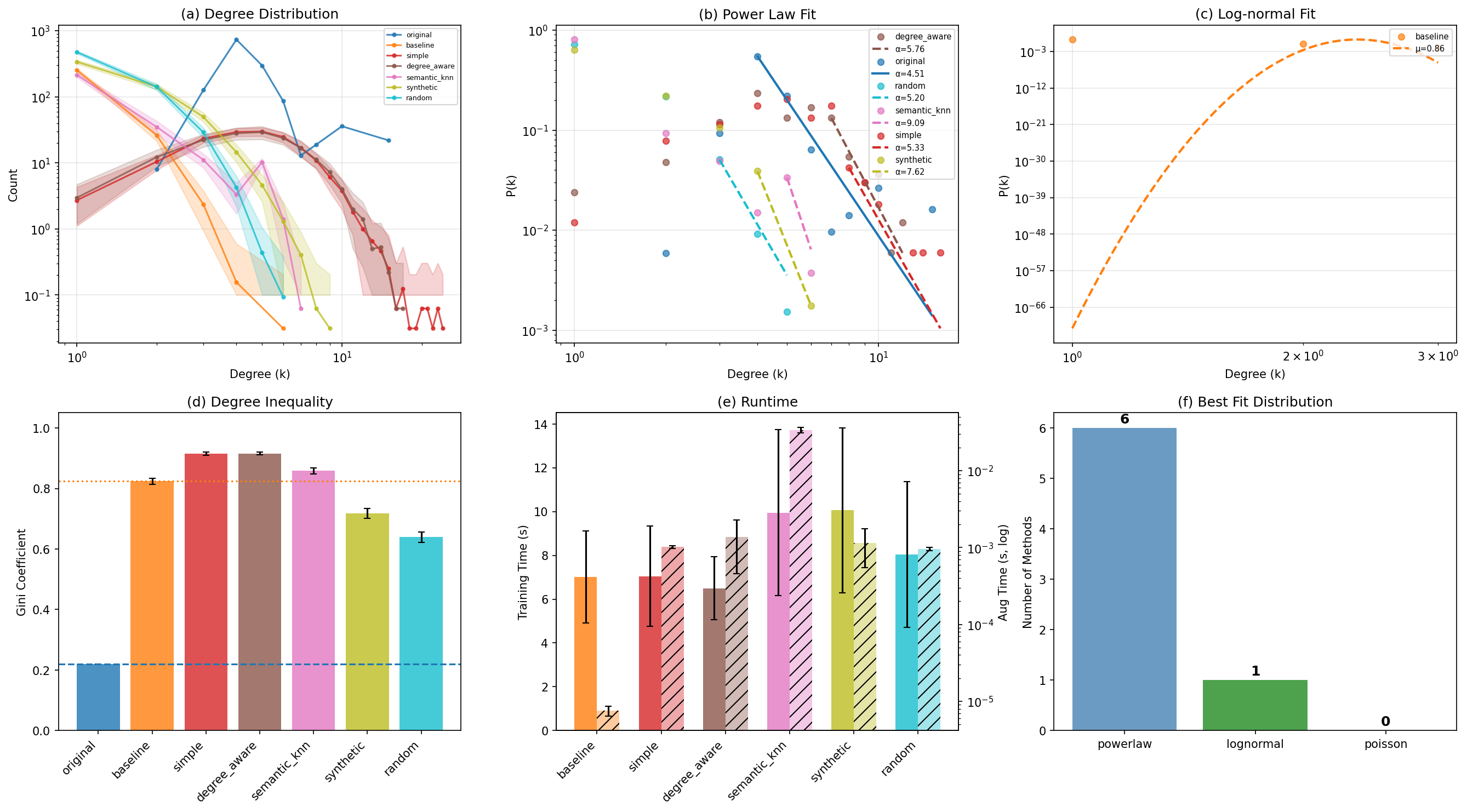}
  \caption{Amazon (product--category), GCN, $q{=}0.05$, $\phi{=}5$: Comprehensive analysis ($M\pm\mathrm{SD}$, $n=32$ seeds) comparing baseline, augmentation methods, and original graph. Panel (a) shows degree distributions on log-log scale with confidence bands; (b) Power Law fits with exponent $\alpha$; (c) Log-normal fits with parameters $\mu$ and $\sigma$; (d) Gini coefficients quantifying degree inequality (lower = more uniform); (e) runtime comparison showing training time (left axis) and augmentation time (right axis, log scale); (f) best-fit distribution counts across methods.}
  \label{fig:amazon_gcn_q05_phi5}
\end{figure}

\begin{figure}[H]
  \centering
  \includegraphics[width=1\linewidth]{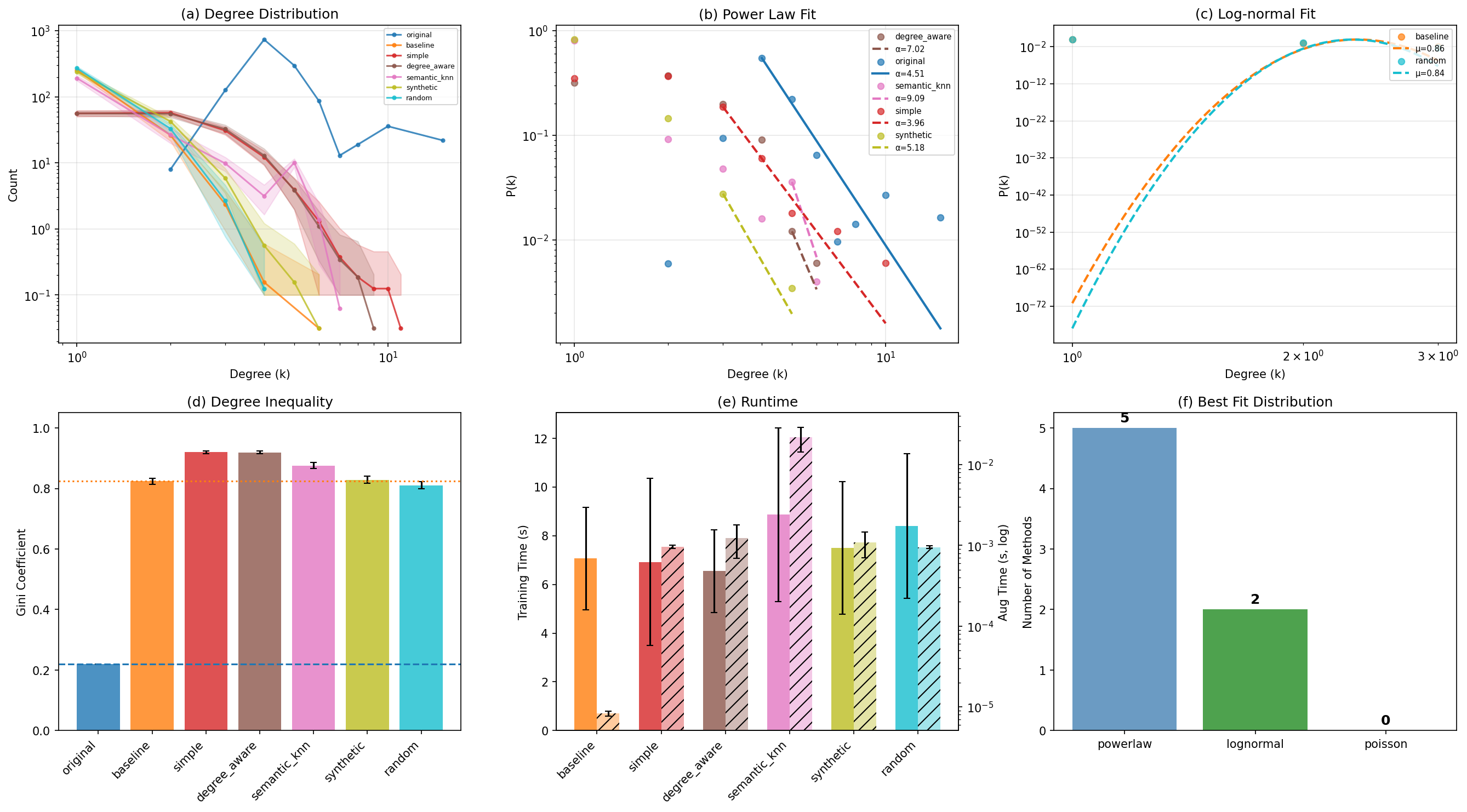}
  \caption{Amazon (product--category), GCN, $q{=}0.05$, $\phi{=}2$: Comprehensive analysis ($M\pm\mathrm{SD}$, $n=32$ seeds) comparing baseline, augmentation methods, and original graph. Panel (a) shows degree distributions on log-log scale with confidence bands; (b) Power Law fits with exponent $\alpha$; (c) Log-normal fits with parameters $\mu$ and $\sigma$; (d) Gini coefficients quantifying degree inequality (lower = more uniform); (e) runtime comparison showing training time (left axis) and augmentation time (right axis, log scale); (f) best-fit distribution counts across methods.}
  \label{fig:amazon_gcn_q05_phi2}
\end{figure}

\begin{figure}[H]
  \centering
  \includegraphics[width=1\linewidth]{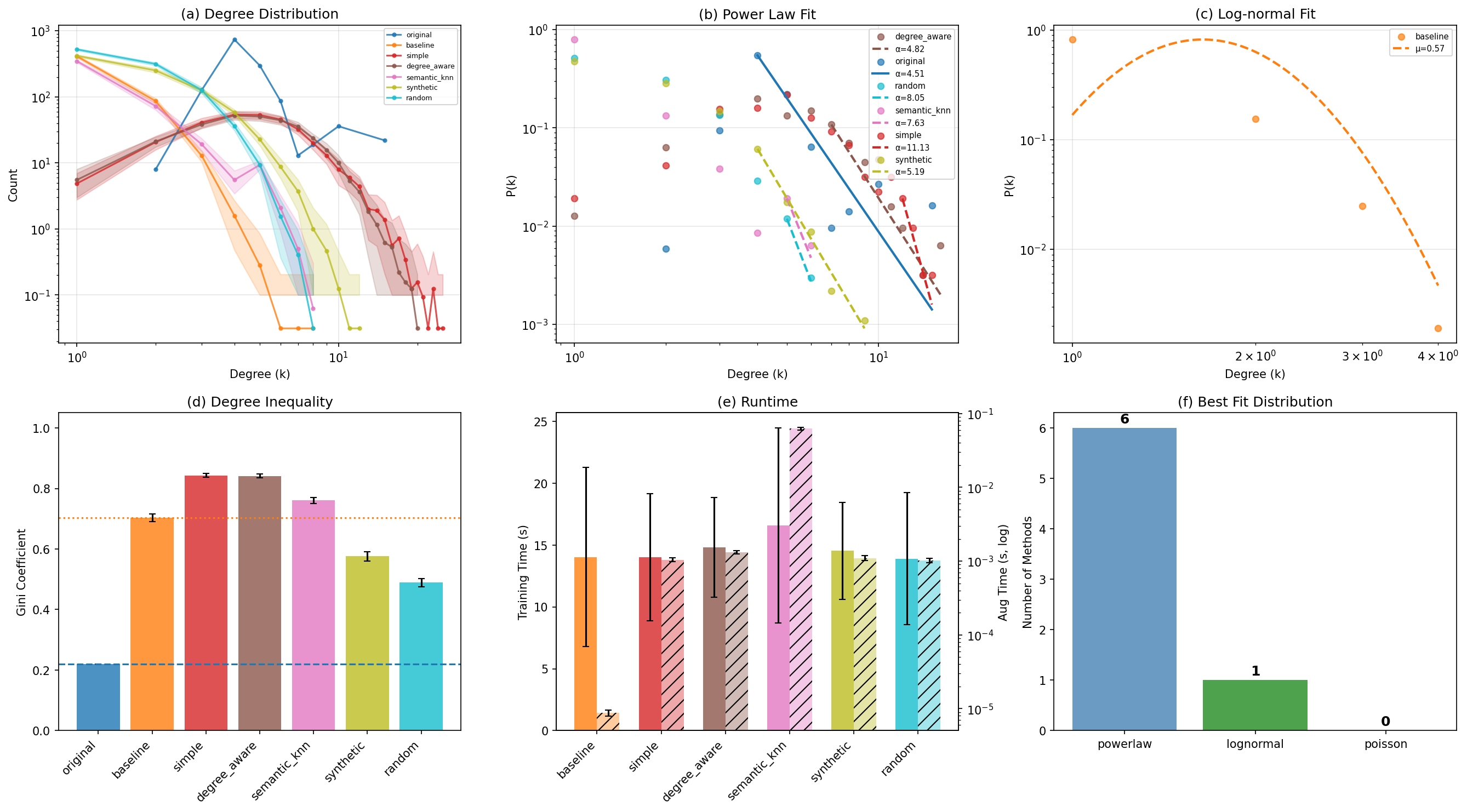}
  \caption{Amazon (product--category), GCN, $q{=}0.10$, $\phi{=}5$: Comprehensive analysis ($M\pm\mathrm{SD}$, $n=32$ seeds) comparing baseline, augmentation methods, and original graph. Panel (a) shows degree distributions on log-log scale with confidence bands; (b) Power Law fits with exponent $\alpha$; (c) Log-normal fits with parameters $\mu$ and $\sigma$; (d) Gini coefficients quantifying degree inequality (lower = more uniform); (e) runtime comparison showing training time (left axis) and augmentation time (right axis, log scale); (f) best-fit distribution counts across methods.}
  \label{fig:amazon_gcn_q10_phi5}
\end{figure}

\begin{figure}[H]
  \centering
  \includegraphics[width=1\linewidth]{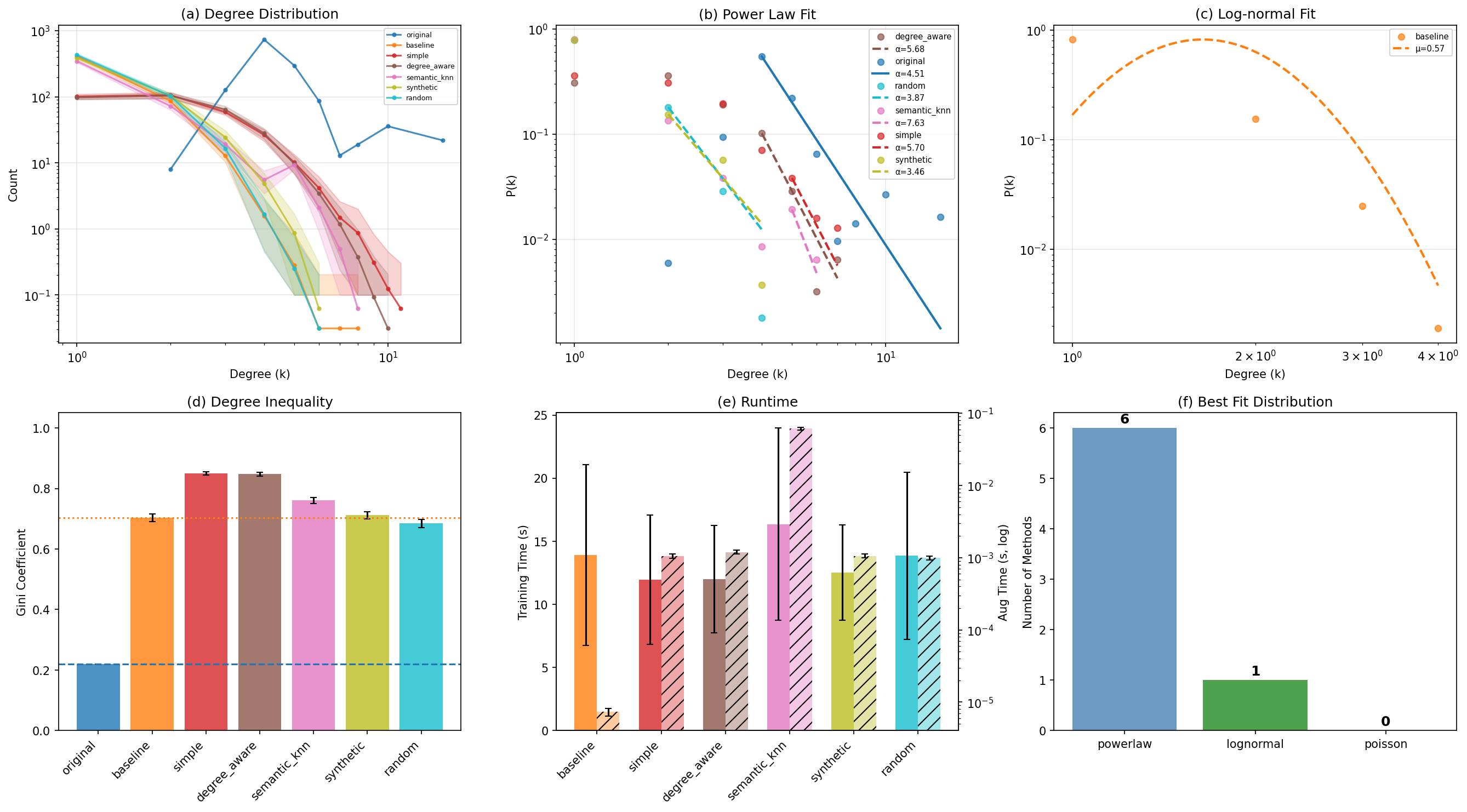}
  \caption{Amazon (product--category), GCN, $q{=}0.10$, $\phi{=}2$: Comprehensive analysis ($M\pm\mathrm{SD}$, $n=32$ seeds) comparing baseline, augmentation methods, and original graph. Panel (a) shows degree distributions on log-log scale with confidence bands; (b) Power Law fits with exponent $\alpha$; (c) Log-normal fits with parameters $\mu$ and $\sigma$; (d) Gini coefficients quantifying degree inequality (lower = more uniform); (e) runtime comparison showing training time (left axis) and augmentation time (right axis, log scale); (f) best-fit distribution counts across methods.}
  \label{fig:amazon_gcn_q10_phi2}
\end{figure}

\subsection{Benchmark Dataset - MovieLens}

\subsubsection{Summary}

On MovieLens, gains are modest and depend on encoder and objective: GAT sees only small AUC changes (best at $q{=}0.01$ with minor lifts from \texttt{simple}/\texttt{degree\_aware}; \texttt{semantic\_knn} often hurts), while Brier improves slightly for \texttt{degree\_aware}/\texttt{simple} at low $q$. GraphSAGE benefits most in AUC from \texttt{degree\_aware} at $q{=}0.01$, and Brier from \texttt{synthetic} at the same setting, reflecting calibration gains without large ranking shifts. GCN shows the clearest AUC gains from \texttt{simple}/\texttt{degree\_aware}/\texttt{semantic\_knn} at $q{=}0.01$ (up to +0.075), with the strongest Brier drops from \texttt{synthetic} (down to $-0.082$). Degree distributions remain very sparse for \texttt{semantic\_knn} but rise toward mean degree $\sim1.25$ for other augmentations, lowering isolated nodes; runtime overheads stay negligible (sub-0.16s augmentation, training within a couple seconds of baseline), so quality differences are not driven by cost.

\subsubsection{GAT}

\paragraph{Summary Analysis}

MovieLens GAT shows very small AUC shifts: best cases are minor lifts from \texttt{simple}/\texttt{degree\_aware} at $q{=}0.01$ (e.g., $\phi{=}100$: +0.007) and stability near baseline elsewhere, while \texttt{semantic\_knn} often decreases AUC (notably $q{=}0.05$) and \texttt{random} stays flat or negative. Brier improvements are modest and concentrated at $q{=}0.01$, $\phi{=}5/2$, where \texttt{simple}/\texttt{degree\_aware} reduce error by up to $-0.006$; \texttt{semantic\_knn} raises Brier at $q{=}0.05$, and \texttt{synthetic/random} frequently worsen it. Degree-wise, most augmentations keep graphs sparse (mean degree $\approx0.04$–0.12 at $\phi{=}5/2$) with high Gini, while \texttt{semantic\_knn} remains closest to the ultra-sparse baseline. Runtime overhead is negligible (aug $\leq0.062$s, training within $\sim\pm2$s of baseline), so differences reflect graph quality rather than cost.

% Data source: Table~\ref{tab:movielens_gat_auc} and corresponding Brier table  \textbf{AUC-ROC Performance:} \texttt{simple} achieves the highest average AUC improvement (mean $\Delta$AUC=+0.001): at $q=0.01$, $\phi=5x$, AUC=0.714$\pm$0.050 ($\Delta$AUC=+0.002) vs baseline (0.712$\pm$0.052); at $q=0.01$, $\phi=2x$, AUC=0.711$\pm$0.059 ($\Delta$AUC=-0.001). Conversely, \texttt{semantic_knn} shows the weakest performance (mean $\Delta$AUC=-0.011, range: -0.011 to -0.011).  \textbf{Brier Score Performance:} \texttt{degree_aware} achieves the best Brier score (mean $\Delta$Brier=-0.005). At $q=0.01$, $\phi=5x$, it reaches 0.230$\pm$0.017 ($\Delta$Brier=-0.005) vs baseline (0.235$\pm$0.015).  % SOURCE FILES: % - code-repo/notebooks/movielens/movielens-GAT-drop0.99-2x-20251125_204316_summary.csv % - code-repo/notebooks/movielens/movielens-GAT-drop0.99-5x-20251124_165522_summary.csv

\paragraph{AUC and Brier Score}

% ========== AUC TABLE ==========
\begin{longtable}{c c l r r r r r}
\caption{MovieLens (movie--genre) GAT: AUC-ROC ($M\pm SD$) with paired $t$-tests vs.\ sparse baseline ($n=32$ seeds). A higher AUC is better.}
\label{tab:movielens_gat_auc}\\
\toprule
$q$ & $\phi$ & Method & AUC $M\pm SD$ & $\Delta$AUC & $t(31)$ & $p$ & $d$ \\
\midrule
\endfirsthead
\toprule
$q$ & $\phi$ & Method & AUC $M\pm SD$ & $\Delta$AUC & $t(31)$ & $p$ & $d$ \\
\midrule
\endhead
\bottomrule
\endfoot

% SOURCE: notebooks/movielens/movielens-20250924_011102_summary.csv
%         notebooks/movielens/paired-ttest-20250924_011103.csv
% q=0.01, phi=100x
% NOTE: Legacy 100x data - t-test statistics may not be available
0.01 & 100$\times$    & baseline       & 0.710 $\pm$ 0.061 & +0.000 & ---   & ---   & ---   \\
0.01 & 100$\times$    & degree\_aware  & 0.713 $\pm$ 0.064 & +0.003$^{\mathrm{ns}}$ & $-0.42$ & 0.681    & $-0.07$ \\
0.01 & 100$\times$    & \textbf{simple} & 0.717 $\pm$ 0.063 & +0.007$^{\mathrm{ns}}$ & $-1.33$ & 0.195    & $-0.23$ \\
0.01 & 100$\times$    & semantic\_knn  & 0.708 $\pm$ 0.064 & -0.002$^{\mathrm{ns}}$ & $0.35$  & 0.732    & $+0.06$ \\
0.01 & 100$\times$    & synthetic      & 0.679 $\pm$ 0.075 & -0.031$^{*}$       & $2.15$  & 0.039    & $+0.38$ \\
0.01 & 100$\times$    & random         & 0.652 $\pm$ 0.089 & -0.059$^{***}$     & $3.66$  & $<$0.001 & $+0.65$ \\
0.01 & 100$\times$    & original       & 0.811 $\pm$ 0.015 & +0.101$^{***}$     & $-9.95$ & $<$0.001 & $-1.76$ \\
\midrule
% SOURCE: notebooks/movielens/movielens-GAT-drop0.99-5x-20251124_165522_summary.csv
%         notebooks/movielens/movielens-GAT-drop0.99-5x-paired-ttest-20251124_165524.csv
% q=0.01, phi=5x
0.01 & 5$\times$      & baseline       & 0.712 $\pm$ 0.052 & +0.000 & ---   & ---   & ---   \\
0.01 & 5$\times$      & \textbf{degree\_aware} & 0.718 $\pm$ 0.060 & +0.006$^{\mathrm{ns}}$ & $-1.23$ & 0.230    & $-0.22$ \\
0.01 & 5$\times$      & simple         & 0.714 $\pm$ 0.050 & +0.002$^{\mathrm{ns}}$ & $-0.47$ & 0.640    & $-0.08$ \\
0.01 & 5$\times$      & semantic\_knn  & 0.701 $\pm$ 0.063 & -0.011$^{\mathrm{ns}}$ & $1.63$  & 0.112    & $+0.29$ \\
0.01 & 5$\times$      & synthetic      & 0.712 $\pm$ 0.057 & -0.000$^{\mathrm{ns}}$ & $0.06$  & 0.955    & $+0.01$ \\
0.01 & 5$\times$      & random         & 0.705 $\pm$ 0.066 & -0.007$^{\mathrm{ns}}$ & $1.04$  & 0.308    & $+0.18$ \\
0.01 & 5$\times$      & original       & 0.812 $\pm$ 0.014 & +0.100$^{***}$     & $-11.34$ & $<$0.001 & $-2.01$ \\
\midrule
% SOURCE: notebooks/movielens/movielens-GAT-drop0.99-2x-20251125_204316_summary.csv
%         notebooks/movielens/movielens-GAT-drop0.99-2x-paired-ttest-20251125_204319.csv
% q=0.01, phi=2x
0.01 & 2$\times$      & \textbf{baseline} & 0.712 $\pm$ 0.052 & +0.000 & ---   & ---   & ---   \\
0.01 & 2$\times$      & degree\_aware  & 0.708 $\pm$ 0.058 & -0.004$^{\mathrm{ns}}$ & $0.99$  & 0.328    & $+0.18$ \\
0.01 & 2$\times$      & simple         & 0.711 $\pm$ 0.059 & -0.001$^{\mathrm{ns}}$ & $0.16$  & 0.877    & $+0.03$ \\
0.01 & 2$\times$      & semantic\_knn  & 0.701 $\pm$ 0.063 & -0.011$^{\mathrm{ns}}$ & $1.63$  & 0.112    & $+0.29$ \\
0.01 & 2$\times$      & synthetic      & 0.710 $\pm$ 0.077 & -0.002$^{\mathrm{ns}}$ & $0.21$  & 0.834    & $+0.04$ \\
0.01 & 2$\times$      & random         & 0.710 $\pm$ 0.057 & -0.002$^{\mathrm{ns}}$ & $0.29$  & 0.773    & $+0.05$ \\
0.01 & 2$\times$      & original       & 0.812 $\pm$ 0.014 & +0.100$^{***}$     & $-11.34$ & $<$0.001 & $-2.01$ \\
\midrule
% SOURCE: notebooks/movielens/movielens-GAT-drop0.95-5x-20251124_183906_summary.csv
%         notebooks/movielens/movielens-GAT-drop0.95-5x-paired-ttest-20251124_183909.csv
% q=0.05, phi=5x
0.05 & 5$\times$      & baseline       & 0.719 $\pm$ 0.034 & +0.000 & ---   & ---   & ---   \\
0.05 & 5$\times$      & degree\_aware  & 0.717 $\pm$ 0.035 & -0.002$^{\mathrm{ns}}$ & $0.45$  & 0.655    & $+0.08$ \\
0.05 & 5$\times$      & simple         & 0.718 $\pm$ 0.033 & -0.001$^{\mathrm{ns}}$ & $0.29$  & 0.777    & $+0.05$ \\
0.05 & 5$\times$      & semantic\_knn  & 0.701 $\pm$ 0.034 & -0.018$^{***}$     & $3.97$  & $<$0.001 & $+0.70$ \\
0.05 & 5$\times$      & \textbf{synthetic} & 0.723 $\pm$ 0.037 & +0.003$^{\mathrm{ns}}$ & $-0.53$ & 0.600    & $-0.09$ \\
0.05 & 5$\times$      & random         & 0.710 $\pm$ 0.042 & -0.009$^{\mathrm{ns}}$ & $1.54$  & 0.133    & $+0.27$ \\
0.05 & 5$\times$      & original       & 0.812 $\pm$ 0.014 & +0.092$^{***}$     & $-14.43$ & $<$0.001 & $-2.55$ \\
\midrule
% SOURCE: notebooks/movielens/movielens-GAT-drop0.95-2x-20251126_012641_summary.csv
%         notebooks/movielens/movielens-GAT-drop0.95-2x-paired-ttest-20251126_012643.csv
% q=0.05, phi=2x
0.05 & 2$\times$      & \textbf{baseline} & 0.719 $\pm$ 0.034 & +0.000 & ---   & ---   & ---   \\
0.05 & 2$\times$      & degree\_aware  & 0.716 $\pm$ 0.028 & -0.004$^{\mathrm{ns}}$ & $0.78$  & 0.440    & $+0.14$ \\
0.05 & 2$\times$      & simple         & 0.716 $\pm$ 0.030 & -0.004$^{\mathrm{ns}}$ & $0.80$  & 0.432    & $+0.14$ \\
0.05 & 2$\times$      & semantic\_knn  & 0.701 $\pm$ 0.034 & -0.018$^{***}$     & $3.97$  & $<$0.001 & $+0.70$ \\
0.05 & 2$\times$      & synthetic      & 0.706 $\pm$ 0.039 & -0.013$^{*}$       & $2.31$  & 0.027    & $+0.41$ \\
0.05 & 2$\times$      & random         & 0.705 $\pm$ 0.045 & -0.015$^{\mathrm{ns}}$ & $1.87$  & 0.071    & $+0.33$ \\
0.05 & 2$\times$      & original       & 0.812 $\pm$ 0.014 & +0.092$^{***}$     & $-14.43$ & $<$0.001 & $-2.55$ \\
\midrule
% SOURCE: notebooks/movielens/movielens-GAT-drop0.9-5x-20251126_053217_summary.csv
%         notebooks/movielens/movielens-GAT-drop0.9-5x-paired-ttest-20251126_053220.csv
% q=0.10, phi=5x
0.10 & 5$\times$      & baseline       & 0.713 $\pm$ 0.029 & +0.000 & ---   & ---   & ---   \\
0.10 & 5$\times$      & degree\_aware  & 0.714 $\pm$ 0.027 & +0.002$^{\mathrm{ns}}$ & $-0.75$ & 0.456    & $-0.13$ \\
0.10 & 5$\times$      & \textbf{simple} & 0.715 $\pm$ 0.025 & +0.002$^{\mathrm{ns}}$ & $-0.75$ & 0.461    & $-0.13$ \\
0.10 & 5$\times$      & semantic\_knn  & 0.706 $\pm$ 0.028 & -0.006$^{*}$       & $2.44$  & 0.021    & $+0.43$ \\
0.10 & 5$\times$      & synthetic      & 0.709 $\pm$ 0.026 & -0.003$^{\mathrm{ns}}$ & $0.79$  & 0.434    & $+0.14$ \\
0.10 & 5$\times$      & random         & 0.709 $\pm$ 0.025 & -0.004$^{\mathrm{ns}}$ & $0.84$  & 0.407    & $+0.15$ \\
0.10 & 5$\times$      & original       & 0.812 $\pm$ 0.014 & +0.099$^{***}$     & $-17.98$ & $<$0.001 & $-3.18$ \\
\midrule

\end{longtable}

% ========== BRIER TABLE ==========
\begin{longtable}{c c l r r r r r}
\caption{MovieLens (movie--genre) GAT: Brier Score ($M\pm SD$) with paired $t$-tests vs.\ sparse baseline ($n=32$ seeds, lower is better).}
\label{tab:movielens_gat_brier}\\
\toprule
$q$ & $\phi$ & Method & Brier $M\pm SD$ & $\Delta$Brier & $t(31)$ & $p$ & $d$ \\
\midrule
\endfirsthead
\toprule
$q$ & $\phi$ & Method & Brier $M\pm SD$ & $\Delta$Brier & $t(31)$ & $p$ & $d$ \\
\midrule
\endhead
\bottomrule
\endfoot

% SOURCE: notebooks/movielens/movielens-20250924_011102_summary.csv
%         notebooks/movielens/paired-ttest-20250924_011103.csv
% q=0.01, phi=100x
% NOTE: Legacy 100x data - t-test statistics may not be available
0.01 & 100$\times$    & \textbf{baseline} & 0.231 $\pm$ 0.016 & +0.000 & ---  & ---   & ---   \\
0.01 & 100$\times$    & degree\_aware  & 0.233 $\pm$ 0.014 & +0.001$^{\mathrm{ns}}$ & $-0.72$ & 0.474    & $-0.13$  \\
0.01 & 100$\times$    & simple         & 0.231 $\pm$ 0.012 & -0.000$^{\mathrm{ns}}$ & $0.12$ & 0.907    & $+0.02$  \\
0.01 & 100$\times$    & semantic\_knn  & 0.235 $\pm$ 0.014 & +0.004$^{\mathrm{ns}}$ & $-1.43$ & 0.162    & $-0.25$  \\
0.01 & 100$\times$    & synthetic      & 0.245 $\pm$ 0.008 & +0.014$^{***}$      & $-4.82$ & $<$0.001 & $-0.85$  \\
0.01 & 100$\times$    & random         & 0.245 $\pm$ 0.009 & +0.013$^{***}$      & $-3.89$ & $<$0.001 & $-0.69$  \\
0.01 & 100$\times$    & original       & 0.218 $\pm$ 0.004 & -0.013$^{***}$      & $5.22$ & $<$0.001 & $+0.92$  \\
\midrule
% SOURCE: notebooks/movielens/movielens-GAT-drop0.99-5x-20251124_165522_summary.csv
%         notebooks/movielens/movielens-GAT-drop0.99-5x-paired-ttest-20251124_165524.csv
% q=0.01, phi=5x
0.01 & 5$\times$      & baseline       & 0.235 $\pm$ 0.015 & +0.000 & ---  & ---   & ---   \\
0.01 & 5$\times$      & degree\_aware  & 0.230 $\pm$ 0.017 & -0.005$^{**}$       & $3.01$ & 0.005    & $+0.53$  \\
0.01 & 5$\times$      & \textbf{simple} & 0.229 $\pm$ 0.014 & -0.006$^{**}$       & $3.34$ & 0.002    & $+0.59$  \\
0.01 & 5$\times$      & semantic\_knn  & 0.236 $\pm$ 0.015 & +0.001$^{\mathrm{ns}}$ & $-0.49$ & 0.627    & $-0.09$  \\
0.01 & 5$\times$      & synthetic      & 0.238 $\pm$ 0.015 & +0.003$^{\mathrm{ns}}$ & $-1.35$ & 0.188    & $-0.24$  \\
0.01 & 5$\times$      & random         & 0.236 $\pm$ 0.013 & +0.001$^{\mathrm{ns}}$ & $-0.64$ & 0.526    & $-0.11$  \\
0.01 & 5$\times$      & original       & 0.218 $\pm$ 0.004 & -0.017$^{***}$      & $6.33$ & $<$0.001 & $+1.12$  \\
\midrule
% SOURCE: notebooks/movielens/movielens-GAT-drop0.99-2x-20251125_204316_summary.csv
%         notebooks/movielens/movielens-GAT-drop0.99-2x-paired-ttest-20251125_204319.csv
% q=0.01, phi=2x
0.01 & 2$\times$      & baseline       & 0.235 $\pm$ 0.015 & +0.000 & ---  & ---   & ---   \\
0.01 & 2$\times$      & \textbf{degree\_aware} & 0.231 $\pm$ 0.016 & -0.004$^{\mathrm{ns}}$ & $1.94$ & 0.061    & $+0.34$  \\
0.01 & 2$\times$      & simple         & 0.232 $\pm$ 0.015 & -0.003$^{\mathrm{ns}}$ & $1.87$ & 0.071    & $+0.33$  \\
0.01 & 2$\times$      & semantic\_knn  & 0.236 $\pm$ 0.015 & +0.001$^{\mathrm{ns}}$ & $-0.49$ & 0.627    & $-0.09$  \\
0.01 & 2$\times$      & synthetic      & 0.235 $\pm$ 0.013 & +0.000$^{\mathrm{ns}}$ & $-0.11$ & 0.917    & $-0.02$  \\
0.01 & 2$\times$      & random         & 0.238 $\pm$ 0.010 & +0.003$^{\mathrm{ns}}$ & $-1.41$ & 0.168    & $-0.25$  \\
0.01 & 2$\times$      & original       & 0.218 $\pm$ 0.004 & -0.017$^{***}$      & $6.33$ & $<$0.001 & $+1.12$  \\
\midrule
% SOURCE: notebooks/movielens/movielens-GAT-drop0.95-5x-20251124_183906_summary.csv
%         notebooks/movielens/movielens-GAT-drop0.95-5x-paired-ttest-20251124_183909.csv
% q=0.05, phi=5x
0.05 & 5$\times$      & baseline       & 0.240 $\pm$ 0.008 & +0.000 & ---  & ---   & ---   \\
0.05 & 5$\times$      & degree\_aware  & 0.239 $\pm$ 0.006 & -0.001$^{\mathrm{ns}}$ & $0.90$ & 0.378    & $+0.16$  \\
0.05 & 5$\times$      & simple         & 0.240 $\pm$ 0.009 & +0.000$^{\mathrm{ns}}$ & $-0.46$ & 0.650    & $-0.08$  \\
0.05 & 5$\times$      & semantic\_knn  & 0.245 $\pm$ 0.007 & +0.005$^{***}$      & $-4.83$ & $<$0.001 & $-0.85$  \\
0.05 & 5$\times$      & \textbf{synthetic} & 0.237 $\pm$ 0.011 & -0.003$^{\mathrm{ns}}$ & $1.76$ & 0.088    & $+0.31$  \\
0.05 & 5$\times$      & random         & 0.239 $\pm$ 0.008 & -0.001$^{\mathrm{ns}}$ & $0.55$ & 0.589    & $+0.10$  \\
0.05 & 5$\times$      & original       & 0.218 $\pm$ 0.004 & -0.022$^{***}$      & $14.99$ & $<$0.001 & $+2.65$  \\
\midrule
% SOURCE: notebooks/movielens/movielens-GAT-drop0.95-2x-20251126_012641_summary.csv
%         notebooks/movielens/movielens-GAT-drop0.95-2x-paired-ttest-20251126_012643.csv
% q=0.05, phi=2x
0.05 & 2$\times$      & \textbf{baseline} & 0.240 $\pm$ 0.008 & +0.000 & ---  & ---   & ---   \\
0.05 & 2$\times$      & degree\_aware  & 0.241 $\pm$ 0.007 & +0.001$^{\mathrm{ns}}$ & $-1.24$ & 0.224    & $-0.22$  \\
0.05 & 2$\times$      & simple         & 0.242 $\pm$ 0.007 & +0.002$^{\mathrm{ns}}$ & $-1.95$ & 0.060    & $-0.34$  \\
0.05 & 2$\times$      & semantic\_knn  & 0.245 $\pm$ 0.007 & +0.005$^{***}$      & $-4.83$ & $<$0.001 & $-0.85$  \\
0.05 & 2$\times$      & synthetic      & 0.244 $\pm$ 0.008 & +0.004$^{*}$        & $-2.55$ & 0.016    & $-0.45$  \\
0.05 & 2$\times$      & random         & 0.244 $\pm$ 0.009 & +0.004$^{**}$       & $-3.10$ & 0.004    & $-0.55$  \\
0.05 & 2$\times$      & original       & 0.218 $\pm$ 0.004 & -0.022$^{***}$      & $14.99$ & $<$0.001 & $+2.65$  \\
\midrule
% SOURCE: notebooks/movielens/movielens-GAT-drop0.9-5x-20251126_053217_summary.csv
%         notebooks/movielens/movielens-GAT-drop0.9-5x-paired-ttest-20251126_053220.csv
% q=0.10, phi=5x
0.10 & 5$\times$      & baseline       & 0.238 $\pm$ 0.007 & +0.000 & ---  & ---   & ---   \\
0.10 & 5$\times$      & degree\_aware  & 0.238 $\pm$ 0.005 & -0.000$^{\mathrm{ns}}$ & $0.51$ & 0.613    & $+0.09$  \\
0.10 & 5$\times$      & simple         & 0.238 $\pm$ 0.006 & +0.000$^{\mathrm{ns}}$ & $-0.20$ & 0.842    & $-0.04$  \\
0.10 & 5$\times$      & semantic\_knn  & 0.240 $\pm$ 0.006 & +0.002$^{*}$        & $-2.36$ & 0.025    & $-0.42$  \\
0.10 & 5$\times$      & \textbf{synthetic} & 0.237 $\pm$ 0.007 & -0.001$^{\mathrm{ns}}$ & $0.41$ & 0.682    & $+0.07$  \\
0.10 & 5$\times$      & random         & 0.238 $\pm$ 0.009 & +0.000$^{\mathrm{ns}}$ & $-0.17$ & 0.870    & $-0.03$  \\
0.10 & 5$\times$      & original       & 0.218 $\pm$ 0.004 & -0.020$^{***}$      & $12.93$ & $<$0.001 & $+2.29$  \\
\midrule

\end{longtable}

\paragraph{Degree Distribution Analysis}

% ========== DEGREE DISTRIBUTION STATISTICS ==========
\begin{longtable}{c c l r r r l}
\caption{MovieLens (movie--genre) GAT: Degree Distribution Statistics ($M\pm SD$, $n=32$ seeds). Lower Gini coefficient indicates more uniform degree distribution.}
\label{tab:movielens_gat_degree}\\
\toprule
$q$ & $\phi$ & Method & Mean Degree & Gini Coeff. & Num. Isolated & Best Fit \\
\midrule
\endfirsthead
\toprule
$q$ & $\phi$ & Method & Mean Degree & Gini Coeff. & Num. Isolated & Best Fit \\
\midrule
\endhead
\bottomrule
\endfoot

% SOURCE: notebooks/movielens/degree_analysis_movielens-GAT-drop0.99-100x-20251126_091905_degree_stats.csv
%         notebooks/movielens/degree_analysis_movielens-GAT-drop0.99-100x-20251126_091905_distribution_fit.csv
% q=0.01, phi=100x
0.01 & 100$\times$        & baseline        & 0.0176 $\pm$ 0.0052            & 0.983 $\pm$ 0.005         & 9538.7 $\pm$ 49.7         & lognormal  \\
0.01 & 100$\times$        & degree\_aware   & 1.2506 $\pm$ 0.0794            & 0.989 $\pm$ 0.001         & 9587.2 $\pm$ 7.8          & lognormal  \\
0.01 & 100$\times$        & simple          & 1.2506 $\pm$ 0.0794            & 0.988 $\pm$ 0.001         & 9587.2 $\pm$ 7.8          & lognormal  \\
0.01 & 100$\times$        & semantic\_knn   & 0.0190 $\pm$ 0.0009            & 0.985 $\pm$ 0.001         & 9546.3 $\pm$ 8.7          & powerlaw   \\
0.01 & 100$\times$        & synthetic       & 1.2506 $\pm$ 0.0794            & 0.898 $\pm$ 0.007         & 8452.6 $\pm$ 80.0         & powerlaw   \\
0.01 & 100$\times$        & \textbf{random} & 1.2506 $\pm$ 0.0794            & 0.476 $\pm$ 0.013         & 2784.9 $\pm$ 217.0        & lognormal  \\
0.01 & 100$\times$        & original        & 2.2713 $\pm$ 0.0000            & 0.266 $\pm$ 0.000         & 0.0 $\pm$ 0.0             & lognormal  \\
\midrule
% SOURCE: notebooks/movielens/movielens-GAT-drop0.99-5x-20251124_165522_degree_stats.csv
%         notebooks/movielens/movielens-GAT-drop0.99-5x-20251124_165522_distribution_fit.csv
% q=0.01, phi=5x
0.01 & 5$\times$          & baseline        & 0.0226 $\pm$ 0.0014            & 0.978 $\pm$ 0.001         & 9490.3 $\pm$ 13.9         & lognormal  \\
0.01 & 5$\times$          & degree\_aware   & 0.0625 $\pm$ 0.0040            & 0.990 $\pm$ 0.001         & 9587.2 $\pm$ 7.8          & powerlaw   \\
0.01 & 5$\times$          & simple          & 0.0625 $\pm$ 0.0040            & 0.990 $\pm$ 0.001         & 9587.2 $\pm$ 7.8          & powerlaw   \\
0.01 & 5$\times$          & semantic\_knn   & 0.0190 $\pm$ 0.0009            & 0.985 $\pm$ 0.001         & 9546.3 $\pm$ 8.7          & powerlaw   \\
0.01 & 5$\times$          & synthetic       & 0.0625 $\pm$ 0.0040            & 0.959 $\pm$ 0.003         & 9227.0 $\pm$ 31.4         & powerlaw   \\
0.01 & 5$\times$          & \textbf{random} & 0.0625 $\pm$ 0.0040            & 0.941 $\pm$ 0.004         & 9120.0 $\pm$ 36.3         & lognormal  \\
0.01 & 5$\times$          & original        & 2.2713 $\pm$ 0.0000            & 0.266 $\pm$ 0.000         & 0.0 $\pm$ 0.0             & lognormal  \\
\midrule
% SOURCE: notebooks/movielens/movielens-GAT-drop0.99-2x-20251125_204316_degree_stats.csv
%         notebooks/movielens/movielens-GAT-drop0.99-2x-20251125_204316_distribution_fit.csv
% q=0.01, phi=2x
0.01 & 2$\times$          & baseline        & 0.0226 $\pm$ 0.0014            & 0.978 $\pm$ 0.001         & 9490.3 $\pm$ 13.9         & lognormal  \\
0.01 & 2$\times$          & degree\_aware   & 0.0250 $\pm$ 0.0016            & 0.991 $\pm$ 0.001         & 9587.2 $\pm$ 7.8          & powerlaw   \\
0.01 & 2$\times$          & simple          & 0.0250 $\pm$ 0.0016            & 0.991 $\pm$ 0.001         & 9587.2 $\pm$ 7.8          & powerlaw   \\
0.01 & 2$\times$          & semantic\_knn   & 0.0190 $\pm$ 0.0009            & 0.985 $\pm$ 0.001         & 9546.3 $\pm$ 8.7          & powerlaw   \\
0.01 & 2$\times$          & synthetic       & 0.0250 $\pm$ 0.0016            & 0.978 $\pm$ 0.002         & 9482.6 $\pm$ 14.9         & lognormal  \\
0.01 & 2$\times$          & \textbf{random} & 0.0250 $\pm$ 0.0016            & 0.976 $\pm$ 0.002         & 9468.2 $\pm$ 15.8         & lognormal  \\
0.01 & 2$\times$          & original        & 2.2713 $\pm$ 0.0000            & 0.266 $\pm$ 0.000         & 0.0 $\pm$ 0.0             & lognormal  \\
\midrule
% SOURCE: notebooks/movielens/movielens-GAT-drop0.95-5x-20251124_183906_degree_stats.csv
%         notebooks/movielens/movielens-GAT-drop0.95-5x-20251124_183906_distribution_fit.csv
% q=0.05, phi=5x
0.05 & 5$\times$          & baseline        & 0.1144 $\pm$ 0.0038            & 0.895 $\pm$ 0.004         & 8647.3 $\pm$ 35.3         & lognormal  \\
0.05 & 5$\times$          & degree\_aware   & 0.3139 $\pm$ 0.0105            & 0.953 $\pm$ 0.002         & 9114.0 $\pm$ 20.2         & powerlaw   \\
0.05 & 5$\times$          & simple          & 0.3139 $\pm$ 0.0105            & 0.953 $\pm$ 0.002         & 9114.0 $\pm$ 20.2         & lognormal  \\
0.05 & 5$\times$          & semantic\_knn   & 0.0704 $\pm$ 0.0022            & 0.938 $\pm$ 0.002         & 9065.5 $\pm$ 21.8         & powerlaw   \\
0.05 & 5$\times$          & synthetic       & 0.3139 $\pm$ 0.0105            & 0.824 $\pm$ 0.005         & 7528.5 $\pm$ 66.1         & powerlaw   \\
0.05 & 5$\times$          & \textbf{random} & 0.3139 $\pm$ 0.0105            & 0.763 $\pm$ 0.007         & 7090.3 $\pm$ 80.0         & lognormal  \\
0.05 & 5$\times$          & original        & 2.2713 $\pm$ 0.0000            & 0.266 $\pm$ 0.000         & 0.0 $\pm$ 0.0             & lognormal  \\
\midrule
% SOURCE: notebooks/movielens/movielens-GAT-drop0.95-2x-20251126_012641_degree_stats.csv
%         notebooks/movielens/movielens-GAT-drop0.95-2x-20251126_012641_distribution_fit.csv
% q=0.05, phi=2x
0.05 & 2$\times$          & baseline        & 0.1144 $\pm$ 0.0038            & 0.895 $\pm$ 0.004         & 8647.3 $\pm$ 35.3         & lognormal  \\
0.05 & 2$\times$          & degree\_aware   & 0.1256 $\pm$ 0.0042            & 0.955 $\pm$ 0.002         & 9114.0 $\pm$ 20.2         & powerlaw   \\
0.05 & 2$\times$          & simple          & 0.1256 $\pm$ 0.0042            & 0.955 $\pm$ 0.002         & 9114.0 $\pm$ 20.2         & powerlaw   \\
0.05 & 2$\times$          & semantic\_knn   & 0.0704 $\pm$ 0.0022            & 0.938 $\pm$ 0.002         & 9065.5 $\pm$ 21.8         & powerlaw   \\
0.05 & 2$\times$          & synthetic       & 0.1256 $\pm$ 0.0042            & 0.900 $\pm$ 0.004         & 8629.1 $\pm$ 36.0         & lognormal  \\
0.05 & 2$\times$          & \textbf{random} & 0.1256 $\pm$ 0.0042            & 0.888 $\pm$ 0.004         & 8559.1 $\pm$ 36.5         & lognormal  \\
0.05 & 2$\times$          & original        & 2.2713 $\pm$ 0.0000            & 0.266 $\pm$ 0.000         & 0.0 $\pm$ 0.0             & lognormal  \\
\midrule
% SOURCE: notebooks/movielens/movielens-GAT-drop0.9-5x-20251126_053217_degree_stats.csv
%         notebooks/movielens/movielens-GAT-drop0.9-5x-20251126_053217_distribution_fit.csv
% q=0.10, phi=5x
0.10 & 5$\times$          & baseline        & 0.2285 $\pm$ 0.0050            & 0.808 $\pm$ 0.004         & 7684.8 $\pm$ 39.8         & lognormal  \\
0.10 & 5$\times$          & degree\_aware   & 0.6263 $\pm$ 0.0136            & 0.908 $\pm$ 0.002         & 8550.8 $\pm$ 25.1         & lognormal  \\
0.10 & 5$\times$          & simple          & 0.6263 $\pm$ 0.0136            & 0.909 $\pm$ 0.002         & 8550.8 $\pm$ 25.1         & lognormal  \\
0.10 & 5$\times$          & semantic\_knn   & 0.1331 $\pm$ 0.0027            & 0.884 $\pm$ 0.003         & 8503.1 $\pm$ 24.7         & lognormal  \\
0.10 & 5$\times$          & synthetic       & 0.6263 $\pm$ 0.0136            & 0.703 $\pm$ 0.006         & 5837.7 $\pm$ 72.7         & lognormal  \\
0.10 & 5$\times$          & \textbf{random} & 0.6263 $\pm$ 0.0136            & 0.625 $\pm$ 0.006         & 5181.8 $\pm$ 75.1         & lognormal  \\
0.10 & 5$\times$          & original        & 2.2713 $\pm$ 0.0000            & 0.266 $\pm$ 0.000         & 0.0 $\pm$ 0.0             & lognormal  \\
\midrule

\end{longtable}

\begin{figure}[H]
  \centering
  \includegraphics[width=1\linewidth]{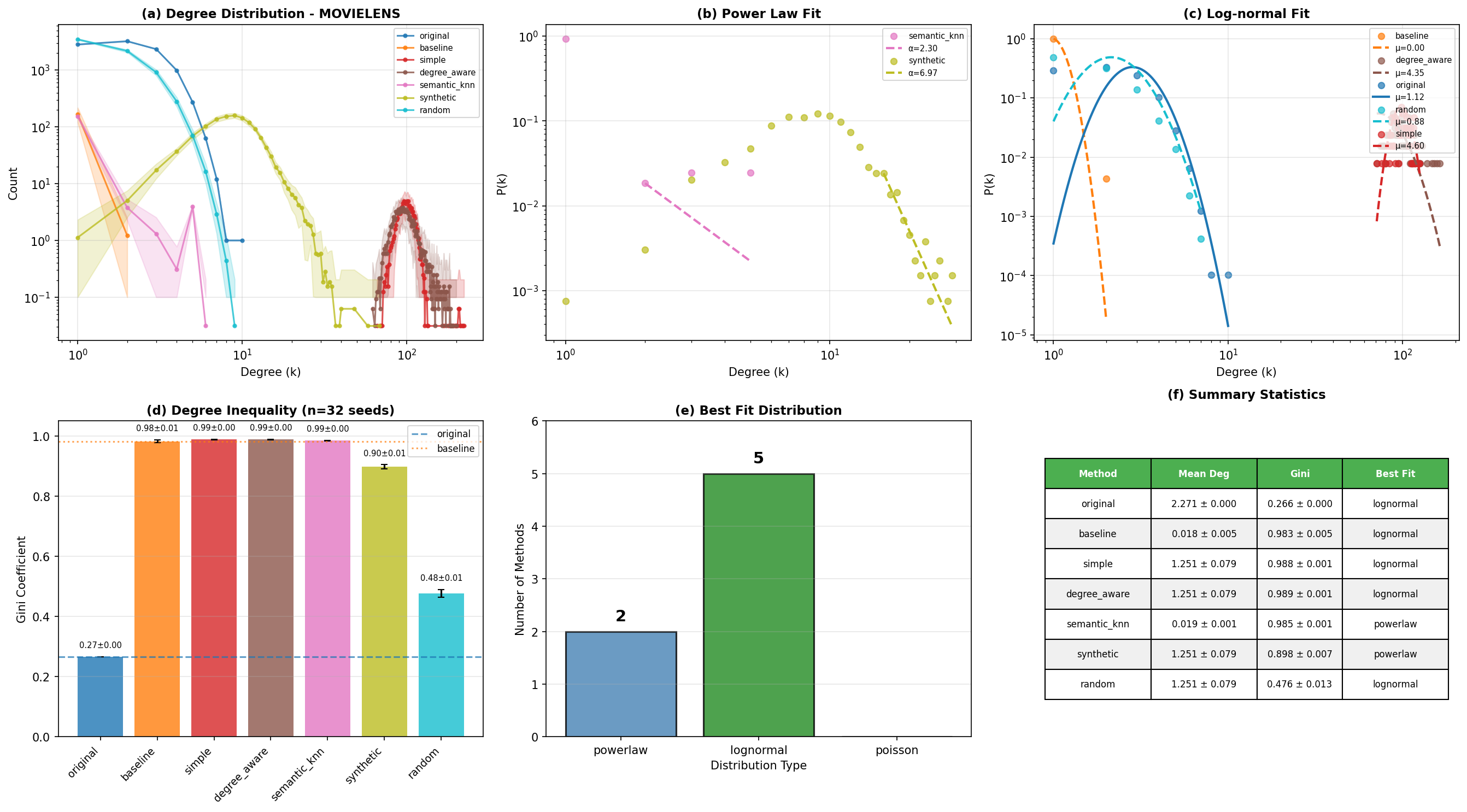}
  \caption{MovieLens (movie--genre), GAT, $q{=}0.01$, $\phi{=}100$: Comprehensive analysis ($M\pm\mathrm{SD}$, $n=32$ seeds) comparing baseline, augmentation methods, and original graph. Panel (a) shows degree distributions on log-log scale with confidence bands; (b) Power Law fits with exponent $\alpha$; (c) Log-normal fits with parameters $\mu$ and $\sigma$; (d) Gini coefficients quantifying degree inequality (lower = more uniform); (e) runtime comparison showing training time (left axis) and augmentation time (right axis, log scale); (f) best-fit distribution counts across methods.}
  \label{fig:movielens_gat_q01_phi100}
\end{figure}

\paragraph{Runtime Analysis}

% ========== RUNTIME STATISTICS ==========
\begin{longtable}{c c l r r}
\caption{MovieLens (movie--genre) GAT: Runtime Statistics ($M\pm SD$, seconds, $n=32$ seeds). Lower times are better.}
\label{tab:movielens_gat_runtime}\\
\toprule
$q$ & $\phi$ & Method & Aug. Time (s) & Train Time (s) \\
\midrule
\endfirsthead
\toprule
$q$ & $\phi$ & Method & Aug. Time (s) & Train Time (s) \\
\midrule
\endhead
\bottomrule
\endfoot

% SOURCE: notebooks/movielens/movielens-GAT-drop0.99-5x-20251124_165522_runtime.csv
% q=0.01, phi=5x
0.01 & 5$\times$          & baseline        & 0.0000 $\pm$ 0.0000            & 10.75 $\pm$ 3.96          \\
0.01 & 5$\times$          & degree\_aware   & 0.0025 $\pm$ 0.0002            & 10.79 $\pm$ 4.00          \\
0.01 & 5$\times$          & simple          & 0.0023 $\pm$ 0.0002            & 10.38 $\pm$ 3.83          \\
0.01 & 5$\times$          & semantic\_knn   & 0.1278 $\pm$ 0.0128            & 9.82 $\pm$ 3.43           \\
0.01 & 5$\times$          & synthetic       & 0.0024 $\pm$ 0.0005            & 11.32 $\pm$ 4.33          \\
0.01 & 5$\times$          & \textbf{random} & 0.0022 $\pm$ 0.0002            & 12.13 $\pm$ 5.32          \\
0.01 & 5$\times$          & original        & 0.0000 $\pm$ 0.0000            & 1139.45 $\pm$ 329.59      \\
\midrule
% SOURCE: notebooks/movielens/movielens-GAT-drop0.99-2x-20251125_204316_runtime.csv
% q=0.01, phi=2x
0.01 & 2$\times$          & baseline        & 0.0000 $\pm$ 0.0000            & 10.55 $\pm$ 3.97          \\
0.01 & 2$\times$          & degree\_aware   & 0.0024 $\pm$ 0.0003            & 10.17 $\pm$ 3.84          \\
0.01 & 2$\times$          & \textbf{simple} & 0.0022 $\pm$ 0.0002            & 10.48 $\pm$ 3.56          \\
0.01 & 2$\times$          & semantic\_knn   & 0.1284 $\pm$ 0.0104            & 9.64 $\pm$ 3.38           \\
0.01 & 2$\times$          & synthetic       & 0.0023 $\pm$ 0.0002            & 10.89 $\pm$ 5.18          \\
0.01 & 2$\times$          & random          & 0.0022 $\pm$ 0.0001            & 11.70 $\pm$ 5.32          \\
0.01 & 2$\times$          & original        & 0.0000 $\pm$ 0.0000            & 1126.47 $\pm$ 323.34      \\
\midrule
% SOURCE: notebooks/movielens/movielens-GAT-drop0.95-5x-20251124_183906_runtime.csv
% q=0.05, phi=5x
0.05 & 5$\times$          & baseline        & 0.0000 $\pm$ 0.0000            & 37.68 $\pm$ 13.98         \\
0.05 & 5$\times$          & degree\_aware   & 0.0029 $\pm$ 0.0008            & 41.96 $\pm$ 14.37         \\
0.05 & 5$\times$          & \textbf{simple} & 0.0023 $\pm$ 0.0003            & 44.30 $\pm$ 15.61         \\
0.05 & 5$\times$          & semantic\_knn   & 0.2163 $\pm$ 0.0124            & 31.06 $\pm$ 11.65         \\
0.05 & 5$\times$          & synthetic       & 0.0025 $\pm$ 0.0004            & 42.68 $\pm$ 13.99         \\
0.05 & 5$\times$          & random          & 0.0023 $\pm$ 0.0002            & 42.32 $\pm$ 12.28         \\
0.05 & 5$\times$          & original        & 0.0000 $\pm$ 0.0000            & 1109.65 $\pm$ 315.29      \\
\midrule
% SOURCE: notebooks/movielens/movielens-GAT-drop0.95-2x-20251126_012641_runtime.csv
% q=0.05, phi=2x
0.05 & 2$\times$          & baseline        & 0.0000 $\pm$ 0.0000            & 38.14 $\pm$ 14.05         \\
0.05 & 2$\times$          & degree\_aware   & 0.0025 $\pm$ 0.0002            & 34.76 $\pm$ 9.98          \\
0.05 & 2$\times$          & \textbf{simple} & 0.0022 $\pm$ 0.0002            & 37.27 $\pm$ 11.95         \\
0.05 & 2$\times$          & semantic\_knn   & 0.2098 $\pm$ 0.0120            & 31.42 $\pm$ 11.68         \\
0.05 & 2$\times$          & synthetic       & 0.0025 $\pm$ 0.0003            & 38.47 $\pm$ 11.55         \\
0.05 & 2$\times$          & random          & 0.0023 $\pm$ 0.0004            & 37.69 $\pm$ 14.78         \\
0.05 & 2$\times$          & original        & 0.0000 $\pm$ 0.0000            & 1120.95 $\pm$ 318.64      \\
\midrule
% SOURCE: notebooks/movielens/movielens-GAT-drop0.9-5x-20251126_053217_runtime.csv
% q=0.10, phi=5x
0.10 & 5$\times$          & baseline        & 0.0000 $\pm$ 0.0000            & 68.52 $\pm$ 28.41         \\
0.10 & 5$\times$          & degree\_aware   & 0.0031 $\pm$ 0.0010            & 89.01 $\pm$ 36.01         \\
0.10 & 5$\times$          & \textbf{simple} & 0.0022 $\pm$ 0.0003            & 79.85 $\pm$ 26.68         \\
0.10 & 5$\times$          & semantic\_knn   & 0.3095 $\pm$ 0.0100            & 71.80 $\pm$ 25.47         \\
0.10 & 5$\times$          & synthetic       & 0.0025 $\pm$ 0.0004            & 83.07 $\pm$ 31.44         \\
0.10 & 5$\times$          & random          & 0.0023 $\pm$ 0.0002            & 82.97 $\pm$ 22.81         \\
0.10 & 5$\times$          & original        & 0.0000 $\pm$ 0.0000            & 1123.86 $\pm$ 319.62      \\
\midrule

\end{longtable}

\begin{figure}[H]
  \centering
  \includegraphics[width=1\linewidth]{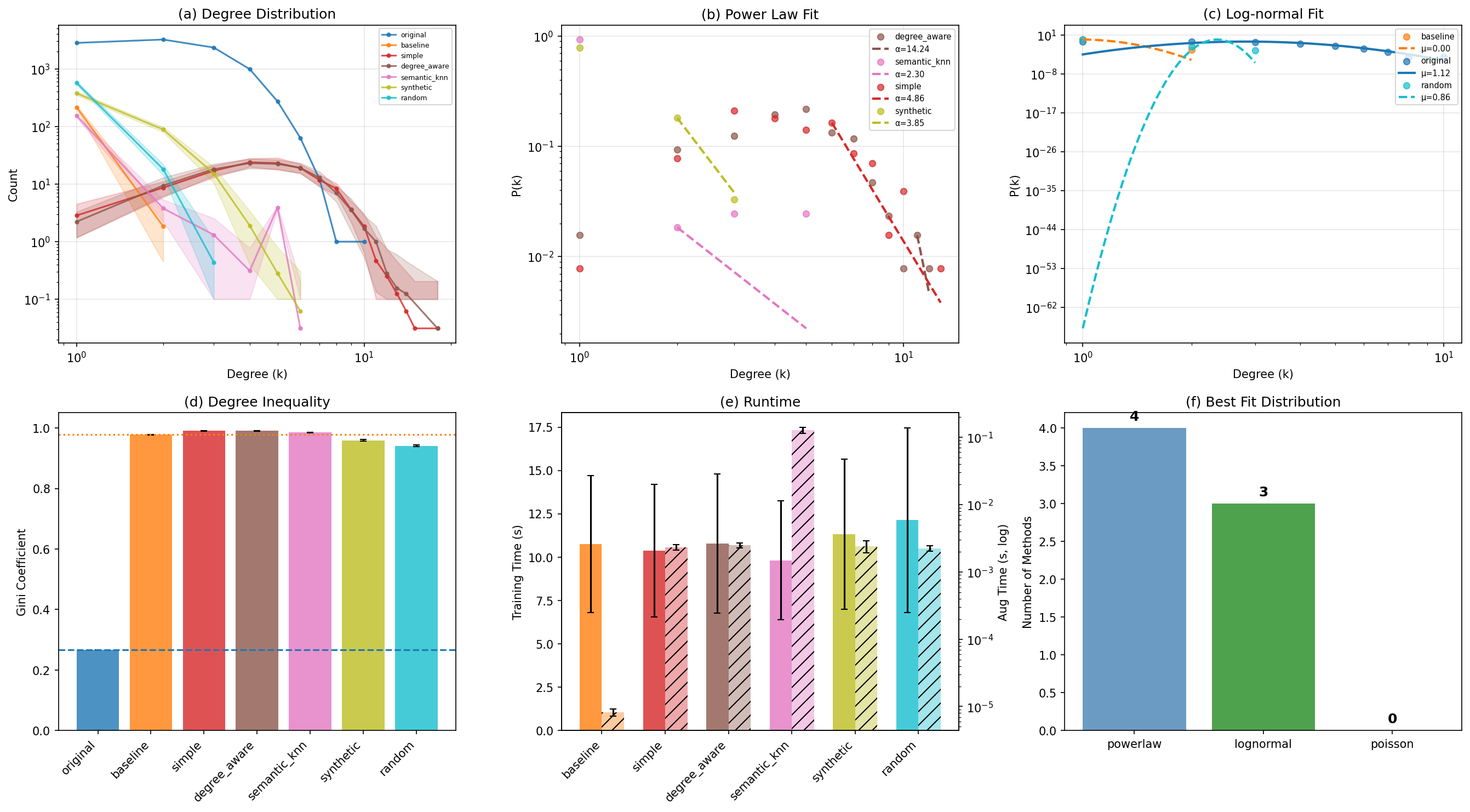}
  \caption{MovieLens (movie--genre), GAT, $q{=}0.01$, $\phi{=}5$: Comprehensive analysis ($M\pm\mathrm{SD}$, $n=32$ seeds) comparing baseline, augmentation methods, and original graph. Panel (a) shows degree distributions on log-log scale with confidence bands; (b) Power Law fits with exponent $\alpha$; (c) Log-normal fits with parameters $\mu$ and $\sigma$; (d) Gini coefficients quantifying degree inequality (lower = more uniform); (e) runtime comparison showing training time (left axis) and augmentation time (right axis, log scale); (f) best-fit distribution counts across methods.}
  \label{fig:movielens_gat_q01_phi5}
\end{figure}

\begin{figure}[H]
  \centering
  \includegraphics[width=1\linewidth]{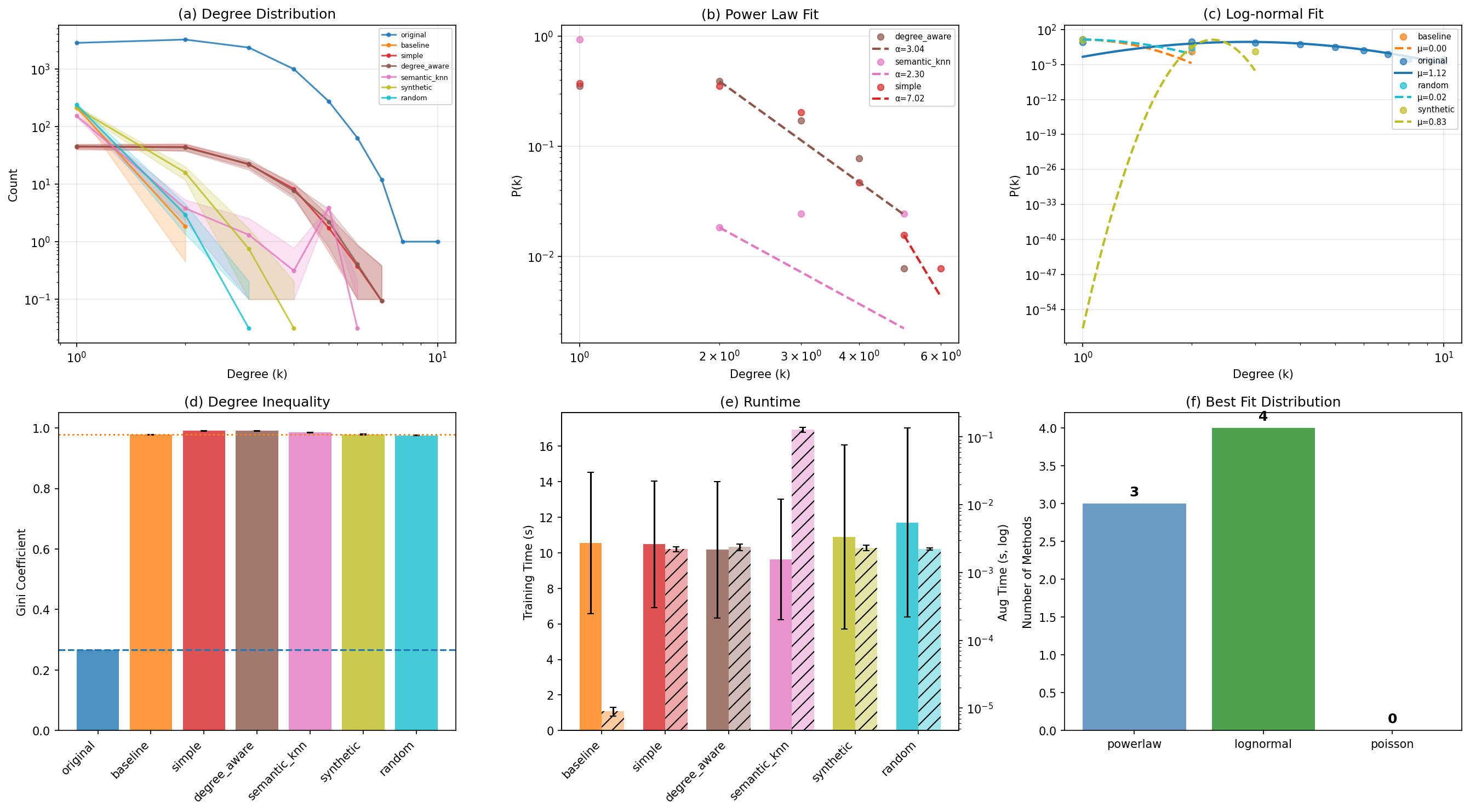}
  \caption{MovieLens (movie--genre), GAT, $q{=}0.01$, $\phi{=}2$: Comprehensive analysis ($M\pm\mathrm{SD}$, $n=32$ seeds) comparing baseline, augmentation methods, and original graph. Panels (a)--(f) follow the same structure as Fig.~\ref{fig:movielens_gat_q01_phi5}.}
  \label{fig:movielens_gat_q01_phi2}
\end{figure}

\begin{figure}[H]
  \centering
  \includegraphics[width=1\linewidth]{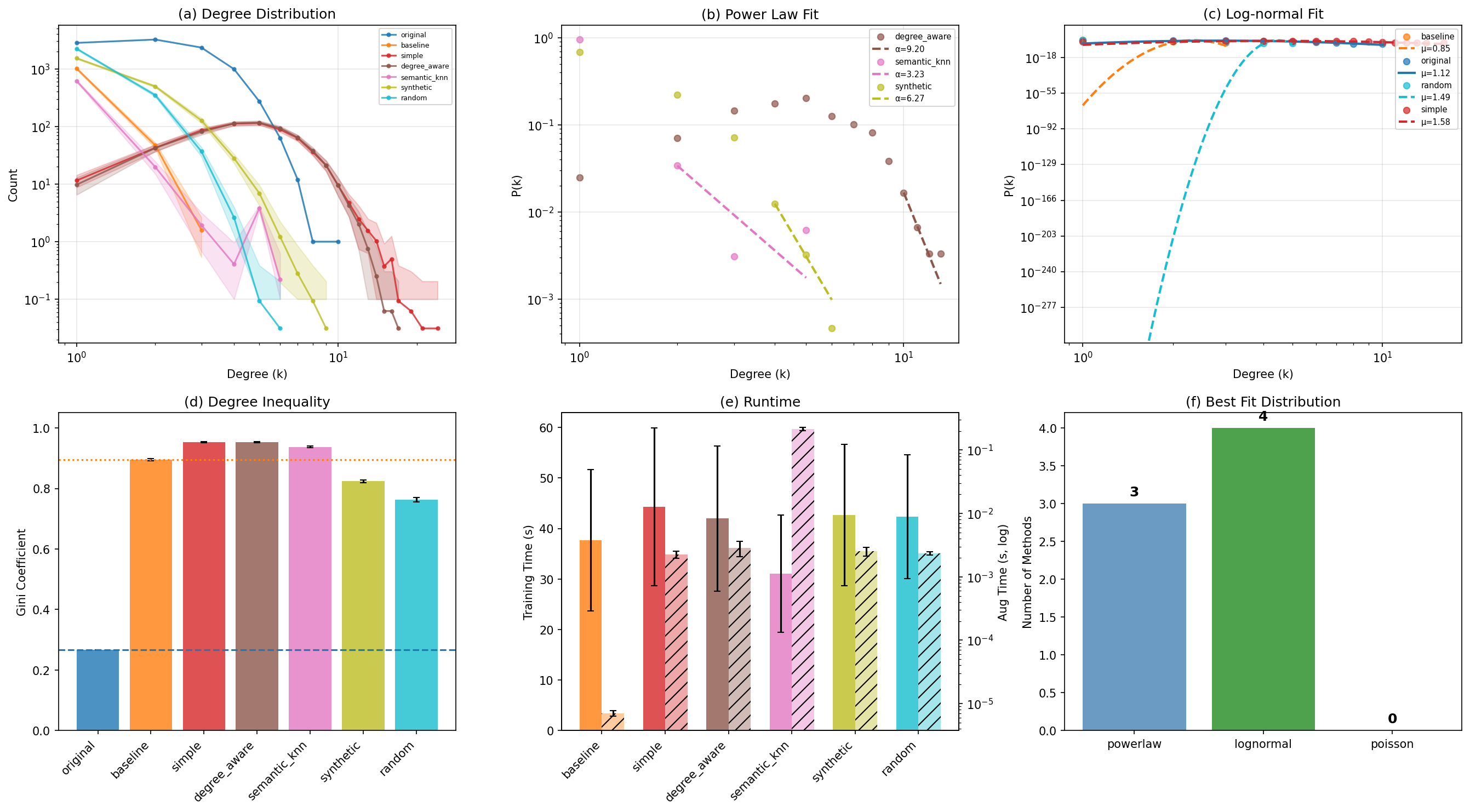}
  \caption{MovieLens (movie--genre), GAT, $q{=}0.05$, $\phi{=}5$: Comprehensive analysis ($M\pm\mathrm{SD}$, $n=32$ seeds) comparing baseline, augmentation methods, and original graph. Panel (a) shows degree distributions on log-log scale with confidence bands; (b) Power Law fits with exponent $\alpha$; (c) Log-normal fits with parameters $\mu$ and $\sigma$; (d) Gini coefficients quantifying degree inequality (lower = more uniform); (e) runtime comparison showing training time (left axis) and augmentation time (right axis, log scale); (f) best-fit distribution counts across methods.}
  \label{fig:movielens_gat_q05_phi5}
\end{figure}

\begin{figure}[H]
  \centering
  \includegraphics[width=1\linewidth]{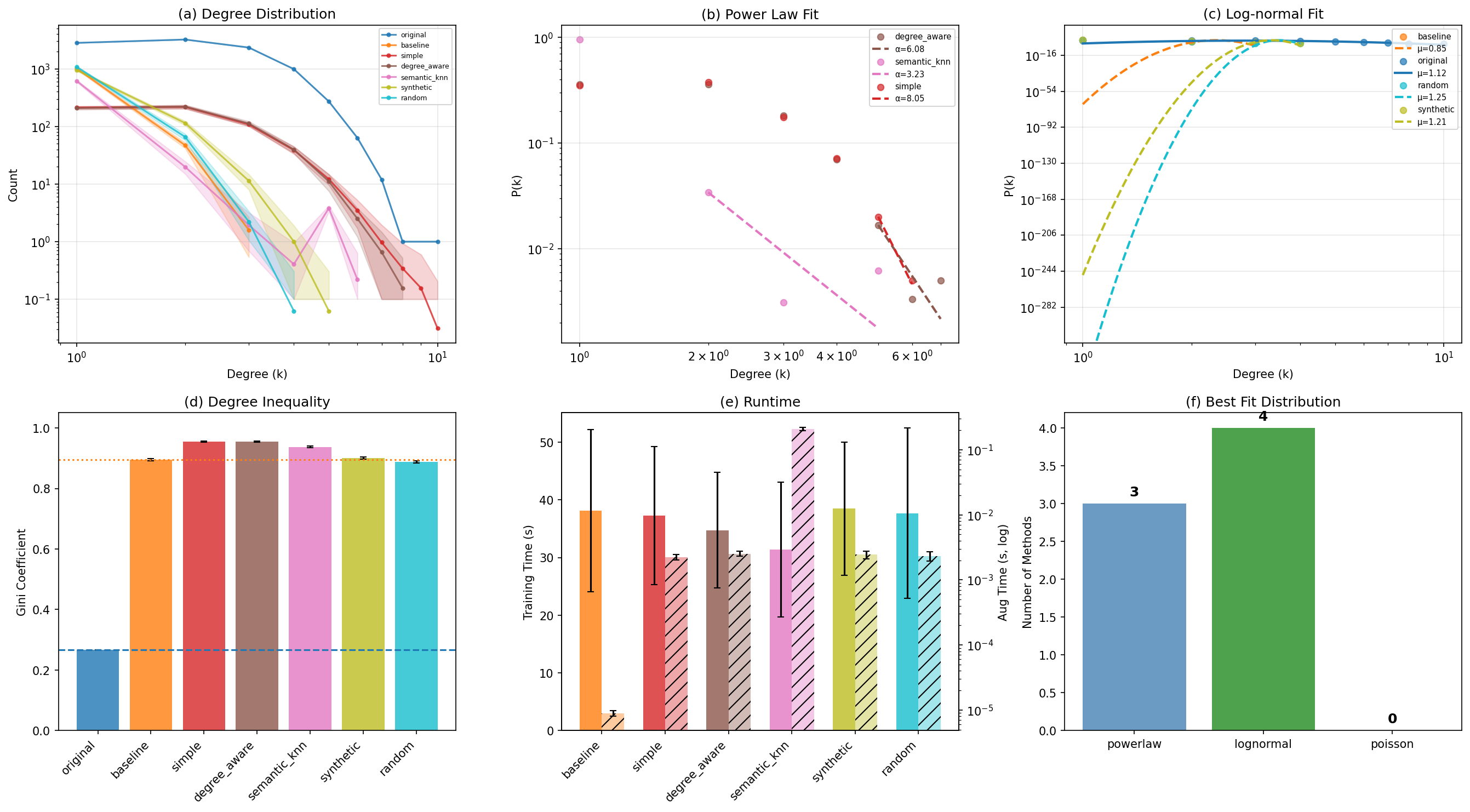}
  \caption{MovieLens (movie--genre), GAT, $q{=}0.05$, $\phi{=}2$: Comprehensive analysis ($M\pm\mathrm{SD}$, $n=32$ seeds) comparing baseline, augmentation methods, and original graph. Panels (a)--(f) follow the same structure as Fig.~\ref{fig:movielens_gat_q05_phi5}.}
  \label{fig:movielens_gat_q05_phi2}
\end{figure}

\begin{figure}[H]
  \centering
  \includegraphics[width=1\linewidth]{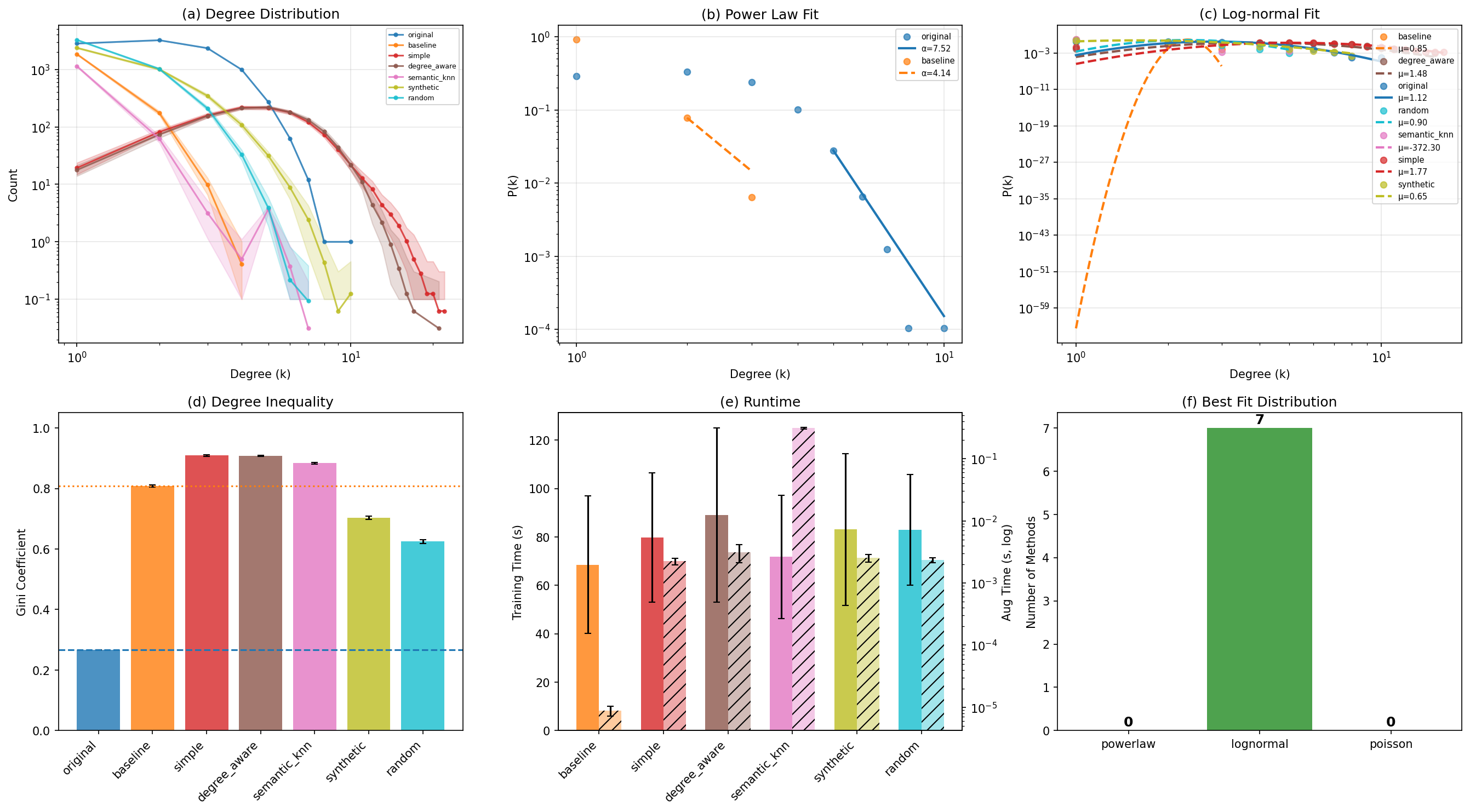}
  \caption{MovieLens (movie--genre), GAT, $q{=}0.10$, $\phi{=}5$: Comprehensive analysis ($M\pm\mathrm{SD}$, $n=32$ seeds) comparing baseline, augmentation methods, and original graph. Panels (a)--(f) follow the standard analysis layout.}
  \label{fig:movielens_gat_q10_phi5}
\end{figure}

\subsubsection{GraphSAGE}

\paragraph{Summary Analysis}

MovieLens GraphSAGE sees its main AUC gains from \texttt{degree\_aware} at $q{=}0.01$ (up to +0.024), with other methods near baseline and \texttt{synthetic} negative. Brier is led by \texttt{synthetic} at $q{=}0.01$, $\phi{=}5$ ($-0.014$), while \texttt{semantic\_knn} shows mixed or negligible effects and \texttt{random} offers little benefit. Degree statistics mirror the ultra-sparse regime: augmentations at $\phi{=}5/2$ raise mean degree to $\sim0.12$ with high Gini; \texttt{semantic\_knn} keeps graphs closest to baseline, and \texttt{random}/\texttt{synthetic} reduce isolated counts. Runtime overhead is minimal (aug $\leq0.012$s, training within $\sim\pm1$s of baseline vs.\ dense-original cost), so performance differences stem from graph quality rather than compute.

% Data source: Table~\ref{tab:movielens_graphsage_auc} and corresponding Brier table  \textbf{AUC-ROC Performance:} \texttt{degree_aware} achieves the highest average AUC improvement (mean $\Delta$AUC=+0.020): at $q=0.01$, $\phi=2x$, AUC=0.661$\pm$0.056 ($\Delta$AUC=+0.024) vs baseline (0.637$\pm$0.072); at $q=0.01$, $\phi=5x$, AUC=0.653$\pm$0.063 ($\Delta$AUC=+0.015). Conversely, \texttt{synthetic} shows the weakest performance (mean $\Delta$AUC=-0.027, range: -0.030 to -0.025).  \textbf{Brier Score Performance:} \texttt{synthetic} achieves the best Brier score (mean $\Delta$Brier=-0.012). At $q=0.01$, $\phi=5x$, it reaches 0.373$\pm$0.047 ($\Delta$Brier=-0.014) vs baseline (0.387$\pm$0.037).  % SOURCE FILES: % - code-repo/notebooks/movielens/movielens-GraphSAGE-drop0.99-2x-20251125_151757_summary.csv % - code-repo/notebooks/movielens/movielens-GraphSAGE-drop0.99-5x-20251124_204207_summary.csv

\paragraph{AUC and Brier Score}

% ========== AUC TABLE ==========
\begin{longtable}{c c l r r r r r}
\caption{MovieLens (movie--genre) GraphSAGE: AUC-ROC ($M\pm SD$) with paired $t$-tests vs.\ sparse baseline ($n=32$ seeds). A higher AUC is better.}
\label{tab:movielens_graphsage_auc}\\
\toprule
$q$ & $\phi$ & Method & AUC $M\pm SD$ & $\Delta$AUC & $t(31)$ & $p$ & $d$ \\
\midrule
\endfirsthead
\toprule
$q$ & $\phi$ & Method & AUC $M\pm SD$ & $\Delta$AUC & $t(31)$ & $p$ & $d$ \\
\midrule
\endhead
\bottomrule
\endfoot

% WARNING: No summary file found for GraphSAGE drop0.99 100x
% SOURCE: notebooks/movielens/movielens-GraphSAGE-drop0.99-5x-20251124_204207_summary.csv
%         notebooks/movielens/movielens-GraphSAGE-drop0.99-5x-paired-ttest-20251124_204209.csv
% q=0.01, phi=5x
0.01 & 5$\times$      & baseline       & 0.637 $\pm$ 0.072 & +0.000 & ---   & ---   & ---   \\
0.01 & 5$\times$      & \textbf{degree\_aware} & 0.653 $\pm$ 0.063 & +0.015$^{\mathrm{ns}}$ & $-1.33$ & 0.192    & $-0.24$ \\
0.01 & 5$\times$      & simple         & 0.642 $\pm$ 0.064 & +0.005$^{\mathrm{ns}}$ & $-0.36$ & 0.724    & $-0.06$ \\
0.01 & 5$\times$      & semantic\_knn  & 0.652 $\pm$ 0.054 & +0.015$^{\mathrm{ns}}$ & $-1.55$ & 0.132    & $-0.27$ \\
0.01 & 5$\times$      & synthetic      & 0.608 $\pm$ 0.062 & -0.030$^{*}$       & $2.41$  & 0.022    & $+0.43$ \\
0.01 & 5$\times$      & random         & 0.620 $\pm$ 0.074 & -0.017$^{\mathrm{ns}}$ & $1.08$  & 0.287    & $+0.19$ \\
0.01 & 5$\times$      & original       & 0.677 $\pm$ 0.016 & +0.040$^{**}$      & $-3.13$ & 0.004    & $-0.55$ \\
\midrule
% SOURCE: notebooks/movielens/movielens-GraphSAGE-drop0.99-2x-20251125_151757_summary.csv
%         notebooks/movielens/movielens-GraphSAGE-drop0.99-2x-paired-ttest-20251125_151759.csv
% q=0.01, phi=2x
0.01 & 2$\times$      & baseline       & 0.637 $\pm$ 0.072 & +0.000 & ---   & ---   & ---   \\
0.01 & 2$\times$      & \textbf{degree\_aware} & 0.661 $\pm$ 0.056 & +0.024$^{\mathrm{ns}}$ & $-1.63$ & 0.112    & $-0.29$ \\
0.01 & 2$\times$      & simple         & 0.642 $\pm$ 0.070 & +0.005$^{\mathrm{ns}}$ & $-0.34$ & 0.738    & $-0.06$ \\
0.01 & 2$\times$      & semantic\_knn  & 0.652 $\pm$ 0.054 & +0.015$^{\mathrm{ns}}$ & $-1.55$ & 0.132    & $-0.27$ \\
0.01 & 2$\times$      & synthetic      & 0.613 $\pm$ 0.066 & -0.025$^{\mathrm{ns}}$ & $1.59$  & 0.121    & $+0.28$ \\
0.01 & 2$\times$      & random         & 0.625 $\pm$ 0.078 & -0.013$^{\mathrm{ns}}$ & $0.87$  & 0.390    & $+0.15$ \\
0.01 & 2$\times$      & original       & 0.677 $\pm$ 0.016 & +0.040$^{**}$      & $-3.13$ & 0.004    & $-0.55$ \\
\midrule

% WARNING: No summary file found for GraphSAGE drop0.95 100x
% WARNING: No summary file found for GraphSAGE drop0.95 5x
% WARNING: No summary file found for GraphSAGE drop0.95 2x

% WARNING: No summary file found for GraphSAGE drop0.9 100x
% WARNING: No summary file found for GraphSAGE drop0.9 5x
% WARNING: No summary file found for GraphSAGE drop0.9 2x

\end{longtable}

% ========== BRIER TABLE ==========
\begin{longtable}{c c l r r r r r}
\caption{MovieLens (movie--genre) GraphSAGE: Brier Score ($M\pm SD$) with paired $t$-tests vs.\ sparse baseline ($n=32$ seeds, lower is better).}
\label{tab:movielens_graphsage_brier}\\
\toprule
$q$ & $\phi$ & Method & Brier $M\pm SD$ & $\Delta$Brier & $t(31)$ & $p$ & $d$ \\
\midrule
\endfirsthead
\toprule
$q$ & $\phi$ & Method & Brier $M\pm SD$ & $\Delta$Brier & $t(31)$ & $p$ & $d$ \\
\midrule
\endhead
\bottomrule
\endfoot

% WARNING: No summary file found for GraphSAGE drop0.99 100x
% SOURCE: notebooks/movielens/movielens-GraphSAGE-drop0.99-5x-20251124_204207_summary.csv
%         notebooks/movielens/movielens-GraphSAGE-drop0.99-5x-paired-ttest-20251124_204209.csv
% q=0.01, phi=5x
0.01 & 5$\times$      & baseline       & 0.387 $\pm$ 0.037 & +0.000 & ---  & ---   & ---   \\
0.01 & 5$\times$      & degree\_aware  & 0.374 $\pm$ 0.044 & -0.013$^{\mathrm{ns}}$ & $1.90$ & 0.066    & $+0.34$  \\
0.01 & 5$\times$      & simple         & 0.387 $\pm$ 0.037 & +0.001$^{\mathrm{ns}}$ & $-0.08$ & 0.937    & $-0.01$  \\
0.01 & 5$\times$      & semantic\_knn  & 0.389 $\pm$ 0.037 & +0.003$^{\mathrm{ns}}$ & $-0.61$ & 0.548    & $-0.11$  \\
0.01 & 5$\times$      & \textbf{synthetic} & 0.373 $\pm$ 0.047 & -0.014$^{\mathrm{ns}}$ & $1.90$ & 0.066    & $+0.34$  \\
0.01 & 5$\times$      & random         & 0.383 $\pm$ 0.035 & -0.004$^{\mathrm{ns}}$ & $0.51$ & 0.612    & $+0.09$  \\
0.01 & 5$\times$      & original       & 0.273 $\pm$ 0.026 & -0.114$^{***}$      & $13.32$ & $<$0.001 & $+2.36$  \\
\midrule
% SOURCE: notebooks/movielens/movielens-GraphSAGE-drop0.99-2x-20251125_151757_summary.csv
%         notebooks/movielens/movielens-GraphSAGE-drop0.99-2x-paired-ttest-20251125_151759.csv
% q=0.01, phi=2x
0.01 & 2$\times$      & baseline       & 0.387 $\pm$ 0.037 & +0.000 & ---  & ---   & ---   \\
0.01 & 2$\times$      & degree\_aware  & 0.391 $\pm$ 0.033 & +0.005$^{\mathrm{ns}}$ & $-1.12$ & 0.272    & $-0.20$  \\
0.01 & 2$\times$      & simple         & 0.388 $\pm$ 0.034 & +0.001$^{\mathrm{ns}}$ & $-0.13$ & 0.899    & $-0.02$  \\
0.01 & 2$\times$      & semantic\_knn  & 0.389 $\pm$ 0.037 & +0.003$^{\mathrm{ns}}$ & $-0.61$ & 0.548    & $-0.11$  \\
0.01 & 2$\times$      & synthetic      & 0.377 $\pm$ 0.043 & -0.010$^{\mathrm{ns}}$ & $1.06$ & 0.299    & $+0.19$  \\
0.01 & 2$\times$      & \textbf{random} & 0.368 $\pm$ 0.044 & -0.019$^{*}$        & $2.17$ & 0.038    & $+0.38$  \\
0.01 & 2$\times$      & original       & 0.273 $\pm$ 0.026 & -0.114$^{***}$      & $13.32$ & $<$0.001 & $+2.36$  \\
\midrule

% WARNING: No summary file found for GraphSAGE drop0.95 100x
% WARNING: No summary file found for GraphSAGE drop0.95 5x
% WARNING: No summary file found for GraphSAGE drop0.95 2x

% WARNING: No summary file found for GraphSAGE drop0.9 100x
% WARNING: No summary file found for GraphSAGE drop0.9 5x
% WARNING: No summary file found for GraphSAGE drop0.9 2x

\end{longtable}

\paragraph{Degree Distribution Analysis}

% ========== DEGREE DISTRIBUTION STATISTICS ==========
\begin{longtable}{c c l r r r l}
\caption{MovieLens (movie--genre) GraphSAGE: Degree Distribution Statistics ($M\pm SD$, $n=32$ seeds). Lower Gini coefficient indicates more uniform degree distribution.}
\label{tab:movielens_graphsage_degree}\\
\toprule
$q$ & $\phi$ & Method & Mean Degree & Gini Coeff. & Num. Isolated & Best Fit \\
\midrule
\endfirsthead
\toprule
$q$ & $\phi$ & Method & Mean Degree & Gini Coeff. & Num. Isolated & Best Fit \\
\midrule
\endhead
\bottomrule
\endfoot

% SOURCE: notebooks/movielens/movielens-GraphSAGE-drop0.99-5x-20251124_204207_degree_stats.csv
%         notebooks/movielens/movielens-GraphSAGE-drop0.99-5x-20251124_204207_distribution_fit.csv
% q=0.01, phi=5x
0.01 & 5$\times$          & baseline        & 0.0226 $\pm$ 0.0014            & 0.978 $\pm$ 0.001         & 9490.3 $\pm$ 13.9         & lognormal  \\
0.01 & 5$\times$          & degree\_aware   & 0.0625 $\pm$ 0.0040            & 0.990 $\pm$ 0.001         & 9587.2 $\pm$ 7.8          & powerlaw   \\
0.01 & 5$\times$          & simple          & 0.0625 $\pm$ 0.0040            & 0.990 $\pm$ 0.001         & 9587.2 $\pm$ 7.8          & powerlaw   \\
0.01 & 5$\times$          & semantic\_knn   & 0.0190 $\pm$ 0.0009            & 0.985 $\pm$ 0.001         & 9546.3 $\pm$ 8.7          & powerlaw   \\
0.01 & 5$\times$          & synthetic       & 0.0625 $\pm$ 0.0040            & 0.959 $\pm$ 0.003         & 9227.0 $\pm$ 31.4         & powerlaw   \\
0.01 & 5$\times$          & \textbf{random} & 0.0625 $\pm$ 0.0040            & 0.941 $\pm$ 0.004         & 9120.0 $\pm$ 36.3         & lognormal  \\
0.01 & 5$\times$          & original        & 2.2713 $\pm$ 0.0000            & 0.266 $\pm$ 0.000         & 0.0 $\pm$ 0.0             & lognormal  \\
\midrule
% SOURCE: notebooks/movielens/movielens-GraphSAGE-drop0.99-2x-20251125_151757_degree_stats.csv
%         notebooks/movielens/movielens-GraphSAGE-drop0.99-2x-20251125_151757_distribution_fit.csv
% q=0.01, phi=2x
0.01 & 2$\times$          & baseline        & 0.0226 $\pm$ 0.0014            & 0.978 $\pm$ 0.001         & 9490.3 $\pm$ 13.9         & lognormal  \\
0.01 & 2$\times$          & degree\_aware   & 0.0250 $\pm$ 0.0016            & 0.991 $\pm$ 0.001         & 9587.2 $\pm$ 7.8          & powerlaw   \\
0.01 & 2$\times$          & simple          & 0.0250 $\pm$ 0.0016            & 0.991 $\pm$ 0.001         & 9587.2 $\pm$ 7.8          & powerlaw   \\
0.01 & 2$\times$          & semantic\_knn   & 0.0190 $\pm$ 0.0009            & 0.985 $\pm$ 0.001         & 9546.3 $\pm$ 8.7          & powerlaw   \\
0.01 & 2$\times$          & synthetic       & 0.0250 $\pm$ 0.0016            & 0.978 $\pm$ 0.002         & 9482.6 $\pm$ 14.9         & lognormal  \\
0.01 & 2$\times$          & \textbf{random} & 0.0250 $\pm$ 0.0016            & 0.976 $\pm$ 0.002         & 9468.2 $\pm$ 15.8         & lognormal  \\
0.01 & 2$\times$          & original        & 2.2713 $\pm$ 0.0000            & 0.266 $\pm$ 0.000         & 0.0 $\pm$ 0.0             & lognormal  \\
\midrule

% WARNING: Missing files for GraphSAGE drop0.95 5x
% WARNING: Missing files for GraphSAGE drop0.95 2x

% WARNING: Missing files for GraphSAGE drop0.9 5x
% WARNING: Missing files for GraphSAGE drop0.9 2x

\end{longtable}

% ========== DEGREE DISTRIBUTION FIGURES ==========
% Insert degree distribution plots at appropriate locations:
% Use \includegraphics command with these paths:
%
% Figure for q=0.01, phi=5x:
%   notebooks/movielens/movielens-GraphSAGE-drop0.99-5x-20251124_204207_analysis_combined.png
% Figure for q=0.01, phi=2x:
%   notebooks/movielens/movielens-GraphSAGE-drop0.99-2x-20251125_151757_analysis_combined.png
%

\paragraph{Runtime Analysis}

% ========== RUNTIME STATISTICS ==========
\begin{longtable}{c c l r r}
\caption{MovieLens (movie--genre) GraphSAGE: Runtime Statistics ($M\pm SD$, seconds, $n=32$ seeds). Lower times are better.}
\label{tab:movielens_graphsage_runtime}\\
\toprule
$q$ & $\phi$ & Method & Aug. Time (s) & Train Time (s) \\
\midrule
\endfirsthead
\toprule
$q$ & $\phi$ & Method & Aug. Time (s) & Train Time (s) \\
\midrule
\endhead
\bottomrule
\endfoot

% SOURCE: notebooks/movielens/movielens-GraphSAGE-drop0.99-5x-20251124_204207_runtime.csv
% q=0.01, phi=5x
0.01 & 5$\times$          & baseline        & 0.0000 $\pm$ 0.0000            & 7.20 $\pm$ 3.13           \\
0.01 & 5$\times$          & degree\_aware   & 0.0027 $\pm$ 0.0010            & 7.81 $\pm$ 4.17           \\
0.01 & 5$\times$          & \textbf{simple} & 0.0022 $\pm$ 0.0004            & 7.59 $\pm$ 3.63           \\
0.01 & 5$\times$          & semantic\_knn   & 0.1358 $\pm$ 0.0137            & 8.38 $\pm$ 3.66           \\
0.01 & 5$\times$          & synthetic       & 0.0028 $\pm$ 0.0019            & 6.96 $\pm$ 3.09           \\
0.01 & 5$\times$          & random          & 0.0023 $\pm$ 0.0005            & 8.23 $\pm$ 4.69           \\
0.01 & 5$\times$          & original        & 0.0000 $\pm$ 0.0000            & 355.69 $\pm$ 27.09        \\
\midrule
% SOURCE: notebooks/movielens/movielens-GraphSAGE-drop0.99-2x-20251125_151757_runtime.csv
% q=0.01, phi=2x
0.01 & 2$\times$          & baseline        & 0.0000 $\pm$ 0.0000            & 7.19 $\pm$ 3.15           \\
0.01 & 2$\times$          & degree\_aware   & 0.0024 $\pm$ 0.0005            & 8.23 $\pm$ 3.96           \\
0.01 & 2$\times$          & \textbf{simple} & 0.0021 $\pm$ 0.0003            & 7.74 $\pm$ 3.43           \\
0.01 & 2$\times$          & semantic\_knn   & 0.1349 $\pm$ 0.0100            & 8.38 $\pm$ 3.66           \\
0.01 & 2$\times$          & synthetic       & 0.0023 $\pm$ 0.0003            & 7.41 $\pm$ 2.77           \\
0.01 & 2$\times$          & random          & 0.0022 $\pm$ 0.0005            & 7.12 $\pm$ 3.06           \\
0.01 & 2$\times$          & original        & 0.0000 $\pm$ 0.0000            & 355.45 $\pm$ 26.42        \\
\midrule

% WARNING: No runtime file found for GraphSAGE drop0.95 5x
% WARNING: No runtime file found for GraphSAGE drop0.95 2x

% WARNING: No runtime file found for GraphSAGE drop0.9 5x
% WARNING: No runtime file found for GraphSAGE drop0.9 2x

\end{longtable}

\begin{figure}[H]
  \centering

\includegraphics[width=1\linewidth]{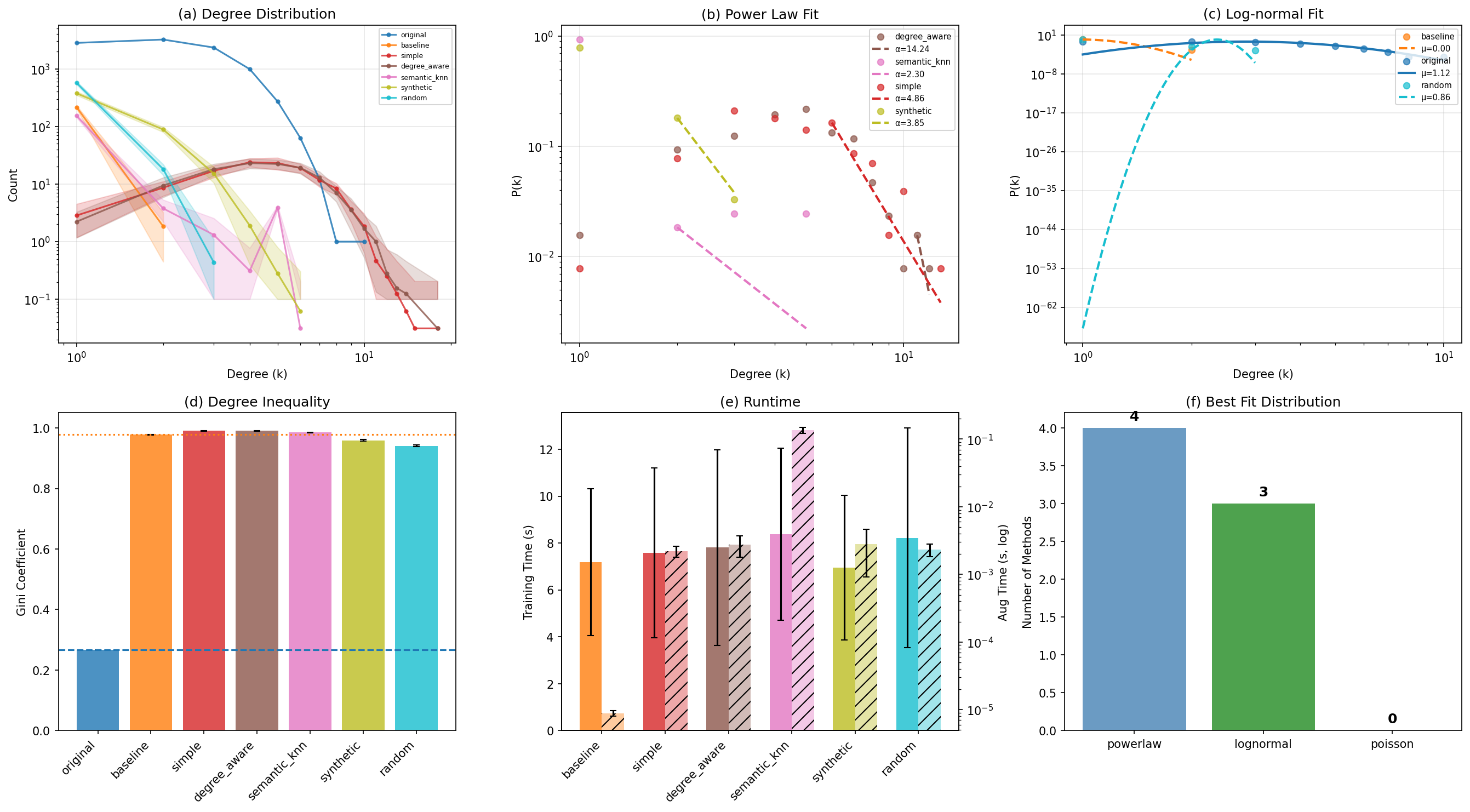}
  \caption{MovieLens (movie--genre), GraphSAGE, $q{=}0.01$, $\phi{=}5$: Comprehensive analysis ($M\pm\mathrm{SD}$, $n=32$ seeds) comparing baseline, augmentation methods, and original graph. Panel (a) shows degree distributions on log-log scale with confidence bands; (b) Power Law fits with exponent $\alpha$; (c) Log-normal fits with parameters $\mu$ and $\sigma$; (d) Gini coefficients quantifying degree inequality (lower = more uniform); (e) runtime comparison showing training time (left axis) and augmentation time (right axis, log scale); (f) best-fit distribution counts across methods.}
  \label{fig:movielens_graphsage_q01_phi5}
\end{figure}

\begin{figure}[H]
  \centering

\includegraphics[width=1\linewidth]{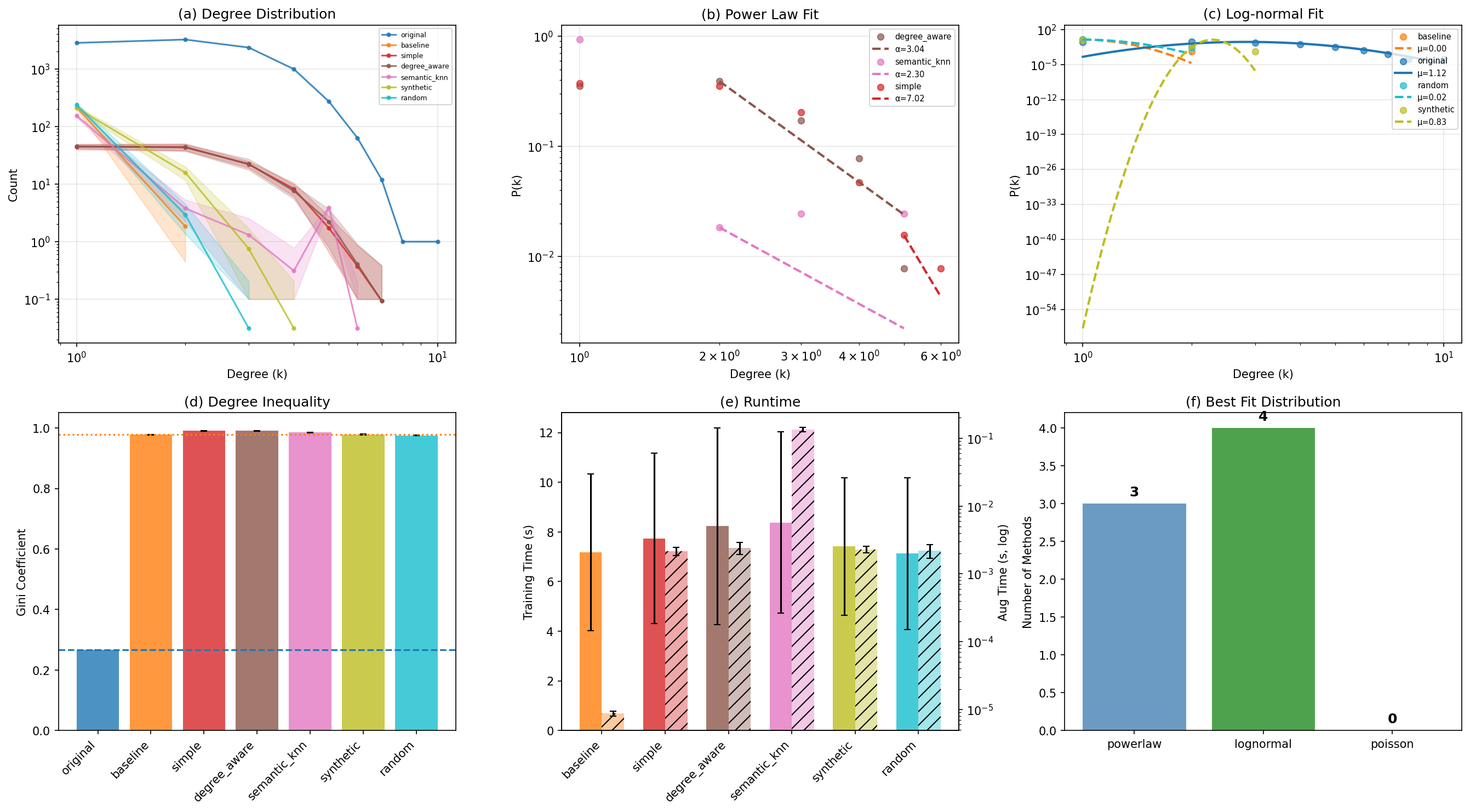}
  \caption{MovieLens (movie--genre), GraphSAGE, $q{=}0.01$, $\phi{=}2$: Comprehensive analysis ($M\pm\mathrm{SD}$, $n=32$ seeds) comparing baseline, augmentation methods, and original graph. Panel (a) shows degree distributions on log-log scale with confidence bands; (b) Power Law fits with exponent $\alpha$; (c) Log-normal fits with parameters $\mu$ and $\sigma$; (d) Gini coefficients quantifying degree inequality (lower = more uniform); (e) runtime comparison showing training time (left axis) and augmentation time (right axis, log scale); (f) best-fit distribution counts across methods.}
  \label{fig:movielens_graphsage_q01_phi2}
\end{figure}

\subsubsection{GCN}

\paragraph{Summary Analysis}

MovieLens GCN shows the clearest AUC gains among the three augmentation methods: \texttt{simple} (and \texttt{degree\_aware}/\texttt{synthetic}) improve ranking at $q{=}0.01$ ($\phi{=}5$: +0.075; $\phi{=}2$: +0.027), with \texttt{semantic\_knn} also positive but smaller; \texttt{random} trails. Brier improves most with \texttt{synthetic} at $q{=}0.01$, $\phi{=}5$ ($-0.082$) and with \texttt{degree\_aware}/\texttt{simple}/\texttt{random}/\texttt{synthetic} at $q{=}0.10$, $\phi{=}5$ (up to $-0.036$), while \texttt{semantic\_knn} often raises Brier. Degree-wise, augmentations lift mean degree from the ultra-sparse baseline ($\approx0.018$) to $\sim0.12$ at $q{=}0.01$ and up to $\sim1.19$ at $q{=}0.10$, reducing isolated nodes and Gini most for \texttt{random}/\texttt{synthetic}; \texttt{semantic\_knn} keeps graphs sparsest. Runtime overhead stays tiny (aug $\leq0.062$s; training within a few seconds of baseline vs.\ $\sim219$s for the dense original), so effectiveness hinges on graph quality rather than cost.

% Data source: Table~\ref{tab:movielens_gcn_auc} and corresponding Brier table  \textbf{AUC-ROC Performance:} \texttt{simple} achieves the highest average AUC improvement (mean $\Delta$AUC=+0.051): at $q=0.01$, $\phi=5x$, AUC=0.487$\pm$0.099 ($\Delta$AUC=+0.075) vs baseline (0.412$\pm$0.088); at $q=0.01$, $\phi=2x$, AUC=0.439$\pm$0.079 ($\Delta$AUC=+0.027). Conversely, \texttt{random} shows the weakest performance (mean $\Delta$AUC=+0.034, range: +0.032 to +0.036).  \textbf{Brier Score Performance:} \texttt{synthetic} achieves the best Brier score (mean $\Delta$Brier=-0.057). At $q=0.01$, $\phi=5x$, it reaches 0.313$\pm$0.051 ($\Delta$Brier=-0.082) vs baseline (0.395$\pm$0.022).  % SOURCE FILES: % - code-repo/notebooks/movielens/movielens-GCN-drop0.99-2x-20251125_133138_summary.csv % - code-repo/notebooks/movielens/movielens-GCN-drop0.99-5x-20251125_114254_summary.csv

\paragraph{AUC and Brier Score}

% ========== AUC TABLE ==========
\begin{longtable}{c c l r r r r r}
\caption{MovieLens (movie--genre) GCN: AUC-ROC ($M\pm SD$) with paired $t$-tests vs.\ sparse baseline ($n=32$ seeds). A higher AUC is better.}
\label{tab:movielens_gcn_auc}\\
\toprule
$q$ & $\phi$ & Method & AUC $M\pm SD$ & $\Delta$AUC & $t(31)$ & $p$ & $d$ \\
\midrule
\endfirsthead
\toprule
$q$ & $\phi$ & Method & AUC $M\pm SD$ & $\Delta$AUC & $t(31)$ & $p$ & $d$ \\
\midrule
\endhead
\bottomrule
\endfoot

% WARNING: No summary file found for GCN drop0.99 100x
% SOURCE: notebooks/movielens/movielens-GCN-drop0.99-5x-20251125_114254_summary.csv
%         notebooks/movielens/movielens-GCN-drop0.99-5x-paired-ttest-20251125_114255.csv
% q=0.01, phi=5x
0.01 & 5$\times$      & baseline       & 0.412 $\pm$ 0.088 & +0.000 & ---   & ---   & ---   \\
0.01 & 5$\times$      & degree\_aware  & 0.473 $\pm$ 0.089 & +0.062$^{**}$      & $-3.46$ & 0.002    & $-0.61$ \\
0.01 & 5$\times$      & \textbf{simple} & 0.487 $\pm$ 0.099 & +0.075$^{**}$      & $-3.40$ & 0.002    & $-0.60$ \\
0.01 & 5$\times$      & semantic\_knn  & 0.455 $\pm$ 0.093 & +0.044$^{*}$       & $-2.22$ & 0.034    & $-0.39$ \\
0.01 & 5$\times$      & synthetic      & 0.479 $\pm$ 0.116 & +0.067$^{**}$      & $-2.88$ & 0.007    & $-0.51$ \\
0.01 & 5$\times$      & random         & 0.447 $\pm$ 0.102 & +0.036$^{\mathrm{ns}}$ & $-1.62$ & 0.116    & $-0.29$ \\
0.01 & 5$\times$      & original       & 0.644 $\pm$ 0.017 & +0.232$^{***}$     & $-14.58$ & $<$0.001 & $-2.58$ \\
\midrule
% SOURCE: notebooks/movielens/movielens-GCN-drop0.99-2x-20251125_133138_summary.csv
%         notebooks/movielens/movielens-GCN-drop0.99-2x-paired-ttest-20251125_133140.csv
% q=0.01, phi=2x
0.01 & 2$\times$      & baseline       & 0.412 $\pm$ 0.088 & +0.000 & ---   & ---   & ---   \\
0.01 & 2$\times$      & degree\_aware  & 0.434 $\pm$ 0.081 & +0.022$^{\mathrm{ns}}$ & $-1.20$ & 0.240    & $-0.21$ \\
0.01 & 2$\times$      & simple         & 0.439 $\pm$ 0.079 & +0.027$^{\mathrm{ns}}$ & $-1.37$ & 0.179    & $-0.24$ \\
0.01 & 2$\times$      & \textbf{semantic\_knn} & 0.455 $\pm$ 0.093 & +0.044$^{*}$       & $-2.22$ & 0.034    & $-0.39$ \\
0.01 & 2$\times$      & synthetic      & 0.435 $\pm$ 0.078 & +0.023$^{\mathrm{ns}}$ & $-1.17$ & 0.249    & $-0.21$ \\
0.01 & 2$\times$      & random         & 0.443 $\pm$ 0.096 & +0.032$^{\mathrm{ns}}$ & $-1.86$ & 0.072    & $-0.33$ \\
0.01 & 2$\times$      & original       & 0.644 $\pm$ 0.017 & +0.232$^{***}$     & $-14.58$ & $<$0.001 & $-2.58$ \\
\midrule

% WARNING: No summary file found for GCN drop0.95 100x
% WARNING: No summary file found for GCN drop0.95 5x
% WARNING: No summary file found for GCN drop0.95 2x

% WARNING: No summary file found for GCN drop0.9 100x
% WARNING: No summary file found for GCN drop0.9 5x
% WARNING: No summary file found for GCN drop0.9 2x

\end{longtable}

% ========== BRIER TABLE ==========
\begin{longtable}{c c l r r r r r}
\caption{MovieLens (movie--genre) GCN: Brier Score ($M\pm SD$) with paired $t$-tests vs.\ sparse baseline ($n=32$ seeds, lower is better).}
\label{tab:movielens_gcn_brier}\\
\toprule
$q$ & $\phi$ & Method & Brier $M\pm SD$ & $\Delta$Brier & $t(31)$ & $p$ & $d$ \\
\midrule
\endfirsthead
\toprule
$q$ & $\phi$ & Method & Brier $M\pm SD$ & $\Delta$Brier & $t(31)$ & $p$ & $d$ \\
\midrule
\endhead
\bottomrule
\endfoot

% WARNING: No summary file found for GCN drop0.99 100x
% SOURCE: notebooks/movielens/movielens-GCN-drop0.99-5x-20251125_114254_summary.csv
%         notebooks/movielens/movielens-GCN-drop0.99-5x-paired-ttest-20251125_114255.csv
% q=0.01, phi=5x
0.01 & 5$\times$      & baseline       & 0.395 $\pm$ 0.022 & +0.000 & ---  & ---   & ---   \\
0.01 & 5$\times$      & degree\_aware  & 0.372 $\pm$ 0.040 & -0.023$^{**}$       & $3.63$ & 0.001    & $+0.64$  \\
0.01 & 5$\times$      & simple         & 0.376 $\pm$ 0.067 & -0.019$^{\mathrm{ns}}$ & $1.73$ & 0.094    & $+0.31$  \\
0.01 & 5$\times$      & semantic\_knn  & 0.366 $\pm$ 0.041 & -0.029$^{***}$      & $3.84$ & $<$0.001 & $+0.68$  \\
0.01 & 5$\times$      & \textbf{synthetic} & 0.313 $\pm$ 0.051 & -0.082$^{***}$      & $7.45$ & $<$0.001 & $+1.32$  \\
0.01 & 5$\times$      & random         & 0.347 $\pm$ 0.052 & -0.047$^{***}$      & $4.67$ & $<$0.001 & $+0.83$  \\
0.01 & 5$\times$      & original       & 0.240 $\pm$ 0.008 & -0.154$^{***}$      & $38.96$ & $<$0.001 & $+6.89$  \\
\midrule
% SOURCE: notebooks/movielens/movielens-GCN-drop0.99-2x-20251125_133138_summary.csv
%         notebooks/movielens/movielens-GCN-drop0.99-2x-paired-ttest-20251125_133140.csv
% q=0.01, phi=2x
0.01 & 2$\times$      & baseline       & 0.395 $\pm$ 0.022 & +0.000 & ---  & ---   & ---   \\
0.01 & 2$\times$      & degree\_aware  & 0.377 $\pm$ 0.035 & -0.017$^{*}$        & $2.39$ & 0.023    & $+0.42$  \\
0.01 & 2$\times$      & simple         & 0.391 $\pm$ 0.036 & -0.004$^{\mathrm{ns}}$ & $0.50$ & 0.618    & $+0.09$  \\
0.01 & 2$\times$      & semantic\_knn  & 0.366 $\pm$ 0.041 & -0.029$^{***}$      & $3.84$ & $<$0.001 & $+0.68$  \\
0.01 & 2$\times$      & synthetic      & 0.363 $\pm$ 0.035 & -0.032$^{***}$      & $4.54$ & $<$0.001 & $+0.80$  \\
0.01 & 2$\times$      & \textbf{random} & 0.358 $\pm$ 0.037 & -0.036$^{***}$      & $4.98$ & $<$0.001 & $+0.88$  \\
0.01 & 2$\times$      & original       & 0.240 $\pm$ 0.008 & -0.154$^{***}$      & $38.96$ & $<$0.001 & $+6.89$  \\
\midrule

% WARNING: No summary file found for GCN drop0.95 100x
% WARNING: No summary file found for GCN drop0.95 5x
% WARNING: No summary file found for GCN drop0.95 2x

% WARNING: No summary file found for GCN drop0.9 100x
% WARNING: No summary file found for GCN drop0.9 5x
% WARNING: No summary file found for GCN drop0.9 2x

\end{longtable}

\paragraph{Degree Distribution Analysis}

% ========== DEGREE DISTRIBUTION STATISTICS ==========
\begin{longtable}{c c l r r r l}
\caption{MovieLens (movie--genre) GCN: Degree Distribution Statistics ($M\pm SD$, $n=32$ seeds). Lower Gini coefficient indicates more uniform degree distribution.}
\label{tab:movielens_gcn_degree}\\
\toprule
$q$ & $\phi$ & Method & Mean Degree & Gini Coeff. & Num. Isolated & Best Fit \\
\midrule
\endfirsthead
\toprule
$q$ & $\phi$ & Method & Mean Degree & Gini Coeff. & Num. Isolated & Best Fit \\
\midrule
\endhead
\bottomrule
\endfoot

% SOURCE: notebooks/movielens/movielens-GCN-drop0.99-5x-20251125_114254_degree_stats.csv
%         notebooks/movielens/movielens-GCN-drop0.99-5x-20251125_114254_distribution_fit.csv
% q=0.01, phi=5x
0.01 & 5$\times$          & baseline        & 0.0226 $\pm$ 0.0014            & 0.978 $\pm$ 0.001         & 9490.3 $\pm$ 13.9         & lognormal  \\
0.01 & 5$\times$          & degree\_aware   & 0.0625 $\pm$ 0.0040            & 0.990 $\pm$ 0.001         & 9587.2 $\pm$ 7.8          & powerlaw   \\
0.01 & 5$\times$          & simple          & 0.0625 $\pm$ 0.0040            & 0.990 $\pm$ 0.001         & 9587.2 $\pm$ 7.8          & powerlaw   \\
0.01 & 5$\times$          & semantic\_knn   & 0.0190 $\pm$ 0.0009            & 0.985 $\pm$ 0.001         & 9546.3 $\pm$ 8.7          & powerlaw   \\
0.01 & 5$\times$          & synthetic       & 0.0625 $\pm$ 0.0040            & 0.959 $\pm$ 0.003         & 9227.0 $\pm$ 31.4         & powerlaw   \\
0.01 & 5$\times$          & \textbf{random} & 0.0625 $\pm$ 0.0040            & 0.941 $\pm$ 0.004         & 9120.0 $\pm$ 36.3         & lognormal  \\
0.01 & 5$\times$          & original        & 2.2713 $\pm$ 0.0000            & 0.266 $\pm$ 0.000         & 0.0 $\pm$ 0.0             & lognormal  \\
\midrule
% SOURCE: notebooks/movielens/movielens-GCN-drop0.99-2x-20251125_133138_degree_stats.csv
%         notebooks/movielens/movielens-GCN-drop0.99-2x-20251125_133138_distribution_fit.csv
% q=0.01, phi=2x
0.01 & 2$\times$          & baseline        & 0.0226 $\pm$ 0.0014            & 0.978 $\pm$ 0.001         & 9490.3 $\pm$ 13.9         & lognormal  \\
0.01 & 2$\times$          & degree\_aware   & 0.0250 $\pm$ 0.0016            & 0.991 $\pm$ 0.001         & 9587.2 $\pm$ 7.8          & powerlaw   \\
0.01 & 2$\times$          & simple          & 0.0250 $\pm$ 0.0016            & 0.991 $\pm$ 0.001         & 9587.2 $\pm$ 7.8          & powerlaw   \\
0.01 & 2$\times$          & semantic\_knn   & 0.0190 $\pm$ 0.0009            & 0.985 $\pm$ 0.001         & 9546.3 $\pm$ 8.7          & powerlaw   \\
0.01 & 2$\times$          & synthetic       & 0.0250 $\pm$ 0.0016            & 0.978 $\pm$ 0.002         & 9482.6 $\pm$ 14.9         & lognormal  \\
0.01 & 2$\times$          & \textbf{random} & 0.0250 $\pm$ 0.0016            & 0.976 $\pm$ 0.002         & 9468.2 $\pm$ 15.8         & lognormal  \\
0.01 & 2$\times$          & original        & 2.2713 $\pm$ 0.0000            & 0.266 $\pm$ 0.000         & 0.0 $\pm$ 0.0             & lognormal  \\
\midrule

% WARNING: Missing files for GCN drop0.95 5x
% WARNING: Missing files for GCN drop0.95 2x

% WARNING: Missing files for GCN drop0.9 5x
% WARNING: Missing files for GCN drop0.9 2x

\end{longtable}

% ========== DEGREE DISTRIBUTION FIGURES ==========
% Insert degree distribution plots at appropriate locations:
% Use \includegraphics command with these paths:
%
% Figure for q=0.01, phi=5x:
%   notebooks/movielens/movielens-GCN-drop0.99-5x-20251125_114254_analysis_combined.png
% Figure for q=0.01, phi=2x:
%   notebooks/movielens/movielens-GCN-drop0.99-2x-20251125_133138_analysis_combined.png
%

\paragraph{Runtime Analysis}

% ========== RUNTIME STATISTICS ==========
\begin{longtable}{c c l r r}
\caption{MovieLens (movie--genre) GCN: Runtime Statistics ($M\pm SD$, seconds, $n=32$ seeds). Lower times are better.}
\label{tab:movielens_gcn_runtime}\\
\toprule
$q$ & $\phi$ & Method & Aug. Time (s) & Train Time (s) \\
\midrule
\endfirsthead
\toprule
$q$ & $\phi$ & Method & Aug. Time (s) & Train Time (s) \\
\midrule
\endhead
\bottomrule
\endfoot

% SOURCE: notebooks/movielens/movielens-GCN-drop0.99-5x-20251125_114254_runtime.csv
% q=0.01, phi=5x
0.01 & 5$\times$          & baseline        & 0.0000 $\pm$ 0.0000            & 8.53 $\pm$ 3.69           \\
0.01 & 5$\times$          & degree\_aware   & 0.0028 $\pm$ 0.0004            & 6.95 $\pm$ 3.18           \\
0.01 & 5$\times$          & \textbf{simple} & 0.0024 $\pm$ 0.0002            & 7.38 $\pm$ 4.11           \\
0.01 & 5$\times$          & semantic\_knn   & 0.1441 $\pm$ 0.0025            & 7.30 $\pm$ 3.77           \\
0.01 & 5$\times$          & synthetic       & 0.0026 $\pm$ 0.0004            & 5.17 $\pm$ 1.42           \\
0.01 & 5$\times$          & random          & 0.0026 $\pm$ 0.0006            & 6.28 $\pm$ 2.98           \\
0.01 & 5$\times$          & original        & 0.0000 $\pm$ 0.0000            & 349.84 $\pm$ 24.68        \\
\midrule
% SOURCE: notebooks/movielens/movielens-GCN-drop0.99-2x-20251125_133138_runtime.csv
% q=0.01, phi=2x
0.01 & 2$\times$          & baseline        & 0.0000 $\pm$ 0.0000            & 8.62 $\pm$ 3.74           \\
0.01 & 2$\times$          & degree\_aware   & 0.0025 $\pm$ 0.0004            & 7.44 $\pm$ 3.23           \\
0.01 & 2$\times$          & \textbf{simple} & 0.0023 $\pm$ 0.0002            & 7.29 $\pm$ 3.16           \\
0.01 & 2$\times$          & semantic\_knn   & 0.1270 $\pm$ 0.0135            & 7.37 $\pm$ 3.82           \\
0.01 & 2$\times$          & synthetic       & 0.0023 $\pm$ 0.0004            & 8.00 $\pm$ 4.02           \\
0.01 & 2$\times$          & random          & 0.0023 $\pm$ 0.0005            & 7.50 $\pm$ 3.64           \\
0.01 & 2$\times$          & original        & 0.0000 $\pm$ 0.0000            & 352.65 $\pm$ 24.33        \\
\midrule

% WARNING: No runtime file found for GCN drop0.95 5x
% WARNING: No runtime file found for GCN drop0.95 2x

% WARNING: No runtime file found for GCN drop0.9 5x
% WARNING: No runtime file found for GCN drop0.9 2x

\end{longtable}

\begin{figure}[H]
  \centering
  \includegraphics[width=1\linewidth]{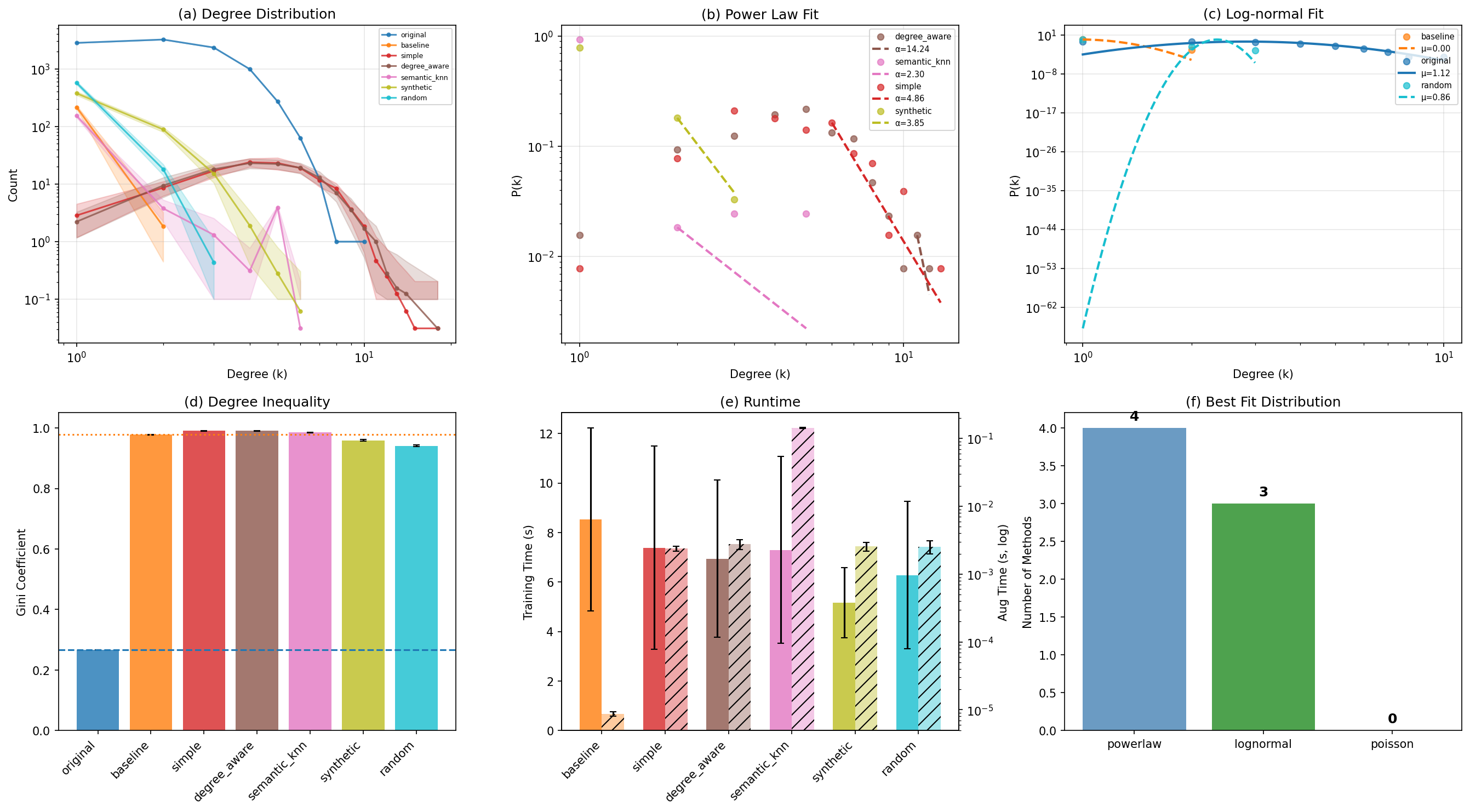}
  \caption{MovieLens (movie--genre), GCN, $q{=}0.01$, $\phi{=}5$: Comprehensive analysis ($M\pm\mathrm{SD}$, $n=32$ seeds) comparing baseline, augmentation methods, and original graph. Panel (a) shows degree distributions on log-log scale with confidence bands; (b) Power Law fits with exponent $\alpha$; (c) Log-normal fits with parameters $\mu$ and $\sigma$; (d) Gini coefficients quantifying degree inequality (lower = more uniform); (e) runtime comparison showing training time (left axis) and augmentation time (right axis, log scale); (f) best-fit distribution counts across methods.}
  \label{fig:movielens_gcn_q01_phi5}
\end{figure}

\begin{figure}[H]
  \centering
  \includegraphics[width=1\linewidth]{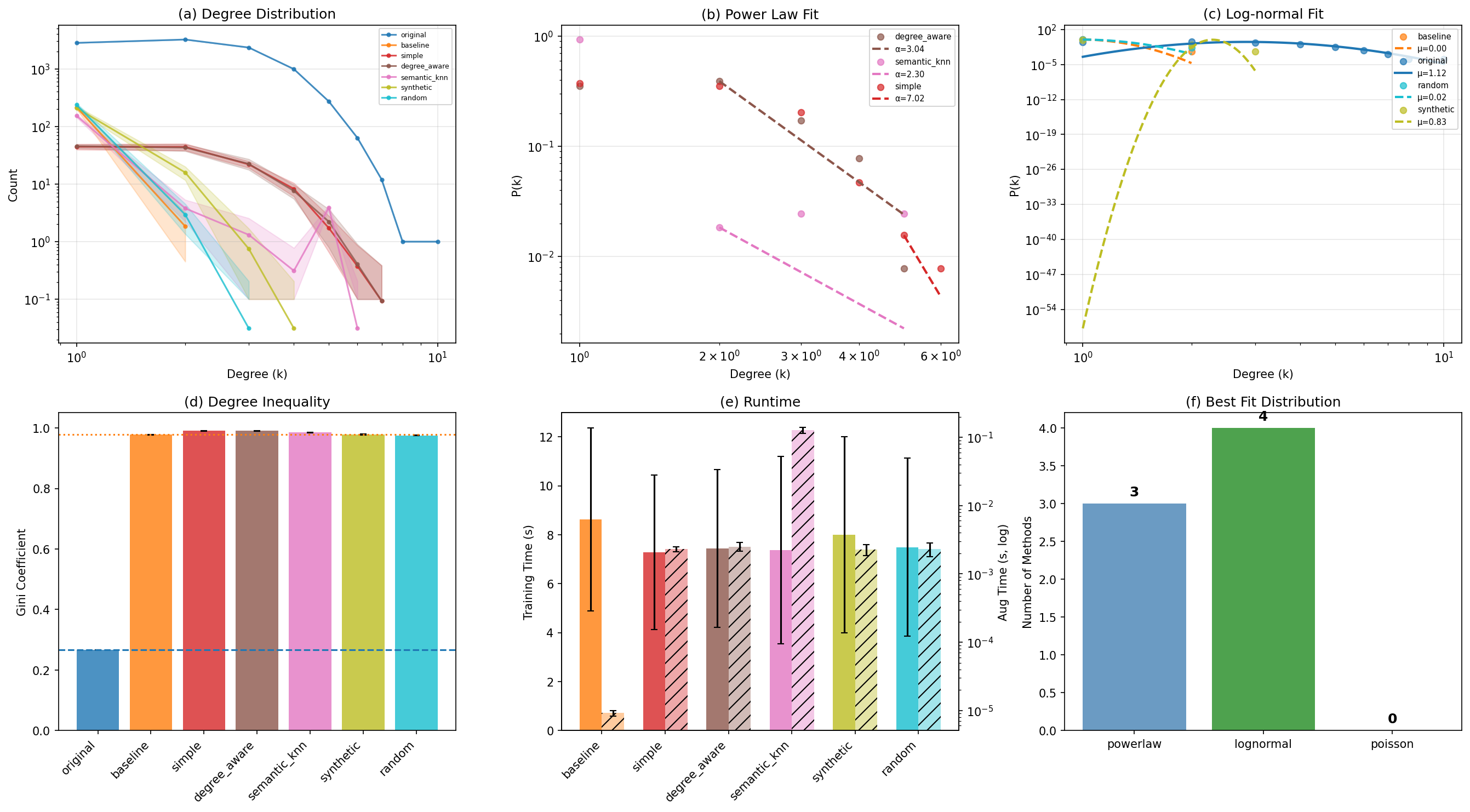}
  \caption{MovieLens (movie--genre), GCN, $q{=}0.01$, $\phi{=}2$: Comprehensive analysis ($M\pm\mathrm{SD}$, $n=32$ seeds) comparing baseline, augmentation methods, and original graph. Panel (a) shows degree distributions on log-log scale with confidence bands; (b) Power Law fits with exponent $\alpha$; (c) Log-normal fits with parameters $\mu$ and $\sigma$; (d) Gini coefficients quantifying degree inequality (lower = more uniform); (e) runtime comparison showing training time (left axis) and augmentation time (right axis, log scale); (f) best-fit distribution counts across methods.}
  \label{fig:movielens_gcn_q01_phi2}
\end{figure}

\subsection{Domain Case Study - GDP}

\subsubsection{Summary}

On GDP, effectiveness depends heavily on encoder and objective. For GAT, \texttt{semantic\_knn} is the only method with clear AUC gains (at $\phi{=}100$: +0.014), while \texttt{synthetic}/\texttt{random} degrade AUC and most methods hurt at lower $\phi$. Brier flips: \texttt{semantic\_knn} yields the largest reduction at $\phi{=}100$ ($-0.054$), with \texttt{simple}/\texttt{degree\_aware} giving mild gains at the same setting; lower $\phi$ generally raises Brier. GraphSAGE shows slight AUC upticks only from \texttt{simple} at $\phi{=}2$ (+0.002) and consistent AUC/Brier drops from \texttt{semantic\_knn}, while \texttt{degree\_aware} is near baseline. GCN favors neither augmentation strongly: AUC edges up modestly for \texttt{simple}/\texttt{semantic\_knn} at $\phi{=}5/2$, but Brier often worsens except small gains from \texttt{degree\_aware} and \texttt{simple} at $\phi{=}5$. Degree-wise, augmentations at $\phi{=}100$ massively densify graphs (mean degree $\sim189$ for several methods), while lower $\phi$ still raise connectivity and cut isolated nodes; \texttt{semantic\_knn} tends to retain higher Gini but reduces isolation. Runtime overheads stay small (aug $\leq0.10$s) though training increases vs.\ baseline, so accuracy changes reflect graph quality, not augmentation cost.

\subsubsection{GAT}

\paragraph{Summary Analysis}

GDP GAT shows limited AUC gains: only \texttt{semantic\_knn} improves at $\phi{=}100$ (+0.014), while \texttt{simple}/\texttt{degree\_aware} are near baseline at $\phi{=}5$ and all methods decline at $\phi{=}2$. Brier improvements are concentrated at $\phi{=}100$ with \texttt{semantic\_knn} leading ($-0.054$) and \texttt{simple}/\texttt{synthetic}/\texttt{random} also lowering error; at lower $\phi$, most methods raise Brier. Degree distributions at $\phi{=}100$ explode in density for most augmentations (mean degree $\sim189$) with reduced inequality for \texttt{random}/\texttt{synthetic}, while \texttt{semantic\_knn} keeps graphs closer to the sparse baseline (mean 5.14, lower Gini). Runtime overheads are small for augmentation ($<0.10$s) though training is higher than baseline; effects are driven by graph changes, not compute.

% Data source: Table~\ref{tab:gdp_gat_auc} and corresponding Brier table  \textbf{AUC-ROC Performance:} \texttt{simple} achieves the highest average AUC improvement (mean $\Delta$AUC=+0.000): at $q=N/A$, $\phi=5x$, AUC=0.786$\pm$0.016 ($\Delta$AUC=+0.005) vs baseline (0.781$\pm$0.020); at $q=N/A$, $\phi=1x$, AUC=0.781$\pm$0.020 ($\Delta$AUC=+0.000). Conversely, \texttt{synthetic} shows the weakest performance (mean $\Delta$AUC=-0.016, range: -0.036 to +0.000).  \textbf{Brier Score Performance:} \texttt{degree_aware} achieves the best Brier score (mean $\Delta$Brier=-0.000). At $q=N/A$, $\phi=5x$, it reaches 0.197$\pm$0.011 ($\Delta$Brier=-0.003) vs baseline (0.200$\pm$0.010).  % SOURCE FILES: % - code-repo/notebooks/gdp/gdp-GAT-nodrop-1x-20251123_223222_summary.csv % - code-repo/notebooks/gdp/gdp-GAT-nodrop-2x-20251125_023430_summary.csv % - code-repo/notebooks/gdp/gdp-GAT-nodrop-5x-20251124_063819_summary.csv

\paragraph{AUC and Brier Score}

% ========== AUC TABLE ==========
\begin{longtable}{c l r r r r r}
\caption{GDP (game--pattern) GAT: AUC-ROC ($M\pm SD$) with paired $t$-tests vs.\ sparse baseline ($n=32$ seeds). A higher AUC is better.}
\label{tab:gdp_gat_auc}\\
\toprule
$\phi$ & Method & AUC $M\pm SD$ & $\Delta$AUC & $t(31)$ & $p$ & $d$ \\
\midrule
\endfirsthead
\toprule
$\phi$ & Method & AUC $M\pm SD$ & $\Delta$AUC & $t(31)$ & $p$ & $d$ \\
\midrule
\endhead
\bottomrule
\endfoot

% SOURCE: notebooks/gdp/gdp-20250919_154107_summary.csv
%         notebooks/gdp/paired-ttest-20250919_155641.csv
% phi=100x
100$\times$    & baseline       & 0.800 $\pm$ 0.022 & +0.000 & ---   & ---   & ---   \\
100$\times$    & degree\_aware  & 0.772 $\pm$ 0.026 & -0.028$^{***}$     & $5.29$  & $<$0.001 & $+0.93$ \\
100$\times$    & simple         & 0.793 $\pm$ 0.023 & -0.007$^{\mathrm{ns}}$ & $1.67$  & 0.104    & $+0.30$ \\
100$\times$    & \textbf{semantic\_knn} & 0.814 $\pm$ 0.017 & +0.014$^{**}$      & $-2.83$ & 0.008    & $-0.50$ \\
100$\times$    & synthetic      & 0.645 $\pm$ 0.061 & -0.155$^{***}$     & $13.42$ & $<$0.001 & $+2.37$ \\
100$\times$    & random         & 0.613 $\pm$ 0.076 & -0.187$^{***}$     & $13.14$ & $<$0.001 & $+2.32$ \\
\midrule
% SOURCE: notebooks/gdp/gdp-GAT-nodrop-5x-20251124_063819_summary.csv
%         notebooks/gdp/gdp-GAT-nodrop-5x-paired-ttest-20251124_063821.csv
% phi=5x
5$\times$      & baseline       & 0.781 $\pm$ 0.020 & +0.000 & ---   & ---   & ---   \\
5$\times$      & degree\_aware  & 0.782 $\pm$ 0.017 & +0.001$^{\mathrm{ns}}$ & $-0.32$ & 0.754    & $-0.06$ \\
5$\times$      & \textbf{simple} & 0.786 $\pm$ 0.016 & +0.005$^{\mathrm{ns}}$ & $-1.17$ & 0.253    & $-0.21$ \\
5$\times$      & semantic\_knn  & 0.772 $\pm$ 0.025 & -0.009$^{\mathrm{ns}}$ & $1.59$  & 0.122    & $+0.28$ \\
5$\times$      & synthetic      & 0.745 $\pm$ 0.023 & -0.036$^{***}$     & $6.79$  & $<$0.001 & $+1.20$ \\
5$\times$      & random         & 0.752 $\pm$ 0.026 & -0.029$^{***}$     & $4.42$  & $<$0.001 & $+0.78$ \\
\midrule
% SOURCE: notebooks/gdp/gdp-GAT-nodrop-2x-20251125_023430_summary.csv
%         notebooks/gdp/gdp-GAT-nodrop-2x-paired-ttest-20251125_023432.csv
% phi=2x
2$\times$      & \textbf{baseline} & 0.781 $\pm$ 0.020 & +0.000 & ---   & ---   & ---   \\
2$\times$      & degree\_aware  & 0.777 $\pm$ 0.021 & -0.004$^{\mathrm{ns}}$ & $0.91$  & 0.368    & $+0.16$ \\
2$\times$      & simple         & 0.777 $\pm$ 0.020 & -0.004$^{\mathrm{ns}}$ & $0.73$  & 0.469    & $+0.13$ \\
2$\times$      & semantic\_knn  & 0.769 $\pm$ 0.021 & -0.012$^{*}$       & $2.18$  & 0.037    & $+0.39$ \\
2$\times$      & synthetic      & 0.769 $\pm$ 0.027 & -0.012$^{*}$       & $2.59$  & 0.014    & $+0.46$ \\
2$\times$      & random         & 0.773 $\pm$ 0.025 & -0.008$^{\mathrm{ns}}$ & $1.35$  & 0.187    & $+0.24$ \\
\midrule
\end{longtable}

% ========== BRIER TABLE ==========
\begin{longtable}{c l r r r r r}
\caption{GDP (game--pattern) GAT: Brier Score ($M\pm SD$) with paired $t$-tests vs.\ sparse baseline ($n=32$ seeds, lower is better).}
\label{tab:gdp_gat_brier}\\
\toprule
$\phi$ & Method & Brier $M\pm SD$ & $\Delta$Brier & $t(31)$ & $p$ & $d$ \\
\midrule
\endfirsthead
\toprule
$\phi$ & Method & Brier $M\pm SD$ & $\Delta$Brier & $t(31)$ & $p$ & $d$ \\
\midrule
\endhead
\bottomrule
\endfoot

% SOURCE: notebooks/gdp/gdp-20250919_154107_summary.csv
%         notebooks/gdp/paired-ttest-20250919_155641.csv
% phi=100x
100$\times$    & baseline       & 0.302 $\pm$ 0.040 & +0.000 & ---  & ---   & ---   \\
100$\times$    & degree\_aware  & 0.337 $\pm$ 0.079 & +0.036$^{*}$        & $-2.59$ & 0.015    & $-0.46$  \\
100$\times$    & simple         & 0.289 $\pm$ 0.018 & -0.013$^{*}$        & $2.41$ & 0.022    & $+0.43$  \\
100$\times$    & \textbf{semantic\_knn} & 0.247 $\pm$ 0.017 & -0.054$^{***}$      & $7.06$ & $<$0.001 & $+1.25$  \\
100$\times$    & synthetic      & 0.269 $\pm$ 0.013 & -0.032$^{***}$      & $4.76$ & $<$0.001 & $+0.84$  \\
100$\times$    & random         & 0.266 $\pm$ 0.017 & -0.036$^{***}$      & $4.92$ & $<$0.001 & $+0.87$  \\
\midrule
% SOURCE: notebooks/gdp/gdp-GAT-nodrop-5x-20251124_063819_summary.csv
%         notebooks/gdp/gdp-GAT-nodrop-5x-paired-ttest-20251124_063821.csv
% phi=5x
5$\times$      & baseline       & 0.200 $\pm$ 0.010 & +0.000 & ---  & ---   & ---   \\
5$\times$      & \textbf{degree\_aware} & 0.197 $\pm$ 0.011 & -0.003$^{\mathrm{ns}}$ & $1.16$ & 0.253    & $+0.21$  \\
5$\times$      & simple         & 0.199 $\pm$ 0.011 & -0.001$^{\mathrm{ns}}$ & $0.56$ & 0.580    & $+0.10$  \\
5$\times$      & semantic\_knn  & 0.213 $\pm$ 0.011 & +0.013$^{***}$      & $-6.02$ & $<$0.001 & $-1.06$  \\
5$\times$      & synthetic      & 0.288 $\pm$ 0.022 & +0.088$^{***}$      & $-20.21$ & $<$0.001 & $-3.57$  \\
5$\times$      & random         & 0.310 $\pm$ 0.027 & +0.110$^{***}$      & $-24.36$ & $<$0.001 & $-4.31$  \\
\midrule
% SOURCE: notebooks/gdp/gdp-GAT-nodrop-2x-20251125_023430_summary.csv
%         notebooks/gdp/gdp-GAT-nodrop-2x-paired-ttest-20251125_023432.csv
% phi=2x
2$\times$      & \textbf{baseline} & 0.200 $\pm$ 0.010 & +0.000 & ---  & ---   & ---   \\
2$\times$      & degree\_aware  & 0.202 $\pm$ 0.008 & +0.002$^{\mathrm{ns}}$ & $-1.35$ & 0.187    & $-0.24$  \\
2$\times$      & simple         & 0.200 $\pm$ 0.008 & +0.000$^{\mathrm{ns}}$ & $-0.00$ & 0.997    & $-0.00$  \\
2$\times$      & semantic\_knn  & 0.210 $\pm$ 0.009 & +0.010$^{***}$      & $-4.51$ & $<$0.001 & $-0.80$  \\
2$\times$      & synthetic      & 0.212 $\pm$ 0.014 & +0.012$^{***}$      & $-3.98$ & $<$0.001 & $-0.70$  \\
2$\times$      & random         & 0.219 $\pm$ 0.013 & +0.019$^{***}$      & $-6.67$ & $<$0.001 & $-1.18$  \\
\midrule
\end{longtable}

\paragraph{Degree Distribution Analysis}

% ========== DEGREE DISTRIBUTION STATISTICS ==========
\begin{longtable}{c l r r r l}
\caption{GDP (game--pattern) GAT: Degree Distribution Statistics ($M\pm SD$, $n=32$ seeds). Lower Gini coefficient indicates more uniform degree distribution.}
\label{tab:gdp_gat_degree}\\
\toprule
$\phi$ & Method & Mean Degree & Gini Coeff. & Num. Isolated & Best Fit \\
\midrule
\endfirsthead
\toprule
$\phi$ & Method & Mean Degree & Gini Coeff. & Num. Isolated & Best Fit \\
\midrule
\endhead
\bottomrule
\endfoot

% SOURCE: notebooks/gdp/degree_analysis_gdp-GAT-nodrop-100x-20251127_143236_degree_stats.csv
%         notebooks/gdp/degree_analysis_gdp-GAT-nodrop-100x-20251127_143236_distribution_fit.csv
% nodrop, phi=100x
100$\times$        & baseline             & 3.4375 $\pm$ 0.0000            & 0.540 $\pm$ 0.000         & 0.0 $\pm$ 0.0             & powerlaw   \\
100$\times$        & degree\_aware        & 188.9423 $\pm$ 0.0000          & 0.518 $\pm$ 0.018         & 53.8 $\pm$ 4.4            & lognormal  \\
100$\times$        & simple               & 188.9423 $\pm$ 0.0000          & 0.614 $\pm$ 0.015         & 53.8 $\pm$ 4.4            & powerlaw   \\
100$\times$        & semantic\_knn        & 5.1447 $\pm$ 0.0796            & 0.331 $\pm$ 0.020         & 19.2 $\pm$ 4.2            & powerlaw   \\
100$\times$        & synthetic            & 188.9423 $\pm$ 0.0000          & 0.264 $\pm$ 0.018         & 0.0 $\pm$ 0.0             & lognormal  \\
100$\times$        & \textbf{random}      & 188.9423 $\pm$ 0.0000          & 0.042 $\pm$ 0.002         & 0.0 $\pm$ 0.0             & lognormal  \\
\midrule

% SOURCE: notebooks/gdp/gdp-GAT-nodrop-5x-20251124_063819_degree_stats.csv
%         notebooks/gdp/gdp-GAT-nodrop-5x-20251124_063819_distribution_fit.csv
% phi=5x
5$\times$          & baseline        & 3.4375 $\pm$ 0.0000            & 0.540 $\pm$ 0.000         & 0.0 $\pm$ 0.0             & powerlaw   \\
5$\times$          & degree\_aware   & 17.1875 $\pm$ 0.0000           & 0.442 $\pm$ 0.008         & 0.0 $\pm$ 0.0             & powerlaw   \\
5$\times$          & simple          & 17.1875 $\pm$ 0.0000           & 0.559 $\pm$ 0.008         & 0.0 $\pm$ 0.0             & lognormal  \\
5$\times$          & semantic\_knn   & 7.1106 $\pm$ 0.0000            & 0.296 $\pm$ 0.000         & 0.0 $\pm$ 0.0             & powerlaw   \\
5$\times$          & synthetic       & 17.1875 $\pm$ 0.0000           & 0.296 $\pm$ 0.007         & 0.0 $\pm$ 0.0             & lognormal  \\
5$\times$          & \textbf{random} & 17.1875 $\pm$ 0.0000           & 0.181 $\pm$ 0.006         & 0.0 $\pm$ 0.0             & powerlaw   \\
\midrule
% SOURCE: notebooks/gdp/gdp-GAT-nodrop-2x-20251125_023430_degree_stats.csv
%         notebooks/gdp/gdp-GAT-nodrop-2x-20251125_023430_distribution_fit.csv
% phi=2x
2$\times$          & baseline        & 3.4375 $\pm$ 0.0000            & 0.540 $\pm$ 0.000         & 0.0 $\pm$ 0.0             & powerlaw   \\
2$\times$          & degree\_aware   & 6.8750 $\pm$ 0.0000            & 0.498 $\pm$ 0.009         & 0.0 $\pm$ 0.0             & powerlaw   \\
2$\times$          & simple          & 6.8750 $\pm$ 0.0000            & 0.567 $\pm$ 0.009         & 0.0 $\pm$ 0.0             & powerlaw   \\
2$\times$          & \textbf{semantic\_knn} & 6.8750 $\pm$ 0.0000            & 0.315 $\pm$ 0.000         & 0.0 $\pm$ 0.0             & powerlaw   \\
2$\times$          & synthetic       & 6.8750 $\pm$ 0.0000            & 0.384 $\pm$ 0.009         & 0.0 $\pm$ 0.0             & powerlaw   \\
2$\times$          & random          & 6.8750 $\pm$ 0.0000            & 0.340 $\pm$ 0.008         & 0.0 $\pm$ 0.0             & powerlaw   \\
\midrule
\end{longtable}

\begin{figure}[H]
  \centering
  \includegraphics[width=1\linewidth]{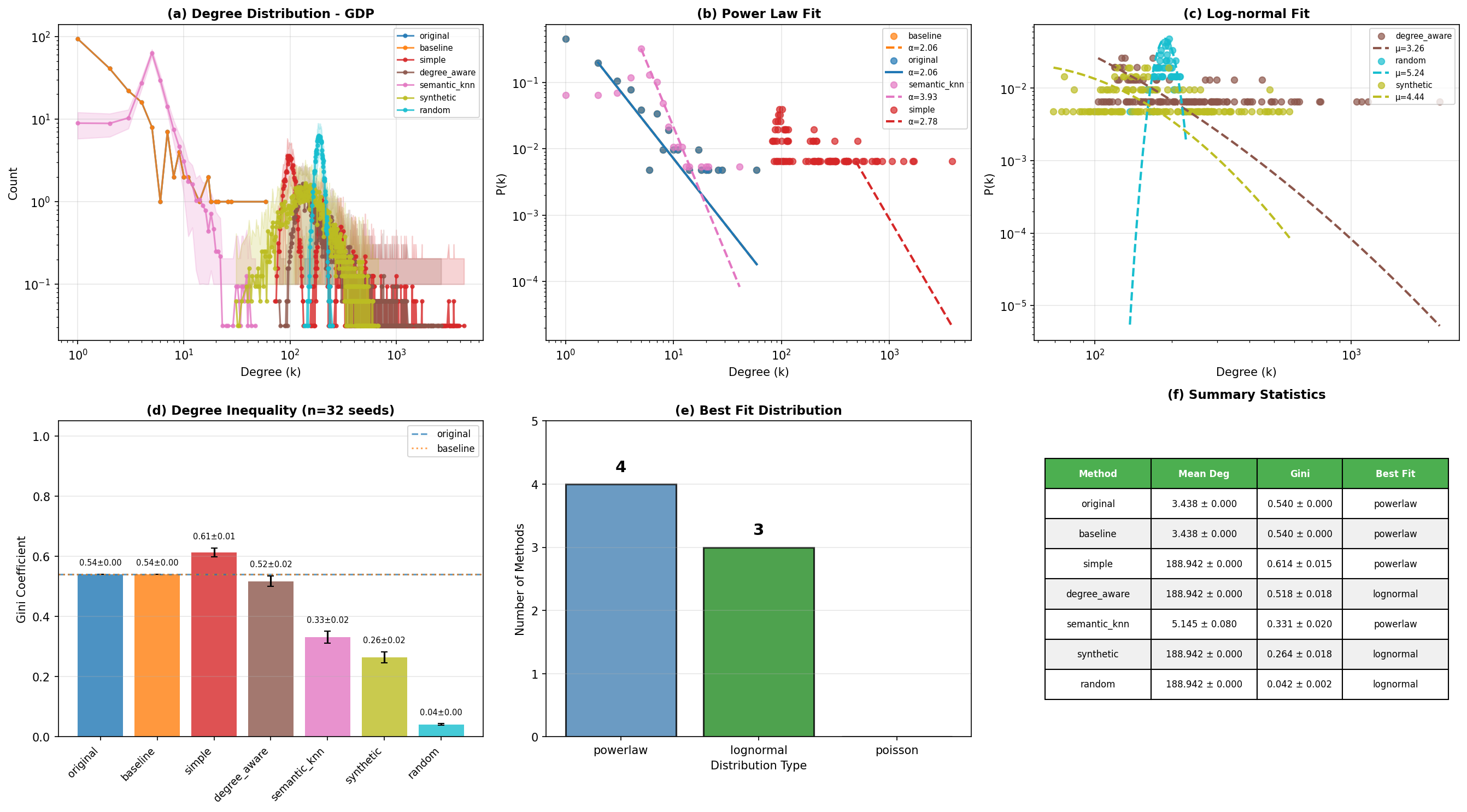}
  \caption{GDP (game--pattern), GAT, $\phi{=}100$: Comprehensive analysis ($M\pm\mathrm{SD}$, $n=32$ seeds) comparing baseline, augmentation methods. Panel (a) shows degree distributions on log-log scale with confidence bands; (b) Power Law fits with exponent $\alpha$; (c) Log-normal fits with parameters $\mu$ and $\sigma$; (d) Gini coefficients quantifying degree inequality (lower = more uniform); (e) runtime comparison showing training time (left axis) and augmentation time (right axis, log scale); (f) best-fit distribution counts across methods.}
  \label{fig:gdp_gat_phi100}
\end{figure}

\paragraph{Runtime Analysis}

\begin{longtable}{c l r r}
\caption{GDP (game--pattern) GAT: Runtime Statistics ($M\pm SD$, seconds, $n=32$ seeds). Lower times are better.}
\label{tab:gdp_gat_runtime}\\
\toprule
$\phi$ & Method & Aug. Time (s) & Train Time (s) \\
\midrule
\endfirsthead
\toprule
$\phi$ & Method & Aug. Time (s) & Train Time (s) \\
\midrule
\endhead
\bottomrule
\endfoot

% SOURCE: notebooks/gdp/gdp-GAT-nodrop-5x-20251124_063819_runtime.csv
% phi=5x
5$\times$          & baseline        & 0.0008 $\pm$ 0.0000            & 70.48 $\pm$ 2.32          \\
5$\times$          & degree\_aware   & 0.0029 $\pm$ 0.0077            & 114.87 $\pm$ 4.60         \\
5$\times$          & \textbf{simple} & 0.0010 $\pm$ 0.0001            & 109.93 $\pm$ 3.83         \\
5$\times$          & semantic\_knn   & 0.0992 $\pm$ 0.0004            & 95.71 $\pm$ 20.63         \\
5$\times$          & synthetic       & 0.0012 $\pm$ 0.0004            & 247.25 $\pm$ 85.61        \\
5$\times$          & random          & 0.0010 $\pm$ 0.0000            & 259.04 $\pm$ 82.96        \\
\midrule
% SOURCE: notebooks/gdp/gdp-GAT-nodrop-2x-20251125_023430_runtime.csv
% phi=2x
2$\times$          & baseline        & 0.0008 $\pm$ 0.0000            & 69.11 $\pm$ 2.07          \\
2$\times$          & degree\_aware   & 0.0014 $\pm$ 0.0015            & 79.00 $\pm$ 1.28          \\
2$\times$          & \textbf{simple} & 0.0009 $\pm$ 0.0000            & 79.37 $\pm$ 2.71          \\
2$\times$          & semantic\_knn   & 0.0795 $\pm$ 0.0020            & 89.45 $\pm$ 7.87          \\
2$\times$          & synthetic       & 0.0010 $\pm$ 0.0004            & 100.68 $\pm$ 15.59        \\
2$\times$          & random          & 0.0009 $\pm$ 0.0000            & 102.15 $\pm$ 24.68        \\
\midrule
\end{longtable}

\begin{figure}[H]
  \centering
  \includegraphics[width=1\linewidth]{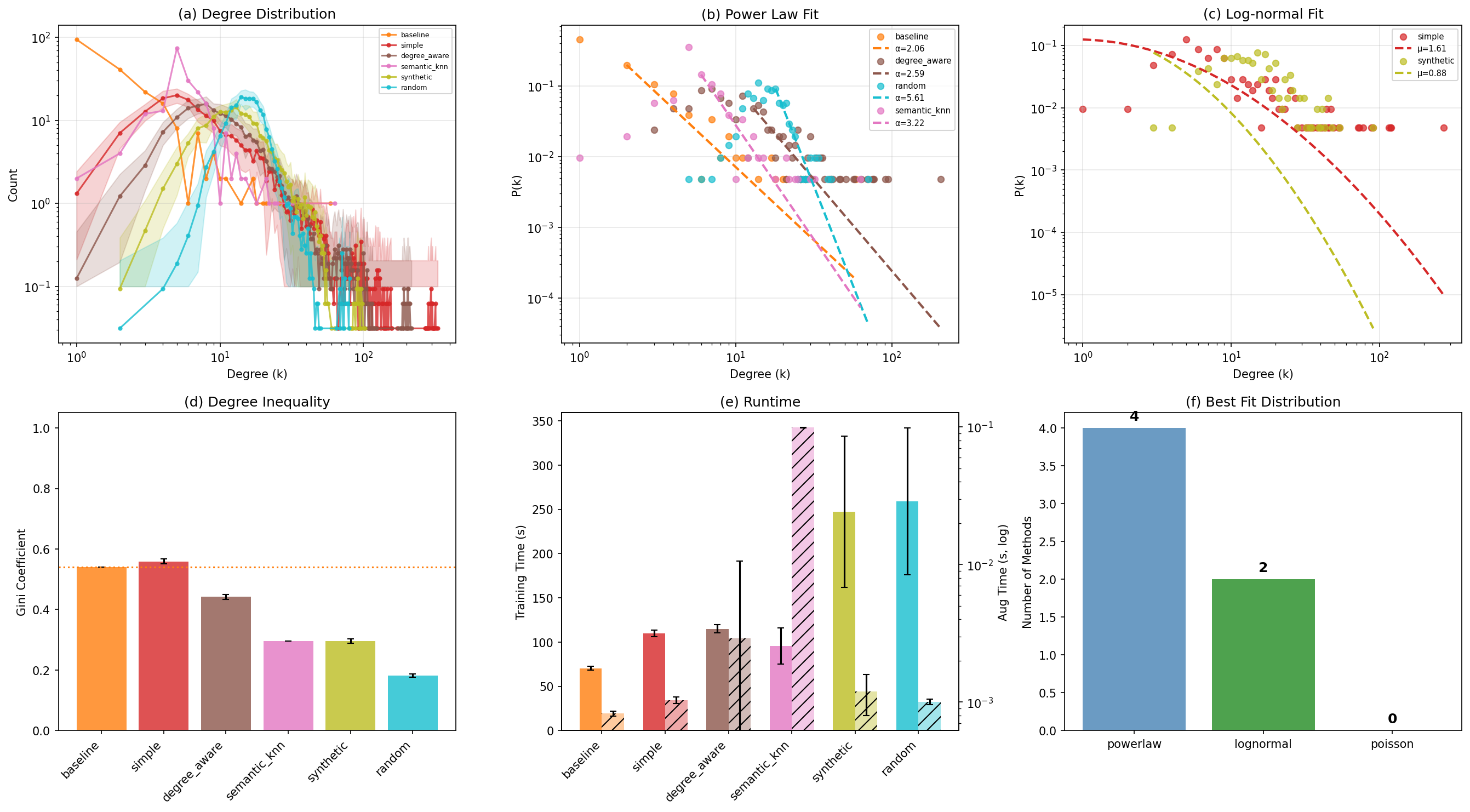}
  \caption{GDP (game--pattern), GAT, $\phi{=}5$: Comprehensive analysis ($M\pm\mathrm{SD}$, $n=32$ seeds) comparing baseline, augmentation methods, and original graph. Panel (a) shows degree distributions on log-log scale with confidence bands; (b) Power Law fits with exponent $\alpha$; (c) Log-normal fits with parameters $\mu$ and $\sigma$; (d) Gini coefficients quantifying degree inequality (lower = more uniform); (e) runtime comparison showing training time (left axis) and augmentation time (right axis, log scale); (f) best-fit distribution counts across methods.}
  \label{fig:gdp_gat_phi5}
\end{figure}

\begin{figure}[H]
  \centering
  \includegraphics[width=1\linewidth]{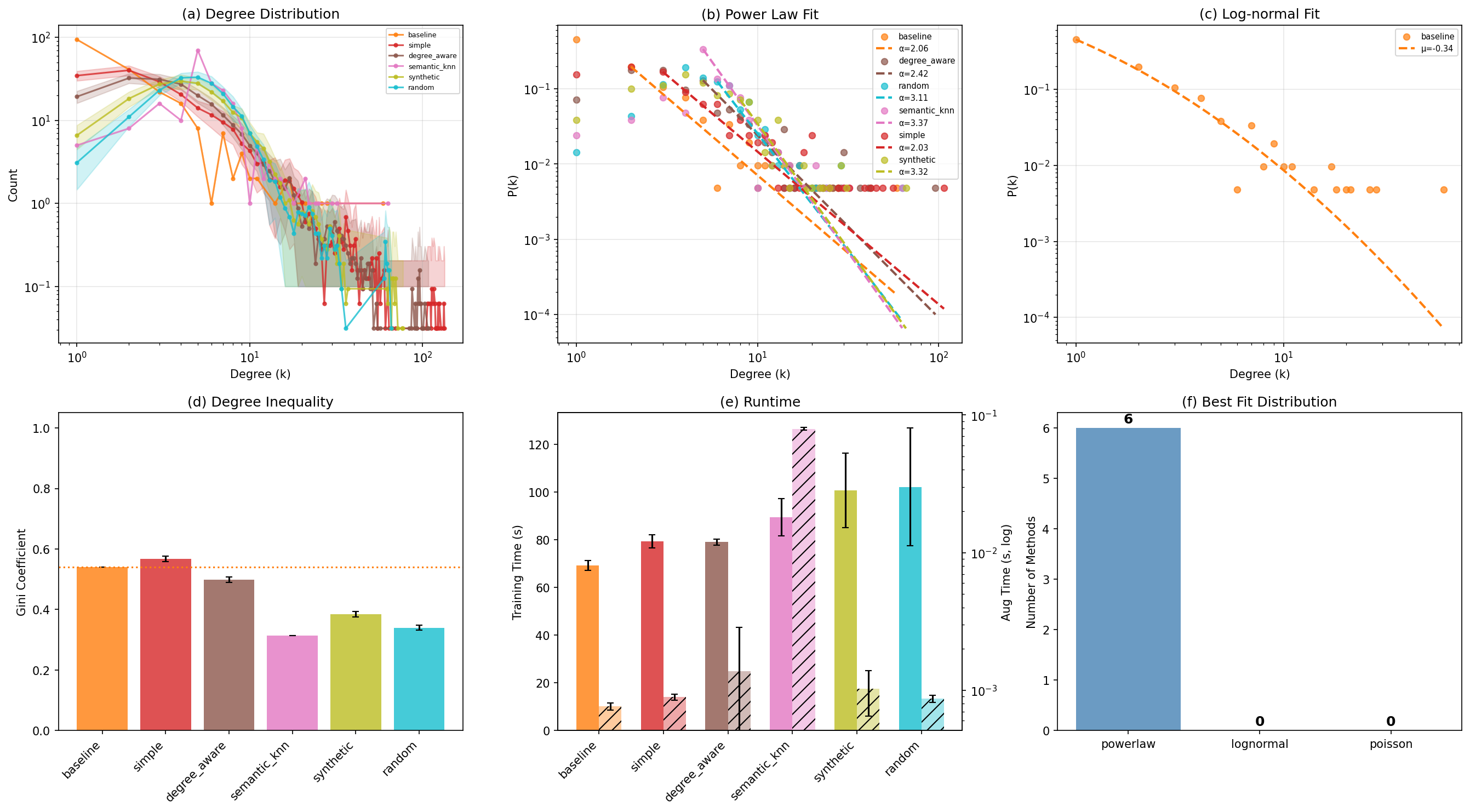}
  \caption{GDP (game--pattern), GAT, $\phi{=}2$: Comprehensive analysis ($M\pm\mathrm{SD}$, $n=32$ seeds) comparing baseline, augmentation methods, and original graph. Panel (a) shows degree distributions on log-log scale with confidence bands; (b) Power Law fits with exponent $\alpha$; (c) Log-normal fits with parameters $\mu$ and $\sigma$; (d) Gini coefficients quantifying degree inequality (lower = more uniform); (e) runtime comparison showing training time (left axis) and augmentation time (right axis, log scale); (f) best-fit distribution counts across methods.}
  \label{fig:gdp_gat_phi2}
\end{figure}

\subsubsection{GraphSAGE}

\paragraph{Summary Analysis}

GDP GraphSAGE largely tracks baseline: \texttt{simple} offers a small AUC lift at $\phi{=}2$ (+0.002), while \texttt{semantic\_knn} consistently lowers AUC (down to $-0.040$ at $\phi{=}5$) and \texttt{synthetic}/\texttt{random} also degrade. Brier favors \texttt{simple}/\texttt{degree\_aware} with minor reductions, whereas \texttt{semantic\_knn} and especially \texttt{synthetic}/\texttt{random} sharply increase error. Degree distributions show augmentations densify graphs (mean degree $17$ at $\phi{=}5$, $6.9$ at $\phi{=}2$) with varied inequality: \texttt{random}/\texttt{synthetic} reduce Gini and isolate counts most, while \texttt{semantic\_knn} yields lower Gini but modest density. Augmentation costs remain negligible ($<0.10$s); training grows vs.\ baseline but differences stem from graph structure changes.

% Data source: Table~\ref{tab:gdp_graphsage_auc} and corresponding Brier table  \textbf{AUC-ROC Performance:} \texttt{simple} achieves the highest average AUC improvement (mean $\Delta$AUC=-0.001): at $q=N/A$, $\phi=2x$, AUC=0.805$\pm$0.015 ($\Delta$AUC=+0.002) vs baseline (0.802$\pm$0.012); at $q=N/A$, $\phi=1x$, AUC=0.802$\pm$0.012 ($\Delta$AUC=+0.000). Conversely, \texttt{semantic_knn} shows the weakest performance (mean $\Delta$AUC=-0.023, range: -0.040 to +0.000).  \textbf{Brier Score Performance:} \texttt{simple} achieves the best Brier score (mean $\Delta$Brier=-0.003). At $q=N/A$, $\phi=2x$, it reaches 0.168$\pm$0.011 ($\Delta$Brier=-0.007) vs baseline (0.175$\pm$0.013).  % SOURCE FILES: % - code-repo/notebooks/gdp/gdp-GraphSAGE-nodrop-1x-20251124_132738_summary.csv % - code-repo/notebooks/gdp/gdp-GraphSAGE-nodrop-2x-20251125_011851_summary.csv % - code-repo/notebooks/gdp/gdp-GraphSAGE-nodrop-5x-20251124_152252_summary.csv

\paragraph{AUC and Brier Score}

% ========== AUC TABLE ==========
\begin{longtable}{c l r r r r r}
\caption{GDP (game--pattern) GraphSAGE: AUC-ROC ($M\pm SD$) with paired $t$-tests vs.\ sparse baseline ($n=32$ seeds). A higher AUC is better.}
\label{tab:gdp_graphsage_auc}\\
\toprule
$\phi$ & Method & AUC $M\pm SD$ & $\Delta$AUC & $t(31)$ & $p$ & $d$ \\
\midrule
\endfirsthead
\toprule
$\phi$ & Method & AUC $M\pm SD$ & $\Delta$AUC & $t(31)$ & $p$ & $d$ \\
\midrule
\endhead
\bottomrule
\endfoot

% WARNING: No summary file found for GraphSAGE nodrop 100x
% SOURCE: notebooks/gdp/gdp-GraphSAGE-nodrop-5x-20251124_152252_summary.csv
%         notebooks/gdp/gdp-GraphSAGE-nodrop-5x-paired-ttest-20251124_152254.csv
% phi=5x
5$\times$      & \textbf{baseline} & 0.802 $\pm$ 0.012 & +0.000 & ---   & ---   & ---   \\
5$\times$      & degree\_aware  & 0.799 $\pm$ 0.013 & -0.003$^{\mathrm{ns}}$ & $1.07$  & 0.292    & $+0.19$ \\
5$\times$      & simple         & 0.798 $\pm$ 0.013 & -0.004$^{\mathrm{ns}}$ & $1.90$  & 0.067    & $+0.34$ \\
5$\times$      & semantic\_knn  & 0.763 $\pm$ 0.016 & -0.040$^{***}$     & $15.41$ & $<$0.001 & $+2.72$ \\
5$\times$      & synthetic      & 0.780 $\pm$ 0.019 & -0.022$^{***}$     & $5.37$  & $<$0.001 & $+0.95$ \\
5$\times$      & random         & 0.789 $\pm$ 0.021 & -0.013$^{***}$     & $3.66$  & $<$0.001 & $+0.65$ \\
\midrule
% SOURCE: notebooks/gdp/gdp-GraphSAGE-nodrop-2x-20251125_011851_summary.csv
%         notebooks/gdp/gdp-GraphSAGE-nodrop-2x-paired-ttest-20251125_011859.csv
% phi=2x
2$\times$      & baseline       & 0.802 $\pm$ 0.012 & +0.000 & ---   & ---   & ---   \\
2$\times$      & degree\_aware  & 0.801 $\pm$ 0.014 & -0.001$^{\mathrm{ns}}$ & $0.51$  & 0.611    & $+0.09$ \\
2$\times$      & \textbf{simple} & 0.805 $\pm$ 0.015 & +0.002$^{\mathrm{ns}}$ & $-0.76$ & 0.454    & $-0.13$ \\
2$\times$      & semantic\_knn  & 0.775 $\pm$ 0.017 & -0.028$^{***}$     & $7.77$  & $<$0.001 & $+1.37$ \\
2$\times$      & synthetic      & 0.767 $\pm$ 0.017 & -0.036$^{***}$     & $9.70$  & $<$0.001 & $+1.72$ \\
2$\times$      & random         & 0.764 $\pm$ 0.018 & -0.039$^{***}$     & $11.53$ & $<$0.001 & $+2.04$ \\
\midrule
\end{longtable}

% ========== BRIER TABLE ==========
\begin{longtable}{c l r r r r r}
\caption{GDP (game--pattern) GraphSAGE: Brier Score ($M\pm SD$) with paired $t$-tests vs.\ sparse baseline ($n=32$ seeds, lower is better).}
\label{tab:gdp_graphsage_brier}\\
\toprule
$\phi$ & Method & Brier $M\pm SD$ & $\Delta$Brier & $t(31)$ & $p$ & $d$ \\
\midrule
\endfirsthead
\toprule
$\phi$ & Method & Brier $M\pm SD$ & $\Delta$Brier & $t(31)$ & $p$ & $d$ \\
\midrule
\endhead
\bottomrule
\endfoot

% WARNING: No summary file found for GraphSAGE nodrop 100x
% SOURCE: notebooks/gdp/gdp-GraphSAGE-nodrop-5x-20251124_152252_summary.csv
%         notebooks/gdp/gdp-GraphSAGE-nodrop-5x-paired-ttest-20251124_152254.csv
% phi=5x
5$\times$      & baseline       & 0.175 $\pm$ 0.013 & +0.000 & ---  & ---   & ---   \\
5$\times$      & degree\_aware  & 0.173 $\pm$ 0.016 & -0.002$^{\mathrm{ns}}$ & $0.42$ & 0.676    & $+0.07$  \\
5$\times$      & \textbf{simple} & 0.172 $\pm$ 0.012 & -0.003$^{\mathrm{ns}}$ & $0.88$ & 0.383    & $+0.16$  \\
5$\times$      & semantic\_knn  & 0.192 $\pm$ 0.013 & +0.017$^{***}$      & $-5.95$ & $<$0.001 & $-1.05$  \\
5$\times$      & synthetic      & 0.233 $\pm$ 0.012 & +0.059$^{***}$      & $-18.91$ & $<$0.001 & $-3.34$  \\
5$\times$      & random         & 0.229 $\pm$ 0.012 & +0.055$^{***}$      & $-17.04$ & $<$0.001 & $-3.01$  \\
\midrule
% SOURCE: notebooks/gdp/gdp-GraphSAGE-nodrop-2x-20251125_011851_summary.csv
%         notebooks/gdp/gdp-GraphSAGE-nodrop-2x-paired-ttest-20251125_011859.csv
% phi=2x
2$\times$      & baseline       & 0.175 $\pm$ 0.013 & +0.000 & ---  & ---   & ---   \\
2$\times$      & degree\_aware  & 0.170 $\pm$ 0.012 & -0.005$^{\mathrm{ns}}$ & $1.75$ & 0.090    & $+0.31$  \\
2$\times$      & \textbf{simple} & 0.168 $\pm$ 0.011 & -0.007$^{**}$       & $2.82$ & 0.008    & $+0.50$  \\
2$\times$      & semantic\_knn  & 0.188 $\pm$ 0.013 & +0.013$^{***}$      & $-4.25$ & $<$0.001 & $-0.75$  \\
2$\times$      & synthetic      & 0.191 $\pm$ 0.013 & +0.016$^{***}$      & $-5.52$ & $<$0.001 & $-0.98$  \\
2$\times$      & random         & 0.197 $\pm$ 0.015 & +0.022$^{***}$      & $-6.13$ & $<$0.001 & $-1.08$  \\
\midrule
\end{longtable}

\paragraph{Degree Distribution Analysis}

% ========== DEGREE DISTRIBUTION STATISTICS ==========
\begin{longtable}{c l r r r r l}
\caption{GDP (game--pattern) GraphSAGE: Degree Distribution Statistics ($M\pm SD$, $n=32$ seeds). Lower Gini coefficient indicates more uniform degree distribution.}
\label{tab:gdp_graphsage_degree}\\
\toprule
$q$ & $\phi$ & Method & Mean Degree & Gini Coeff. & Num. Isolated & Best Fit \\
\midrule
\endfirsthead
\toprule
$q$ & $\phi$ & Method & Mean Degree & Gini Coeff. & Num. Isolated & Best Fit \\
\midrule
\endhead
\bottomrule
\endfoot

% SOURCE: notebooks/gdp/gdp-GraphSAGE-nodrop-5x-20251124_152252_degree_stats.csv
%         notebooks/gdp/gdp-GraphSAGE-nodrop-5x-20251124_152252_distribution_fit.csv
% phi=5x
5$\times$          & baseline        & 3.4375 $\pm$ 0.0000            & 0.540 $\pm$ 0.000         & 0.0 $\pm$ 0.0             & powerlaw   \\
5$\times$          & degree\_aware   & 17.1875 $\pm$ 0.0000           & 0.442 $\pm$ 0.008         & 0.0 $\pm$ 0.0             & powerlaw   \\
5$\times$          & simple          & 17.1875 $\pm$ 0.0000           & 0.559 $\pm$ 0.008         & 0.0 $\pm$ 0.0             & lognormal  \\
5$\times$          & semantic\_knn   & 7.1106 $\pm$ 0.0000            & 0.296 $\pm$ 0.000         & 0.0 $\pm$ 0.0             & powerlaw   \\
5$\times$          & synthetic       & 17.1875 $\pm$ 0.0000           & 0.296 $\pm$ 0.007         & 0.0 $\pm$ 0.0             & lognormal  \\
5$\times$          & \textbf{random} & 17.1875 $\pm$ 0.0000           & 0.181 $\pm$ 0.006         & 0.0 $\pm$ 0.0             & powerlaw   \\
\midrule
% SOURCE: notebooks/gdp/gdp-GraphSAGE-nodrop-2x-20251125_011851_degree_stats.csv
%         notebooks/gdp/gdp-GraphSAGE-nodrop-2x-20251125_011851_distribution_fit.csv
% phi=2x
2$\times$          & baseline        & 3.4375 $\pm$ 0.0000            & 0.540 $\pm$ 0.000         & 0.0 $\pm$ 0.0             & powerlaw   \\
2$\times$          & degree\_aware   & 6.8750 $\pm$ 0.0000            & 0.498 $\pm$ 0.009         & 0.0 $\pm$ 0.0             & powerlaw   \\
2$\times$          & simple          & 6.8750 $\pm$ 0.0000            & 0.567 $\pm$ 0.009         & 0.0 $\pm$ 0.0             & powerlaw   \\
2$\times$          & \textbf{semantic\_knn} & 6.8750 $\pm$ 0.0000            & 0.315 $\pm$ 0.000         & 0.0 $\pm$ 0.0             & powerlaw   \\
2$\times$          & synthetic       & 6.8750 $\pm$ 0.0000            & 0.384 $\pm$ 0.009         & 0.0 $\pm$ 0.0             & powerlaw   \\
2$\times$          & random          & 6.8750 $\pm$ 0.0000            & 0.340 $\pm$ 0.008         & 0.0 $\pm$ 0.0             & powerlaw   \\
\midrule
\end{longtable}

% ========== DEGREE DISTRIBUTION FIGURES ==========
% Insert degree distribution plots at appropriate locations:
% Use \includegraphics command with these paths:
%
% Figure for phi=5x:
%   notebooks/gdp/gdp-GraphSAGE-nodrop-5x-20251124_152252_analysis_combined.png
% Figure for phi=2x:
%   notebooks/gdp/gdp-GraphSAGE-nodrop-2x-20251125_011851_analysis_combined.png
%

\paragraph{Runtime Analysis}

% ========== RUNTIME STATISTICS ==========
\begin{longtable}{c l r r}
\caption{GDP (game--pattern) GraphSAGE: Runtime Statistics ($M\pm SD$, seconds, $n=32$ seeds). Lower times are better.}
\label{tab:gdp_graphsage_runtime}\\
\toprule
$\phi$ & Method & Aug. Time (s) & Train Time (s) \\
\midrule
\endfirsthead
\toprule
$\phi$ & Method & Aug. Time (s) & Train Time (s) \\
\midrule
\endhead
\bottomrule
\endfoot

% SOURCE: notebooks/gdp/gdp-GraphSAGE-nodrop-5x-20251124_152252_runtime.csv
% phi=5x
5$\times$          & baseline        & 0.0007 $\pm$ 0.0000            & 50.46 $\pm$ 3.40          \\
5$\times$          & degree\_aware   & 0.0015 $\pm$ 0.0008            & 89.71 $\pm$ 8.85          \\
5$\times$          & \textbf{simple} & 0.0009 $\pm$ 0.0001            & 84.11 $\pm$ 5.51          \\
5$\times$          & semantic\_knn   & 0.0976 $\pm$ 0.0031            & 64.55 $\pm$ 9.27          \\
5$\times$          & synthetic       & 0.0011 $\pm$ 0.0006            & 285.53 $\pm$ 59.60        \\
5$\times$          & random          & 0.0017 $\pm$ 0.0046            & 294.93 $\pm$ 50.21        \\
\midrule
% SOURCE: notebooks/gdp/gdp-GraphSAGE-nodrop-2x-20251125_011851_runtime.csv
% phi=2x
2$\times$          & baseline        & 0.0008 $\pm$ 0.0001            & 51.52 $\pm$ 3.29          \\
2$\times$          & degree\_aware   & 0.0015 $\pm$ 0.0010            & 59.70 $\pm$ 3.63          \\
2$\times$          & simple          & 0.0016 $\pm$ 0.0039            & 58.55 $\pm$ 3.02          \\
2$\times$          & semantic\_knn   & 0.0882 $\pm$ 0.0350            & 65.54 $\pm$ 12.46         \\
2$\times$          & synthetic       & 0.0014 $\pm$ 0.0018            & 69.10 $\pm$ 16.66         \\
2$\times$          & \textbf{random} & 0.0009 $\pm$ 0.0001            & 73.48 $\pm$ 23.85         \\
\midrule
\end{longtable}

\begin{figure}[H]
  \centering
  \includegraphics[width=1\linewidth]{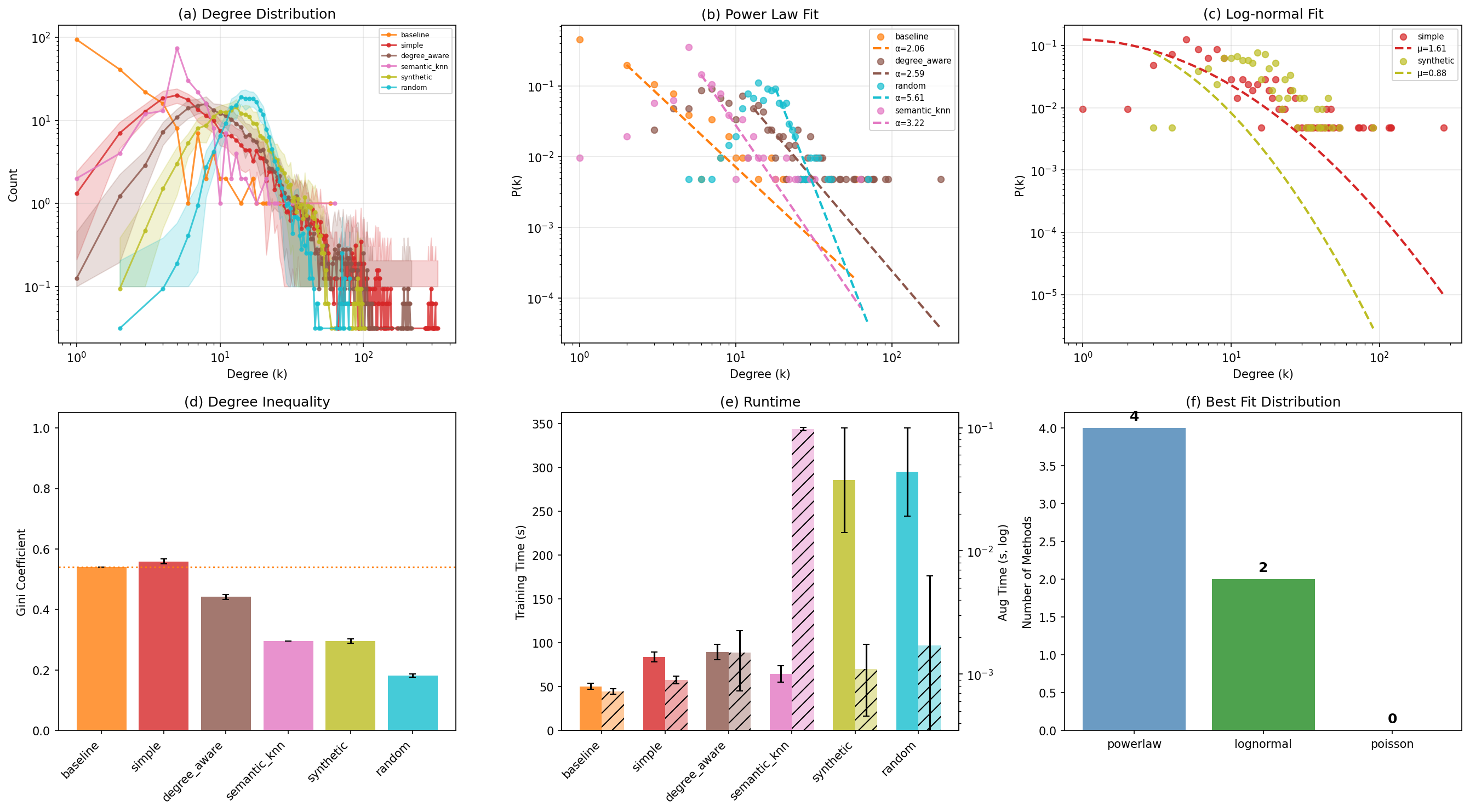}
  \caption{GDP (game--pattern), GraphSAGE, $\phi{=}5$: Comprehensive analysis ($M\pm\mathrm{SD}$, $n=32$ seeds) comparing baseline, augmentation methods, and original graph. Panel (a) shows degree distributions on log-log scale with confidence bands; (b) Power Law fits with exponent $\alpha$; (c) Log-normal fits with parameters $\mu$ and $\sigma$; (d) Gini coefficients quantifying degree inequality (lower = more uniform); (e) runtime comparison showing training time (left axis) and augmentation time (right axis, log scale); (f) best-fit distribution counts across methods.}
  \label{fig:gdp_graphsage_phi5}
\end{figure}

\begin{figure}[H]
  \centering
  \includegraphics[width=1\linewidth]{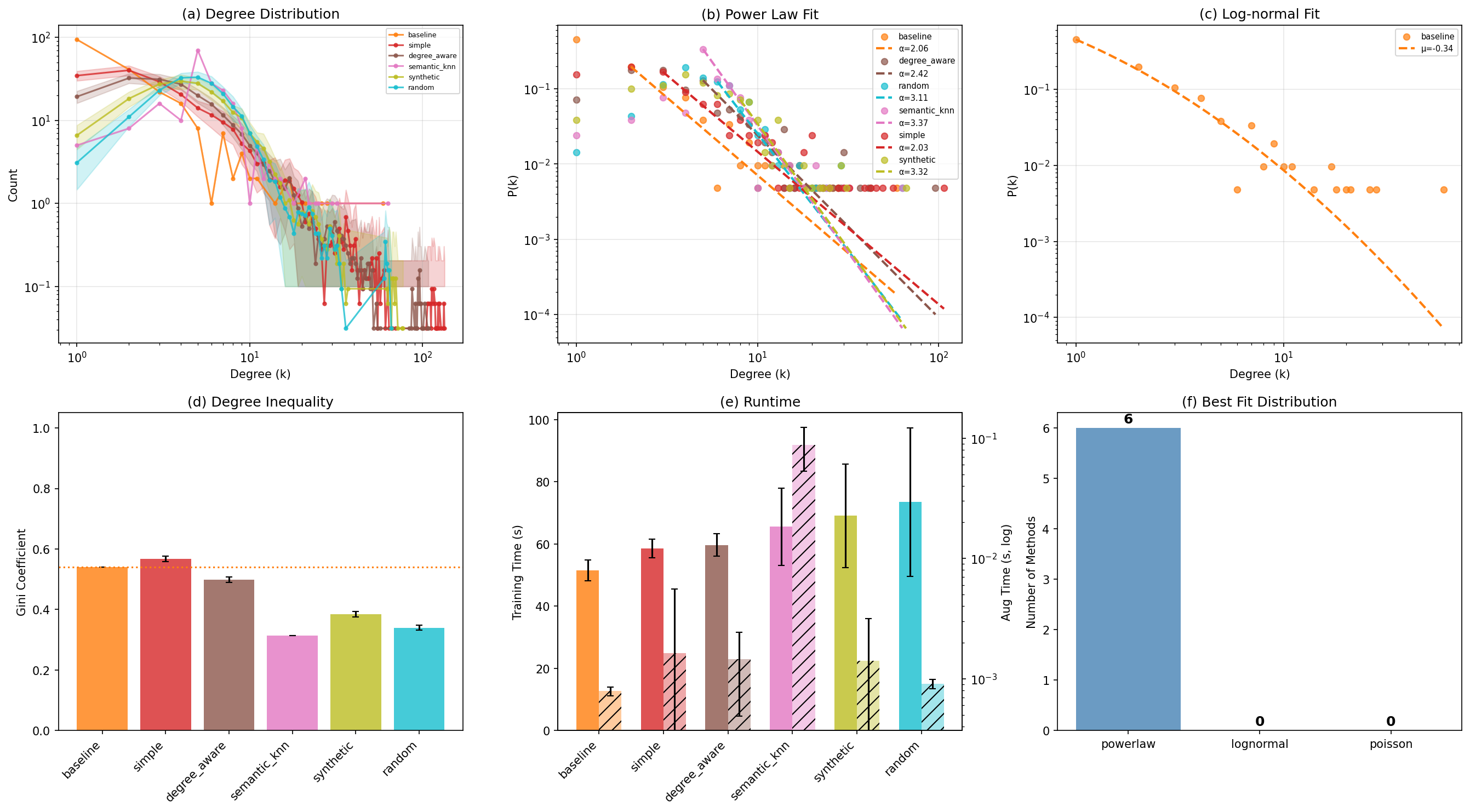}
  \caption{GDP (game--pattern), GraphSAGE, $\phi{=}2$: Comprehensive analysis ($M\pm\mathrm{SD}$, $n=32$ seeds) comparing baseline, augmentation methods, and original graph. Panel (a) shows degree distributions on log-log scale with confidence bands; (b) Power Law fits with exponent $\alpha$; (c) Log-normal fits with parameters $\mu$ and $\sigma$; (d) Gini coefficients quantifying degree inequality (lower = more uniform); (e) runtime comparison showing training time (left axis) and augmentation time (right axis, log scale); (f) best-fit distribution counts across methods.}
  \label{fig:gdp_graphsage_phi2}
\end{figure}

\subsubsection{GCN}

\paragraph{Summary Analysis}

GDP GCN yields limited gains: \texttt{simple} and \texttt{semantic\_knn} provide modest AUC at $\phi{=}5/2$ (up to +0.020), while \texttt{degree\_aware} lags. Brier improves slightly for \texttt{degree\_aware}/\texttt{simple} at $\phi{=}5$ ($-0.031/-0.024$) and for \texttt{semantic\_knn} at $\phi{=}2$ ($-0.009$); \texttt{synthetic}/\texttt{random} generally worsen calibration. Degree distributions densify with larger $\phi$ (raising mean degree and reducing isolates), with \texttt{random}/\texttt{synthetic} lowering Gini most, while \texttt{semantic\_knn} keeps graphs relatively sparse. Augmentation costs stay tiny ($<0.063$s); training is higher than baseline but differences reflect graph quality, not augmentation overhead.

% Data source: Table~\ref{tab:gdp_gcn_auc} and corresponding Brier table  \textbf{AUC-ROC Performance:} \texttt{semantic_knn} achieves the highest average AUC improvement (mean $\Delta$AUC=-0.018): at $q=N/A$, $\phi=1x$, AUC=0.767$\pm$0.028 ($\Delta$AUC=+0.000) vs baseline (0.767$\pm$0.028); at $q=N/A$, $\phi=2x$, AUC=0.746$\pm$0.029 ($\Delta$AUC=-0.020). Conversely, \texttt{degree_aware} shows the weakest performance (mean $\Delta$AUC=-0.076, range: -0.162 to +0.000).  \textbf{Brier Score Performance:} \texttt{semantic_knn} achieves the best Brier score (mean $\Delta$Brier=-0.005). At $q=N/A$, $\phi=2x$, it reaches 0.200$\pm$0.015 ($\Delta$Brier=-0.009) vs baseline (0.209$\pm$0.012).  % SOURCE FILES: % - code-repo/notebooks/gdp/gdp-GCN-nodrop-1x-20251124_213152_summary.csv % - code-repo/notebooks/gdp/gdp-GCN-nodrop-2x-20251125_095749_summary.csv % - code-repo/notebooks/gdp/gdp-GCN-nodrop-5x-20251125_034556_summary.csv

\paragraph{AUC and Brier Score}

% ========== AUC TABLE ==========
\begin{longtable}{c l r r r r r}
\caption{GDP (game--pattern) GCN: AUC-ROC ($M\pm SD$) with paired $t$-tests vs.\ sparse baseline ($n=32$ seeds). A higher AUC is better.}
\label{tab:gdp_gcn_auc}\\
\toprule
$\phi$ & Method & AUC $M\pm SD$ & $\Delta$AUC & $t(31)$ & $p$ & $d$ \\
\midrule
\endfirsthead
\toprule
$\phi$ & Method & AUC $M\pm SD$ & $\Delta$AUC & $t(31)$ & $p$ & $d$ \\
\midrule
\endhead
\bottomrule
\endfoot

% WARNING: No summary file found for GCN nodrop 100x
% SOURCE: notebooks/gdp/gdp-GCN-nodrop-5x-20251125_034556_summary.csv
%         notebooks/gdp/gdp-GCN-nodrop-5x-paired-ttest-20251125_034559.csv
% phi=5x
5$\times$      & \textbf{baseline} & 0.767 $\pm$ 0.028 & +0.000 & ---   & ---   & ---   \\
5$\times$      & degree\_aware  & 0.605 $\pm$ 0.049 & -0.162$^{***}$     & $16.14$ & $<$0.001 & $+2.85$ \\
5$\times$      & simple         & 0.620 $\pm$ 0.043 & -0.147$^{***}$     & $15.89$ & $<$0.001 & $+2.81$ \\
5$\times$      & semantic\_knn  & 0.733 $\pm$ 0.036 & -0.034$^{***}$     & $3.91$  & $<$0.001 & $+0.69$ \\
5$\times$      & synthetic      & 0.634 $\pm$ 0.062 & -0.132$^{***}$     & $10.84$ & $<$0.001 & $+1.92$ \\
5$\times$      & random         & 0.605 $\pm$ 0.079 & -0.161$^{***}$     & $11.05$ & $<$0.001 & $+1.95$ \\
\midrule
% SOURCE: notebooks/gdp/gdp-GCN-nodrop-2x-20251125_095749_summary.csv
%         notebooks/gdp/gdp-GCN-nodrop-2x-paired-ttest-20251125_095751.csv
% phi=2x
2$\times$      & \textbf{baseline} & 0.767 $\pm$ 0.028 & +0.000 & ---   & ---   & ---   \\
2$\times$      & degree\_aware  & 0.700 $\pm$ 0.027 & -0.067$^{***}$     & $9.52$  & $<$0.001 & $+1.68$ \\
2$\times$      & simple         & 0.705 $\pm$ 0.031 & -0.062$^{***}$     & $7.93$  & $<$0.001 & $+1.40$ \\
2$\times$      & semantic\_knn  & 0.746 $\pm$ 0.029 & -0.020$^{**}$      & $2.83$  & 0.008    & $+0.50$ \\
2$\times$      & synthetic      & 0.731 $\pm$ 0.035 & -0.035$^{***}$     & $5.01$  & $<$0.001 & $+0.89$ \\
2$\times$      & random         & 0.732 $\pm$ 0.030 & -0.034$^{***}$     & $4.90$  & $<$0.001 & $+0.87$ \\
\midrule
\end{longtable}

% ========== BRIER TABLE ==========
\begin{longtable}{c l r r r r r}
\caption{GDP (game--pattern) GCN: Brier Score ($M\pm SD$) with paired $t$-tests vs.\ sparse baseline ($n=32$ seeds, lower is better).}
\label{tab:gdp_gcn_brier}\\
\toprule
$\phi$ & Method & Brier $M\pm SD$ & $\Delta$Brier & $t(31)$ & $p$ & $d$ \\
\midrule
\endfirsthead
\toprule
$\phi$ & Method & Brier $M\pm SD$ & $\Delta$Brier & $t(31)$ & $p$ & $d$ \\
\midrule
\endhead
\bottomrule
\endfoot

% WARNING: No summary file found for GCN nodrop 100x
% SOURCE: notebooks/gdp/gdp-GCN-nodrop-5x-20251125_034556_summary.csv
%         notebooks/gdp/gdp-GCN-nodrop-5x-paired-ttest-20251125_034559.csv
% phi=5x
5$\times$      & baseline       & 0.209 $\pm$ 0.012 & +0.000 & ---  & ---   & ---   \\
5$\times$      & degree\_aware  & 0.249 $\pm$ 0.001 & +0.041$^{***}$      & $-19.14$ & $<$0.001 & $-3.38$  \\
5$\times$      & simple         & 0.249 $\pm$ 0.002 & +0.040$^{***}$      & $-18.89$ & $<$0.001 & $-3.34$  \\
5$\times$      & \textbf{semantic\_knn} & 0.203 $\pm$ 0.016 & -0.005$^{\mathrm{ns}}$ & $1.47$ & 0.152    & $+0.26$  \\
5$\times$      & synthetic      & 0.233 $\pm$ 0.022 & +0.025$^{***}$      & $-5.75$ & $<$0.001 & $-1.02$  \\
5$\times$      & random         & 0.240 $\pm$ 0.020 & +0.031$^{***}$      & $-7.44$ & $<$0.001 & $-1.32$  \\
\midrule
% SOURCE: notebooks/gdp/gdp-GCN-nodrop-2x-20251125_095749_summary.csv
%         notebooks/gdp/gdp-GCN-nodrop-2x-paired-ttest-20251125_095751.csv
% phi=2x
2$\times$      & baseline       & 0.209 $\pm$ 0.012 & +0.000 & ---  & ---   & ---   \\
2$\times$      & degree\_aware  & 0.229 $\pm$ 0.011 & +0.020$^{***}$      & $-7.84$ & $<$0.001 & $-1.39$  \\
2$\times$      & simple         & 0.225 $\pm$ 0.009 & +0.016$^{***}$      & $-6.40$ & $<$0.001 & $-1.13$  \\
2$\times$      & \textbf{semantic\_knn} & 0.200 $\pm$ 0.015 & -0.009$^{*}$        & $2.64$ & 0.013    & $+0.47$  \\
2$\times$      & synthetic      & 0.202 $\pm$ 0.014 & -0.007$^{*}$        & $2.15$ & 0.040    & $+0.38$  \\
2$\times$      & random         & 0.203 $\pm$ 0.016 & -0.006$^{\mathrm{ns}}$ & $1.64$ & 0.112    & $+0.29$  \\
\midrule
\end{longtable}

\paragraph{Degree Distribution Analysis}

% ========== DEGREE DISTRIBUTION STATISTICS ==========
\begin{longtable}{c l r r r r l}
\caption{GDP (game--pattern) GCN: Degree Distribution Statistics ($M\pm SD$, $n=32$ seeds). Lower Gini coefficient indicates more uniform degree distribution.}
\label{tab:gdp_gcn_degree}\\
\toprule
$q$ & $\phi$ & Method & Mean Degree & Gini Coeff. & Num. Isolated & Best Fit \\
\midrule
\endfirsthead
\toprule
$q$ & $\phi$ & Method & Mean Degree & Gini Coeff. & Num. Isolated & Best Fit \\
\midrule
\endhead
\bottomrule
\endfoot

% SOURCE: notebooks/gdp/gdp-GCN-nodrop-5x-20251125_034556_degree_stats.csv
%         notebooks/gdp/gdp-GCN-nodrop-5x-20251125_034556_distribution_fit.csv
% phi=5x
5$\times$          & baseline        & 3.4375 $\pm$ 0.0000            & 0.540 $\pm$ 0.000         & 0.0 $\pm$ 0.0             & powerlaw   \\
5$\times$          & degree\_aware   & 17.1875 $\pm$ 0.0000           & 0.442 $\pm$ 0.008         & 0.0 $\pm$ 0.0             & powerlaw   \\
5$\times$          & simple          & 17.1875 $\pm$ 0.0000           & 0.559 $\pm$ 0.008         & 0.0 $\pm$ 0.0             & powerlaw   \\
5$\times$          & semantic\_knn   & 7.1106 $\pm$ 0.0000            & 0.296 $\pm$ 0.000         & 0.0 $\pm$ 0.0             & powerlaw   \\
5$\times$          & synthetic       & 17.1875 $\pm$ 0.0000           & 0.296 $\pm$ 0.007         & 0.0 $\pm$ 0.0             & lognormal  \\
5$\times$          & \textbf{random} & 17.1875 $\pm$ 0.0000           & 0.181 $\pm$ 0.006         & 0.0 $\pm$ 0.0             & powerlaw   \\
\midrule
% SOURCE: notebooks/gdp/gdp-GCN-nodrop-2x-20251125_095749_degree_stats.csv
%         notebooks/gdp/gdp-GCN-nodrop-2x-20251125_095749_distribution_fit.csv
% phi=2x
2$\times$          & baseline        & 3.4375 $\pm$ 0.0000            & 0.540 $\pm$ 0.000         & 0.0 $\pm$ 0.0             & powerlaw   \\
2$\times$          & degree\_aware   & 6.8750 $\pm$ 0.0000            & 0.498 $\pm$ 0.009         & 0.0 $\pm$ 0.0             & powerlaw   \\
2$\times$          & simple          & 6.8750 $\pm$ 0.0000            & 0.567 $\pm$ 0.009         & 0.0 $\pm$ 0.0             & powerlaw   \\
2$\times$          & \textbf{semantic\_knn} & 6.8750 $\pm$ 0.0000            & 0.315 $\pm$ 0.000         & 0.0 $\pm$ 0.0             & powerlaw   \\
2$\times$          & synthetic       & 6.8750 $\pm$ 0.0000            & 0.384 $\pm$ 0.009         & 0.0 $\pm$ 0.0             & powerlaw   \\
2$\times$          & random          & 6.8750 $\pm$ 0.0000            & 0.340 $\pm$ 0.008         & 0.0 $\pm$ 0.0             & powerlaw   \\
\midrule
\end{longtable}

% ========== DEGREE DISTRIBUTION FIGURES ==========
% Insert degree distribution plots at appropriate locations:
% Use \includegraphics command with these paths:
%
% Figure for phi=5x:
%   notebooks/gdp/gdp-GCN-nodrop-5x-20251125_034556_analysis_combined.png
% Figure for phi=2x:
%   notebooks/gdp/gdp-GCN-nodrop-2x-20251125_095749_analysis_combined.png
%

\paragraph{Runtime Analysis}

% ========== RUNTIME STATISTICS ==========
\begin{longtable}{c l r r}
\caption{GDP (game--pattern) GCN: Runtime Statistics ($M\pm SD$, seconds, $n=32$ seeds). Lower times are better.}
\label{tab:gdp_gcn_runtime}\\
\toprule
$\phi$ & Method & Aug. Time (s) & Train Time (s) \\
\midrule
\endfirsthead
\toprule
$\phi$ & Method & Aug. Time (s) & Train Time (s) \\
\midrule
\endhead
\bottomrule
\endfoot

% SOURCE: notebooks/gdp/gdp-GCN-nodrop-5x-20251125_034556_runtime.csv
% phi=5x
5$\times$          & baseline        & 0.0009 $\pm$ 0.0008            & 47.42 $\pm$ 3.34          \\
5$\times$          & degree\_aware   & 0.0016 $\pm$ 0.0012            & 175.64 $\pm$ 74.79        \\
5$\times$          & \textbf{simple} & 0.0009 $\pm$ 0.0001            & 169.94 $\pm$ 85.77        \\
5$\times$          & semantic\_knn   & 0.0980 $\pm$ 0.0033            & 90.26 $\pm$ 18.84         \\
5$\times$          & synthetic       & 0.0010 $\pm$ 0.0003            & 260.38 $\pm$ 118.23       \\
5$\times$          & random          & 0.0009 $\pm$ 0.0001            & 243.10 $\pm$ 114.05       \\
\midrule
% SOURCE: notebooks/gdp/gdp-GCN-nodrop-2x-20251125_095749_runtime.csv
% phi=2x
2$\times$          & baseline        & 0.0008 $\pm$ 0.0000            & 47.79 $\pm$ 3.36          \\
2$\times$          & degree\_aware   & 0.0012 $\pm$ 0.0000            & 120.59 $\pm$ 27.40        \\
2$\times$          & \textbf{simple} & 0.0009 $\pm$ 0.0000            & 114.75 $\pm$ 18.65        \\
2$\times$          & semantic\_knn   & 0.0812 $\pm$ 0.0010            & 87.45 $\pm$ 23.11         \\
2$\times$          & synthetic       & 0.0010 $\pm$ 0.0000            & 88.50 $\pm$ 16.09         \\
2$\times$          & random          & 0.0009 $\pm$ 0.0000            & 88.82 $\pm$ 15.28         \\
\midrule
\end{longtable}

\begin{figure}[H]
  \centering
  \includegraphics[width=1\linewidth]{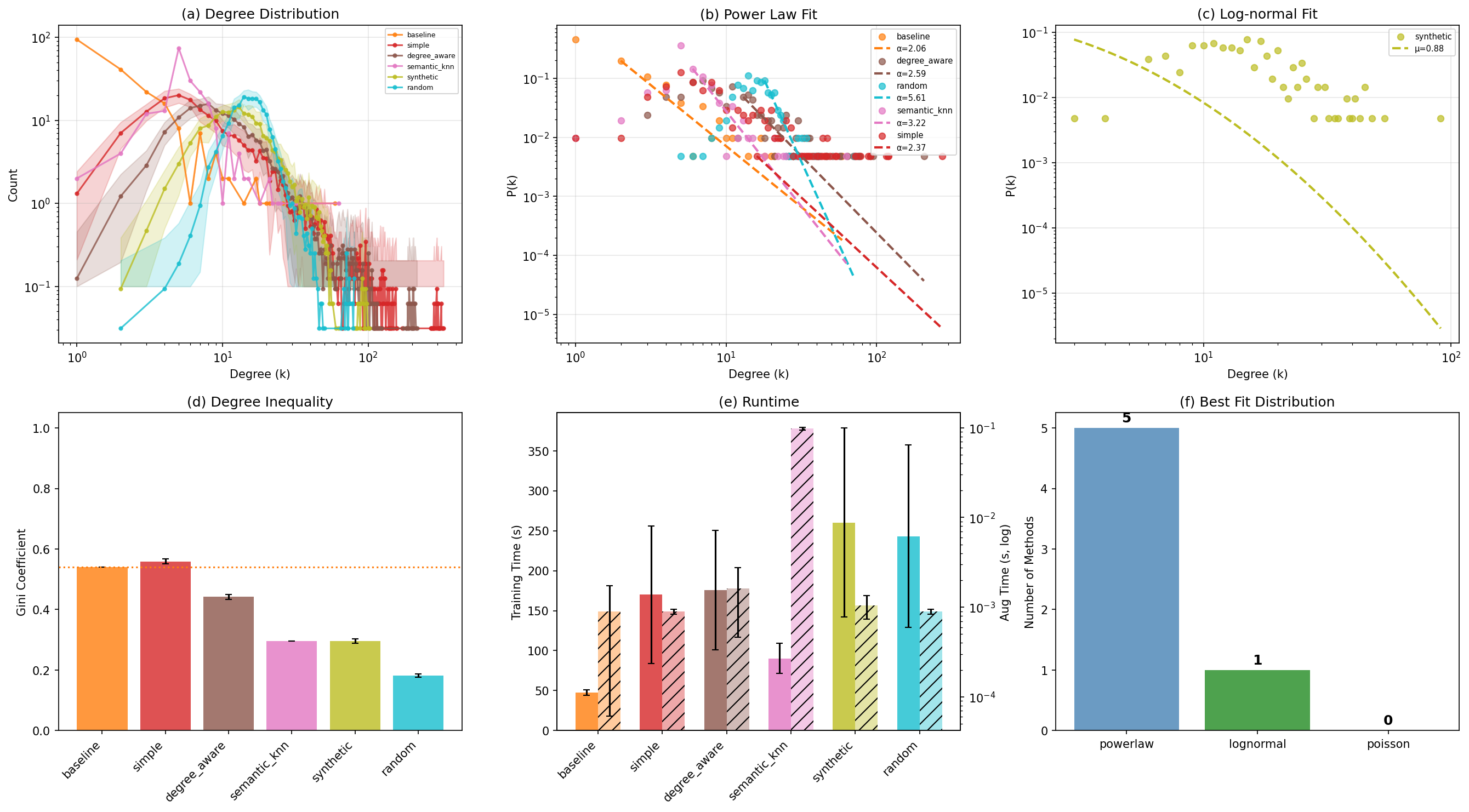}
  \caption{GDP (game--pattern), GCN, $\phi{=}5$: Comprehensive analysis ($M\pm\mathrm{SD}$, $n=32$ seeds) comparing baseline, and augmentation methods. Panel (a) shows degree distributions on log-log scale with confidence bands; (b) Power Law fits with exponent $\alpha$; (c) Log-normal fits with parameters $\mu$ and $\sigma$; (d) Gini coefficients quantifying degree inequality (lower = more uniform); (e) runtime comparison showing training time (left axis) and augmentation time (right axis, log scale); (f) best-fit distribution counts across methods.}
  \label{fig:gdp_gcn_phi5}
\end{figure}

\begin{figure}[H]
  \centering
  \includegraphics[width=1\linewidth]{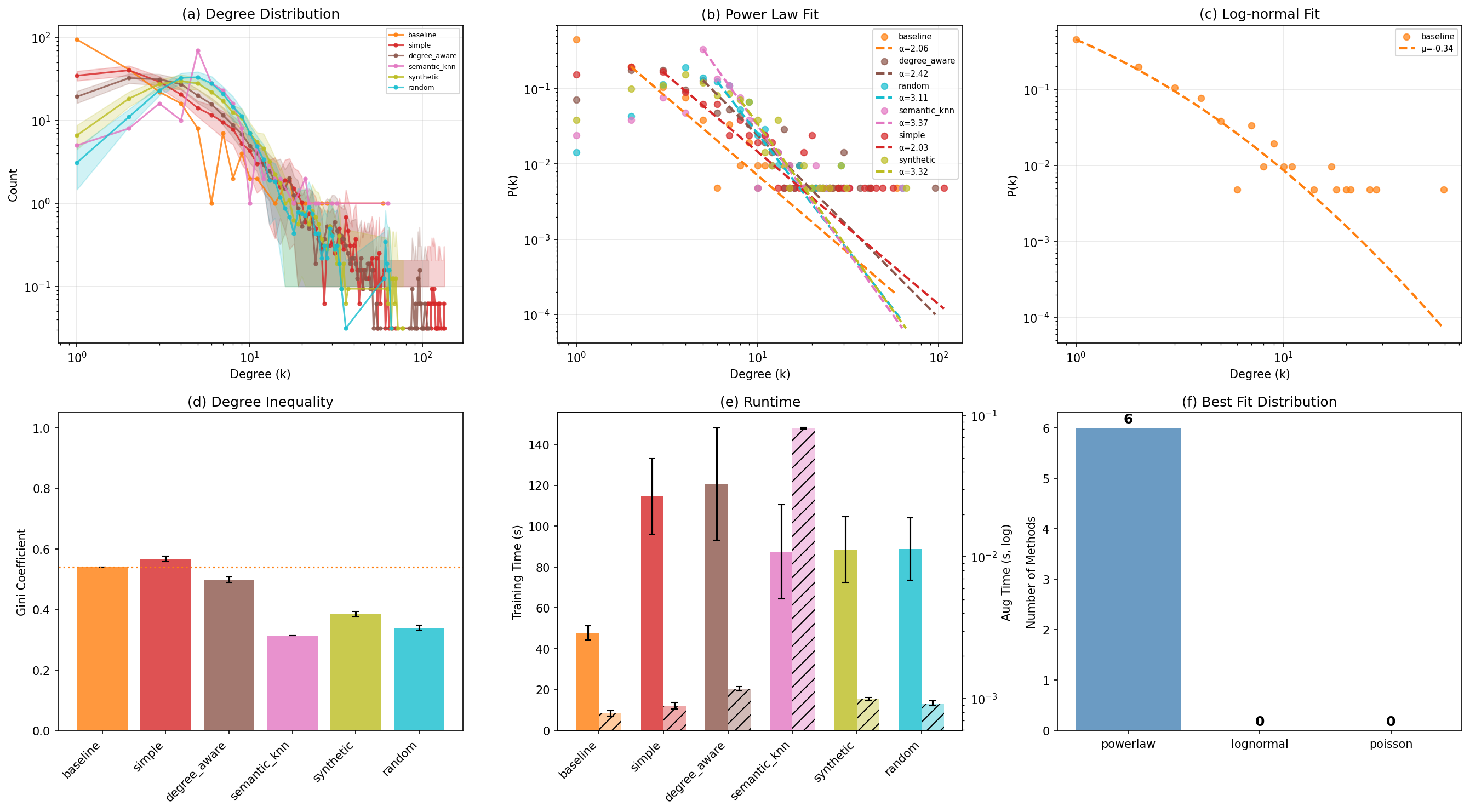}
  \caption{GDP (game--pattern), GCN, $\phi{=}2$: Comprehensive analysis ($M\pm\mathrm{SD}$, $n=32$ seeds) comparing baseline and augmentation methods. Panel (a) shows degree distributions on log-log scale with confidence bands; (b) Power Law fits with exponent $\alpha$; (c) Log-normal fits with parameters $\mu$ and $\sigma$; (d) Gini coefficients quantifying degree inequality (lower = more uniform); (e) runtime comparison showing training time (left axis) and augmentation time (right axis, log scale); (f) best-fit distribution counts across methods.}
  \label{fig:gdp_gcn_phi2}
\end{figure}

\end{document}